\documentclass[runningheads]{llncs}

\usepackage{eccv}
\usepackage{xcolor}
\usepackage{multirow}
\usepackage{wrapfig} 
\usepackage{float}
\definecolor{customblue}{RGB}{0,114,189}

\usepackage{eccvabbrv}

\usepackage{graphicx}
\usepackage{booktabs}
\usepackage{makecell}  

\usepackage[accsupp]{axessibility}  

\usepackage[pagebackref,breaklinks,colorlinks,citecolor=eccvblue]{hyperref}

\usepackage{orcidlink}

\begin{document}

\title{FM-SIREN \& FM-FINER: Implicit Neural Representation Using Nyquist-based Orthogonality}

\titlerunning{FM-SIREN \& FM-FINER}

\author{M. Alsakabi\textsuperscript{\rm 1}, W. Mobeirek\textsuperscript{\rm 2}, J. M. Dolan\textsuperscript{\rm 1}, O. K. Tonguz\textsuperscript{\rm 1}  \vspace{.5em} \\
}

\authorrunning{M.~Alsakabi et al.}
\institute{Carnegie Mellon University \and
Google\\
}
\maketitle

\begin{abstract}
Existing periodic activation-based implicit neural representation (INR) networks, such as SIREN and FINER, suffer from hidden feature redundancy, where neurons within a layer capture overlapping frequency components due to the use of a fixed frequency multiplier. This redundancy limits the expressive capacity of multilayer perceptrons (MLPs). Drawing inspiration from classical signal processing methods such as the Discrete Sine Transform (DST), in this paper, we propose FM-SIREN and FM-FINER, which assign Nyquist-informed, neuron-specific frequency multipliers to periodic activations. Contrary to existing approaches, our design introduces frequency diversity without requiring hyperparameter tuning or additional network depth. This simple yet principled approach reduces the redundancy of features by nearly $50\%$ and consistently improves signal reconstruction across diverse INR tasks, such as fitting 1D audio, 2D image and 3D shape, and video, outperforming their baseline counterparts while maintaining efficiency.
\end{abstract}
\section{Introduction}
\vspace{-5pt}

\par Implicit neural representations (INRs) have emerged as a powerful paradigm for continuous signal modeling, where a multilayer perceptron (MLP) is trained to map coordinates directly to signal values, enabling continuous signal representations that are resolution-independent \cite{essakine2024we, popescu2009multilayer}. Their ability to represent complex signals continuously and compactly has driven significant interest across a wide range of applications, including image and video compression, 3D scene reconstruction, novel view synthesis, and medical imaging \cite{sitzmann2019scene, mildenhall2021nerf, martel2021acorn, molaei2023implicit, strumpler2022implicit, lee2023entropy}. Yet despite their flexibility and iterative optimization, a single-layer periodic-activation INR, structurally equivalent to the Discrete Sine Transform (DST) in parameter count and architecture, paradoxically fails to match the reconstruction quality of this classical, closed-form method that requires no iterative optimization \cite{ahmed2006discrete}.

\par Among the most prominent INR architectures are those with periodic activations, such as SIREN \cite{sitzmann2019siren} and FINER \cite{liu2024finer}, which use sinusoidal activations as basis functions \cite{vetterli2014foundations} to represent high-frequency signal components. Their simple design enables smooth, continuous high frequency signal fitting, making them easy to adopt across diverse INR tasks. However, as shown in Figure \ref{fig:all_subfigures_jpg}, a classical DST achieves higher reconstruction quality than a single-layer SIREN or FINER, revealing a fundamental bottleneck: despite their iterative optimization and architectural flexibility, these networks do not fully exploit the expressive capacity available within each layer. Since each layer in an MLP builds upon the representations of the previous one, this per-layer inefficiency accumulates through the network, acting as a persistent bottleneck that deeper architectures can only partially mitigate.

\begin{figure}[t]
    \centering
    \begin{subfigure}{0.31\textwidth}
        \centering
        \includegraphics[width=\linewidth]{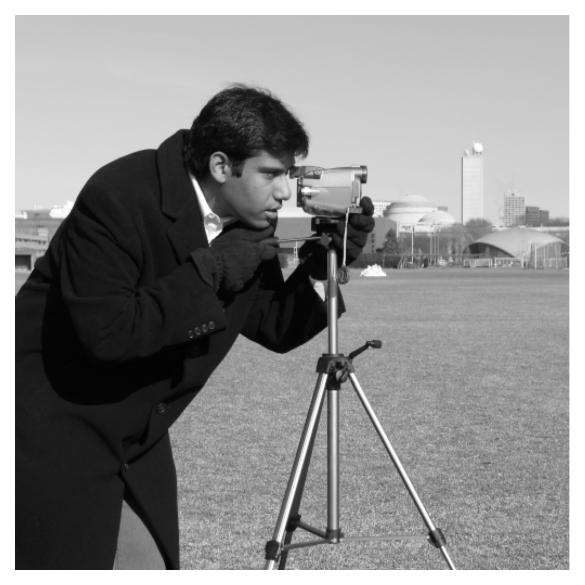}
        \caption{Ground Truth}
        \label{fig:sub1}
    \end{subfigure}
    \hfill
    \begin{subfigure}{0.31\textwidth}
        \centering
        \includegraphics[width=\linewidth]{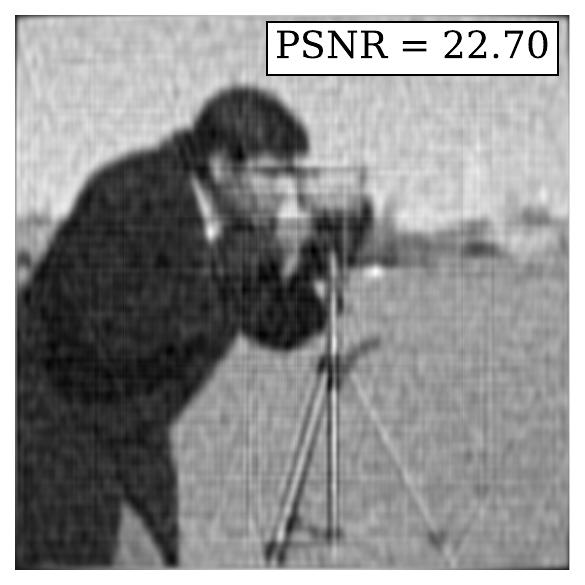}
        \caption{DST}
        \label{fig:sub2}
    \end{subfigure}
    \hfill
    \begin{subfigure}{0.31\textwidth}
        \centering
        \includegraphics[width=\linewidth]{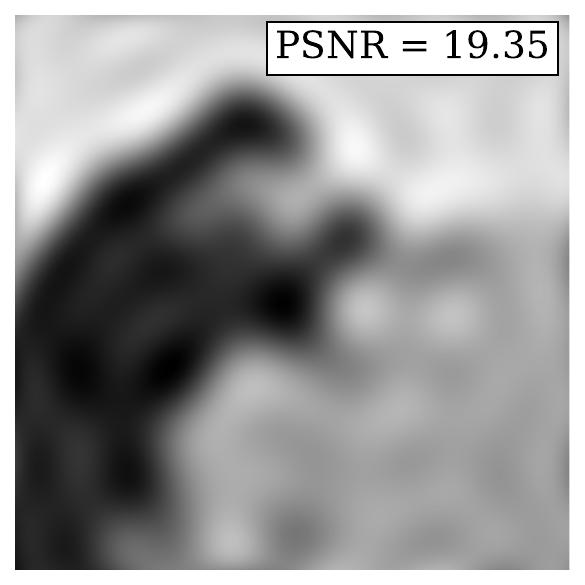}
        \caption{SIREN}
        \label{fig:sub3}
    \end{subfigure}
    \begin{subfigure}{0.31\textwidth}
        \centering
        \includegraphics[width=\linewidth]{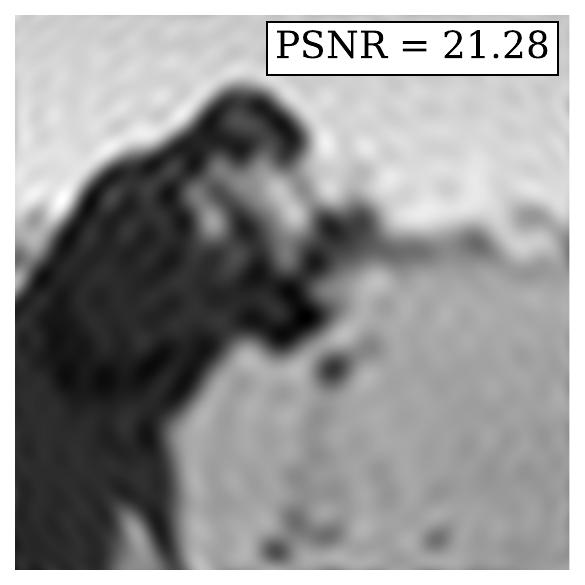}
        \caption{FINER}
        \label{fig:sub4}
    \end{subfigure}
    \hfill
    \begin{subfigure}{0.31\textwidth}
        \centering
        \includegraphics[width=\linewidth]{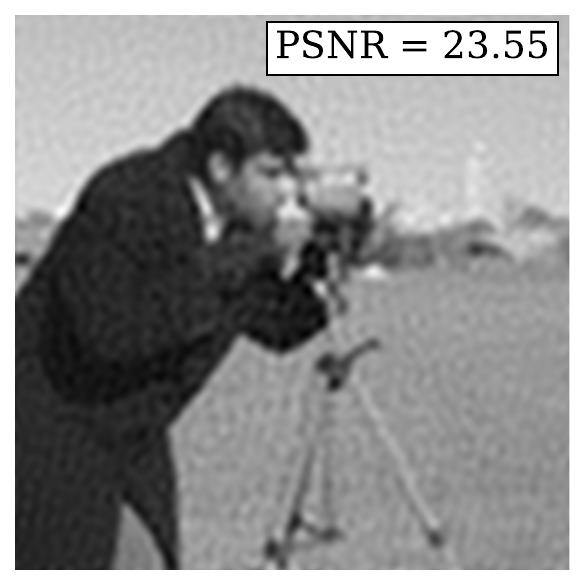}
        \caption{FM-SIREN}
        \label{fig:sub5}
    \end{subfigure}
    \hfill
    \begin{subfigure}{0.31\textwidth}
        \centering
        \includegraphics[width=\linewidth]{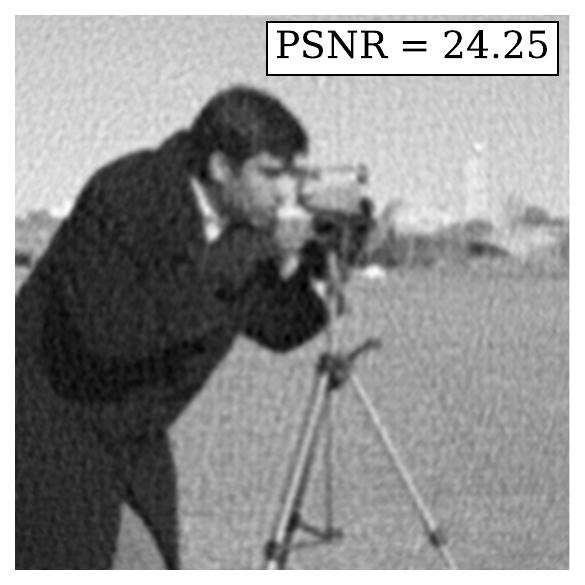}
        \caption{FM-FINER}
        \label{fig:sub6}
    \end{subfigure}
    \caption{Comparison of reconstruction quality using different methods for the cameraman image from the USC-SIPI Image Database \cite{usc-sipi-database}. Subfigures (c--f) show INR-based reconstructions with single-layer networks, designed to match the linear DST baseline in (b). All models were trained for 500 epochs using the Adam optimizer. FM-SIREN and FM-FINER yield the highest PSNR values, outperforming both SIREN and FINER as well as the classical DST.}
    \label{fig:all_subfigures}
    \vspace{-20pt}
\end{figure}

\par We identify the root cause of this bottleneck as hidden feature redundancy, a consequence of using a fixed frequency multiplier shared among all neurons within a layer. In conventional INR architectures such as SIREN and FINER, all neurons are assigned the same frequency scaling, meaning they effectively act as replicas of the same basis function, differing only in their learned weights and biases. This stands in stark contrast to the DST, where basis functions are harmonically spaced, mutually orthogonal frequencies by construction, ensuring each component captures a unique part of the signal with no overlap \cite{zhou2016interpretations}. In INRs, the optimizer must enforce this diversity implicitly through weight updates --- a task it cannot reliably accomplish given the non-convex nature of MLP optimization \cite{bottou2018optimization}. This motivates a principled redesign of the frequency assignment scheme in periodic-activation INRs, one that explicitly induces frequency diversity within each layer rather than leaving it to chance.


\par The solution lies in classical signal processing. Orthogonal transforms such as the DST assign each basis function a distinct, harmonically spaced frequency based on the Nyquist sampling theorem, ensuring maximum frequency diversity within a fixed parameter budget \cite{ahmed2012orthogonal, por2019nyquist}. We propose to incorporate this principle into INR design by replacing the fixed, shared frequency multiplier with a Nyquist-informed frequency vector that assigns each neuron a unique, per-neuron frequency multiplier spanning the signal's full spectral range. This explicitly induces frequency diversity among neurons within a single layer, without requiring additional parameters, increased network depth, hyperparameter tuning, or computational overhead --- directly addressing the root cause.


\par Building on the principle of Nyquist-informed orthogonality, we introduce two new architectures: \textbf{Frequency Multiplier-SIREN (FM-SIREN)} and \textbf{Frequency Multiplier-FINER (FM-FINER)}, which apply Nyquist-informed, per-neuron frequency assignment to SIREN and FINER activations, respectively. In this sense, FM-SIREN and FM-FINER can be interpreted as stacking DST-like layers to form a deep MLP, where each layer performs an orthogonal frequency assignment informed by the Nyquist theorem. As shown in figures \ref{fig:all_subfigures_jpg} and \ref{fig:correlation maps}, FM-SIREN and FM-FINER outperform their baselines and the classical DST in reconstruction quality, while reducing hidden feature redundancy by 49.92\% and 50.43\%, respectively, as measured by the Frobenius norm of the hidden embedding covariance matrix \cite{lancaster1984covariance, bottcher2008frobenius}. By maximizing the representational capacity of each layer, this frequency assignment delivers high-fidelity reconstructions with compact 2--3 layer networks, without relying on increased depth to compensate for per-layer redundancy, translating into consistent performance gains across diverse INR tasks.


\par Our contributions are as follows: (1) We draw a formal analogy between classical linear signal reconstruction and MLPs, connecting orthogonal frequency assignment from the DST to the design of periodic-activation INRs, identifying hidden feature redundancy as a fundamental bottleneck. (2) We propose a principled, Nyquist-informed scheme for designing neuron-specific frequency multipliers that increases frequency diversity without increasing network width, depth, or computational overhead, and eliminates the need for hyperparameter tuning. (3) We introduce FM-SIREN and FM-FINER, two new INR architectures that implement this scheme, reducing hidden feature redundancy by approximately 50\%. (4) We present comprehensive experiments across 1D audio, 2D image, 3D shape, and video fitting, demonstrating consistent improvements of up to 9.44 dB in PSNR and 55.15\% reduction in Chamfer Distance over state-of-the-art architectures such as SIREN, FINER, and other competing methods.
\section{Related Work}
INR techniques differ primarily in their choice of activation function or the higher-level framework used to configure them. We organize the most relevant methods into two groups.

\subsection{Activation-based INRs}
SIREN \cite{sitzmann2019siren} introduced sinusoidal activations with a fixed frequency multiplier $\beta$, enabling smooth continuous signal fitting with stable training dynamics. However, all neurons in a layer share the same frequency $\beta$, limiting frequency diversity and reducing representational capacity on spectrally rich signals. FINER \cite{liu2024finer} extends SIREN with a variable-frequency activation $\sigma(x) = \sin(\beta(|x|+1)x)$, broadening the representable frequency spectrum, yet the shared-$\beta$ design leaves hidden feature redundancy and its sensitivity to bias initialization unresolved, and adds training complexity. Gauss \cite{ramasinghe2022beyond} replaces periodic activations with a Gaussian $\sigma(x) = e^{-(sx)^2}$, motivated by links between Lipschitz smoothness \cite{beliakov2007smoothing} and hidden representation rank. However, its localized, non-periodic nature restricts its spectral coverage, making it less effective on signals with rich high-frequency content. WIRE \cite{saragadam2023wire} employs Gabor wavelet activations \cite{lee1996image} to achieve joint space-frequency localization, improving robustness to noise and high-frequency detail modeling. However, its complex-valued formulation doubles the parameter count and computational cost, while still applying a uniform $\beta$ across all neurons. SPDER \cite{shah2024spder} introduces a semiperiodic damping activation $\sigma(x) = \sin(\beta x) \cdot \delta(x)$, adding computational overhead while retaining the shared-$\beta$ design, and requires disproportionately deep networks to achieve strong performance (e.g., 5 layers for images and 12 for video).

\subsection{Input and Architecture-level INRs}
Positional Encoding (PE) \cite{tancik2020fourier} addresses the spectral bias of ReLU MLPs by lifting input coordinates into a higher-dimensional sinusoidal embedding $\gamma(\mathbf{x}) = [\sin(2^j\pi \mathbf{x}),\, \cos(2^j\pi \mathbf{x})]_{j=0}^{L-1}$, improving high-frequency signal fitting. However, it provides no mechanism for inducing frequency diversity within hidden layers, is sensitive to the choice of $L$, and increases the total parameter count by expanding input dimensionality. FreSh \cite{kania2024fresh} automatically selects a global frequency scale $\beta$ by minimizing the Wasserstein distance \cite{villani2009wasserstein} between the model's initial output spectrum and the target signal's spectrum, eliminating costly grid searches. However, it uniformly applies the chosen scale across all neurons, does not address hidden feature redundancy, and its reliance on 2D spectral analysis restricts it to INRs with 2D outputs. TUNER \cite{novello2025tuning} derives a spectral initialization and weight bounding scheme from an amplitude-phase expansion of sinusoidal MLPs, achieving fast convergence by freezing input frequencies after initialization. However, all neurons still share the same $\beta$, meaning per-neuron frequency diversity is never established, its analysis is restricted to image representation, and frozen frequencies limit adaptability to signals outside the initialization distribution. MIRE \cite{jayasundara2025mire} uses a dictionary-matching procedure to select the best activation function per layer, improving layer-wise expressivity. However, it does not address frequency diversity \textit{within} a single layer, and its sequential search incurs significant computational overhead before training begins. Neural Experts \cite{ben2024neural} decomposes the representation into a mixture of subnetworks \cite{masoudnia2014mixture}, and Fourier Reparameterization \cite{shi2024improved} recasts MLP weights in a fixed Fourier basis to reduce spectral bias, both improving upon standard INRs through architectural restructuring rather than activation design, and neither addressing per-neuron frequency diversity.

None of the above methods explicitly guarantees frequency diversity or orthogonality among neurons within the same layer, a property that classical transforms such as the DST achieve by construction. In contrast, our work targets the activation function itself, introducing a principled Nyquist-informed frequency assignment that induces diversity directly within each layer without requiring architectural search, additional components, or weight reparameterization. In this sense, FM-SIREN and FM-FINER can be interpreted as stacking DST-like layers to form a deep MLP, where each layer performs an orthogonal frequency assignment informed by the Nyquist theorem. This design reduces hidden feature redundancy by nearly 50\% and consistently improves signal reconstruction across diverse INR tasks.
\section{Problem Formulation}
In this section, we formulate the signal construction problem in INR and MLPs by providing the necessary background on signal reconstruction using periodic basis functions, the Nyquist sampling theorem for proper sampling, and by elaborating on how to extend these concepts to signal reconstruction with MLPs.

\subsection{Linear Signal Reconstruction}
According to \cite{stearns1990digital}, let a desired function $f(x)$ be approximated as closely as possible by another function $f^*(c,x)$, where $c$ denotes a set of adjustable parameters used to minimize the error between the two functions. The approximation $f^*(c,x)$ is expressed as a linear combination of a set of $M$ periodic bases with different frequencies $[\phi_0(x), \phi_1(x), \dots, \phi_{M-1}(x)]$, with the corresponding amplitude coefficients $[c_0, c_1, \dots, c_{M-1}]$. Formally, $f^*(c,x)$ can be written as:

\begin{equation}
    f^*(c,x)=\sum_{m=0}^{M-1} c_m\phi_{m}(x)
    \label{eq:reconstruction}
\end{equation}

\noindent The optimal approximation $f^*(x)$ is obtained by minimizing the least squares error between $f^*(c,x)$ and $f(x)$. A central property in this formulation is the orthogonality of the basis functions, which ensures that each coefficient can be computed independently of the others. In particular, orthogonality arises when the bases $[\phi_0(x), \phi_1(x), \dots, \phi_{M-1}(x)]$ correspond to harmonically spaced frequencies over the sampling interval, as in the DST construction, and the optimal coefficients $c_m$ are easily computed via normalized cross correlation as:

\begin{equation}
    c_m = \frac{\sum_{n=0}^{N-1} f(x_n)\,\phi_{m}(x_n)}{\sum_{n=0}^{N-1} \phi_{m}^2(x_n)}
    \label{eq: coefficients}
\end{equation}

The orthogonality of the basis functions is what makes classical transforms efficient and non-redundant (i.e., proportional representational capacity with the number of basis functions), a property that conventional MLPs fail to guarantee due to their over-reliance on parameter optimization despite the non-convexity of the loss function.


\subsection{Nyquist Frequency for Periodic Basis Functions}
The Nyquist frequency specifies the maximum frequency that can be captured without aliasing when sampling a continuous-time signal of periodic basis functions \cite{shannon2006communication}. According to the Nyquist sampling theorem, the sampling frequency must be at least twice the highest frequency component present in the signal to ensure a perfect reconstruction. Sampling below this threshold causes frequency components to overlap in the spectral domain, resulting in distortion and loss of information \cite{oppenheim1997signals}, \cite{lyons2011understanding}. Formally, the Nyquist frequency is defined as:
\begin{equation}
    f_{\text{Nyq}} = \frac{f_\text{s}}{2}, \quad  f_\text{s}\geq f_{\text{max}}
\end{equation}
\noindent where $f_\text{s}$ is the sampling frequency and $f_{\text{max}}$ denotes the maximum frequency present in the signal. For example, if a signal is sampled at 2 kHz, the Nyquist frequency is 1 kHz; any signal component exceeding this limit is aliased. For higher-dimensional signals, such as 2D images and 3D volumes, each dimension maintains its own Nyquist frequency determined by the spatial sampling interval \cite{gonzalez2008digital}. In a digital image where the sampling rate is 1 pixel, the Nyquist frequency is $0.5$ cycles/pixel. Consequently, for an image of size $512 \times 512$ pixels, the highest resolvable spatial frequency is $256$ cycles per image width and height \cite{jain1989fundamentals}.

\par The Nyquist frequency is then used to determine the frequency limits of the periodic basis function: sines and cosines. Specifically, the Nyquist frequency determines the spectral span of a signal as follows:
\begin{equation}
    f_{\text{span}} = [0, f_{\text{Nyq}}]
    \label{eq: f_span}
\end{equation}

Then, the spectral span is uniformly sampled given the number of bases $M$, where each frequency corresponds to a distinct basis function that is orthogonal to the others. This makes it possible to compute the basis coefficient using equation \ref{eq: coefficients}. The expression of $M$ sinusoidal basis functions with frequency $\beta$ is as follows:
\vspace{-8pt}
\begin{equation}
    \phi_m(x) = \sin(\beta_m x), \qquad \beta_m = \frac{m f_{\text{Nyq}}}{M}, \quad m = 0,1,2,\dots, M
\end{equation}

\subsection{MLP Relevance to Linear Signal Reconstruction}
MLPs are composed of sequential linear layers (with non-linear activation), which can be interpreted as linear and hierarchical reconstruction embeddings \cite{goodfellow2016deep}. For simplicity, consider a single hidden linear layer where both the input $x$ and the output $y$ are one-dimensional. The output $y$ of this layer with respect to the input $x$ can be written as:
\begin{equation}
    y_n = \sum_{m=0}^{M-1} w_{m}^{\text{out}} \, 
    \sigma_{m}\!\left(w_{m}^{\text{in}} x_n\right)
    \label{eq:mlp_layer}
\end{equation}

\noindent where $w_{m}^{\text{in}}$, $w_{m}^{\text{out}}$, and $\sigma_{m}$ denote the input weights, the output weights, and the activation function, respectively. This representation 
resembles the linear reconstruction in Equation \ref{eq:reconstruction}. However, classical signal reconstruction assigns basis functions $\phi_m$ at harmonically spaced frequencies that are mutually orthogonal, avoiding redundancy. In contrast, conventional INRs assign the same activation function to all neurons in a layer, effectively creating frequency replicas that violate orthogonality.

\par
\begin{figure}[t]
    \centering
    \begin{subfigure}{0.225\textwidth}
        \centering
        \includegraphics[width=\linewidth]{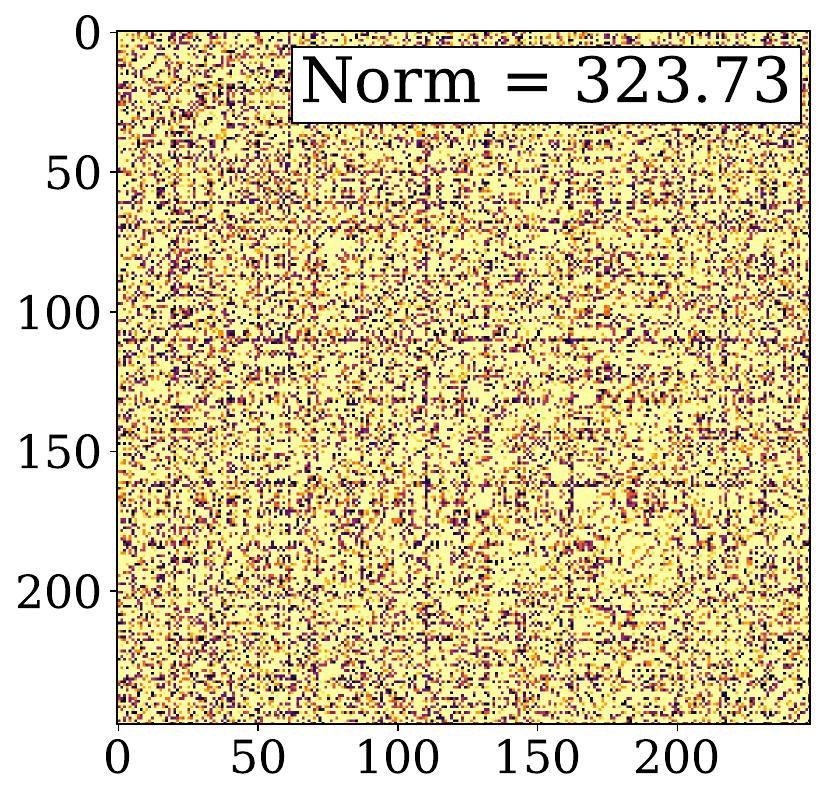}
        \caption{SIREN}
        \label{fig:maps1}
    \end{subfigure}
    \begin{subfigure}{0.225\textwidth}
        \centering
        \includegraphics[width=\linewidth]{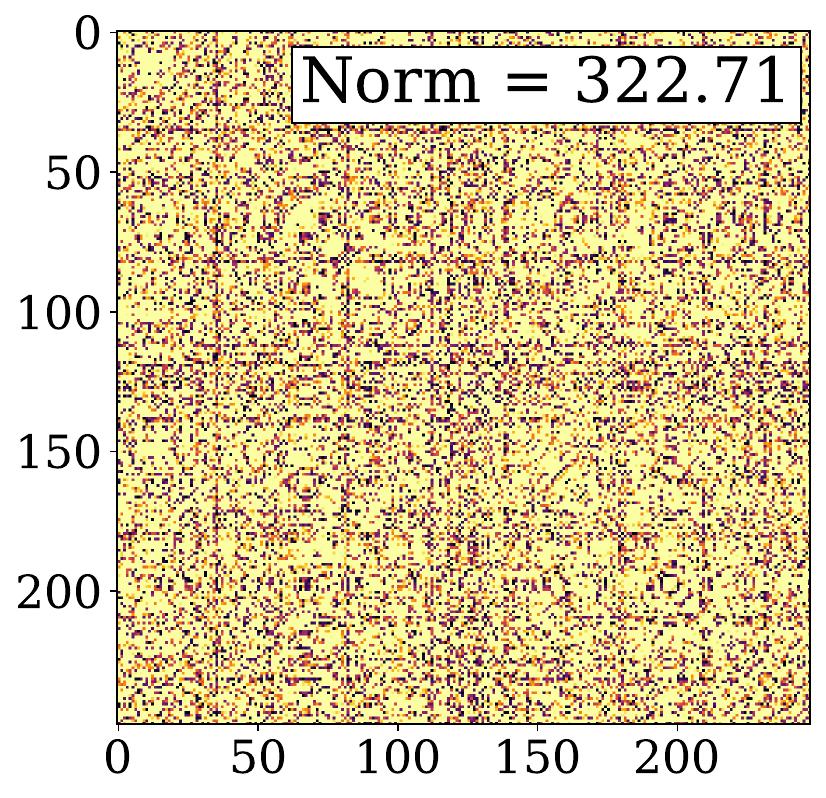}
        \caption{FINER}
        \label{fig:maps2}
    \end{subfigure}
    \begin{subfigure}{0.225\textwidth}
        \centering
        \includegraphics[width=\linewidth]{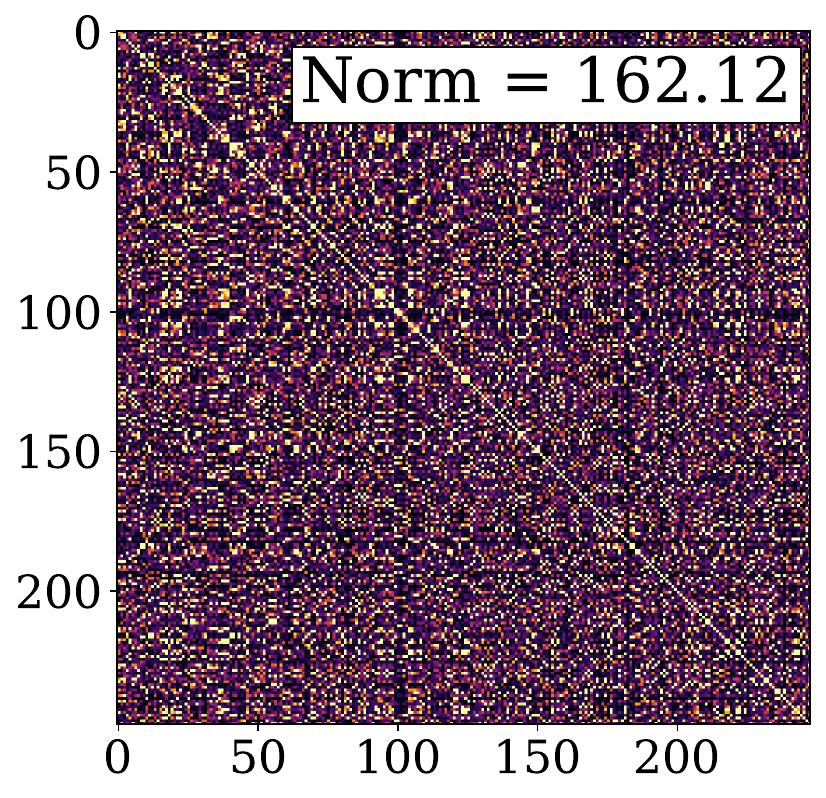}
        \caption{FM-SIREN}
        \label{fig:maps3}
    \end{subfigure}
    \begin{subfigure}{0.225\textwidth}
        \centering
        \includegraphics[width=\linewidth]{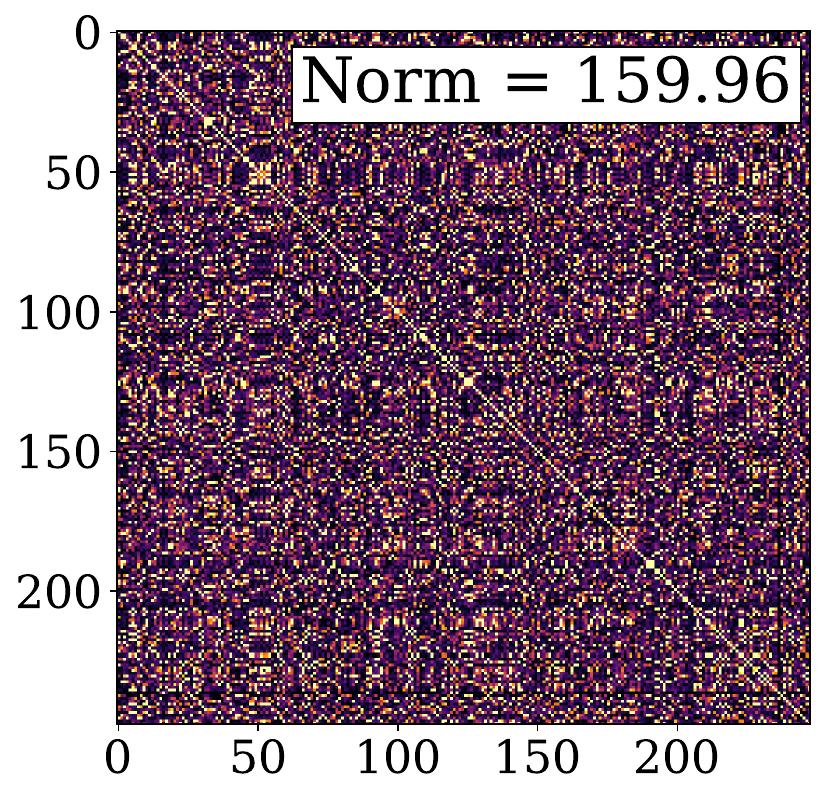}
        \caption{FM-FINER}
        \label{fig:maps4}
    \end{subfigure}%
    \raisebox{0.19\height}{%
        \includegraphics[height=0.207\textwidth]{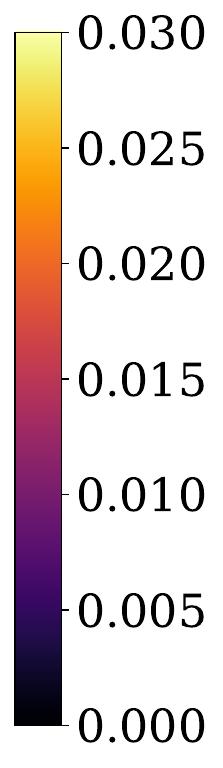}%
    }
    \caption{Covariance maps of hidden embeddings with the Frobenius norm shown in the top-right of each subfigure. $\text{x}$ and $\text{y}$ axes represent neuron index. Observe that the yellow color represents higher correlation while the black color denotes lower/zero correlation. FM-SIREN (c) and FM-FINER (d) yield substantially lower covariance norms than SIREN (a) and FINER (b), reflecting improved frequency diversity. All maps are derived from the networks used to reconstruct images in Figure \ref{fig:all_subfigures}.}
    \label{fig:correlation maps}
    
\end{figure}

 As a result, the optimizer must enforce orthogonality through weight updates, which is not guaranteed due to the strongly non-convex nature of MLPs \cite{bottou2018optimization}. This is evident in the covariance maps of hidden embeddings shown in Figures \ref{fig:maps1} and \ref{fig:maps2}, where SIREN and FINER exhibit substantial feature correlation. Hidden embeddings in models such as SIREN and FINER therefore remain correlated, reducing overall MLP capacity. We propose to address this by explicitly orthogonalizing the hidden embeddings through a novel frequency assignment, as described in the next section.

    

\section{Proposed Solution}
Building on the problem formulation and the analysis in Section 3, we propose a principled frequency multiplier scheme that explicitly induces frequency diversity among neurons within a single layer, as described in the following subsections.
\subsection{Nyquist-Informed Frequency Multiplier from Input Signal}
Given any input signal, the spectral bounds of its periodic basis functions can be explicitly determined from its size and sampling rate for each dimension. Specifically, given a spatial $N$-dimensional signal with size $D_1\times D_2\times \dots\times D_N$, its sampling rates for each dimension are simply $D_1, D_2, \dots, D_N$, respectively. Therefore, the Nyquist frequency of its $n$-th dimension becomes:
\begin{equation}
    f_{s,D_n}=D_n \implies f_{\text{Nyq},D_n}=\frac{D_n}{2}
\end{equation}
\noindent Hence, the vector of frequency multipliers for $M$ basis functions for the dimension $n$ (act as activations in a linear layer with $M$ neurons) is defined as follows:


\begin{equation}
    \boldsymbol{\beta}_{D_n} = \left[\frac{f_{\text{Nyq},D_n}}{M}, \frac{2f_{\text{Nyq},D_n}}{M}, \cdots, \frac{Mf_{\text{Nyq},D_n}}{M}\right] = \left[\frac{D_n}{2M}, \frac{2D_n}{2M}, \cdots, \frac{D_n}{2}\right] 
\label{eq: omega1}
\end{equation}

\noindent For temporal dimension (e.g., 1D audio and temporal dimension in video), the sampling is determined by the temporal dimension length $L$ (number of samples) over the total time period: $f_s=\frac{L}{T} \implies \boldsymbol{\beta} = \left[\frac{L}{2MT}, \frac{2L}{2MT}, \dots, \frac{L}{2T}\right]$.


\begin{figure}[t]
    \centering
    \makebox[\textwidth]{\includegraphics[width=1\textwidth]{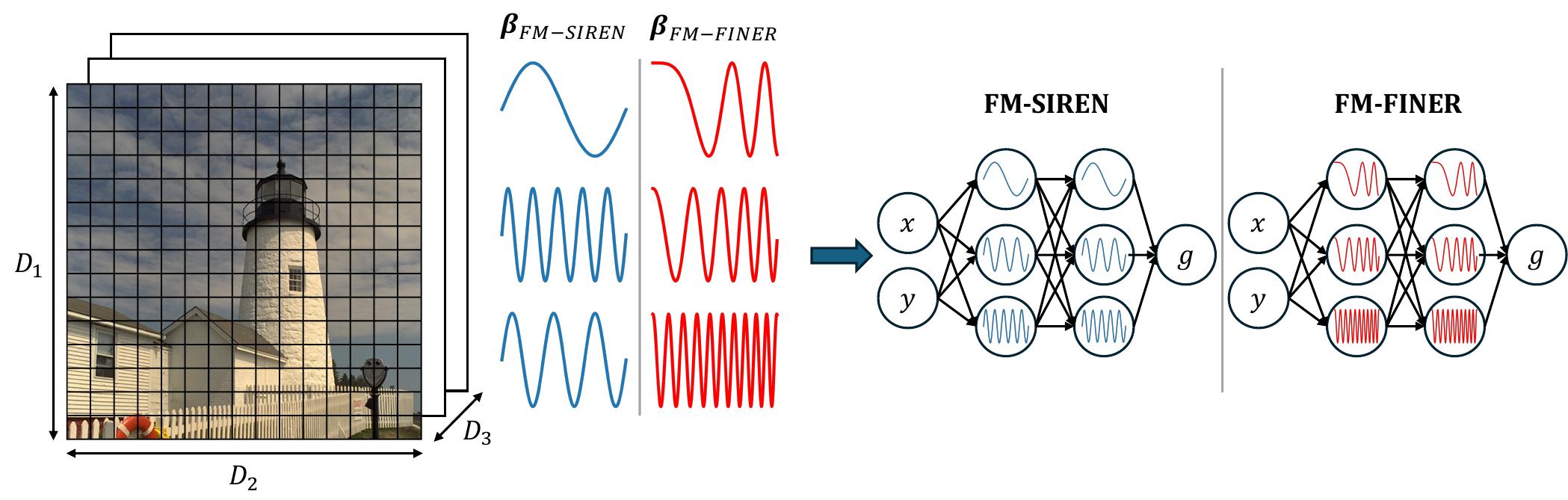}}
    \caption{FM-SIREN and FM-FINER frequency assignment framework for an input signal of dimensions $D_1 \times D_2 \times D_3$. The dimensions inform $\boldsymbol{\beta}_{FM-SIREN}$ and $\boldsymbol{\beta}_{FM-FINER}$ as described in equations \ref{eq: omega1} and \ref{eq: omega2}, yeilding the activation functions for the FM-SIREN and FM-FINER architectures.}
    \label{fig: process}
\end{figure}

\subsection{FM-SIREN \& FM-FINER}
As illustrated in Figure \ref{fig: process}, FM-SIREN and FM-FINER implement the Nyquist-informed frequency assignment across all layers, effectively stacking DST-like layers to form a deep MLP where each layer performs an orthogonal frequency assignment informed by the Nyquist theorem. FM-SIREN alters the frequency multipliers for all neurons in each layer following Equation \ref{eq: omega1} with the highest dimensional sampling rate $\max(D_n)$, introducing a higher degree of frequency diversity than SIREN. Using the maximum dimension's sampling rate ensures that the frequency vector spans the widest spectral range present in the signal, avoiding under-coverage in asymmetric signals.
\begin{equation}
    \boldsymbol{\beta}_{FM-SIREN} = \left[\frac{\text{max}(D_n)}{2M}, 
    \frac{2\text{max}(D_n)}{2M}, \dots, 
    \frac{\text{max}(D_n)}{2}\right]
    \label{eq: omega1}
\end{equation}
FM-FINER applies the same principle, but requires a modified frequency 
range to account for the variable-frequency nature of the FINER 
activation. Specifically, since INR inputs are normalized to 
$x \in [-1, 1]$, the FINER activation $\sin(\beta(|x|+1)x)$ has an 
instantaneous frequency of $\beta(|x|+1)$, which reaches a maximum of 
$2\beta$ at $|x|=1$. To prevent aliasing, the effective frequency must 
satisfy $2\beta_m \leq f_{\text{Nyquist}}$, which gives $\beta_m \leq 
\frac{f_{\text{Nyquist}}}{2}$. Therefore, the frequency range is scaled 
down by a factor of 2.
\begin{equation}
    \boldsymbol{\beta}_{FM-FINER} = \left[\frac{\text{max}(D_n)}{4M}, 
    \frac{2\text{max}(D_n)}{4M}, \dots, 
    \frac{\text{max}(D_n)}{4}\right]
    \label{eq: omega2}
    \vspace{-10pt}
\end{equation}

\begin{figure}[t]
    \centering
    \begin{subfigure}{0.2425\textwidth}
        \centering
        \includegraphics[width=\linewidth]{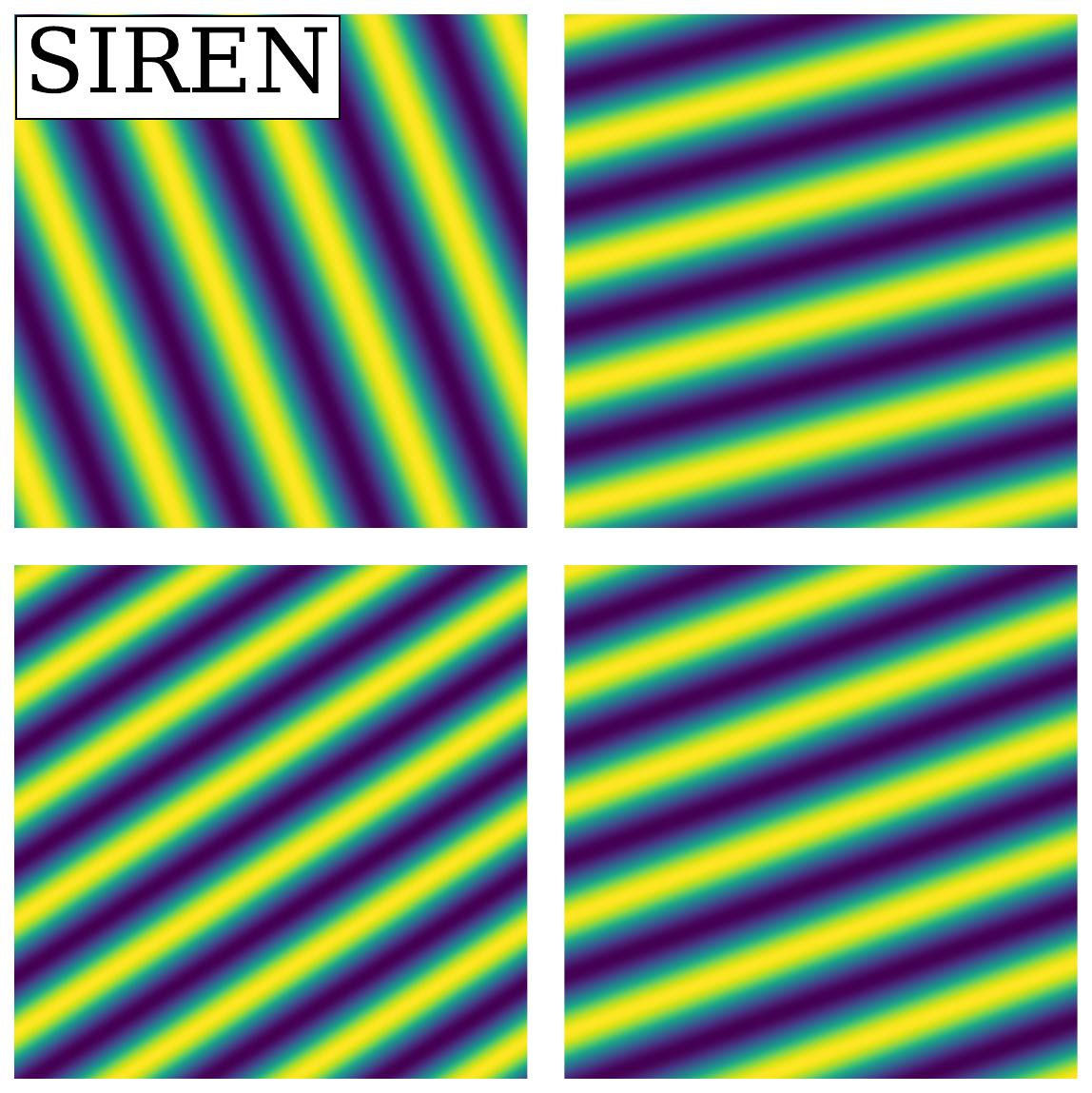}
        \label{fig:siren_hidden}
    \end{subfigure}
    \begin{subfigure}{0.2425\textwidth}
        \centering
        \includegraphics[width=\linewidth]{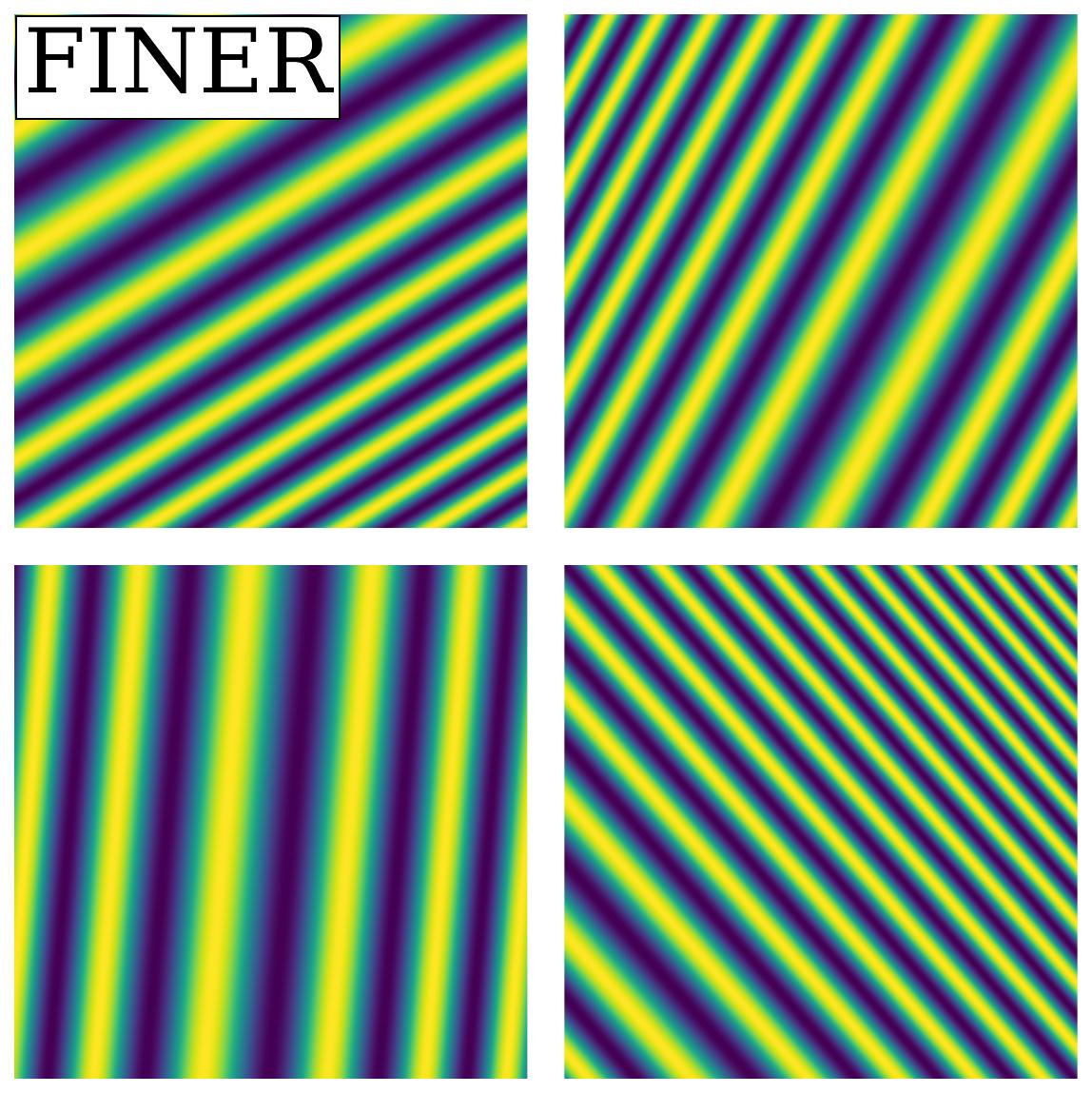}
        \label{fig:finer_hidden}
    \end{subfigure}
    \begin{subfigure}{0.2425\textwidth}
        \centering
        \includegraphics[width=\linewidth]{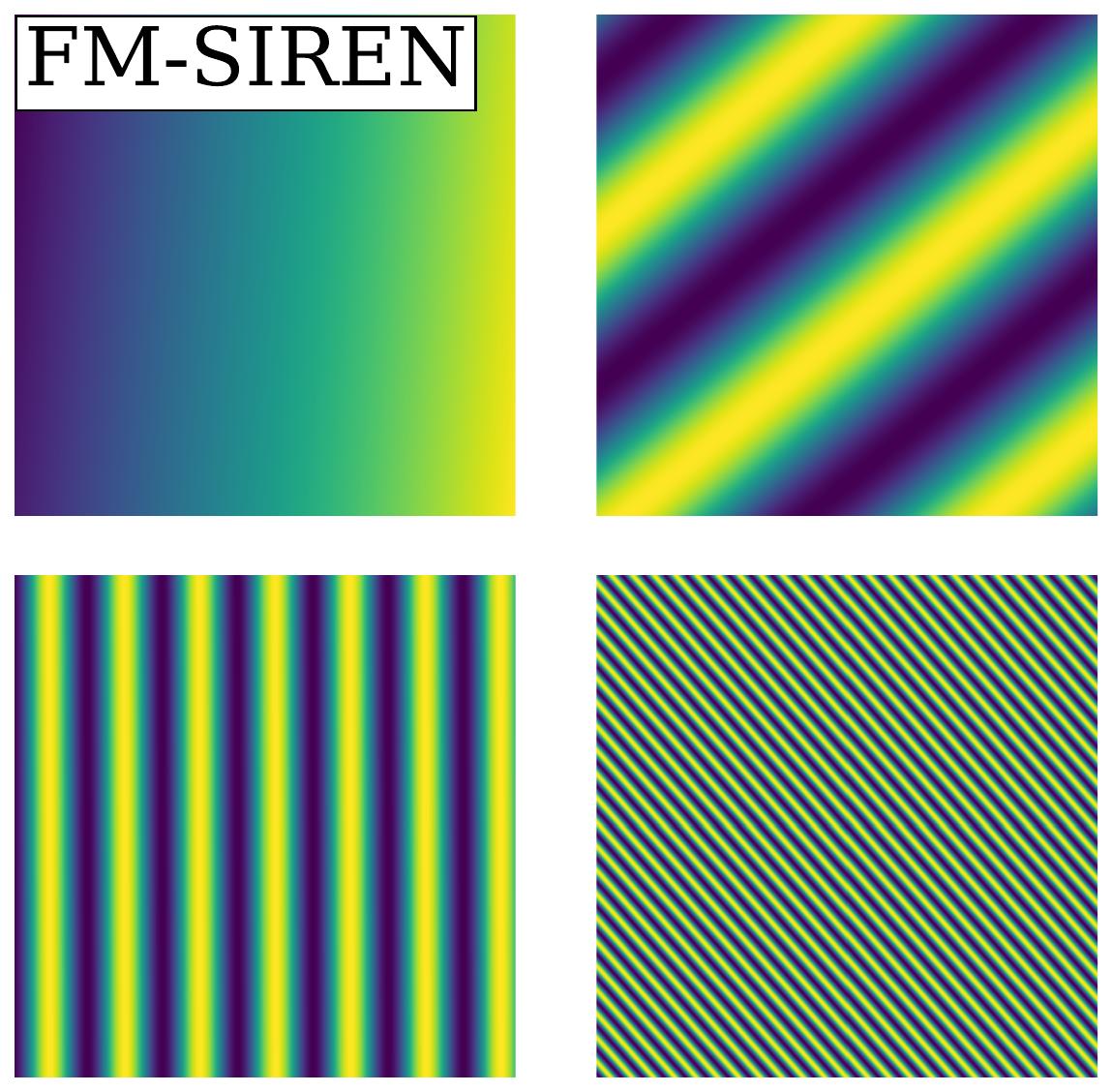}
        \label{fig:esiren_hidden}
    \end{subfigure}
    \begin{subfigure}{0.2425\textwidth}
        \centering
        \includegraphics[width=\linewidth]{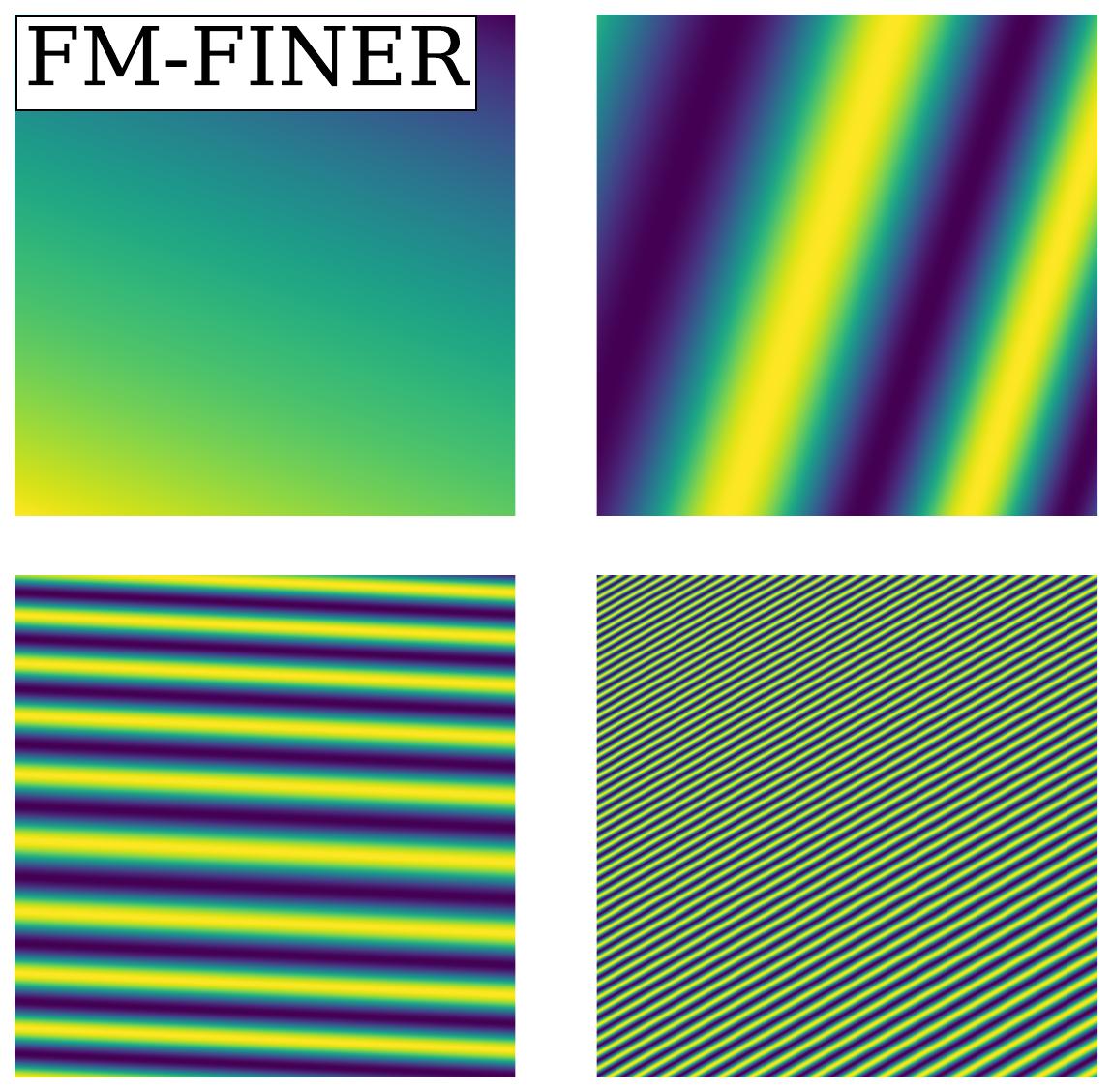}
        \label{fig:efiner_hidden}
    \end{subfigure}
    
    \begin{subfigure}{0.2425\textwidth}
        \centering
        \includegraphics[width=\linewidth]{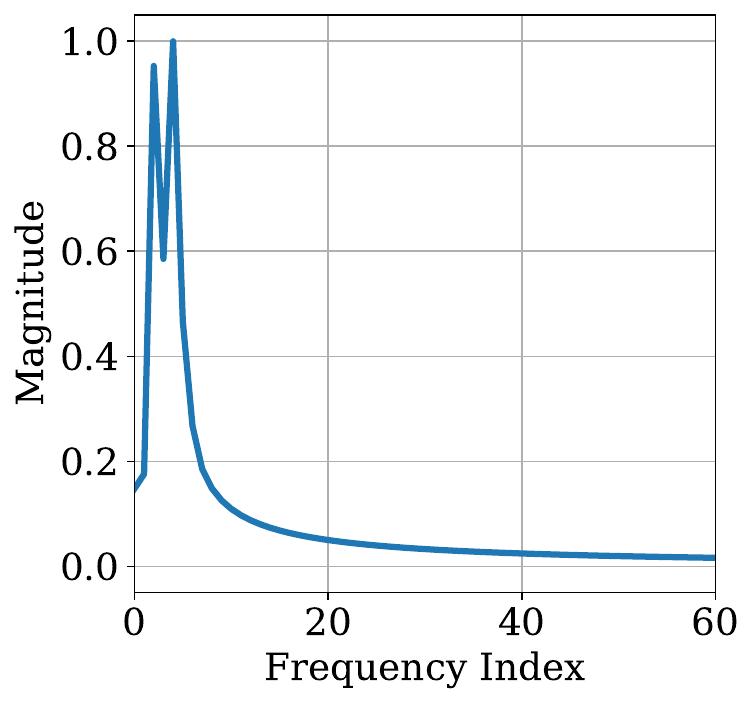}
        \caption{SIREN}
        \label{fig:siren_spectrum}
    \end{subfigure}
    \begin{subfigure}{0.2425\textwidth}
        \centering
        \includegraphics[width=\linewidth]{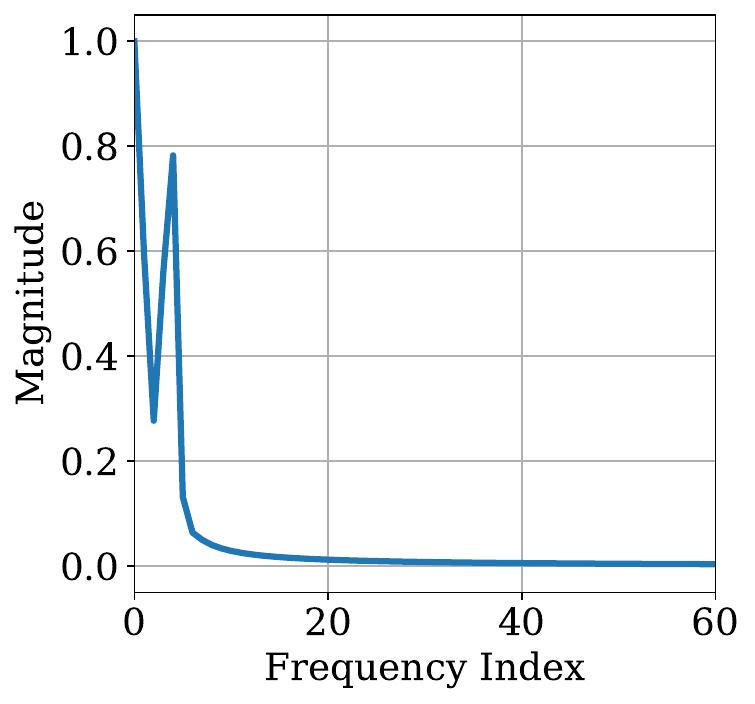}
        \caption{FINER}
        \label{fig:finer_spectrum}
    \end{subfigure}
    \begin{subfigure}{0.2425\textwidth}
        \centering
        \includegraphics[width=\linewidth]{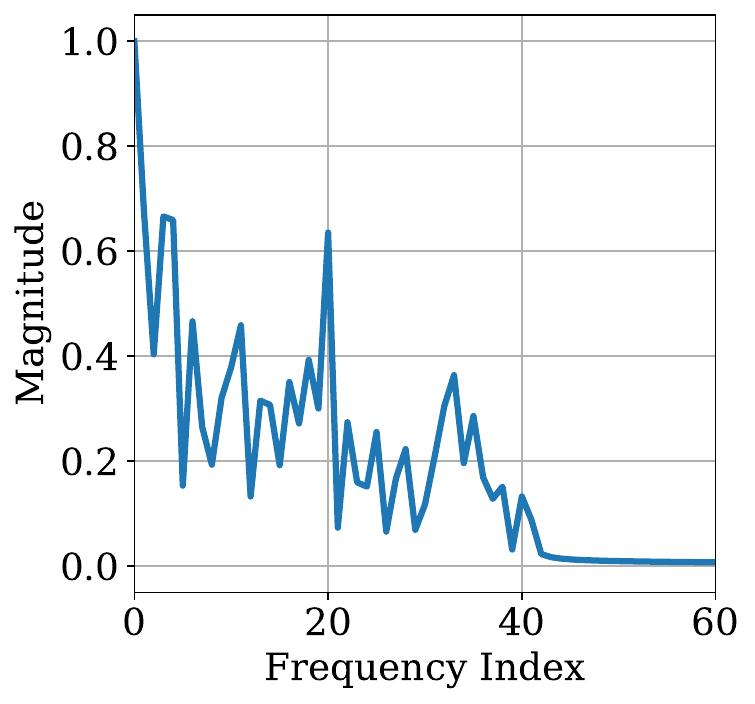}
        \caption{FM-SIREN}
        \label{fig:esiren_spectrum}
    \end{subfigure}
    \begin{subfigure}{0.2425\textwidth}
        \centering
        \includegraphics[width=\linewidth]{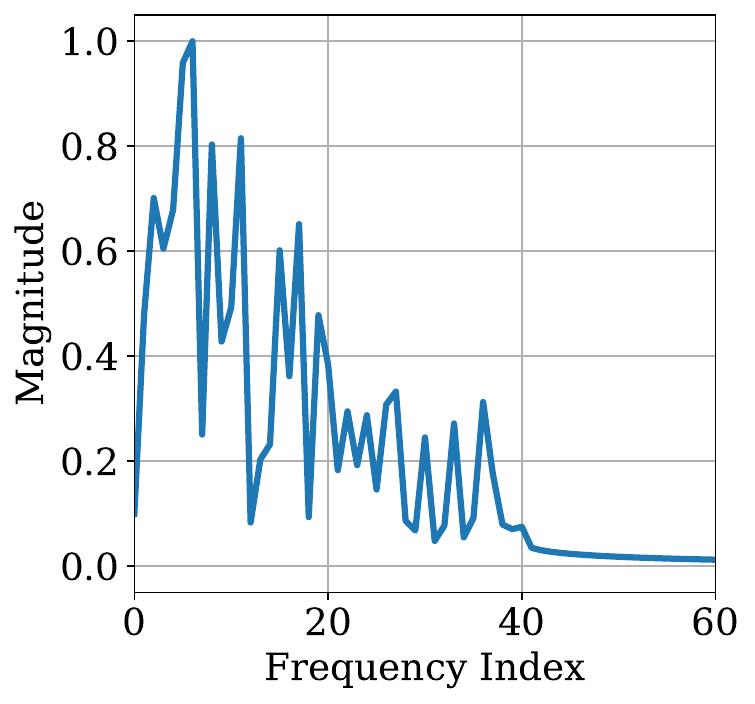}
        \caption{FM-FINER}
        \label{fig:efiner_spectrum}
    \end{subfigure}
    \caption{Hidden embeddings from uniformly-sampled four neurons (top) and their magnitude spectra (bottom). SIREN and FINER produce repetitive patterns concentrated at low frequencies, while FM-SIREN and FM-FINER exhibit more diverse embeddings spanning a broader spectral range. The magnitude spectra of the proposed FM-SIREN and FM-FINER clearly show that the proposed approach significantly increases the frequency range of the representation, thus increasing the expressiveness of the INR.}
    \vspace{-18pt}
    \label{fig:embeddings}
\end{figure}


\subsection{Feature Redundancy}
To quantify feature redundancy, we generate a distinct feature vector of length 10,000 from each neuron, compute the covariance matrix of those feature vectors, and measure the Frobenius norm of the resulting covariance matrix. Figures \ref{fig:maps1} and \ref{fig:maps2} show the resulting covariance maps for SIREN and FINER, revealing substantial feature redundancy within a single layer. Ideally, the off-diagonal elements of the covariance matrix should approach zero, indicating orthogonality among neuron embeddings, but this is rarely achieved in practice due to the non-convex nature of MLP optimization. In contrast, FM-SIREN and FM-FINER substantially reduce feature redundancy, as shown in Figures \ref{fig:maps3} and \ref{fig:maps4}, achieving a 49.92\% and 50.43\% improvement in the Frobenius norm of the covariance matrix compared to their baselines.

This is further illustrated in Figure \ref{fig:embeddings}, which shows the hidden embeddings and their magnitude spectra for each method. While SIREN and FINER produce repetitive embedding patterns concentrated at similar frequencies, FM-SIREN and FM-FINER exhibit visibly more diverse embeddings spanning a broader spectral range, consistent with the reduced covariance norms in Figure \ref{fig:correlation maps}.

\begin{figure}[t]
    \centering
    \begin{subfigure}{0.24\textwidth}
        \includegraphics[width=\linewidth]{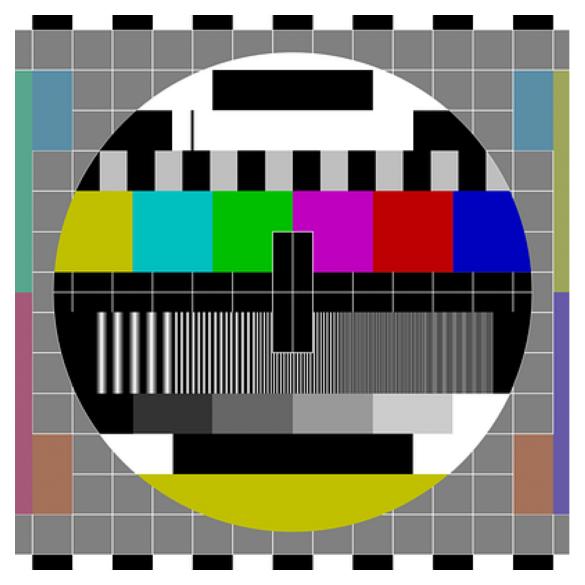}
        \caption{Ground Truth}
    \end{subfigure}
    \begin{subfigure}{0.24\textwidth}
        \includegraphics[width=\linewidth]{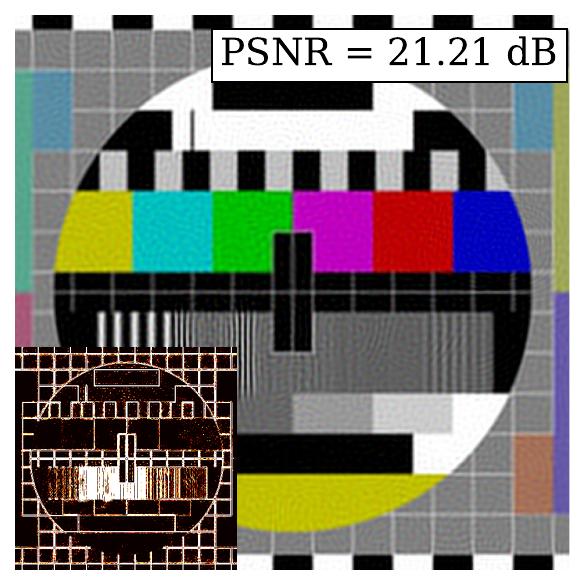}
        \caption{SIREN}
    \end{subfigure}
    \begin{subfigure}{0.24\textwidth}
        \includegraphics[width=\linewidth]{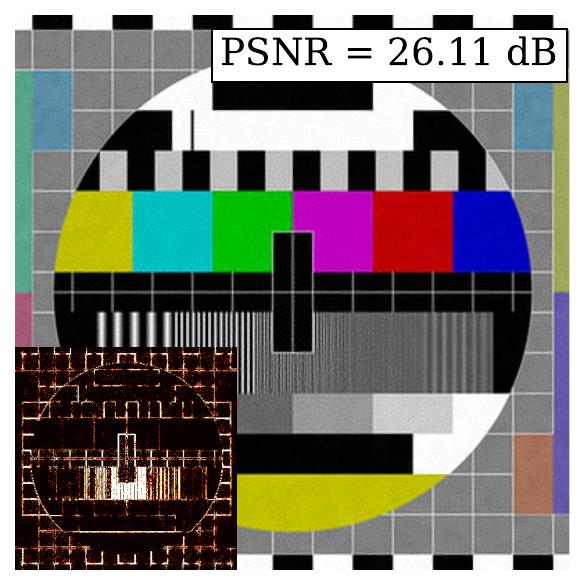}
        \caption{FINER}
    \end{subfigure}
    \begin{subfigure}{0.24\textwidth}
        \includegraphics[width=\linewidth]{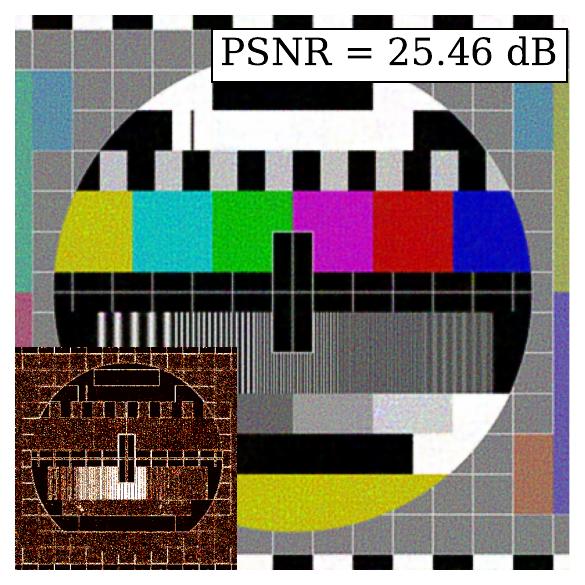}
        \caption{PE}
    \end{subfigure}

    \begin{subfigure}{0.24\textwidth}
        \includegraphics[width=\linewidth]{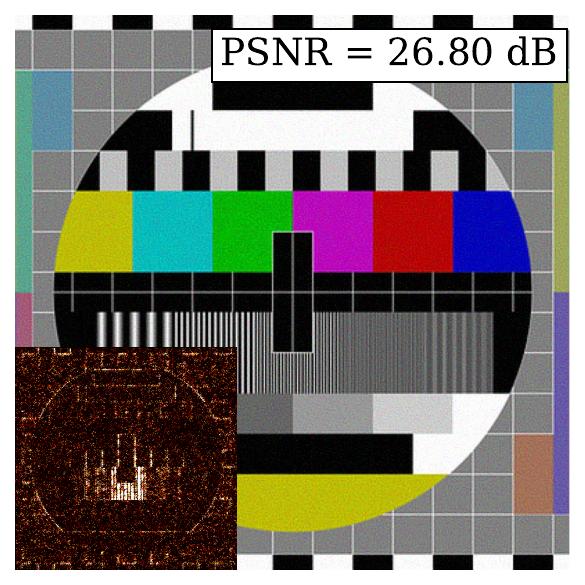}
        \caption{TUNER}
    \end{subfigure}
    \begin{subfigure}{0.24\textwidth}
        \includegraphics[width=\linewidth]{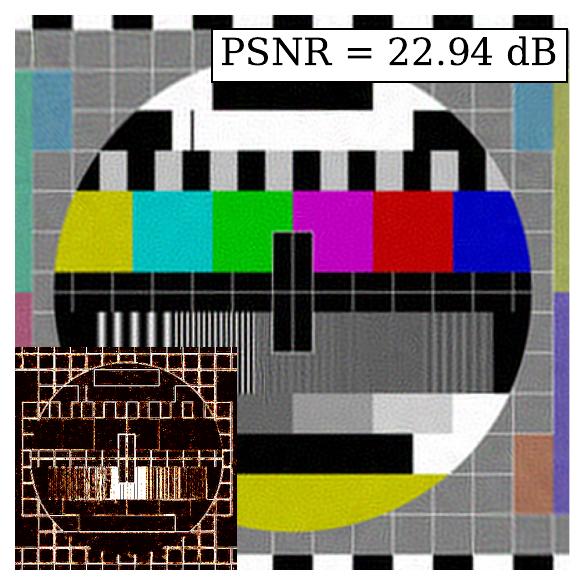}
        \caption{WIRE}
    \end{subfigure}
    \begin{subfigure}{0.24\textwidth}
        \includegraphics[width=\linewidth]{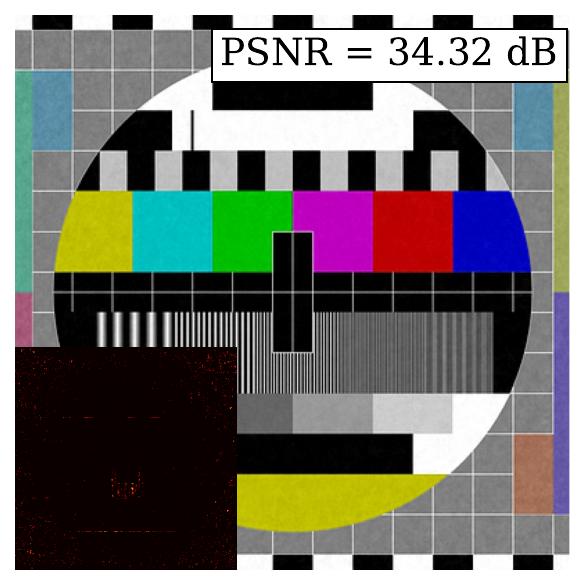}
        \caption{FM-SIREN}
    \end{subfigure}
    \begin{subfigure}{0.24\textwidth}
        \includegraphics[width=\linewidth]{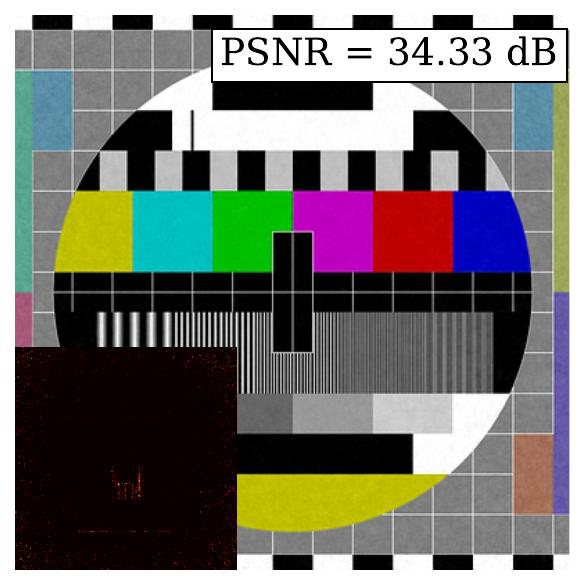}
        \caption{FM-FINER}
    \end{subfigure}
    \caption{Qualitative image reconstruction results of the Philips Circle Pattern \cite{philips_circle_wiki} using two-layer networks. The PSNR (dB) of each reconstruction is reported in the top-right corner and the error map in the bottom-left corner of its subfigure. FM-SIREN and FM-FINER achieve the highest PSNR values and produce visibly sharper reconstructions compared to the baselines. While TUNER attains the best performance among the baseline methods, its reconstructions appear noisier than ours, as highlighted in the zoomed-in slices. Observe that FM-FINER and FM-SIREN improve the PSNR of Positional Encoding scheme which is the closest competitor by about 7.5 dB.}
    \label{fig:image_test}

\end{figure}

\section{Experimental Results}
\vspace{-10pt}

\par In this section, we evaluate the proposed FM-SIREN and FM-FINER on four representation tasks: image fitting, image inpainting, 3D shape, and video. To further demonstrate generalization across signal modalities, we present 1D audio fitting and neural radiance fields (NeRF) results in the supplementary material. For fair comparisons, we adopt the hyperparameters recommended in the original works of the respective baselines, and use the same layer architecture across all models, with the exception of WIRE, which employs a complex-valued formulation requiring a modified layer configuration \cite{saragadam2023wire}. Training configurations are task-dependent and detailed in the supplementary material, along with full training infrastructure, hyperparameter lists, and an ablation study on frequency multiplier design. All experiments are conducted on an NVIDIA H200 GPU using PyTorch \cite{paszke2019pytorch}.
\subsection{Image Fitting}
\vspace{-5pt}

\begin{figure}[t]
    \centering
    \begin{subfigure}{0.15\textwidth}
        \centering
        \includegraphics[width=\linewidth]{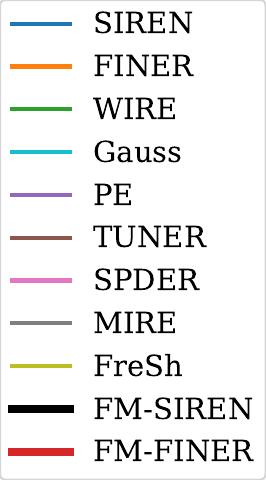}
    \end{subfigure}
    \hfill
    \begin{subfigure}{0.4\textwidth}
        \centering
        \includegraphics[width=\linewidth]{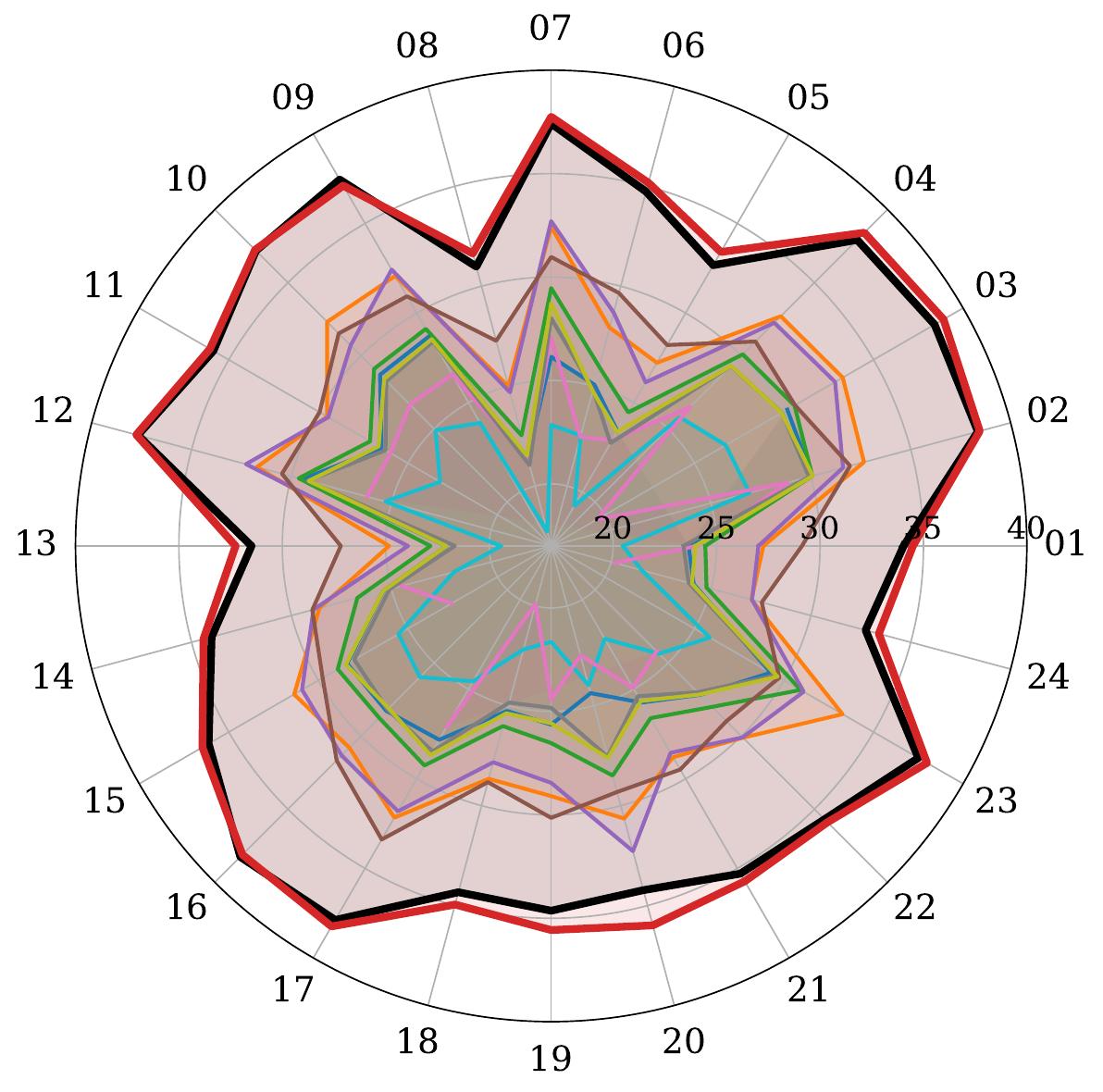}
        \caption{Kodak}
        \label{fig:subplot1}
    \end{subfigure}
    \hfill
    \begin{subfigure}{0.4\textwidth}
        \centering
        \includegraphics[width=\linewidth]{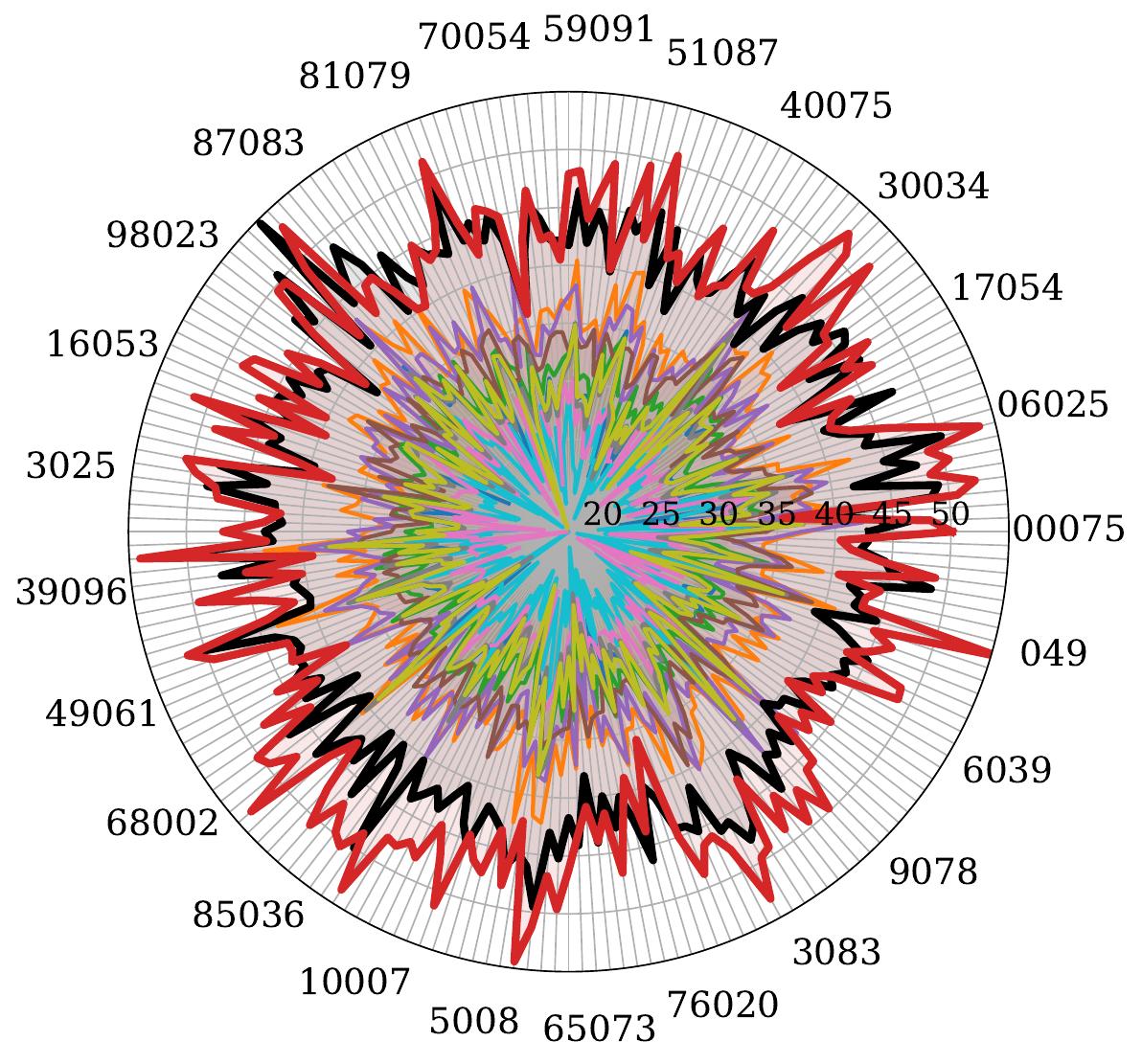}
        \caption{BSDS500}
        \label{fig:subplot2}
    \end{subfigure}
    \vspace{-8pt}
    \caption{Image fitting results on Kodak and BSDS500 datasets. Each axis corresponds to a specific image from the dataset and the radial distance indicates PSNR (dB). FM-SIREN and FM-FINER consistently outperform all baselines by a significant margin across every image in both datasets (see Table~\ref{tab:image_results}). The radar plots reflect uniform performance gains across all images without increased variance, suggesting robustness to image content diversity.}
    \label{fig:radar_plots}
    \vspace{-20pt}
\end{figure}

\begin{wrapfigure}{r}{0.5\textwidth}
    \vspace{-35pt}
    \centering
    \includegraphics[width=0.48\textwidth]{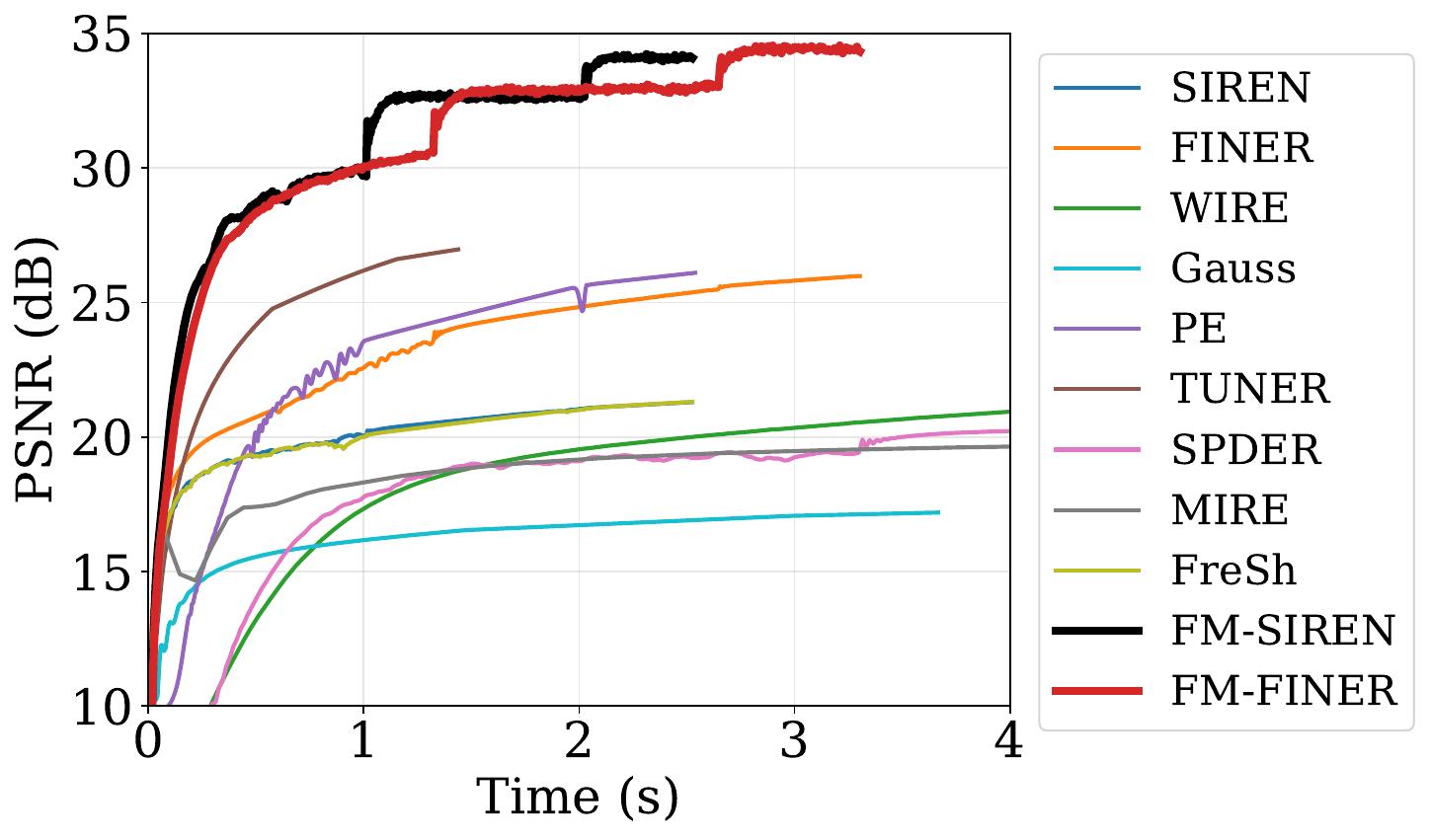}
    \vspace{-7pt}
    \caption{PSNR vs. training time over 500 training epochs on PCP. FM-SIREN and FM-FINER consistently achieve higher PSNR than all baselines throughout training, demonstrating both accuracy and efficiency gains. Notably, TUNER converges fastest among the baselines, yet FM-SIREN and FM-FINER already exceed its final PSNR at TUNER's convergence.}
    \vspace{-20pt}
    \label{fig:psnr_vs_time}
\end{wrapfigure}

\par We evaluate all models on the Kodak Lossless True Color Image Suite \cite{kodak} and the BSDS500 dataset \cite{bsds500}, which correspond to informed Nyquist frequencies of $256$ and $160.5$ cycles/image, respectively. Performance is assessed using two standard metrics: the peak signal-to-noise ratio (PSNR) \cite{korhonen2012peak} and the structural similarity index (SSIM) \cite{wang2004image}. All models are implemented with two layers of $512$ neurons each. As shown in Table~\ref{tab:image_results}, FM-SIREN and FM-FINER consistently achieve the highest average PSNR and SSIM, outperforming all baseline methods by up to $5.64$~dB and $6.20$~dB in PSNR on Kodak, and up to $7.36$~dB and $9.44$~dB on BSDS500, respectively. Figure~\ref{fig:radar_plots} shows the PSNR results per image in both datasets. In addition, we evaluate the models on the Philips Circle Pattern (PCP) \cite{philips_circle_wiki}, a benchmark with both sharp and smooth geometric structures ideal for testing reconstruction fidelity with a Nyquist frequency of $200$ cycles/image. Figure~\ref{fig:image_test} shows qualitative results, where FM-SIREN and FM-FINER exceed the strongest baseline by $7.52$~dB and $7.53$~dB in terms of PSNR, respectively. Figure~\ref{fig:psnr_vs_time} shows the PSNR curve for each model, where FM-SIREN and FM-FINER achieve substantial performance gains from the early stages of training, surpassing the fastest converging model at convergence. These results highlight the ability of our models to deliver high-fidelity image reconstructions using only two-layer networks. Additional reconstruction results and error distributions are provided in the supplementary material.

\begin{table}[H]
    \centering
    \caption{\label{tab:image_results} Average reconstruction performance on the Kodak \cite{kodak}, BSDS500 \cite{bsds500}, and PCP datasets in PSNR and SSIM. \colorbox{Goldenrod}{Best} and \colorbox{Dandelion}{Second Best} results are highlighted.}
    \vspace{-10pt}
    \resizebox{0.75\textwidth}{!}{%
    \begin{tabular}{@{}l c cccccc@{}}
        \toprule
        \multirow{2}{*}{\textbf{Model}} & \multirow{2}{*}{\textbf{Train Time} (min)} & \multicolumn{3}{c}{\textbf{PSNR} (dB) $\uparrow$} & \multicolumn{3}{c}{\textbf{SSIM} $\uparrow$} \\
        \cmidrule(lr){3-5} \cmidrule(lr){6-8}
        & & Kodak & BSDS500 & PCP & Kodak & BSDS500 & PCP \\
        \midrule
        FINER   & 9.24  & 29.98 & 36.05 & 26.11 & 0.860 & 0.959 & 0.777 \\
        SIREN   & 7.14  & 25.98 & 29.01 & 21.21 & 0.696 & 0.826 & 0.644 \\
        Gauss   & 10.13 & 23.28 & 24.50 & 17.19 & 0.553 & 0.624 & 0.492 \\
        WIRE    & 33.43 & 27.52 & 31.30 & 22.94 & 0.747 & 0.874 & 0.642 \\
        PE      & 6.76  & 29.71 & 34.68 & 25.46 & 0.837 & 0.937 & 0.648 \\
        MIRE    & 40.09 & 26.48 & 29.57 & 20.41 & 0.719 & 0.858 & 0.594 \\
        TUNER   & 4.15  & 29.86 & 33.17 & 26.80 & 0.835 & 0.938 & 0.610 \\
        SPDER   & 22.50 & 22.90 & 25.29 & 20.54 & 0.564 & 0.684 & 0.462 \\
        FreSh   & 10.22 & 26.76 & 29.79 & 21.20 & 0.733 & 0.855 & 0.650 \\
        \midrule
        \textbf{FM-SIREN} & 7.14  & \colorbox{Dandelion}{35.62} & \colorbox{Dandelion}{43.41} & \colorbox{Dandelion}{34.32} & \colorbox{Goldenrod}{0.933} & \colorbox{Dandelion}{0.990} & \colorbox{Goldenrod}{0.8771} \\
        \textbf{FM-FINER} & 9.24  & \colorbox{Goldenrod}{36.06} & \colorbox{Goldenrod}{45.49} & \colorbox{Goldenrod}{34.33} & \colorbox{Dandelion}{0.938} & \colorbox{Goldenrod}{0.993} & \colorbox{Dandelion}{0.8768} \\
        \midrule
        \multirow{2}{*}{\textbf{Improvement}} & \textbf{FM-SIREN} & 5.64dB & 7.36dB & 7.52dB & 8.51\% & 3.24\% & 12.75\% \\
        & \textbf{FM-FINER} & 6.08dB & 9.44dB & 7.53dB & 9.03\% & 3.51\% & 12.71\% \\
        \bottomrule
    \end{tabular}%
    }
    \vspace{-35pt}
\end{table}

\begin{figure}[t]
    \centering
    \begin{subfigure}{0.18\textwidth}
        \includegraphics[width=\textwidth]{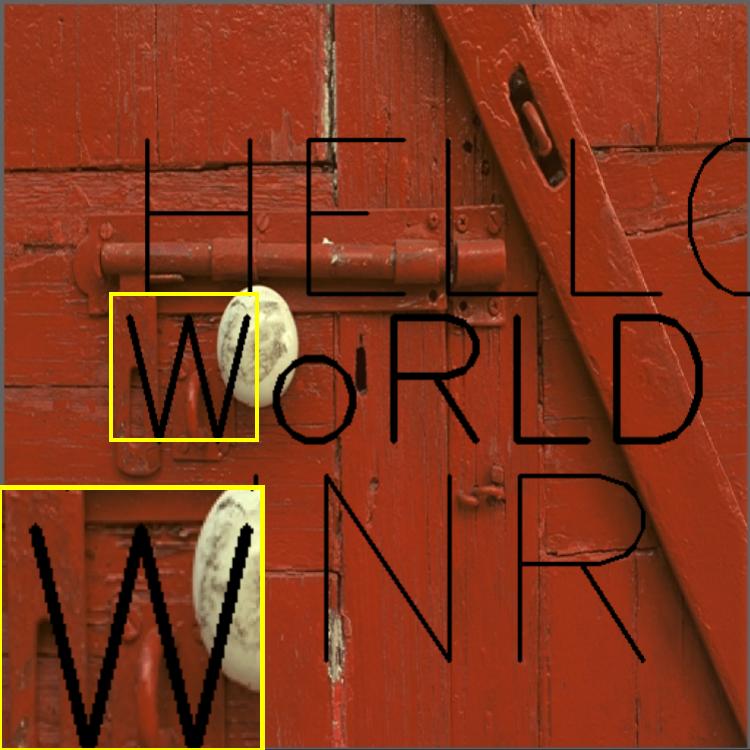}
        \caption{Masked image}
    \end{subfigure}
    \hfill
    \begin{subfigure}{0.18\textwidth}
        \includegraphics[width=\textwidth]{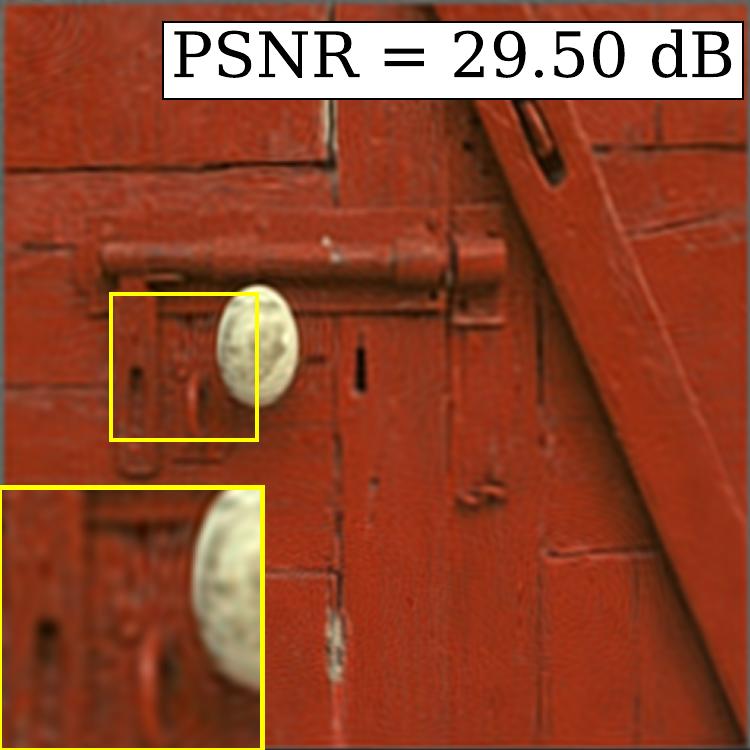}
        \caption{SIREN}
    \end{subfigure}
    \hfill
    \begin{subfigure}{0.18\textwidth}
        \includegraphics[width=\textwidth]{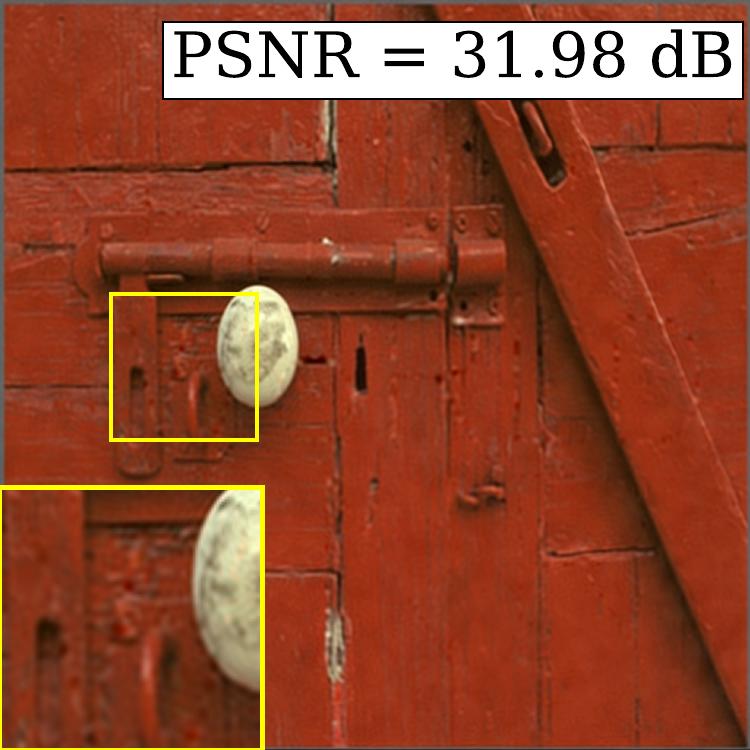}
        \caption{FINER}
    \end{subfigure}
    \hfill
    \begin{subfigure}{0.18\textwidth}
        \includegraphics[width=\textwidth]{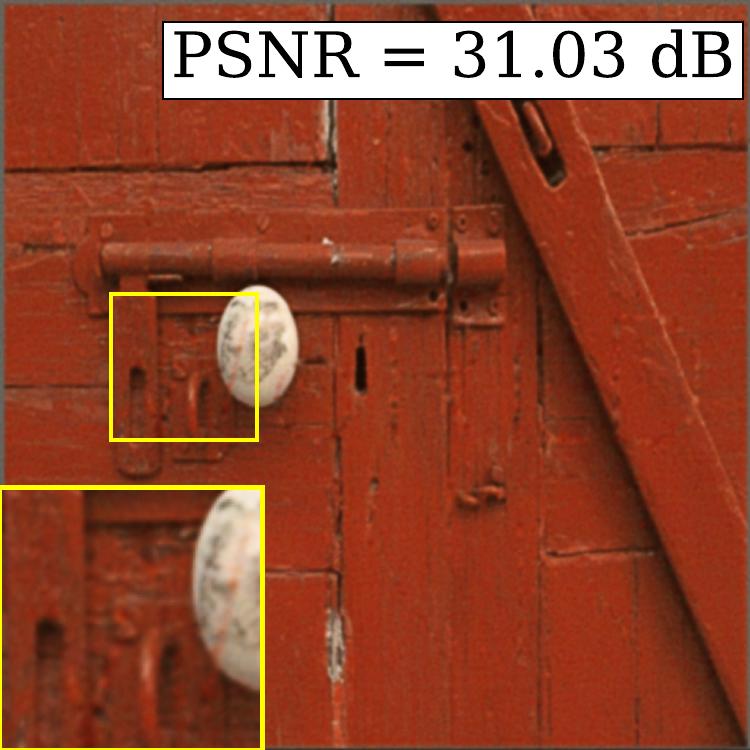}
        \caption{PE}
    \end{subfigure}
    \hfill
    \begin{subfigure}{0.18\textwidth}
        \includegraphics[width=\textwidth]{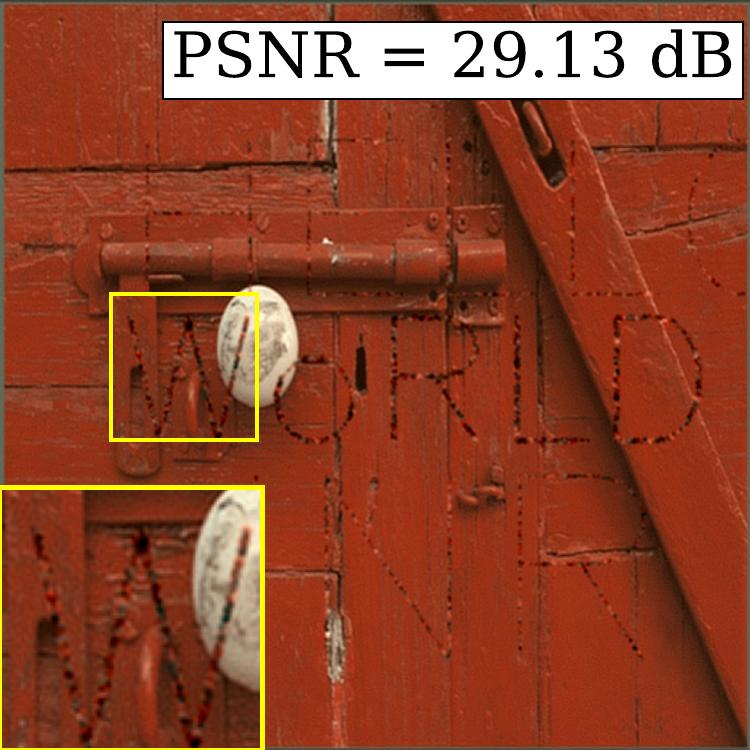}
        \caption{TUNER}
    \end{subfigure}

    \vspace{0.5em}

    \begin{subfigure}{0.18\textwidth}
        \includegraphics[width=\textwidth]{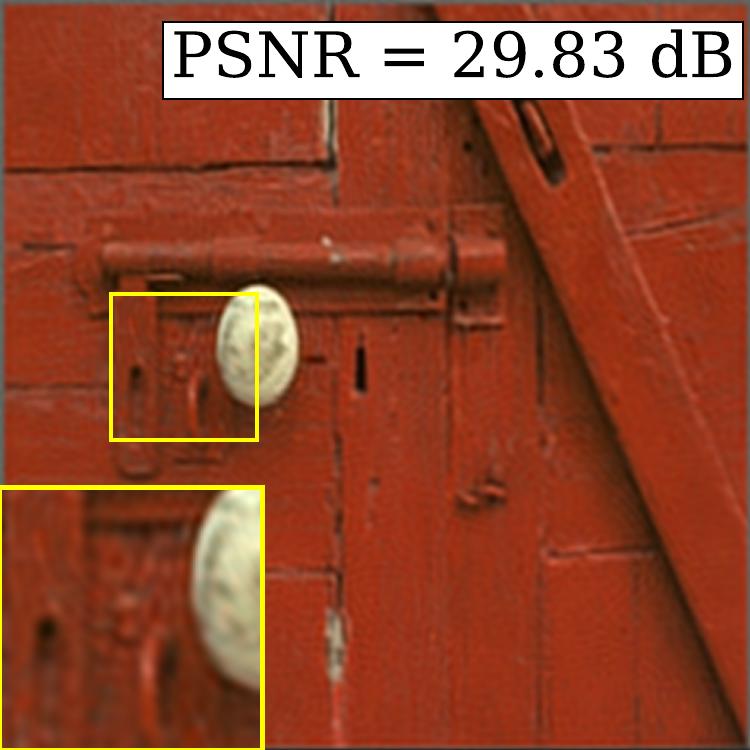}
        \caption{MIRE}
    \end{subfigure}
    \hfill
    \begin{subfigure}{0.18\textwidth}
        \includegraphics[width=\textwidth]{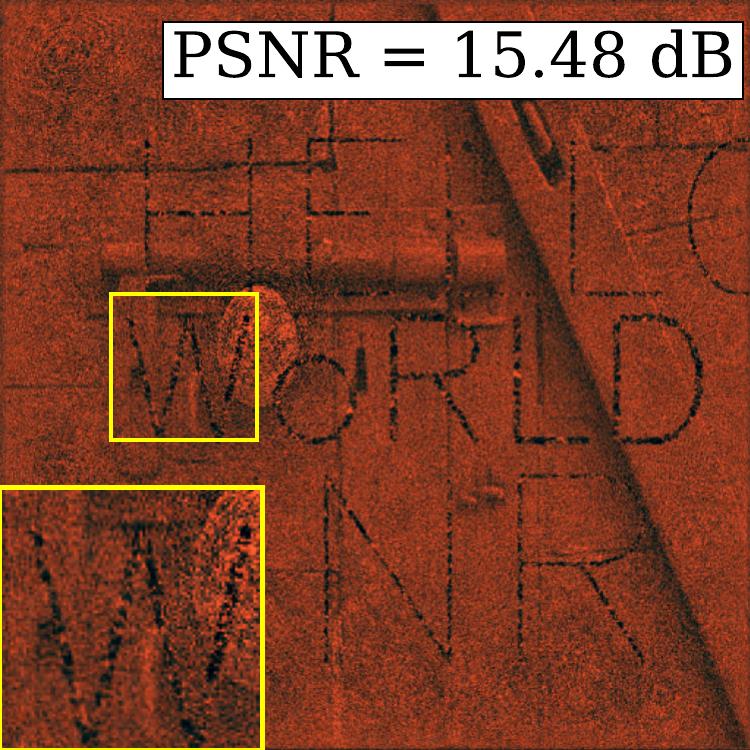}
        \caption{SPDER}
    \end{subfigure}
    \hfill
    \begin{subfigure}{0.18\textwidth}
        \includegraphics[width=\textwidth]{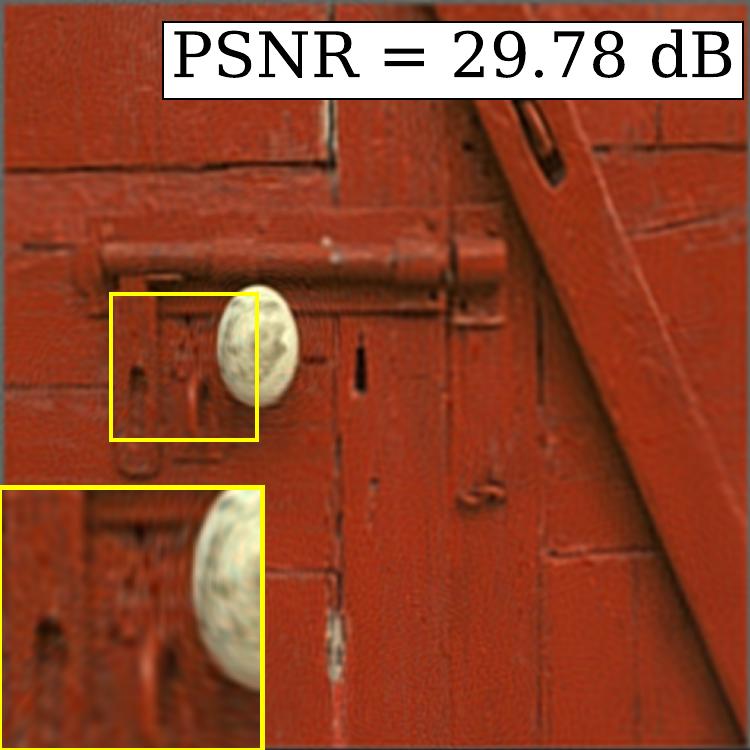}
        \caption{FreSh}
    \end{subfigure}
    \hfill
    \begin{subfigure}{0.18\textwidth}
        \includegraphics[width=\textwidth]{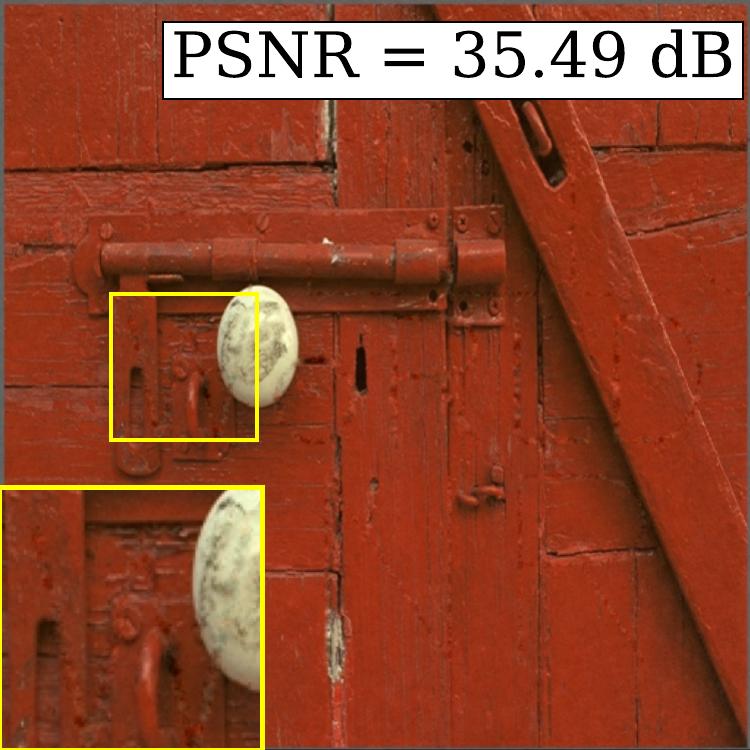}
        \caption{FM-SIREN}
    \end{subfigure}
    \hfill
    \begin{subfigure}{0.18\textwidth}
        \includegraphics[width=\textwidth]{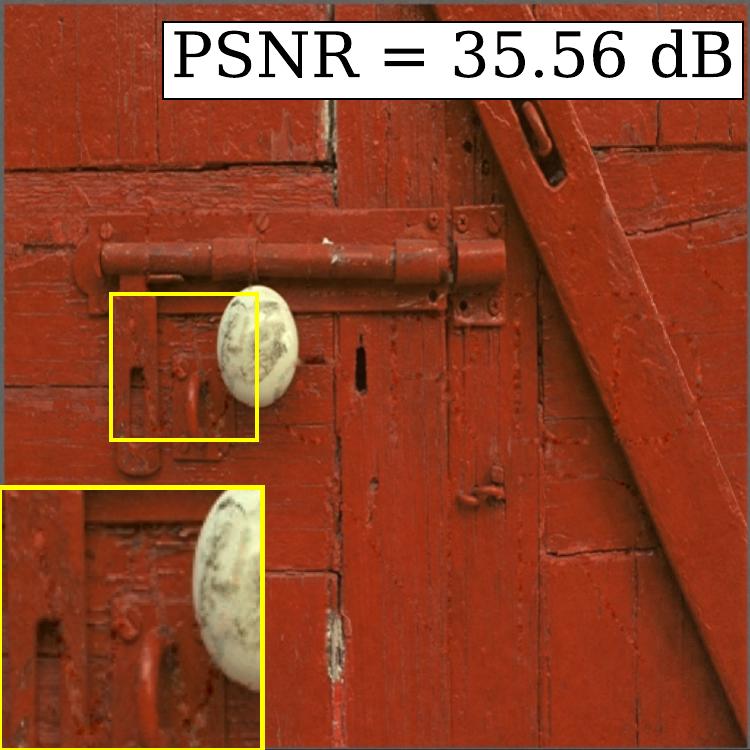}
        \caption{FM-FINER}
    \end{subfigure}
    \vspace{-8pt}
    \caption{Qualitative image inpainting results on the Kodak dataset using two-layer networks. A text mask is applied to the input image following \cite{jayasundara2025mire}. The PSNR (dB) of each reconstruction is reported in the top-right corner of its subfigure. FM-SIREN and FM-FINER recover finer details compared to all baseline methods.}
    \vspace{-15pt}
    \label{fig:inpainting}
\end{figure}

\subsection{Image Inpainting}
\vspace{-5pt}
\par Image inpainting evaluates the ability of INR models to reconstruct missing regions from partial observations, serving as a measure of generalization beyond the training coordinates. Following the protocol of~\cite{jayasundara2025mire}, we apply a text mask to the input image, removing the corrupted pixels from the input coordinates, and train all models to recover the complete image from the remaining visible pixels. We evaluate on the Kodak dataset~\cite{kodak} using PSNR and SSIM as evaluation metrics. As shown in Figure~\ref{fig:inpainting}, FM-SIREN and FM-FINER consistently outperform all baseline methods, demonstrating that Nyquist-informed frequency diversity improves generalization to unobserved image regions while retaining excellent reconstruction. Additional results are provided in the supplementary material.
\subsection{3D Shape Fitting}
\vspace{-5pt}
\par
\begin{figure}[t]
    \centering
    \begin{subfigure}{0.23\textwidth}
        \includegraphics[width=\linewidth, trim=0 0 0 45, clip]{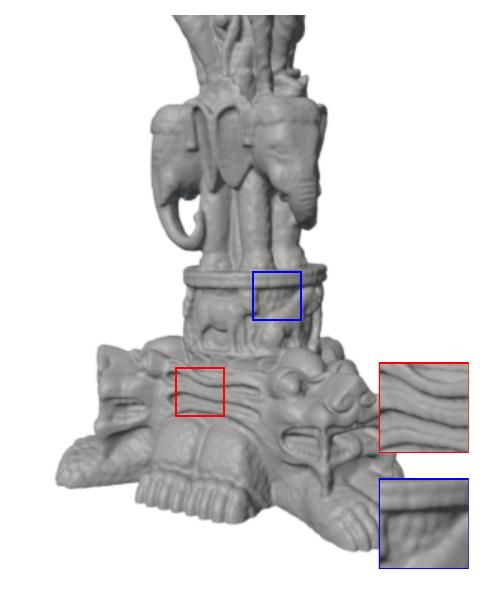}
        \caption{Ground Truth}
    \end{subfigure}
    \begin{subfigure}{0.23\textwidth}
        \includegraphics[width=\linewidth, trim=0 0 0 45, clip]{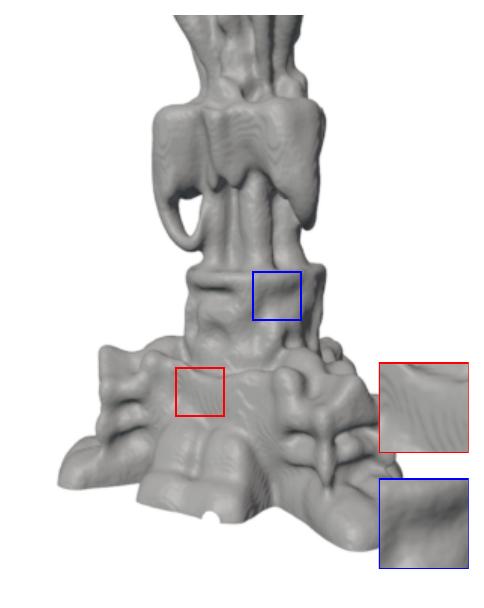}
        \caption{SIREN}
    \end{subfigure}
    \begin{subfigure}{0.23\textwidth}
        \includegraphics[width=\linewidth, trim=0 0 0 45, clip]{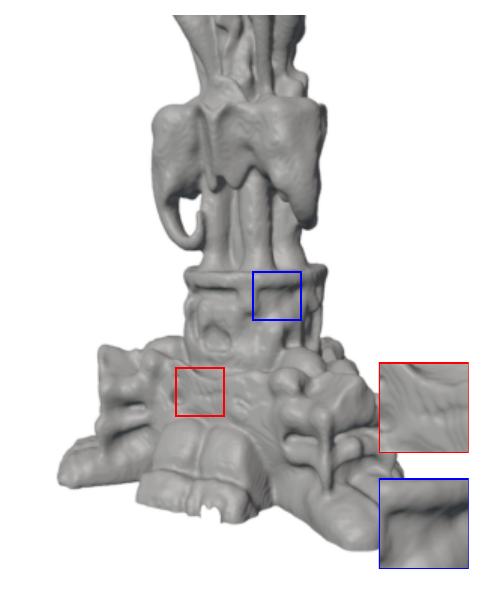}
        \caption{FINER}
    \end{subfigure}
    \begin{subfigure}{0.23\textwidth}
        \includegraphics[width=\linewidth, trim=0 0 0 45, clip]{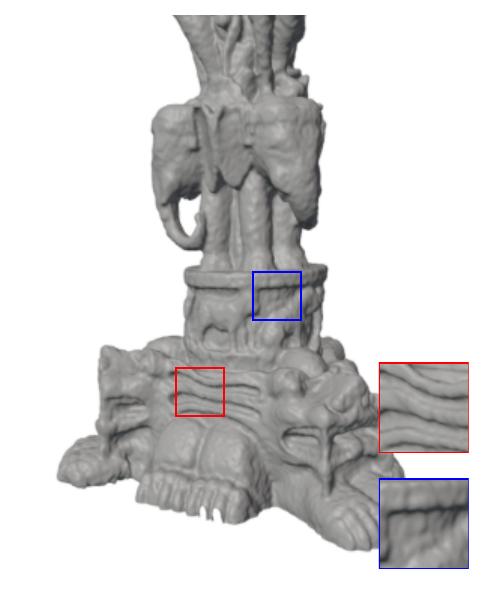}
        \caption{PE}
    \end{subfigure}
    

    \begin{subfigure}{0.23\textwidth}
        \includegraphics[width=\linewidth, trim=0 0 0 45, clip]{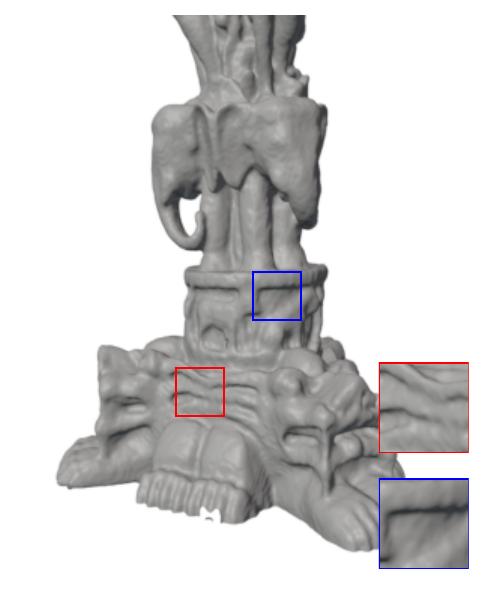}
        \caption{Gauss}
    \end{subfigure}
    \begin{subfigure}{0.23\textwidth}
        \includegraphics[width=\linewidth, trim=0 0 0 45, clip]{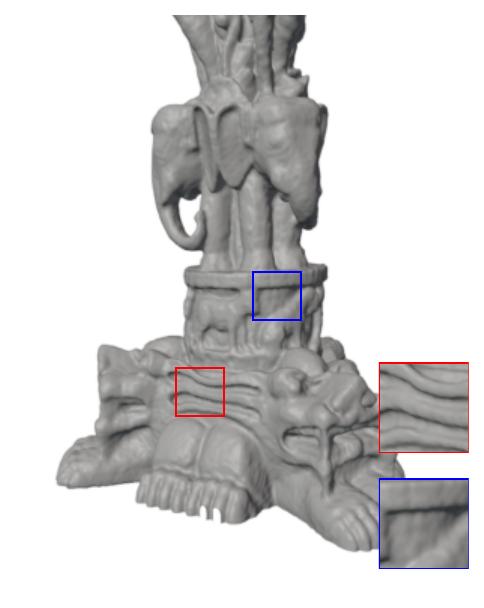}
        \caption{WIRE}
    \end{subfigure}
    \begin{subfigure}{0.23\textwidth}
        \includegraphics[width=\linewidth, trim=0 0 0 45, clip]{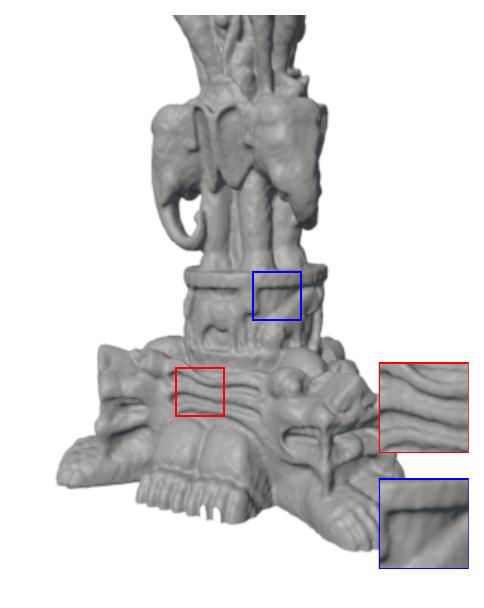}
        \caption{FM-SIREN}
    \end{subfigure}
    \begin{subfigure}{0.23\textwidth}
        \includegraphics[width=\linewidth, trim=0 0 0 45, clip]{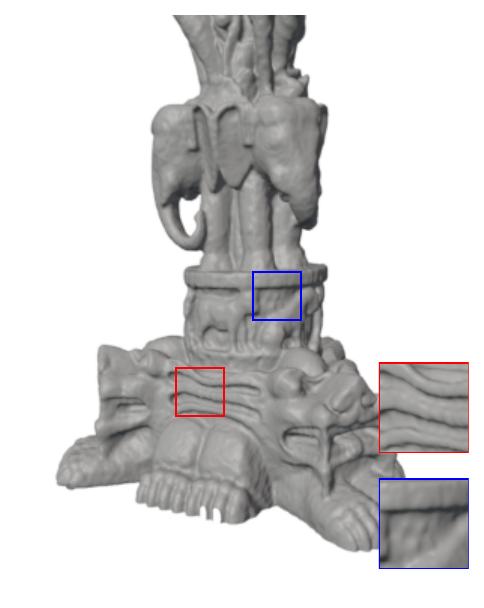}
        \caption{FM-FINER}
    \end{subfigure}

    \caption{Qualitative reconstruction results for the Thai Statue 3D scene using three-layer networks. FM-SIREN and FM-FINER deliver higher-fidelity reconstructions with visibly sharper details compared to baselines. The zoomed-in slices highlight that fine structures are better preserved by our models. Notably, FM-SIREN and FM-FINER achieve lower CD error by 6.67\% and 27.68\% compared to the best baseline, respectively, indicating better reconstruction fidelity.}
    \label{fig:3Dfit_main}
\end{figure}

 For 3D shape fitting, we follow the framework of~\cite{saragadam2023wire}, sampling shapes over a $512 \times 512 \times 512$ grid with a corresponding Nyquist frequency of $256$ cycles per volume. All models use three layers with $256$ neurons each, and performance is evaluated using IoU~\cite{rezatofighi2019generalized} and Chamfer Distance (CD)~\cite{wu2021balanced} on four shapes from the Stanford 3D Scanning Repository~\cite{stanford_3d_scan}. As shown in Table~\ref{tab:3DIoU_main}, FM-SIREN and FM-FINER consistently outperform all baselines across both metrics. The gains in CD are particularly striking --- FM-FINER reduces it by up to $55.15\%$ over the strongest baseline (WIRE), demonstrating that Nyquist-informed frequency diversity and orthogonality better preserves fine geometric detail. This is further confirmed qualitatively in Figure~\ref{fig:3Dfit_main}, where FM-SIREN and FM-FINER recover thin structures. Note that IoU is naturally biased towards 1 since the shapes nearly occupy the complete 3D grid; CD is therefore the more accurate metric for measuring surface-level reconstruction fidelity. Notably, these accuracy gains come at no additional cost: FM-SIREN and FM-FINER match the training times of their respective baselines exactly ($30.03$ vs. $30.02$~min and $32.34$ vs. $32.35$~min), while running $1.6\times$ faster than WIRE.

\begin{table}[t]
    \centering
    \caption{\label{tab:3DIoU_main} Train time, IoU, and CD results for shape fitting on scenes from the Stanford 3D Scanning Repository dataset \cite{stanford_3d_scan}. \colorbox{Goldenrod}{Best} and \colorbox{Dandelion}{Second Best} results are highlighted.}
    \resizebox{1\textwidth}{!}{%
    \begin{tabular}{@{}l c cccc cccc@{}}
        \toprule
        \multirow{2}{*}{\textbf{Model}} & \multirow{2}{*}{\textbf{Train Time (min)}} & \multicolumn{4}{c}{\textbf{IoU $\uparrow$}} & \multicolumn{4}{c}{\textbf{Chamfer Distance $\downarrow$}} \\
        \cmidrule(lr){3-6} \cmidrule(lr){7-10}
         &  & \textbf{Thai Statue} & \textbf{Armadillo} & \textbf{Dragon} & \textbf{Asian Dragon} & \textbf{Thai Statue} & \textbf{Armadillo} & \textbf{Dragon} & \textbf{Asian Dragon} \\
        \midrule
        FINER               & $32.35$ & $0.9763$ & $0.9897$ & $0.9906$ & $0.9760$ & $0.7234$ & $0.5489$ & $0.3892$ & $0.5928$ \\
        Gauss               & $34.34$ & $0.9819$ & $0.9925$ & $0.9558$ & $0.9815$ & $0.5634$ & $0.3976$ & $10.1334$ & $0.4654$ \\
        PE & $29.33$ & $0.9826$ & $0.9949$ & $0.9778$ & $0.9801$ & $0.7684$ & $0.3504$ & $2.3198$ & $0.6088$ \\
        SIREN               & $30.02$ & $0.9598$ & $0.9803$ & $0.9810$ & $0.9508$ & $1.2119$ & $1.0178$ & $0.7627$ & $1.2076$ \\
        WIRE                & $48.47$ & $0.9877$ & $0.9939$ & $0.9932$ & $0.9842$ & $0.3715$ & $0.3144$ & $0.2608$ & $0.4132$ \\
        MIRE                & $70.14$  & $0.9422$ & $0.9799$ & $0.9715$ & $0.9611$ & $1.7462$ & $1.0327$ & $1.1436$ & $0.9482$ \\
        SPDER               & $47.60$ & $0.9791$ & $0.9911$ & $0.9925$ & $0.9789$ & $0.6412$ & $0.4734$ & $0.2965$ & $0.5226$ \\
        \midrule
        \textbf{FM-SIREN}   & $30.03$ & \colorbox{Dandelion}{$0.9888$} & \colorbox{Dandelion}{$0.9965$} & \colorbox{Dandelion}{$0.9950$} & \colorbox{Dandelion}{$0.9875$} & \colorbox{Dandelion}{$0.3467$} & \colorbox{Dandelion}{$0.1846$} & \colorbox{Dandelion}{$0.1981$} & \colorbox{Dandelion}{$0.3137$} \\
        \textbf{FM-FINER}   & $32.34$ & \colorbox{Goldenrod}{$0.9913$} & \colorbox{Goldenrod}{$0.9973$} & \colorbox{Goldenrod}{$0.9961$} & \colorbox{Goldenrod}{$0.9909$} & \colorbox{Goldenrod}{$0.2687$} & \colorbox{Goldenrod}{$0.1410$} & \colorbox{Goldenrod}{$0.1533$} & \colorbox{Goldenrod}{$0.2277$} \\
        \midrule
        \multirow{2}{*}{\textbf{Improvement}} & \textbf{FM-SIREN} & $0.11\%$ & $0.26\%$ & $0.18\%$ & $0.34\%$ & $6.67\%$ & $41.28\%$ & $24.04\%$ & $24.08\%$ \\
                                              & \textbf{FM-FINER} & $0.36\%$ & $0.34\%$ & $0.29\%$ & $0.68\%$ & $27.68\%$ & $55.15\%$ & $41.22\%$ & $44.89\%$ \\
        \bottomrule
    \end{tabular}%
    }

\end{table}

\begin{figure}[t]
    \centering
    \begin{subfigure}{0.18\textwidth}
        \includegraphics[width=\textwidth]{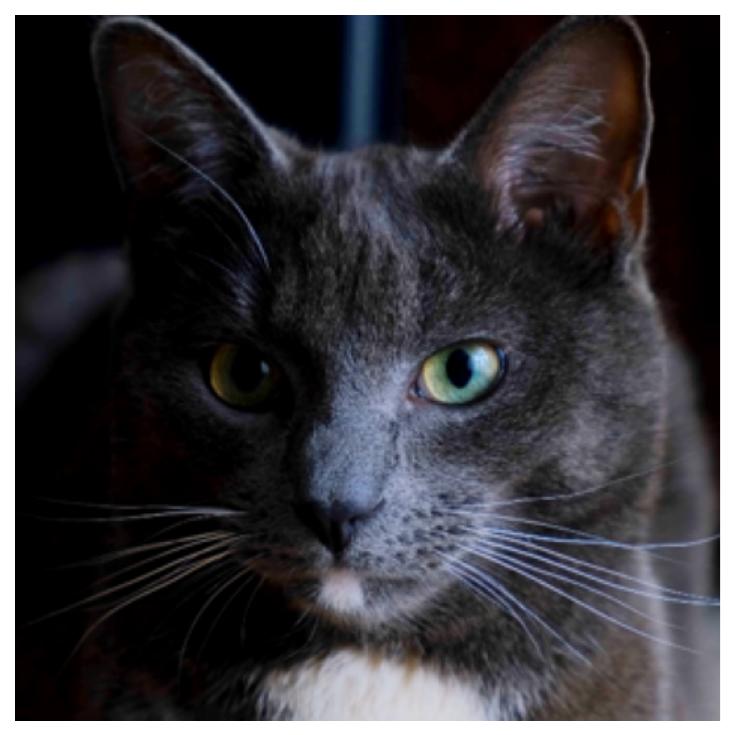}
        \caption{Ground Truth}
    \end{subfigure}
    \hfill
    \begin{subfigure}{0.18\textwidth}
        \includegraphics[width=\textwidth]{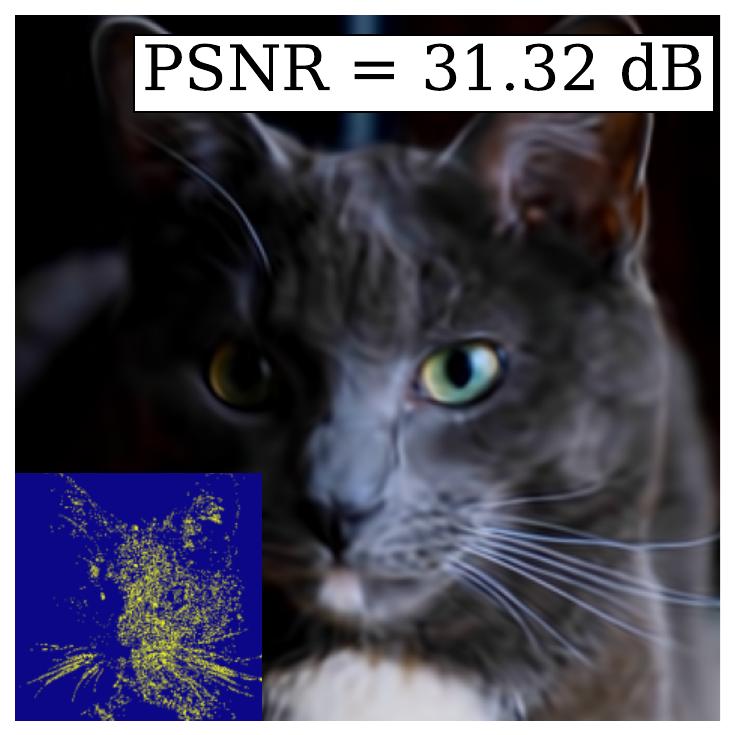}
        \caption{SIREN}
    \end{subfigure}
    \hfill
    \begin{subfigure}{0.18\textwidth}
        \includegraphics[width=\textwidth]{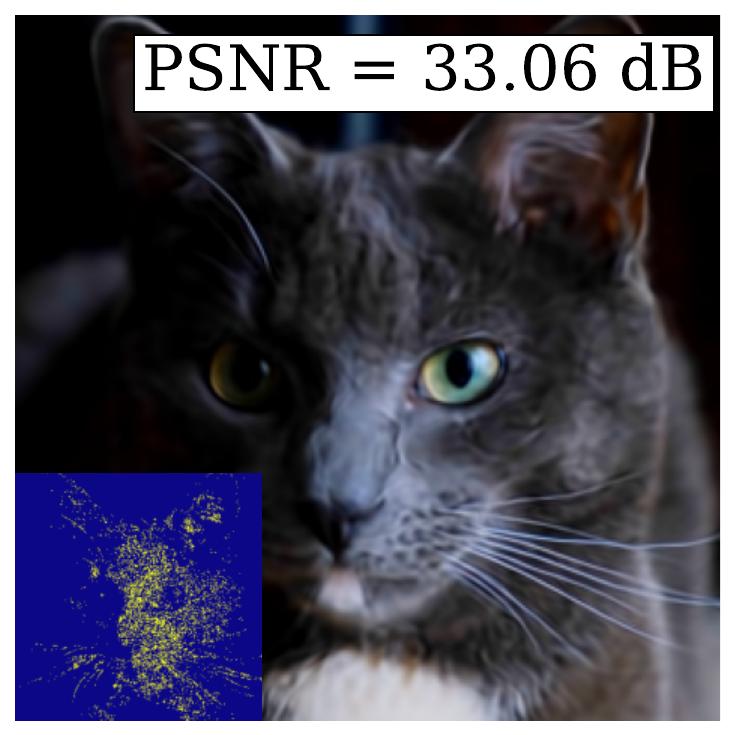}
        \caption{FINER}
    \end{subfigure}
    \hfill
    \begin{subfigure}{0.18\textwidth}
        \includegraphics[width=\textwidth]{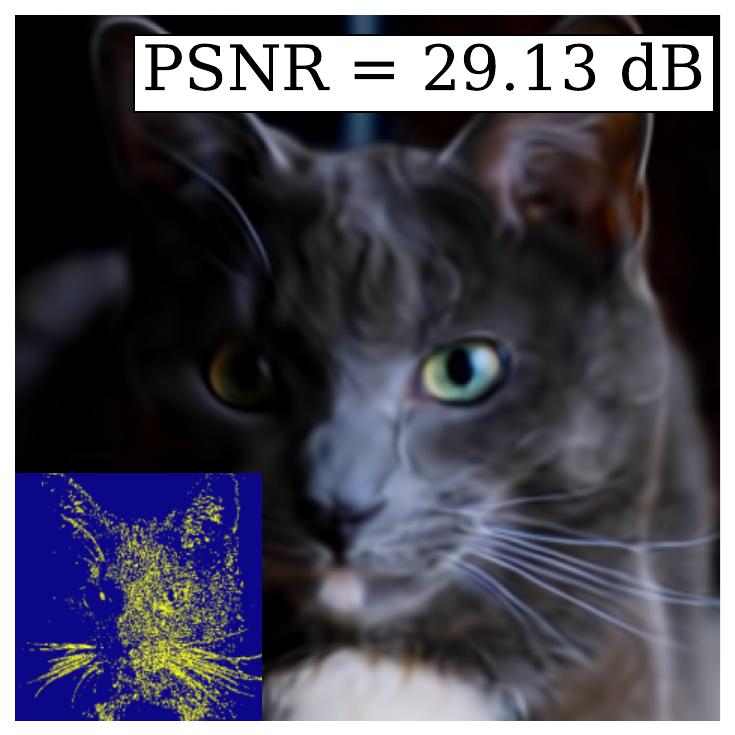}
        \caption{MIRE}
    \end{subfigure}
    \hfill
    \begin{subfigure}{0.18\textwidth}
        \includegraphics[width=\textwidth]{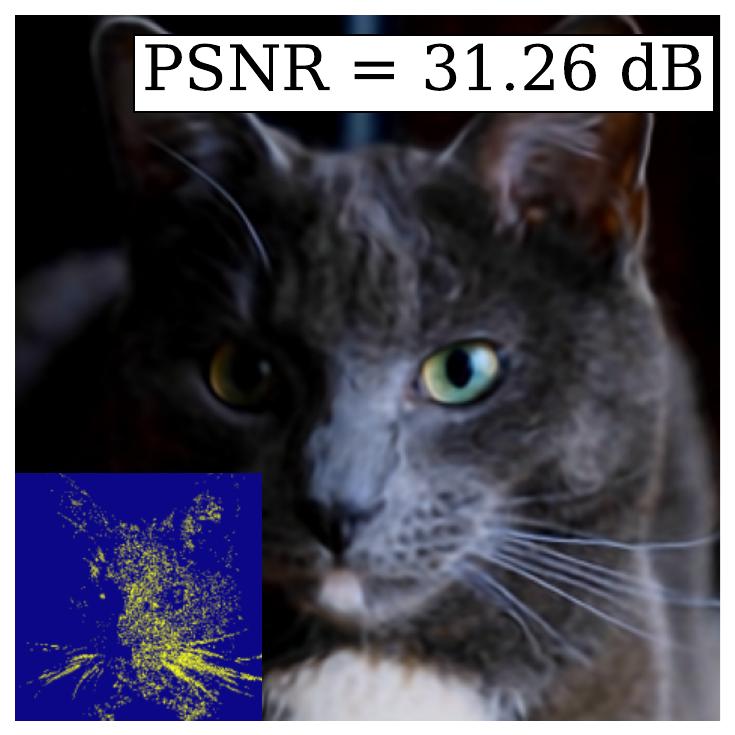}
        \caption{SPDER}
    \end{subfigure}

    \vspace{-0.1em}

    \begin{subfigure}{0.18\textwidth}
        \includegraphics[width=\textwidth]{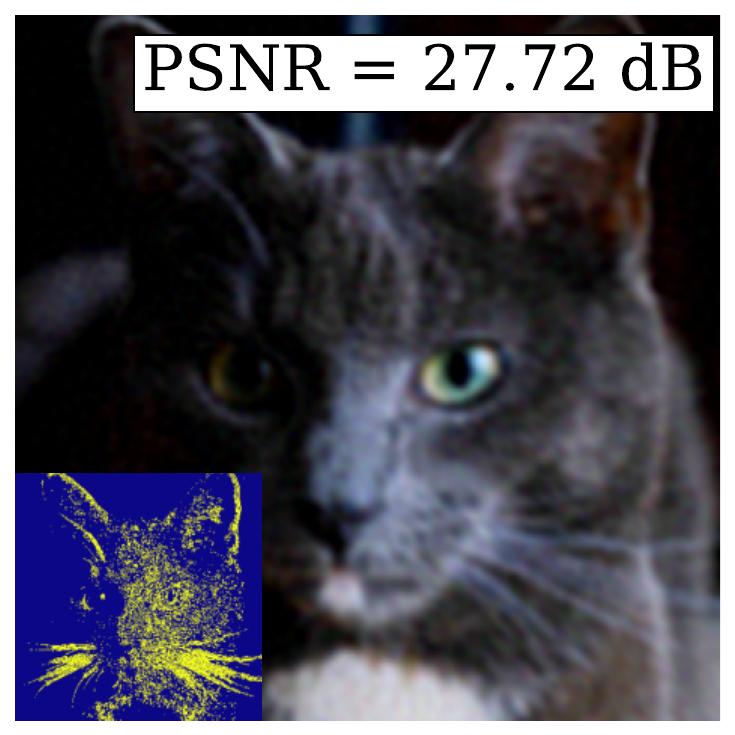}
        \caption{Gauss}
    \end{subfigure}
    \hfill
    \begin{subfigure}{0.18\textwidth}
        \includegraphics[width=\textwidth]{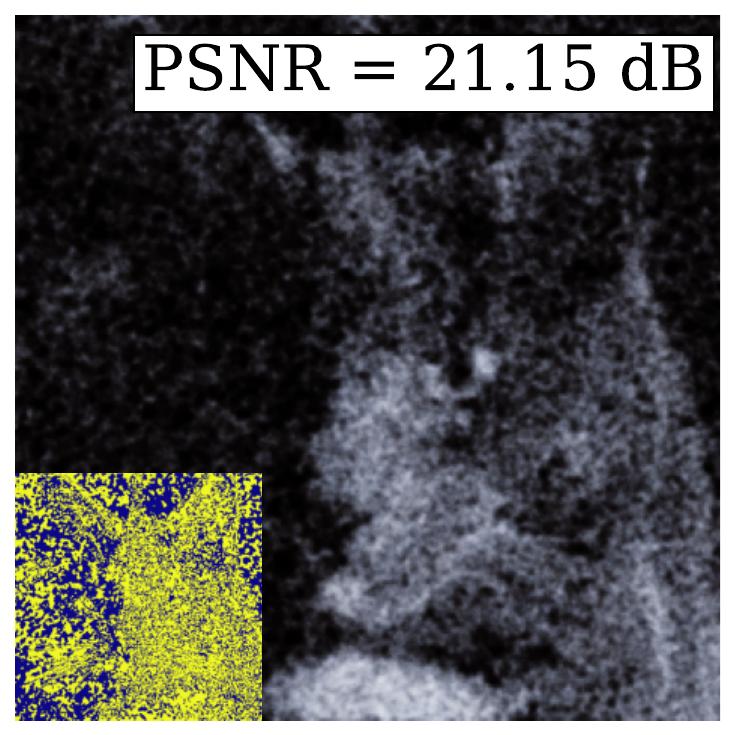}
        \caption{PE}
    \end{subfigure}
    \hfill
    \begin{subfigure}{0.18\textwidth}
        \includegraphics[width=\textwidth]{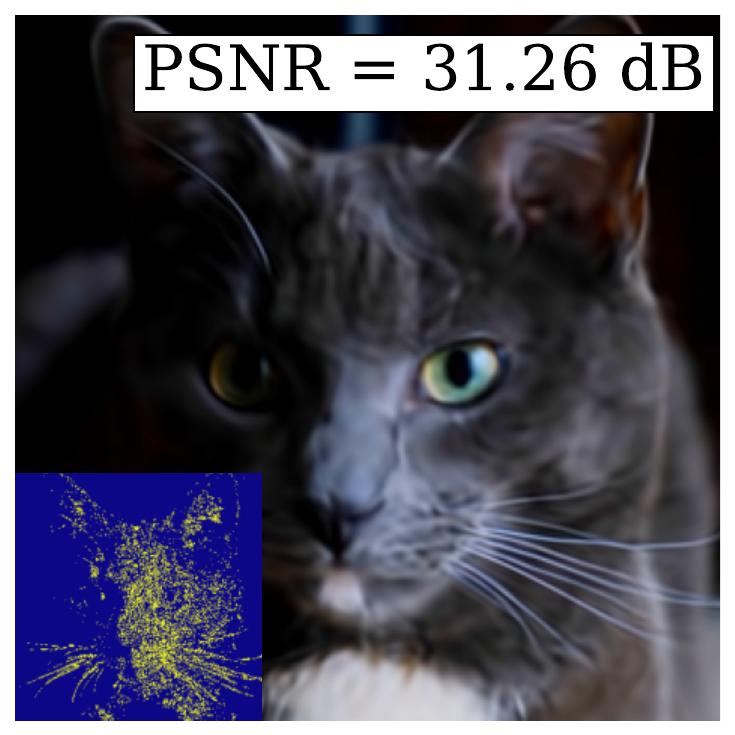}
        \caption{FreSh}
    \end{subfigure}
    \hfill
    \begin{subfigure}{0.18\textwidth}
        \includegraphics[width=\textwidth]{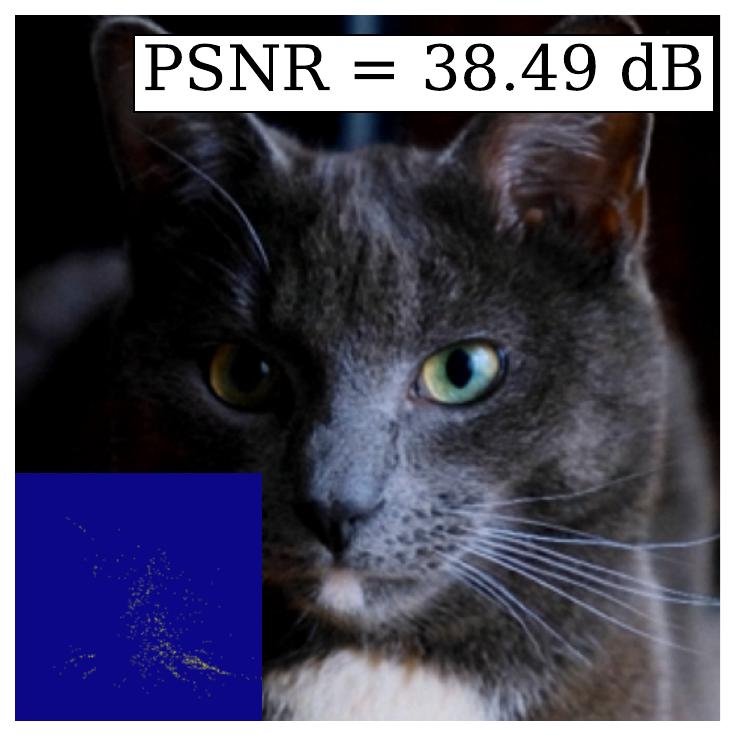}
        \caption{FM-SIREN}
    \end{subfigure}
    \hfill
    \begin{subfigure}{0.18\textwidth}
        \includegraphics[width=\textwidth]{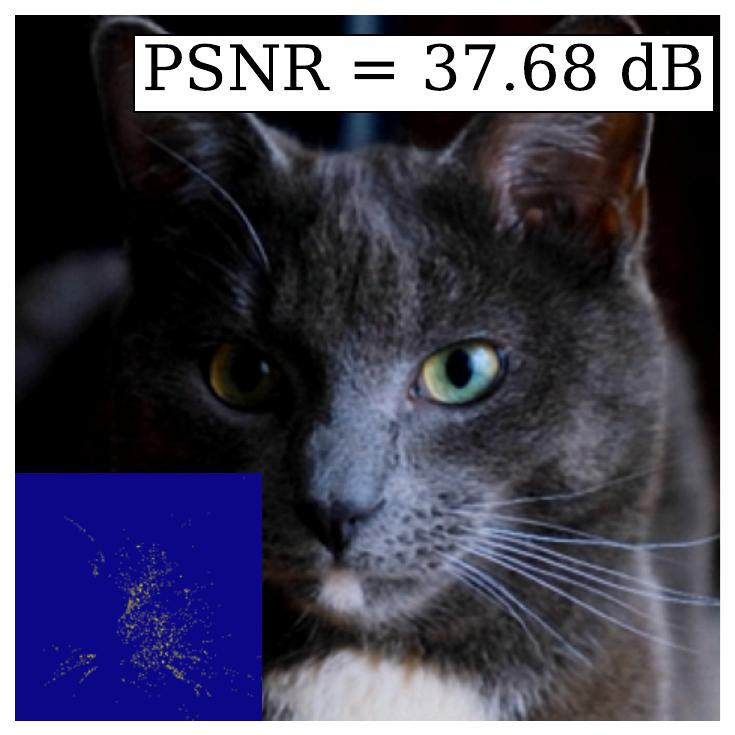}
        \caption{FM-FINER}
    \end{subfigure}
    \caption{Qualitative results on video reconstruction from the ScikitVid dataset~\cite{scikitvideo}. The PSNR (dB) of each reconstruction is reported in the top-right corner of its subfigure. FM-SIREN and FM-FINER produce the sharpest and most faithful reconstructions across all baselines, consistent with the quantitative results in Table~\ref{tab:video}.}
    \label{fig:video}
\end{figure}

\subsection{Video Fitting}
\par We evaluate FM-SIREN and FM-FINER on fitting videos from the ScikitVid dataset~\cite{scikitvideo} and the commonly used cat video from~\cite{sitzmann2019siren}, following the same evaluation protocol. Input videos are sampled over a 3D volume of shape $300 \times 300 \times N$, where $N$ is the number of frames, yielding Nyquist frequencies of $150$, $150$ and $N/2$ for the horizontal, vertical and temporal dimensions, respectively. Although the temporal dimension tends to exhibit low-frequency content due to the gradual nature of frame-to-frame changes, FM-SIREN and FM-FINER achieve significant performance improvements over all baseline methods, as shown in Table~\ref{tab:video}. Figure~\ref{fig:video} shows a frame snapshot of each reconstructed video, where FM-SIREN and FM-FINER produce the sharpest and most faithful reconstructions. The frame interpolation results are provided in the supplementary material.

\begin{table}[t]
    \centering
    \caption{PSNR results in dB for video fitting for different models. \colorbox{Goldenrod}{Best} and \colorbox{Dandelion}{Second Best} results are highlighted.}
    \resizebox{1\textwidth}{!}{%
    \begin{tabular}{@{}l cccccccc  cc  cc@{}}
        \toprule
        \multirow{2}{*}{\textbf{Dataset}} & \multicolumn{10}{c}{\textbf{PSNR (dB)} $\uparrow$} & \multicolumn{2}{c}{\textbf{Improvement}} \\
        \cmidrule(lr){2-11} \cmidrule(lr){12-13}
        & FINER & Gauss & PE & SIREN & WIRE & SPDER & MIRE & FreSh & \textbf{FM-SIREN} & \textbf{FM-FINER} & \textbf{FM-SIREN} & \textbf{FM-FINER} \\

        \midrule
        Cat                 & $34.43$ & $27.25$ & $21.57$ & $33.09$ & $13.50$ & $32.42$ & $30.87$ & $33.19$ & \colorbox{Goldenrod}{$37.54$}  & \colorbox{Dandelion}{$37.31$}  & $3.11$~dB & $2.88$~dB \\
        Bikes               & $38.56$ & $26.15$ & $20.31$ & $35.01$ & $10.59$ & $37.23$ & $29.26$ & $36.15$ & \colorbox{Dandelion}{$43.10$}  & \colorbox{Goldenrod}{$44.09$}  & $4.54$~dB & $5.53$~dB \\
        BigBuckBunny        & $30.54$ & $25.70$ & $19.12$ & $28.83$ & $12.11$ & $29.53$ & $27.09$ & $28.50$ & \colorbox{Dandelion}{$34.71$}  & \colorbox{Goldenrod}{$37.12$}  & $4.17$~dB & $6.58$~dB \\
        Carphone Distorted  & $36.44$ & $28.34$ & $21.53$ & $33.96$ & $10.61$ & $35.84$ & $30.59$ & $24.12$ & \colorbox{Dandelion}{$42.13$}  & \colorbox{Goldenrod}{$43.57$}  & $5.69$~dB & $7.13$~dB \\
        \midrule
        Train Time (min)    & $8.59$  & $9.22$  & $6.41$  & $7.27$  & $42.52$ & $17.04$ & $11.80$ & $8.42$ & $7.21$ & $9.03$ & --- & --- \\
        \bottomrule
    \end{tabular}%
    \label{tab:video}
    }
    \vspace{-20pt}
\end{table}

\section{Discussion}

The consistent and significant improvements with of FM-SIREN and FM-FINER (up to 9.44 and 7.13 dB in images and videos, and 55.15\% reduction in CD in 3D shapes) provide compelling evidence that hidden feature redundancy is a fundamental bottleneck in periodic-activation INRs, and that Nyquist-informed orthogonality is an effective and principled solution. One limitation of the proposed solution is the dependence on Cartesian sampling: in non-Cartesian settings such as NeRF, input coordinates do not correspond to a regular grid, so the Nyquist-informed frequency vector does not align with the hidden feature spectral structure, thus limiting redundancy reduction. Although performance improvement is observed with NeRF, extending this design to irregular coordinate systems remains an open research problem.
\section{Conclusion}

\par In this work, we identified hidden feature redundancy, caused by fixed, shared frequency multipliers, as a fundamental bottleneck in periodic-activation INRs. FM-SIREN and FM-FINER architectures address this by assigning Nyquist-informed, neuron-specific frequency multipliers, reducing redundancy by nearly 50\% and achieving consistent improvements across 1D audio, 2D image, 3D shape, and video fitting with compact networks. Beyond reconstruction quality, the compactness of our representations offers a promising path toward accurate and efficient neural codecs and data compression, bridging classical signal processing and modern implicit neural representations.
\vspace{-15pt}

\newpage
\bibliographystyle{splncs04}
\bibliography{main}

\newpage
\setcounter{figure}{0}
\setcounter{table}{0}
\section{Experimental Setup}
\subsection{Experimentation Infrastructure}
For all experiments, we used PyTorch \cite{paszke2019pytorch} with the Adam optimizer \cite{zhang2018improved} and a StepLR scheduler that decayed the learning rate by a factor of 0.1 every 100 epochs. All experiments were conducted on an NVIDIA H200 (80GB VRAM, 256GB RAM).

\subsection{Evaluation Metrics}
We used mean square error (MSE) as the training loss function and to evaluate reconstruction performance for audio fitting:
\begin{equation}
    \operatorname{MSE}=\frac{\sum_{i=1}^{N}\left(x_{i}-\hat{x}_{i}\right)^{2}}{N}
\end{equation}
\noindent where $x_{i}$ is the ground-truth value, $\hat{x}_{i}$ is the predicted value, and $N$ is the number of samples.

For image reconstruction, we report the Peak Signal-to-Noise Ratio (PSNR) and the Structural Similarity Index (SSIM). PSNR measures reconstruction quality in decibels (dB), where higher values indicate better fidelity:
\begin{equation}
    \operatorname{PSNR} = 10 \cdot \log_{10} \left( \frac{MAX_{I}^2}{\operatorname{MSE}} \right)
\end{equation}
\noindent where $MAX_I$ is the maximum possible pixel value (255 for 8-bit images). SSIM evaluates perceptual similarity between two images based on luminance, contrast, and structure, ranging from $-1$ to $1$, with $1$ indicating perfect similarity:
\begin{equation}
    \operatorname{SSIM}(x, y) = \frac{(2\mu_x\mu_y + c_1)(2\sigma_{xy} + c_2)}{(\mu_x^2 + \mu_y^2 + c_1)(\sigma_x^2 + \sigma_y^2 + c_2)}
\end{equation}
\noindent where $\mu_x$, $\mu_y$ are the mean pixel values, $\sigma_x$, $\sigma_y$ are the standard deviations, $\sigma_{xy}$ is the cross-covariance, and $c_1$, $c_2$ are stabilizing constants.

For 3D shape fitting, we report two metrics, the Intersection over Union (IoU) and the Chamfer Distance (CD). IoU measures the volumetric overlap between the predicted shape $V_p$ and ground-truth $V_{gt}$:
\begin{equation}
    \operatorname{IoU} = \frac{|V_p \cap V_{gt}|}{|V_p \cup V_{gt}|}
\end{equation}
\noindent where $\cap$ and $\cup$ denote the intersection and the union over occupied voxels, respectively. A higher IoU indicates better reconstruction. However, since shapes tend to occupy a large portion of the voxel grid, IoU can be biased toward high values; we therefore also report CD, which operates on surface point clouds:
\begin{equation}
    \operatorname{CD}(S_p, S_{gt}) = \frac{1}{|S_p|}\sum_{x \in S_p} \min_{y \in S_{gt}} \|x - y\|_2^2
    + \frac{1}{|S_{gt}|}\sum_{y \in S_{gt}} \min_{x \in S_p} \|x - y\|_2^2
\end{equation}
\noindent where $S_p$ and $S_{gt}$ are the predicted and ground-truth point sets, respectively. The two terms penalize coverage errors in both directions, and a lower CD value indicates better surface reconstruction.

For analyzing hidden feature redundancy, we compute the Frobenius norm \cite{bottcher2008frobenius} of the covariance matrix of the hidden embeddings.
Given activations $\mathbf{Z} \in \mathbb{R}^{N \times d}$ collected from a hidden layer over $N$ input coordinates,
the covariance matrix $\mathbf{C} \in \mathbb{R}^{d \times d}$ is given by:
\begin{equation}
    \mathbf{C} = \frac{1}{N-1}(\mathbf{Z} - \bar{\mathbf{Z}})^\top(\mathbf{Z} - \bar{\mathbf{Z}})
\end{equation}
\noindent where T denotes Transpose of a matrix and $\bar{\mathbf{Z}}$ is the column-wise mean. The Frobenius norm of $\mathbf{C}$ is then:
\begin{equation}
    \|\mathbf{C}\|_F = \sqrt{\sum_{i=1}^{d}\sum_{j=1}^{d} C_{ij}^2}
\end{equation}
\noindent where a larger $\|\mathbf{C}\|_F$ indicates higher inter-neuron correlation and thus greater redundancy in the hidden representations, while a smaller value suggests more decorrelated, information-dense activations.
\section{Ablation Study}

\begin{figure}[t]
    \centering
    \begin{subfigure}{0.11\textwidth}
        \centering
        \includegraphics[width=\linewidth]{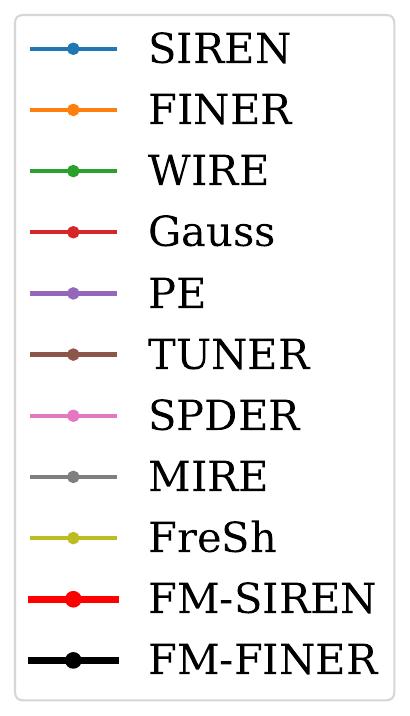}
        \label{fig:ablation_legend}
    \end{subfigure}
    \hspace{-0.5em}
    \begin{subfigure}{0.28\textwidth}
        \centering
        \includegraphics[width=\linewidth]{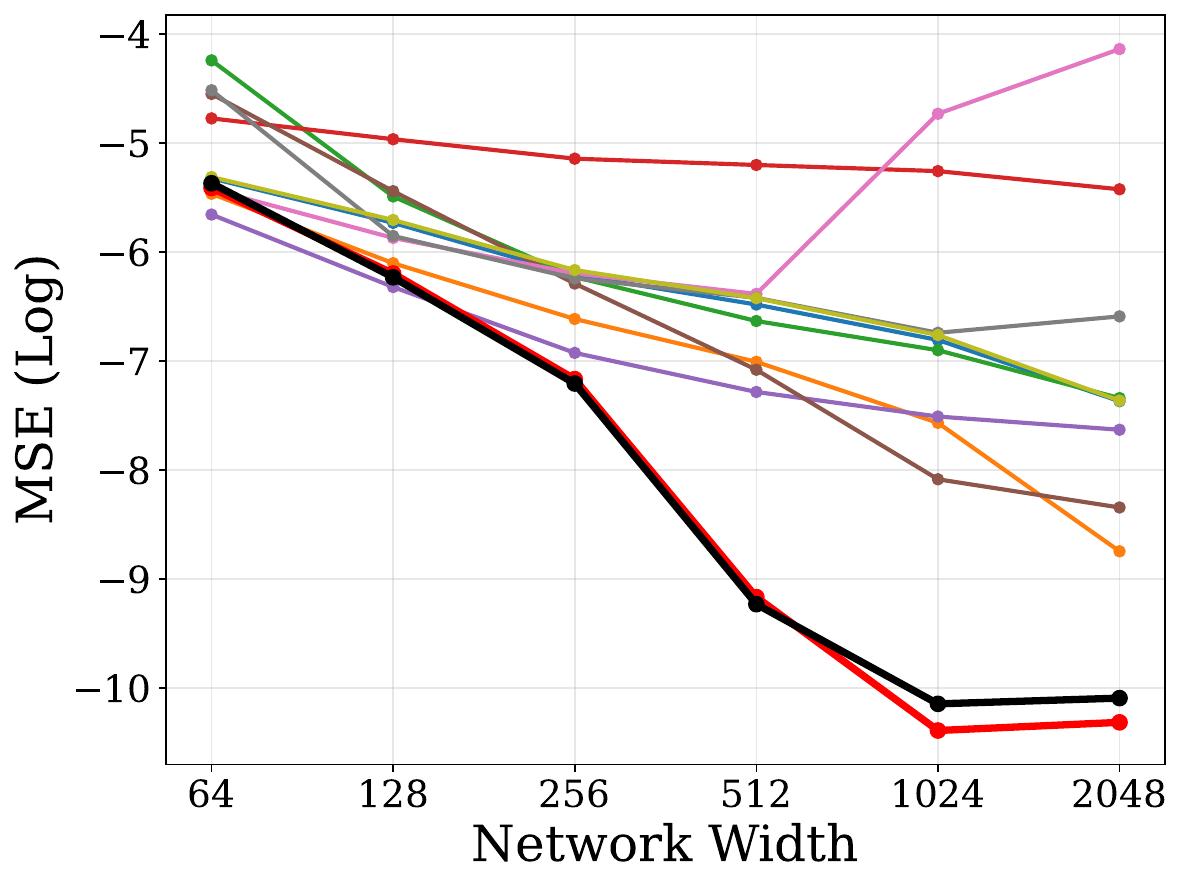}
        \caption{Network Width}
        \label{fig:ablation1}
    \end{subfigure}
    \hspace{-0.5em}
    \begin{subfigure}{0.28\textwidth}
        \centering
        \includegraphics[width=\linewidth]{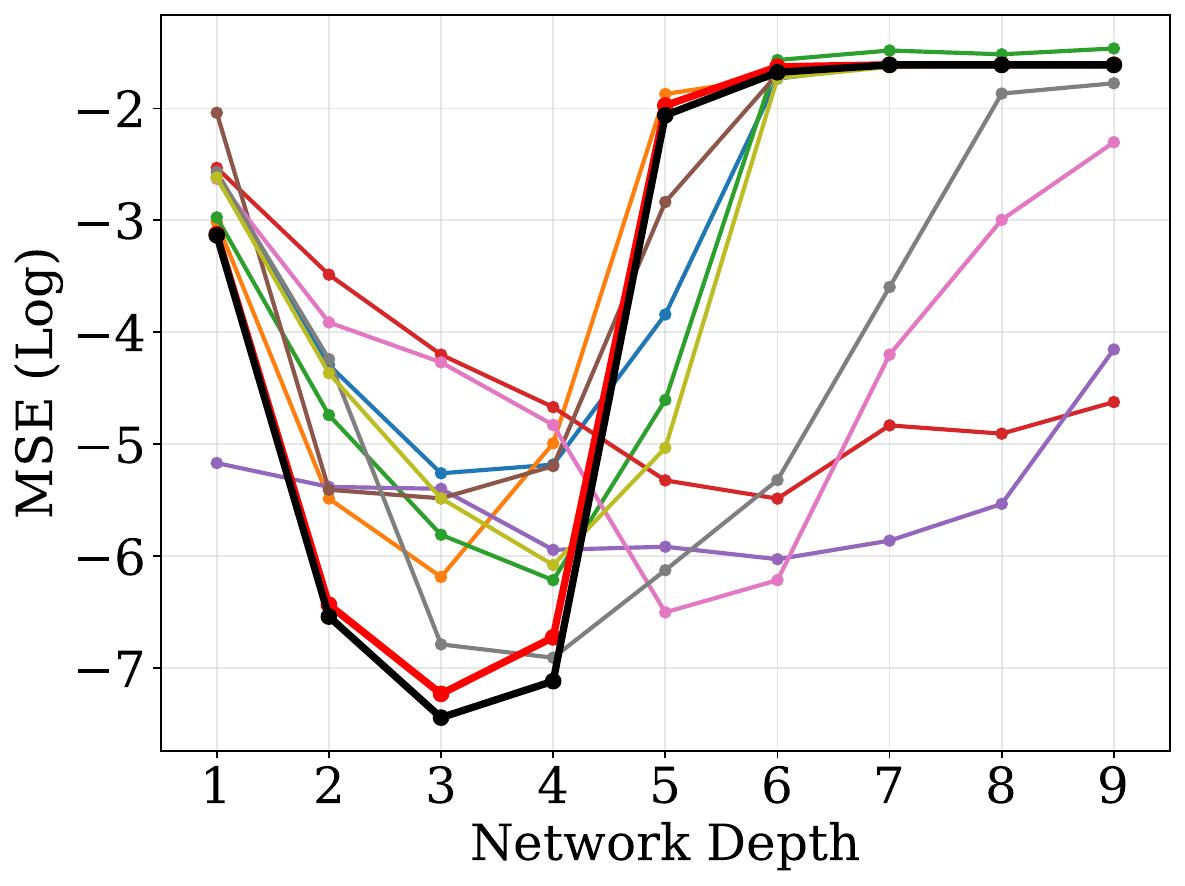}
        \caption{Network Depth}
        \label{fig:ablation2}
    \end{subfigure}
    \hspace{-0.5em}
    \begin{subfigure}{0.28\textwidth}
        \centering
        \includegraphics[width=\linewidth]{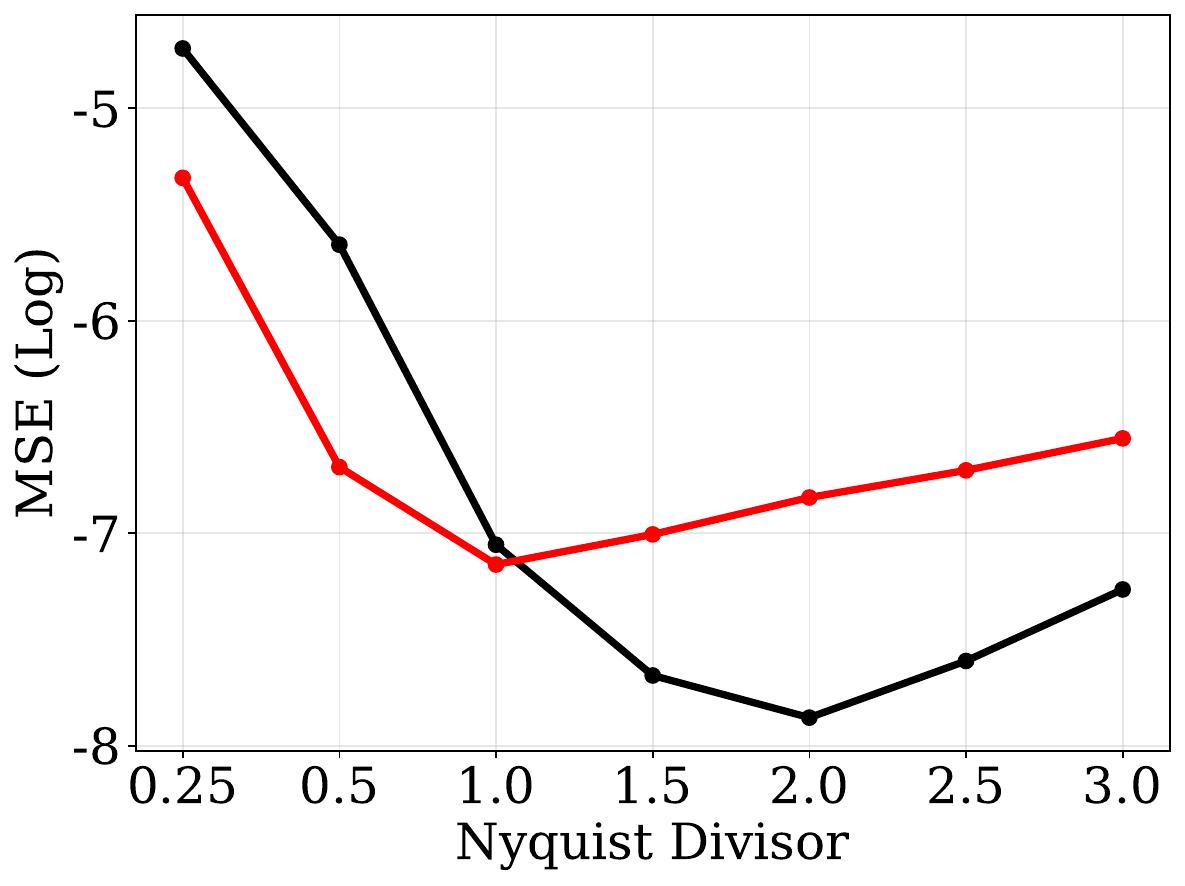}
        \caption{Nyquist Factor}
        \label{fig:ablation3}
    \end{subfigure}
    
    \caption{Ablation study for different factors in the network design.}
    \label{fig:ablation}
\end{figure}

We conducted ablation experiments to assess the impact of three key design factors: network width, network depth, and the Nyquist divisor (i.e., the maximum frequency multiplier relative to the Nyquist frequency).

\subsection{Network Width}
We varied the number of neurons per layer over $\{128, 256, 512, 1024, 2048\}$, fixing the depth to two layers. As shown in Figure~\ref{fig:ablation1}, FM-SIREN and FM-FINER consistently outperform all baselines on most of the width setting. However, performance plateaus beyond $1024$ neurons, which we attribute to increased redundancy in hidden representations: neuron-specific frequency multipliers are capped by the Nyquist range, causing frequency components to overlap at larger widths. Nonetheless, both models maintain a clear performance advantage over baselines.

\subsection{Network Depth}
We varied the number of layers over $\{2, 3, 4, 5, 6, 7, 8, 9\}$, fixing the network width
to 256 neurons. As shown in Figure~\ref{fig:ablation2}, all models exhibit an oscillatory
behavior with respect to depth: MSE decreases up to a model-specific optimal depth, and
then increases monotonically beyond that point, consistent with known optimization challenges in deep networks \cite{goodfellow2016deep}. The key difference across models lies in
when this degradation begins and how severe it is.

For FM-SIREN and FM-FINER, peak performance is achieved at three layers, after which the MSE begins to rise. We attribute this to the interaction between frequency-modulated activations and depth: in shallow networks, the Nyquist-constrained frequency multipliers assign diverse, non-overlapping frequency components across neurons, enabling compact and efficient signal representation. As depth increases beyond three layers, the accumulated frequency components from earlier layers propagate through subsequent layers as high-frequency noise, compounding representational errors rather than refining them. While other baselines (MIRE, SPDER, Gauss and PE) may sustain performance gains up to five or six layers before degrading, their overall MSE at any depth remains higher than that of FM-SIREN and FM-FINER at their respective optima. Importantly, this behavior does not constitute a practical limitation since shallow networks are more efficient to store, train, and deploy. This makes FM-SIREN and FM-FINER particularly well-suited for resource-constrained INR applications such as data compression, signal transfer, and image super-resolution. Crucially, at every depth tested, the three-layer FM-SIREN and FM-FINER outperform the best-performing baseline at its optimal depth, demonstrating that our models are not only more accurate, but also more depth-efficient than competing methods.

\subsection{Nyquist Divisor}
We studied the effect of the Nyquist divisor by testing frequency multiplier upper bounds at $f_{Nyquist} / k$ for $k \in \{0.25, 0.5, 1.0, 1.5, 2.0, 2.5, 3.0\}$, where $f_{Nyquist} = \max(D_n) / 2$. As shown in Figure~\ref{fig:ablation3}, FM-SIREN achieves peak performance at $k = 1.0$, i.e., when the multiplier equals exactly the Nyquist limit, and degrades at larger $k$ as the representable frequency range becomes overly restricted. FM-FINER, whose activations inherently span a broader instantaneous bandwidth, achieves peak performance at $k = 2.0$ (as presented in Equation 10), beyond which the restricted frequency range begins to limit its representational capacity. Taken together with the width and depth studies, these results confirm that Nyquist-informed frequency diversity and orthogonality is the key driver of improved performance in FM-SIREN and FM-FINER.

\begin{figure}[h]
    \centering
    \begin{subfigure}{0.9\linewidth}
        \includegraphics[width=\linewidth]{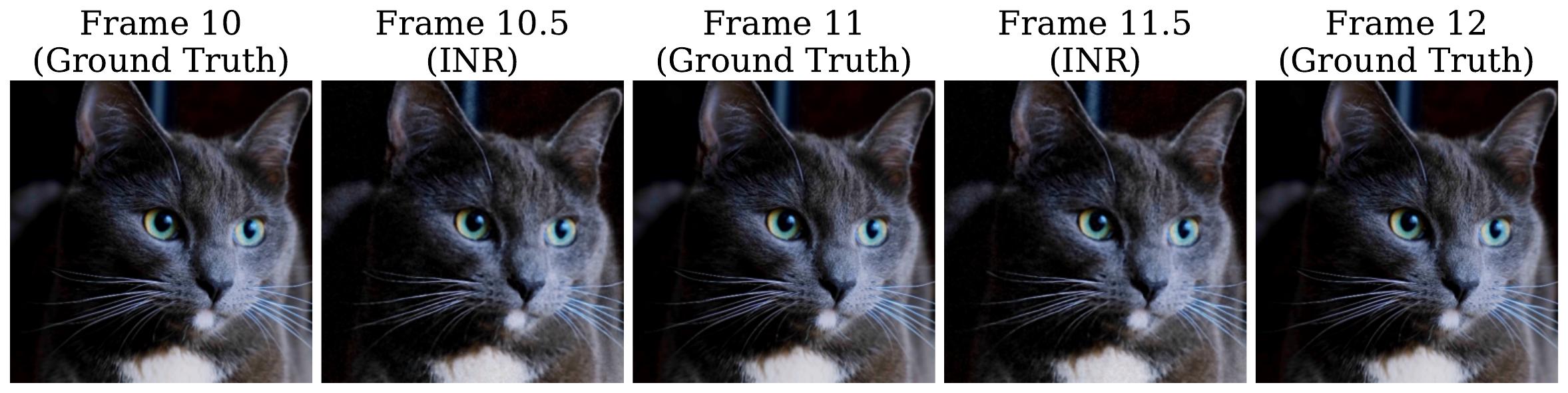}
    \end{subfigure}


    \begin{subfigure}{0.9\linewidth}
        \includegraphics[width=\linewidth]{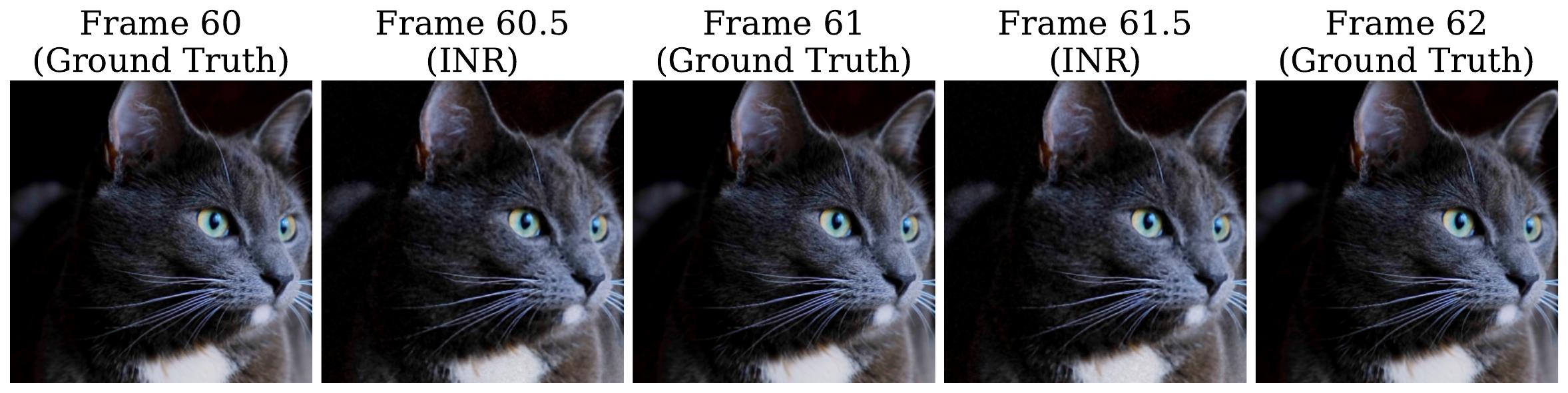}
        \caption{FM-SIREN}
    \end{subfigure}


    \begin{subfigure}{0.9\linewidth}
        \includegraphics[width=\linewidth]{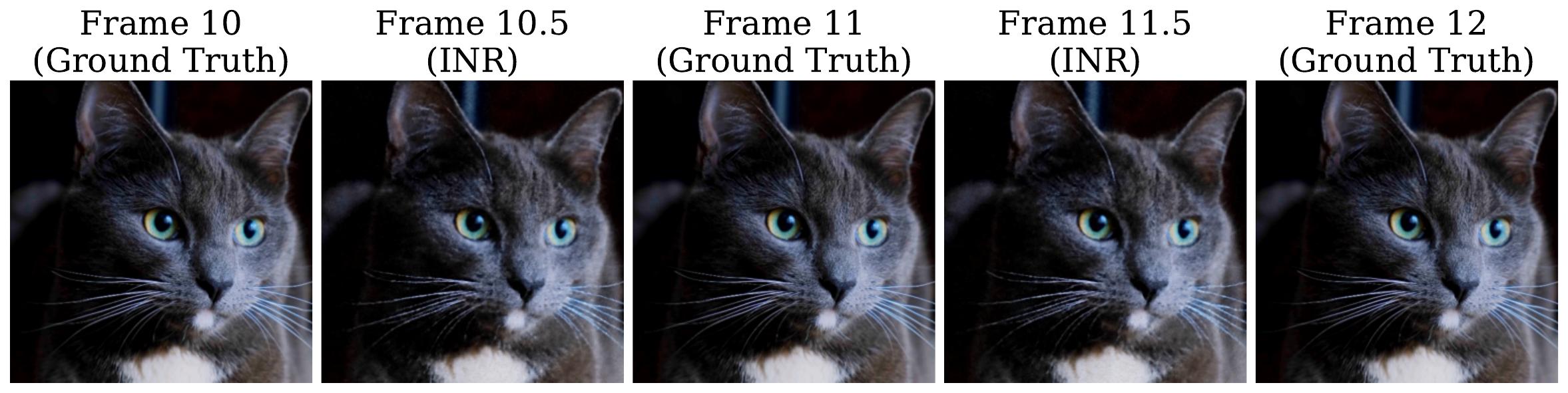}
    \end{subfigure}


    \begin{subfigure}{0.9\linewidth}
        \includegraphics[width=\linewidth]{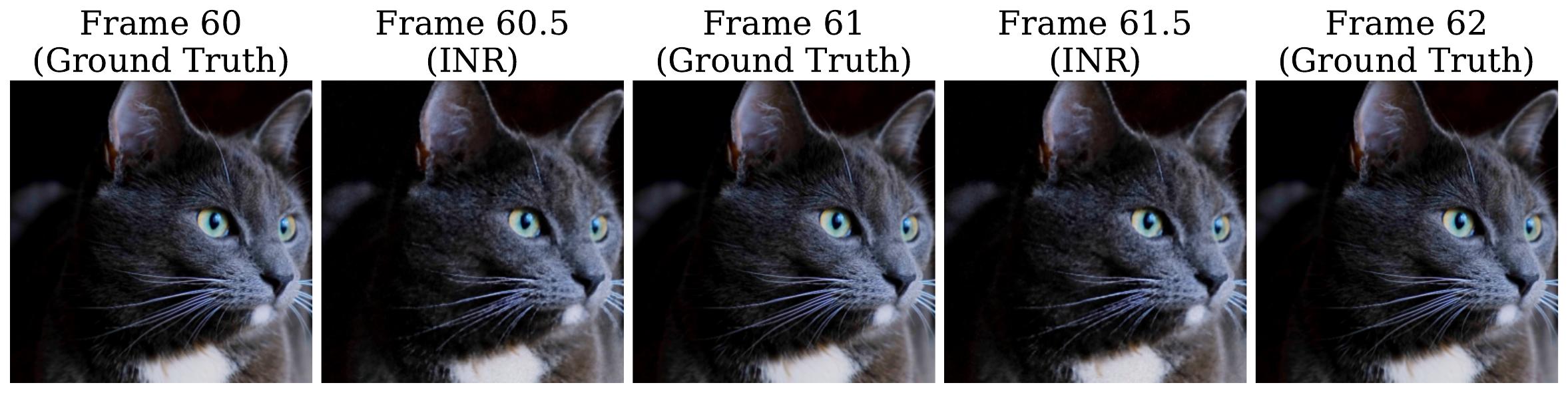}
        \caption{FM-FINER}
    \end{subfigure}
    \caption{Video frame interpolation. Top two rows show FM-SIREN's results where bottom two rows show FM-FINER's results. Reconstruction is achieved with the same networks of video fitting experiments.}
    \label{fig:video_interpolation}
\end{figure}

\section{Additional Experiments}
\subsection{Video Interpolation}

Beyond static image fitting, INRs can represent full videos by extending the coordinate domain to include time. Given a video of $T$ frames with spatial resolution $H \times W$, we train the network to map coordinates$(x, y, t) \in [0,1]^3$ to RGB values, where $t$ is normalized to $[0,1]$. A key advantage of this continuous formulation is the ability to query the network at arbitrary temporal coordinates, enabling sub-frame interpolation without any additional mechanism. To evaluate this capability, we train FM-SIREN and FM-FINER on a video sequence and query the models at half-integer time steps $t = n + 0.5$ for $n \in \{0, 1, \ldots, T-2\}$, corresponding to frames never seen during training. Figure~\ref{fig:video_interpolation} shows qualitative results in which interpolated frames exhibit strong reconstruction between adjacent ground-truth frames without temporal artifacts, demonstrating that frequency-modulated activations learn a temporally coherent representation of the video signal.


\subsection{Audio}

\par
\begin{figure}[t]
    \centering
    \begin{subfigure}{0.20\textwidth}
        \includegraphics[width=\linewidth]{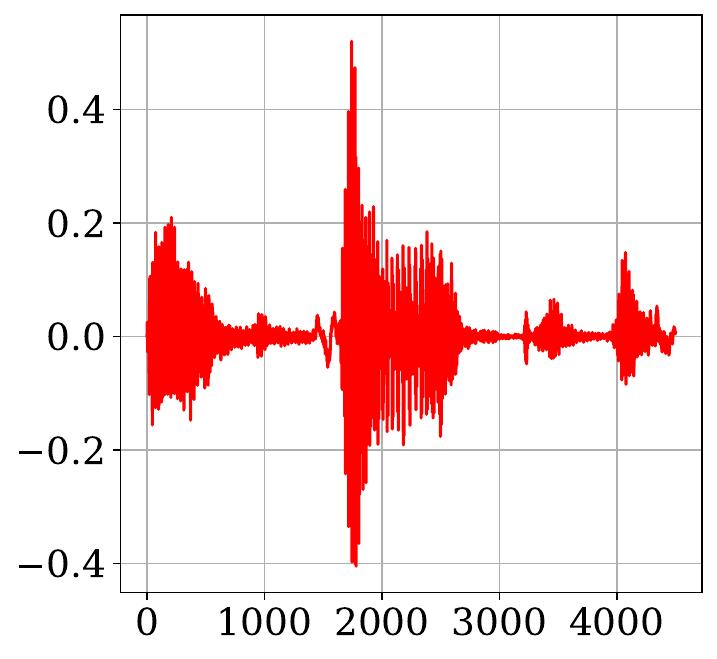}
        \caption{Ground Truth}
    \end{subfigure}
    \hspace{-0.5em}
    \begin{subfigure}{0.20\textwidth}
        \includegraphics[width=\linewidth]{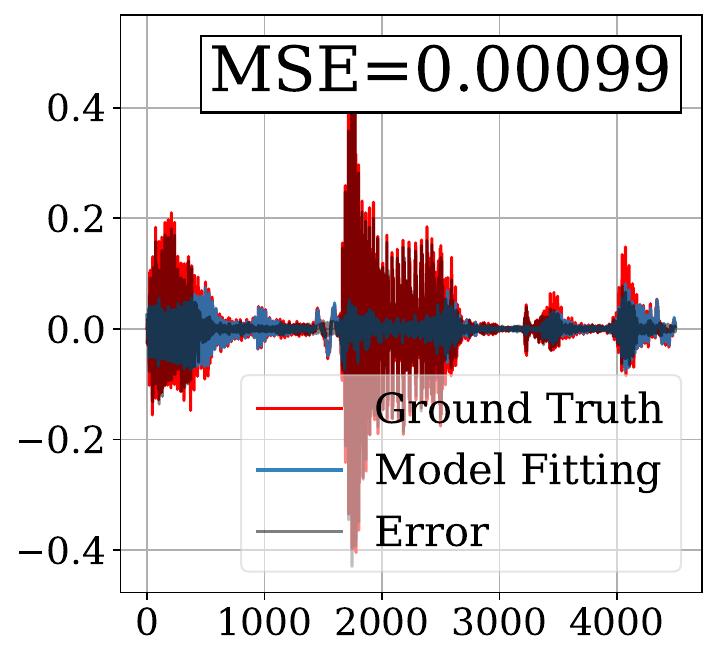}
        \caption{SIREN}
    \end{subfigure}\hspace{-0.5em}
    \begin{subfigure}{0.20\textwidth}
        \includegraphics[width=\linewidth]{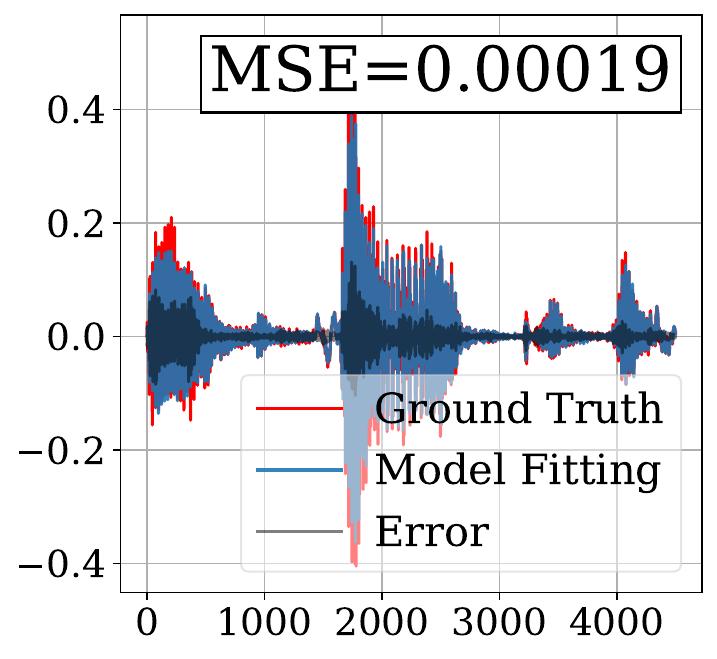}
        \caption{FINER}
    \end{subfigure}\hspace{-0.5em}
    \begin{subfigure}{0.20\textwidth}
        \includegraphics[width=\linewidth]{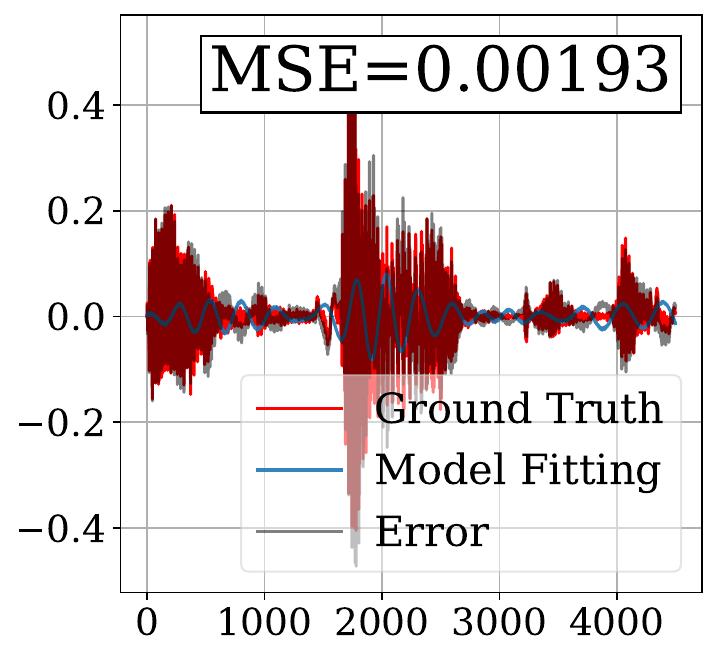}
        \caption{WIRE}
    \end{subfigure}\hspace{-0.5em}
    \begin{subfigure}{0.20\textwidth}
        \includegraphics[width=\linewidth]{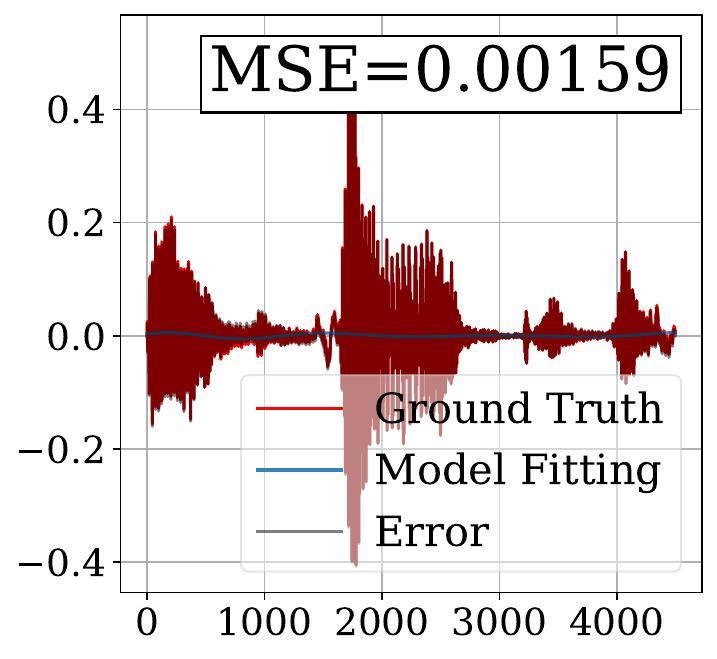}
        \caption{Gauss}
    \end{subfigure}

    \vspace{0.5em}

    \begin{subfigure}{0.20\textwidth}
        \includegraphics[width=\linewidth]{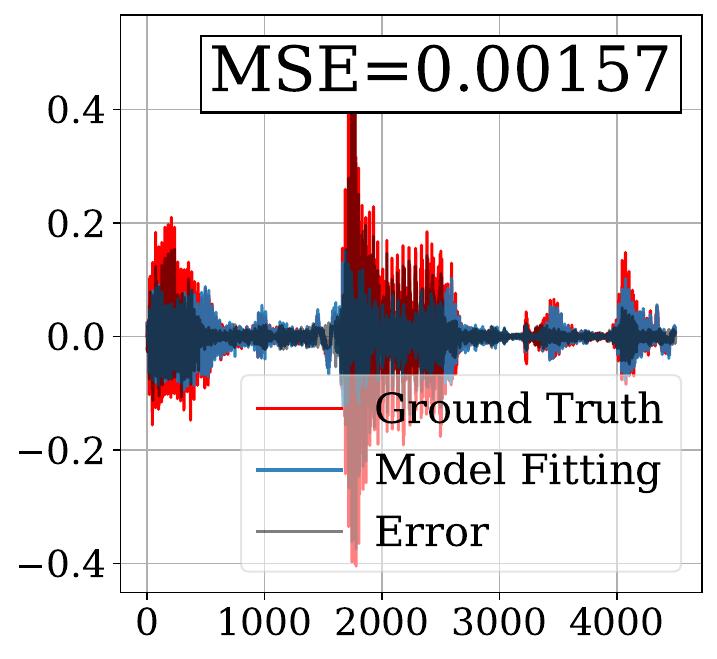}
        \caption{MIRE}
    \end{subfigure}\hspace{-0.5em}
    \begin{subfigure}{0.20\textwidth}
        \includegraphics[width=\linewidth]{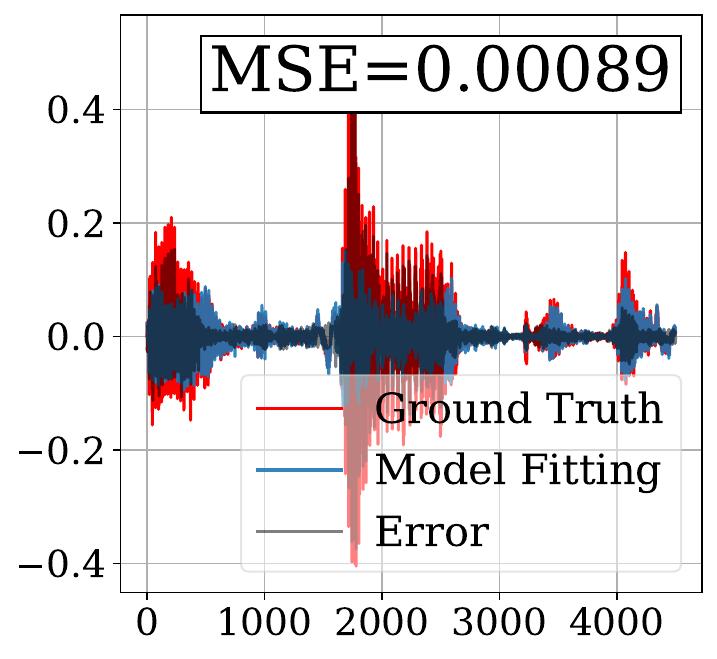}
        \caption{SPDER}
    \end{subfigure}\hspace{-0.5em}
    \begin{subfigure}{0.20\textwidth}
        \includegraphics[width=\linewidth]{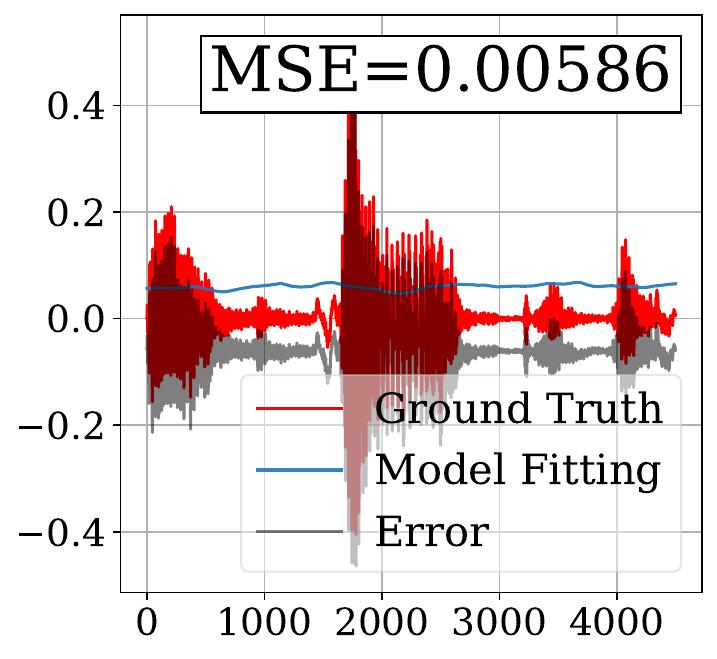}
        \caption{PE}
    \end{subfigure}\hspace{-0.5em}
    \begin{subfigure}{0.20\textwidth}
        \includegraphics[width=\linewidth]{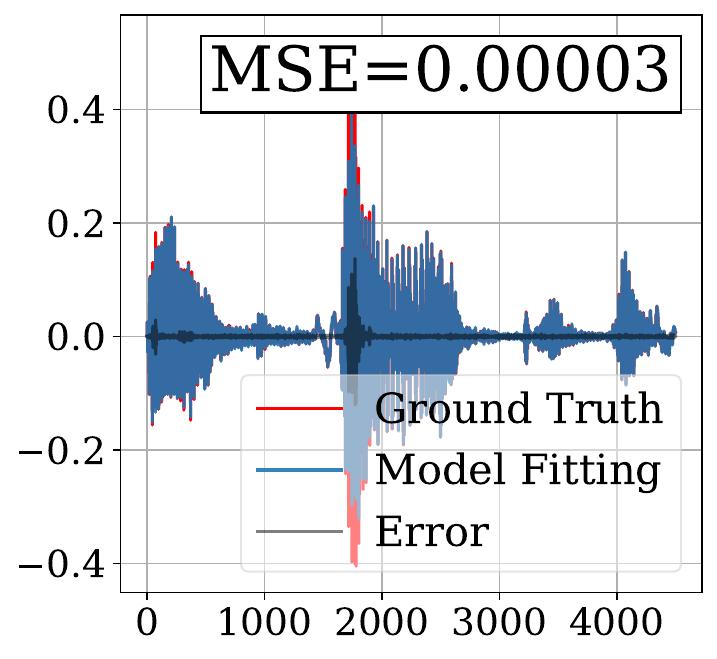}
        \caption{FM-SIREN}
    \end{subfigure}\hspace{-0.5em}
    \begin{subfigure}{0.20\textwidth}
        \includegraphics[width=\linewidth]{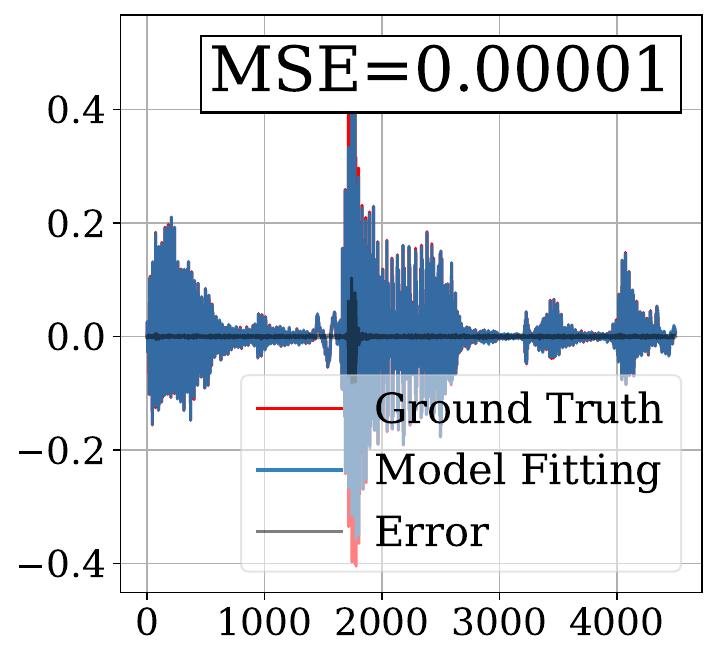}
        \caption{FM-FINER}
    \end{subfigure}

    \caption{Qualitative results of one-second audio reconstructions for the 1995 Pacific Grand Prix clip in the Spoken English Wikipedia dataset \cite{spokenwikipedia}, using two-layer networks with different approaches. The one-second MSE of each reconstruction is reported in the top-right corner of its subfigure. FM-SIREN and FM-FINER achieve visibly closer alignment to the ground truth and substantially lower MSE compared to the baselines. \colorbox{red}{\textcolor{white}{Red}}, \colorbox{customblue}{\textcolor{white}{medium blue}}, and \colorbox{black}{\textcolor{white}{black}} lines correspond to ground truth, reconstructed signal, and error signal, respectively.}
    \label{fig:audio-fit-main}
\end{figure}

 We evaluate audio reconstruction using two-layer models with $256$ neurons in each layer. We used the Spoken English Wikipedia dataset \cite{spokenwikipedia} as a reference, which contains 1,313 clips sampled at $f_\text{s} = 4\,\text{kHz}$, informing a Nyquist frequency of $f_\text{Nyquist} = 2\,\text{kHz}$,. For each clip, we fit the first $10$ seconds of audio and report performance in terms of MSE, averaged across all clips. As summarized in Table~\ref{fig:AudioMse}, FM-SIREN and FM-FINER achieve substantially lower errors than all baselines, and FM-FINER achieves the best overall performance. This demonstrates the effectiveness of Nyquist-informed frequency diversity in capturing fine-grained temporal structures. Figure~\ref{fig:audio-fit-main} provides qualitative results on a one-second segment of the 1995 Pacific Grand Prix clip, where our models exhibit a markedly lower reconstruction error compared to SIREN, FINER, and other baselines.

\begin{table}[h]
    \centering
    \caption{\label{fig:AudioMse} Average MSE for audio fitting on the Spoken English Wikipedia dataset \cite{spokenwikipedia}, comparing different INR models. \colorbox{Goldenrod}{Best} and \colorbox{Dandelion}{Second Best} results are highlighted.}
    \vspace{-5pt}
    \resizebox{1\textwidth}{!}{%
    \begin{tabular}{lccccccc|ccc}
        \toprule
        \textbf{Model} & FINER & Gauss & PE & SIREN & WIRE & SPDER & MIRE & \textbf{FM-SIREN} & \textbf{FM-FINER} \\
        \midrule
        \textbf{MSE ($\times 10^{-3})$} $\downarrow$ & $0.263$ & $5.948$ & $6.884$ & $0.513$ & $5.812$ & $1.476$ & $6.117$ & \colorbox{Dandelion}{${0.047}$} & \colorbox{Goldenrod}{${0.040}$} \\
        \textbf{Train Time (min)} & $13.69$ & $15.78$ & $14.32$ & $12.67$ & $36.57$ & $30.00$ & $73.47$ & $12.66$ & $13.69$ \\
        \bottomrule
    \end{tabular}%
    }
    \vspace{-10pt}
\end{table}


\subsection{Neural Radiance Fields}

\par
\begin{figure}[h]
    \centering
    \begin{subfigure}{0.19\textwidth}
        \includegraphics[width=\linewidth]{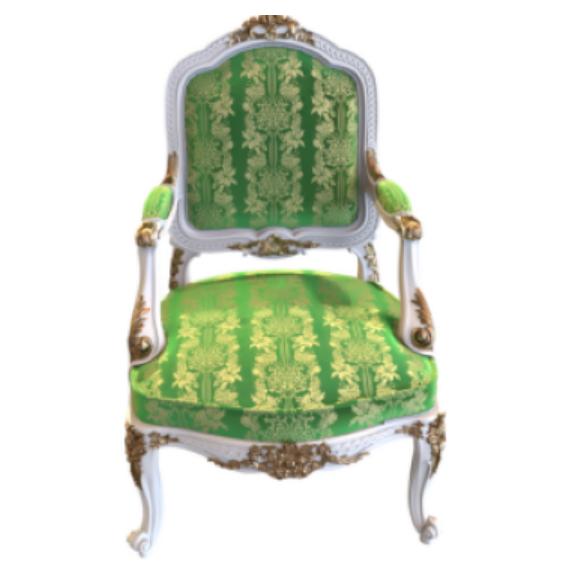}
        \caption{Ground Truth}
    \end{subfigure}
    \begin{subfigure}{0.19\textwidth}
        \includegraphics[width=\linewidth]{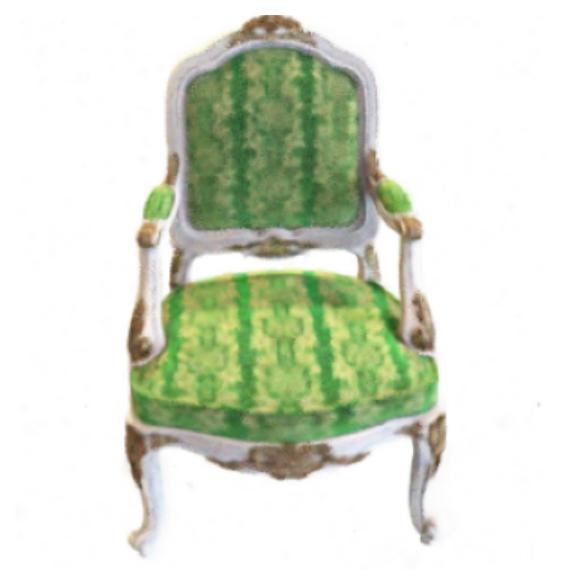}
        \caption{SIREN}
    \end{subfigure}
    \begin{subfigure}{0.19\textwidth}
        \includegraphics[width=\linewidth]{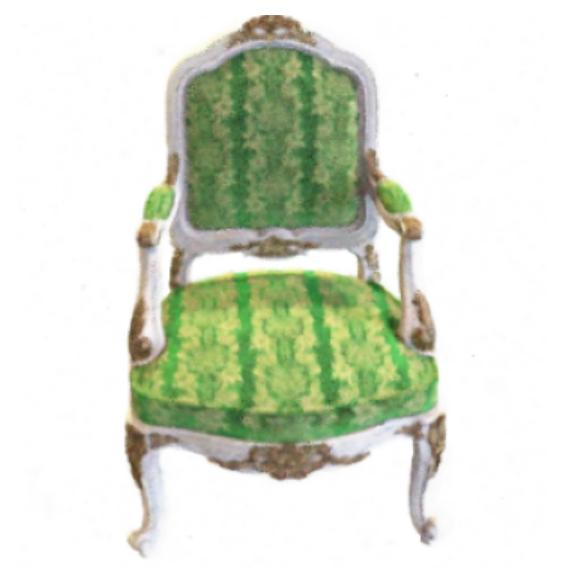}
        \caption{FINER}
    \end{subfigure}
    \begin{subfigure}{0.19\textwidth}
        \includegraphics[width=\linewidth]{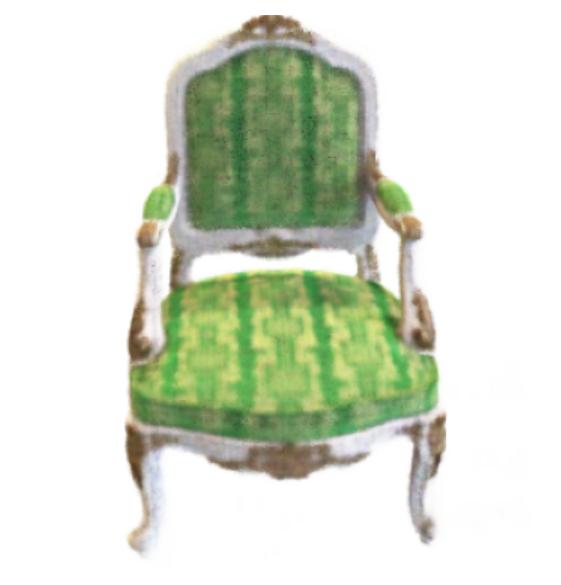}
        \caption{PE}
    \end{subfigure}
    \begin{subfigure}{0.19\textwidth}
        \includegraphics[width=\linewidth]{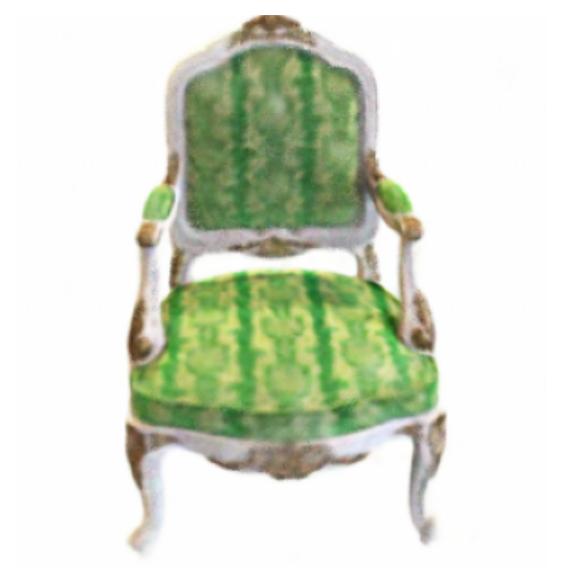}
        \caption{Gauss}
    \end{subfigure}
    \\
    \begin{subfigure}{0.19\textwidth}
        \includegraphics[width=\linewidth]{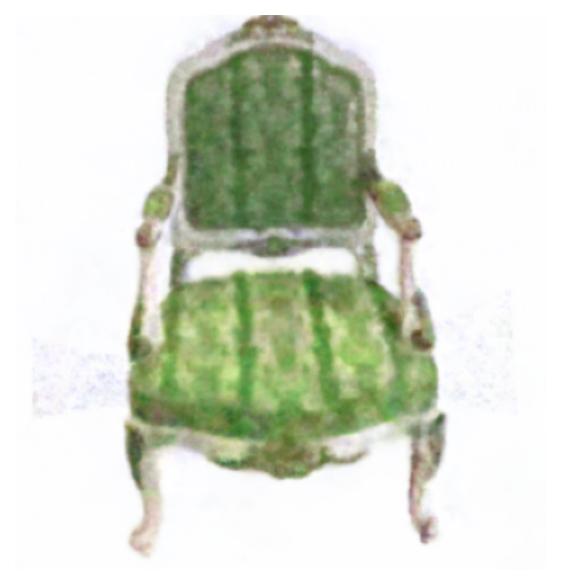}
        \caption{WIRE}
    \end{subfigure}
    \begin{subfigure}{0.19\textwidth}
        \includegraphics[width=\linewidth]{figures_jpg/experimental_results/NeRF/SIREN_NeRF_chair.jpg}
        \caption{FreSh}
    \end{subfigure}
    \begin{subfigure}{0.19\textwidth}
        \includegraphics[width=\linewidth]{figures_jpg/experimental_results/NeRF/Gauss_NeRF_chair.jpg}
        \caption{MIRE}
    \end{subfigure}
    \begin{subfigure}{0.19\textwidth}
        \includegraphics[width=\linewidth]{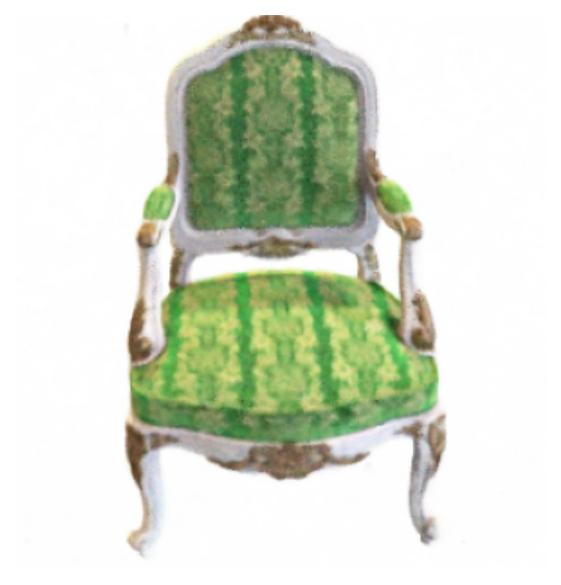}
        \caption{FM-SIREN}
    \end{subfigure}
    \begin{subfigure}{0.19\textwidth}
        \includegraphics[width=\linewidth]{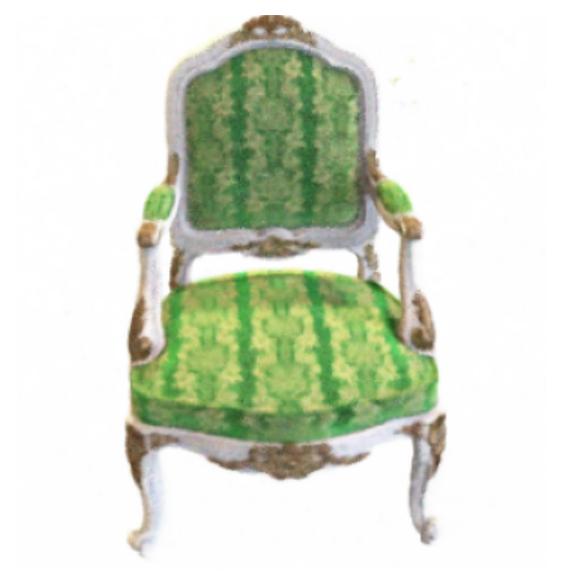}
        \caption{FM-FINER}
    \end{subfigure}
    \vspace{-5pt}
    \caption{Qualitative reconstruction results of the Chair scene from the Blender dataset \cite{mildenhall2021nerf}, using five-layer networks (three layers for the first block, one layer for density, and one layer for color). FM-SIREN and FM-FINER achieve improvements over the baselines, with exact PSNR values reported in Table~\ref{tab:NeRF}.}
    \vspace{-10pt}
    \label{fig:NeRF_main}
\end{figure}

 NeRF synthesizes novel 3D scene views from sparse 2D images by learning a continuous volumetric representation that is optimized via ray-based rendering \cite{mildenhall2020nerfrepresentingscenesneural}. In our experiments, the volume is discretized at a resolution of $100 \times 100 \times 100$, corresponding to a Nyquist frequency of $50$ cycles/volume. We adopt the concise implementation of \cite{vandegar2025nerf100lines} on the Blender dataset \cite{mildenhall2021nerf}, using five-layer networks that model density and color. Performance is evaluated as the average PSNR across all 200 test images per scene, each at a resolution of $400 \times 400$. As reported in Table~\ref{tab:NeRF} and illustrated in Figure~\ref{fig:NeRF_main}, FM-SIREN and FM-FINER achieve higher PSNR than baseline models in most scenes.

\begin{table}[h!]
    \centering
    \caption{Train time and PSNR for NeRF fitting on the Blender dataset \cite{mildenhall2021nerf}. \colorbox{Goldenrod}{Best} and \colorbox{Dandelion}{Second Best} results are highlighted.}
    \vspace{-5pt}
    \resizebox{1\textwidth}{!}{%
    \begin{tabular}{@{}lc cccccccc@{}}
        \toprule
        \multirow{2}{*}{\textbf{Model}} & 
        \multirow{2}{*}{\textbf{Train Time (min)}} & 
        \multicolumn{8}{c}{\textbf{PSNR} $\uparrow$} \\
        \cmidrule(lr){3-10}
         & & \textbf{Lego} & \textbf{Ship} & \textbf{Chair} & \textbf{Mic} & \textbf{Materials} & \textbf{Hotdog} & \textbf{Ficus} & \textbf{Drums} \\
        \midrule
        FINER & $227.92$ & \colorbox{Dandelion}{$28.01$} & $33.09$ & $34.53$ & $34.44$ & $33.38$ & $37.16$ & $33.75$ & $32.51$ \\
        FreSh & $247.89$ & $27.99$ & $33.11$ & $34.50$ & $34.44$ & $33.35$ & $37.32$ & $33.52$ & $32.52$ \\
        Gauss & $210.52$ & $27.58$ & $32.91$ & $34.23$ & \colorbox{Dandelion}{$34.48$} & $33.21$ & $36.66$ & $33.33$ & $32.31$ \\
        MIRE & $448.96$ & $25.95$ & $28.82$ & $33.77$ & $30.82$ & $29.57$ & $33.63$ & $27.41$ & $29.67$ \\
        PE & $269.50$ & $27.63$ & $32.48$ & $34.13$ & $34.24$ & $32.86$ & $36.17$ & $32.82$ & $31.72$ \\
        SIREN & $260.52$ & $27.81$ & $33.08$ & $34.42$ & $34.45$ & $33.38$ & $37.00$ & \colorbox{Goldenrod}{$33.79$} & $32.48$ \\
        WIRE & $298.75$ & $23.36$ & $20.34$ & $25.21$ & $26.31$ & $23.07$ & $27.50$ & $23.19$ & $21.11$ \\
        \midrule
        \textbf{FM-SIREN} & 222.66 & $27.55$ & \colorbox{Goldenrod}{$33.17$} & \colorbox{Dandelion}{$34.55$} & \colorbox{Goldenrod}{$34.64$} & \colorbox{Goldenrod}{$33.45$} & \colorbox{Dandelion}{$37.37$} & \colorbox{Dandelion}{$33.76$} & \colorbox{Goldenrod}{$32.54$} \\
        \textbf{FM-FINER} & $189.34$ & \colorbox{Goldenrod}{$28.11$} & \colorbox{Dandelion}{$33.14$} & \colorbox{Goldenrod}{$34.56$} & $29.81$ & \colorbox{Dandelion}{$33.41$} & \colorbox{Goldenrod}{$37.41$} & $33.73$ & \colorbox{Dandelion}{$32.53$} \\
        \bottomrule
    \end{tabular}%
    }
    \label{tab:NeRF}
\end{table}
\section{Hyperparameters}

All experiments share the same model-specific hyperparameters recommended in each baseline's reference with an exception for $\beta_0$. We used the best performing $\beta_0$ for audio fitting since not all references presented audio fitting experiments. Table \ref{tab:audio_hyperparameters} presents global, model-specific hyperparameters, whereas \ref{tab:experiments_hyperparameters} presents experiment-specific hyperparameters.


\begin{table}[h]
\centering
\caption{Model-specific hyperparameters for the all experiments.}
\label{tab:audio_hyperparameters}
\resizebox{\textwidth}{!}{%
\begin{tabular}{@{}lccccccccccc@{}}
\toprule
\textbf{Parameter} & \textbf{SIREN} & \textbf{FINER} & \textbf{WIRE} & \textbf{Gauss} & \textbf{PE} & \textbf{MIRE} & \textbf{FreSh} & \textbf{TUNER} & \textbf{SPDER} & \textbf{FM-SIREN} & \textbf{FM-FINER} \\
\midrule
First $\beta_0$      & 30.0 & 30.0 & 16 & - & - & 30.0 & - & - & 30.0 & as defined & as defined \\
Hidden $\beta_0$     & 30.0 & 30.0 & 16 & - & - & 30.0 & - & 30.0 & 30.0 & as defined & as defined \\
Outermost Linear     & \checkmark & - & - & - & - & - & \checkmark & \checkmark & \checkmark & \checkmark & - \\
Bias                 & \checkmark & \checkmark & \checkmark & \checkmark & \checkmark & \checkmark & \checkmark & \checkmark & \checkmark & \checkmark & \checkmark \\
Scale                & - & - & 8 & 16 & - & - & - & - & - & - & - \\
Embedding Size       & - & - & - & - & 256 & - & - & - & - & - & - \\
Frequency Shifting      & - & - & - & - & - & - & \checkmark & - & - & - & - \\
Nyquist Divisor      & - & - & - & - & - & - & - & - & - & 1 & 2 \\
\bottomrule
\end{tabular}%
}
\end{table}

\begin{table}[h]
\centering
\caption{Experiment-specific hyperparameters.}
\label{tab:experiments_hyperparameters}
\begin{tabular}{@{}lccccc@{}}
\toprule
\textbf{Parameter} & \textbf{Audio} & \textbf{Image} & \textbf{3D Shape} & \textbf{Video} & \textbf{NeRF} \\
\midrule
FINER $\beta_0$          & $700$ & $30$ & $30$ & $30$ & $30$ \\
SIREN $\beta_0$          & $800$ & $30$ & $30$ & $30$ & $30$ \\
Learning Rate            & $0.0001$ & $0.001$ & $0.0001$ & $0.0001$ & $0.0001$ \\
Number of Hidden Layers  & $2$ & $2$ & $3$ & $3$ & $5$ \\
Number of Epochs         & $500$ & $500$ & $75$ & $2000$ & $16$ \\
\bottomrule
\end{tabular}
\end{table}





\section{Additional Results}

\begin{figure}[t]
    \centering
    \begin{subfigure}{0.20\textwidth}
        \includegraphics[width=\linewidth]{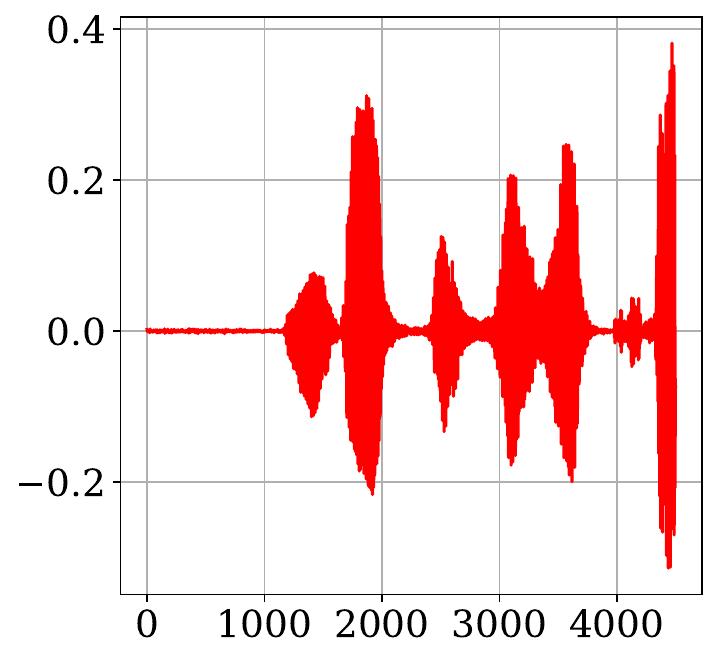}
        \caption{Ground Truth}
    \end{subfigure}
    \hspace{-0.5em}
    \begin{subfigure}{0.20\textwidth}
        \includegraphics[width=\linewidth]{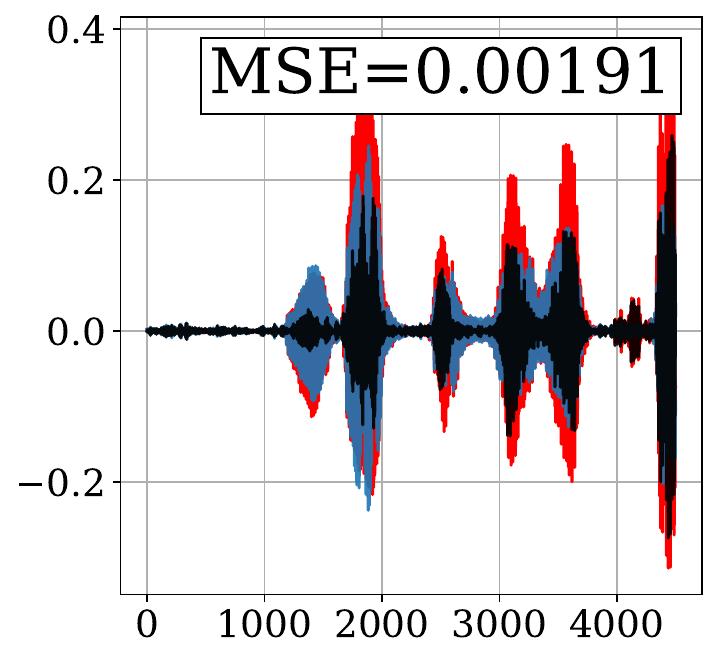}
        \caption{SIREN}
    \end{subfigure}\hspace{-0.5em}
    \begin{subfigure}{0.20\textwidth}
        \includegraphics[width=\linewidth]{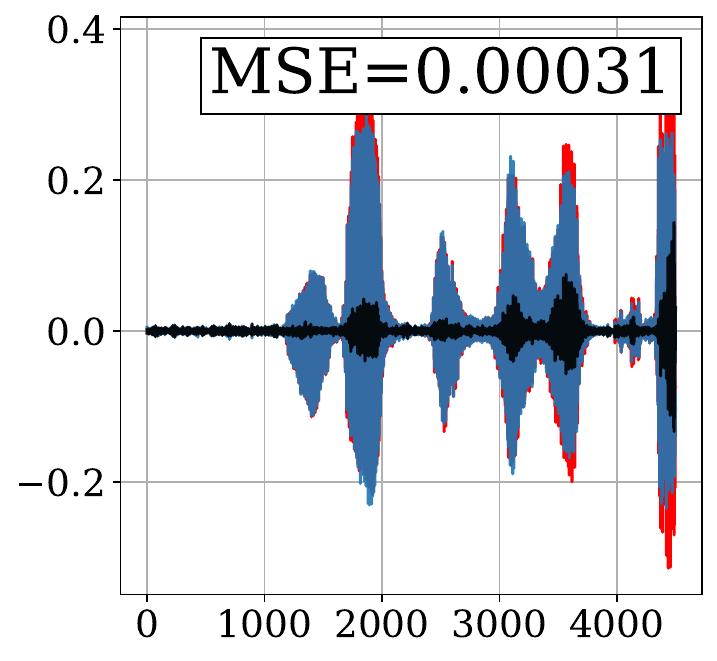}
        \caption{FINER}
    \end{subfigure}\hspace{-0.5em}
    \begin{subfigure}{0.20\textwidth}
        \includegraphics[width=\linewidth]{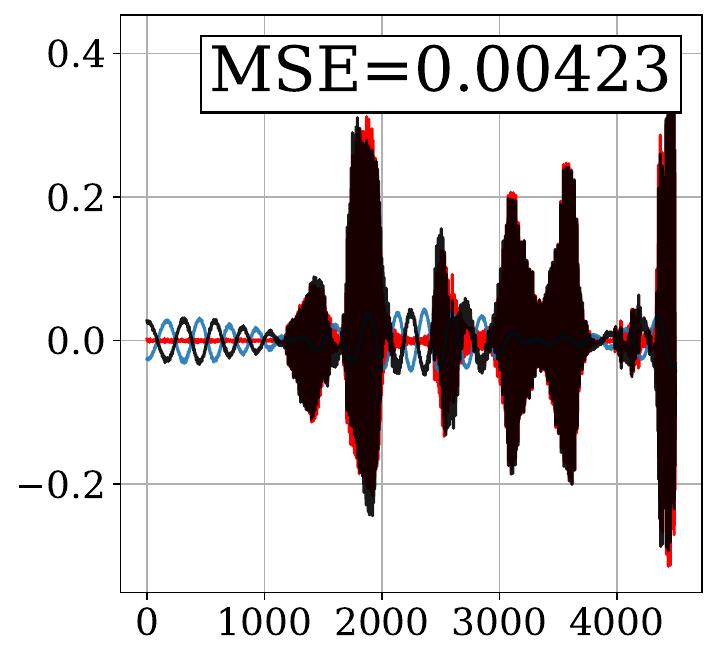}
        \caption{WIRE}
    \end{subfigure}\hspace{-0.5em}
    \begin{subfigure}{0.20\textwidth}
        \includegraphics[width=\linewidth]{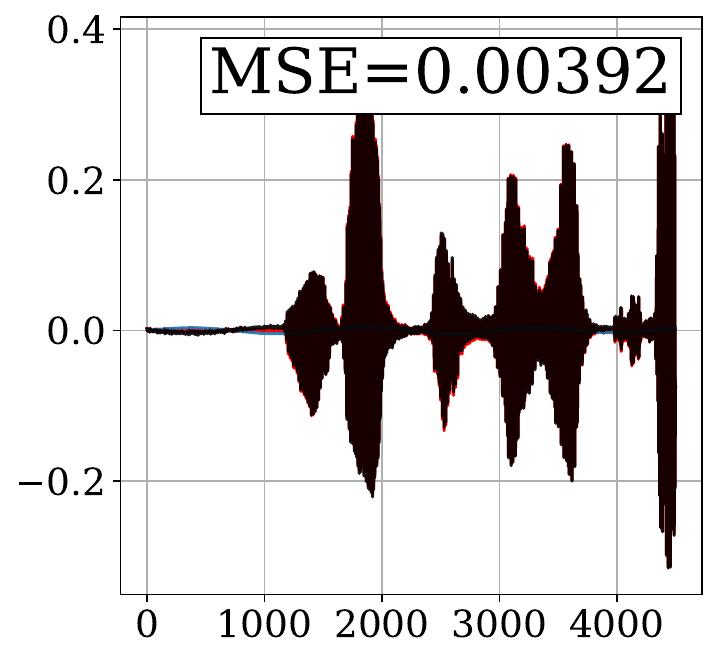}
        \caption{Gauss}
    \end{subfigure}

    \vspace{0.5em}

    \begin{subfigure}{0.20\textwidth}
        \includegraphics[width=\linewidth]{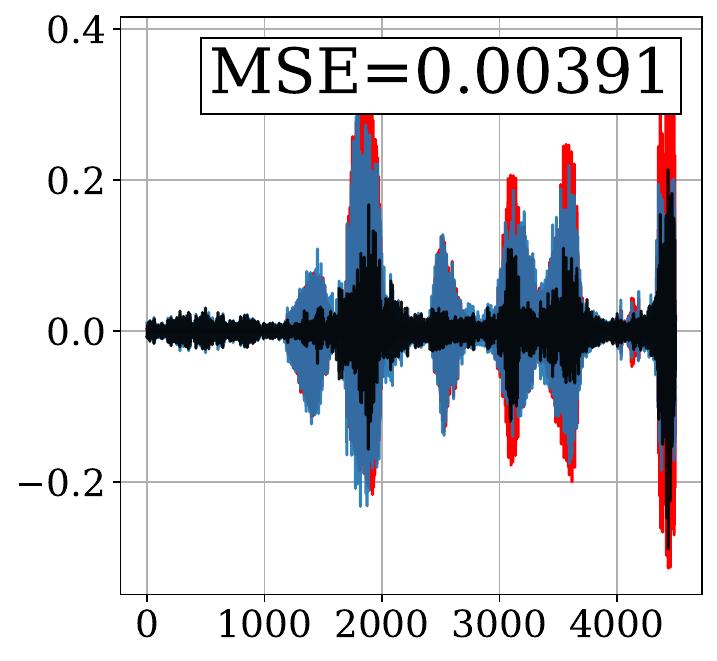}
        \caption{MIRE}
    \end{subfigure}\hspace{-0.5em}
    \begin{subfigure}{0.20\textwidth}
        \includegraphics[width=\linewidth]{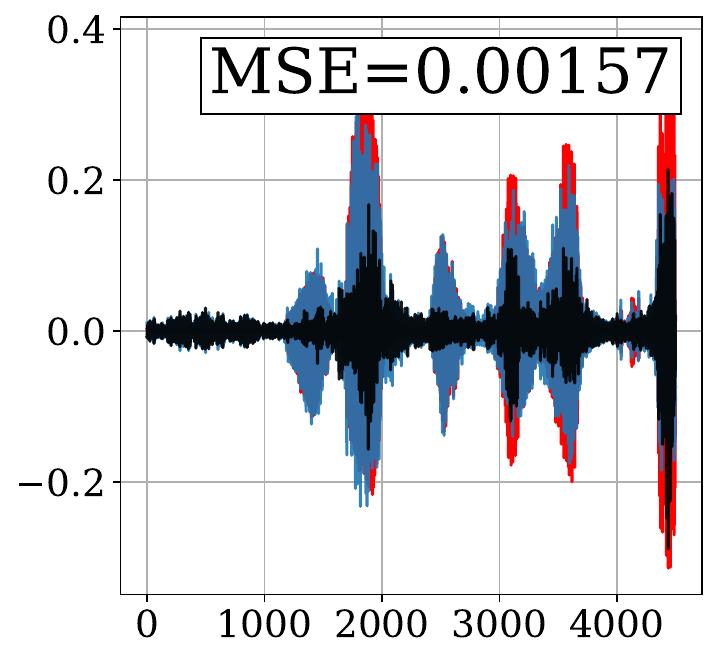}
        \caption{SPDER}
    \end{subfigure}\hspace{-0.5em}
    \begin{subfigure}{0.20\textwidth}
        \includegraphics[width=\linewidth]{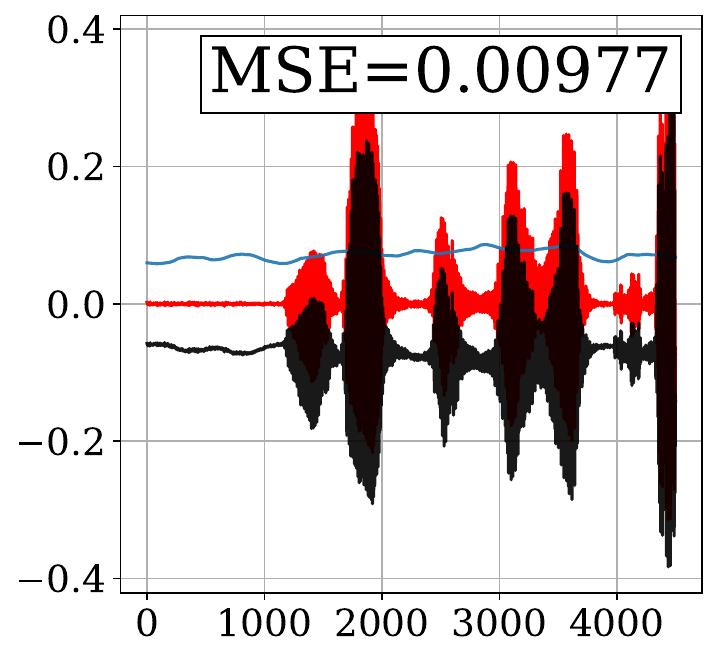}
        \caption{PE}
    \end{subfigure}\hspace{-0.5em}
    \begin{subfigure}{0.20\textwidth}
        \includegraphics[width=\linewidth]{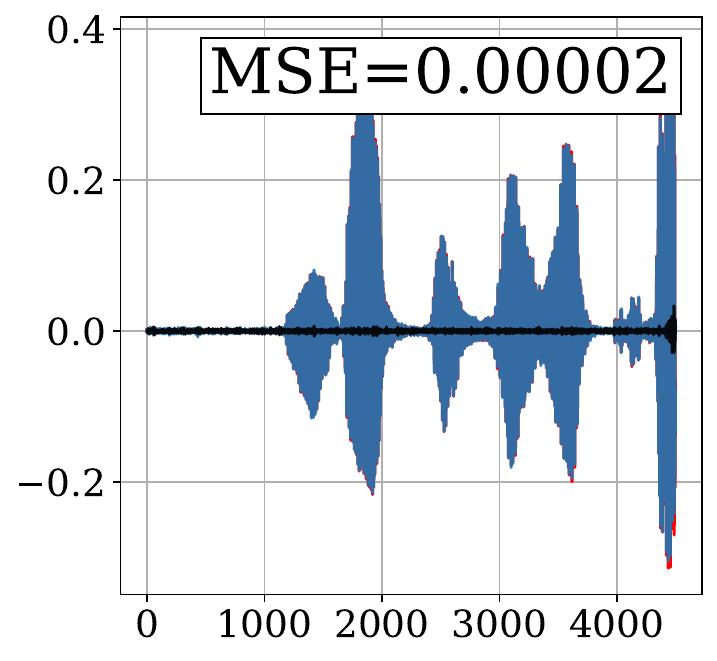}
        \caption{FM-SIREN}
    \end{subfigure}\hspace{-0.5em}
    \begin{subfigure}{0.20\textwidth}
        \includegraphics[width=\linewidth]{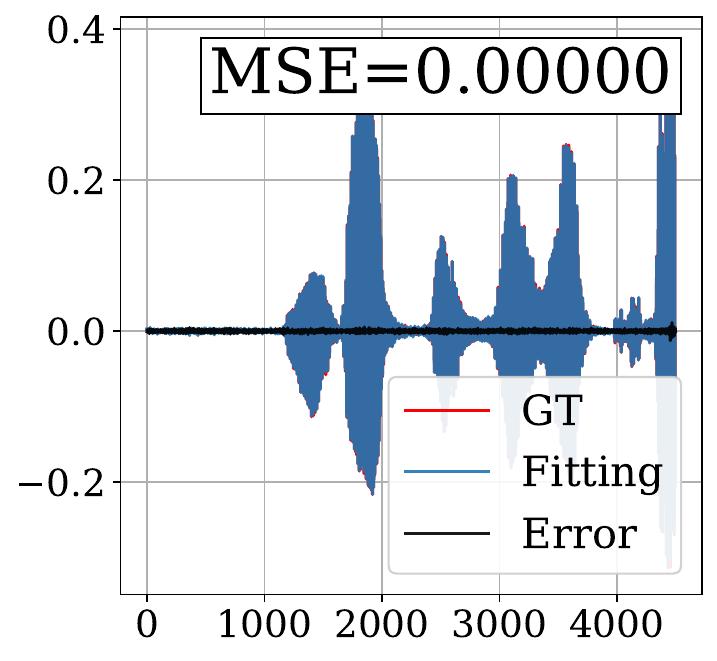}
        \caption{FM-FINER}
    \end{subfigure}

    \caption{Qualitative results of one-second audio reconstructions for the Alleyway clip in the Spoken English Wikipedia dataset \cite{spokenwikipedia}, using two-layer networks with different approaches. \colorbox{red}{\textcolor{white}{Red}}, \colorbox{customblue}{\textcolor{white}{medium blue}}, and \colorbox{black}{\textcolor{white}{black}} lines correspond to ground truth, reconstructed signal, and error signal, respectively. FM-SIREN and FM-FINER demonstrate superior performance in terms of MSE, as shown on the top-right of each subfigure.}
    \label{fig:audio-fit-1}
\end{figure}

\begin{figure}[t]
    \centering
    \begin{subfigure}{0.20\textwidth}
        \includegraphics[width=\linewidth]{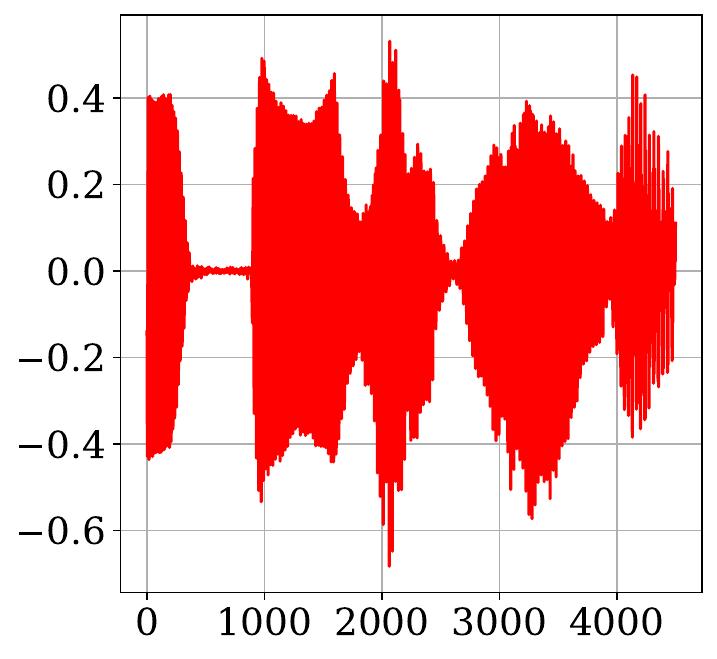}
        \caption{Ground Truth}
    \end{subfigure}
    \hspace{-0.5em}
    \begin{subfigure}{0.20\textwidth}
        \includegraphics[width=\linewidth]{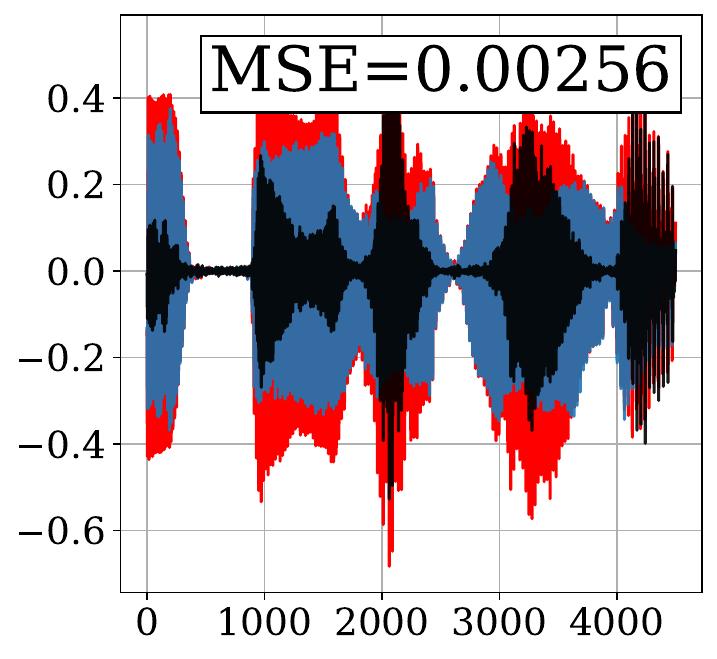}
        \caption{SIREN}
    \end{subfigure}\hspace{-0.5em}
    \begin{subfigure}{0.20\textwidth}
        \includegraphics[width=\linewidth]{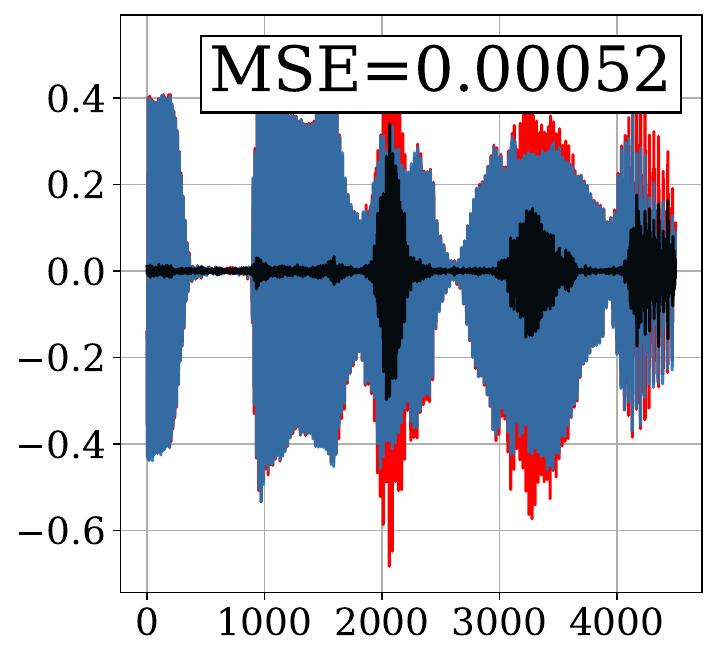}
        \caption{FINER}
    \end{subfigure}\hspace{-0.5em}
    \begin{subfigure}{0.20\textwidth}
        \includegraphics[width=\linewidth]{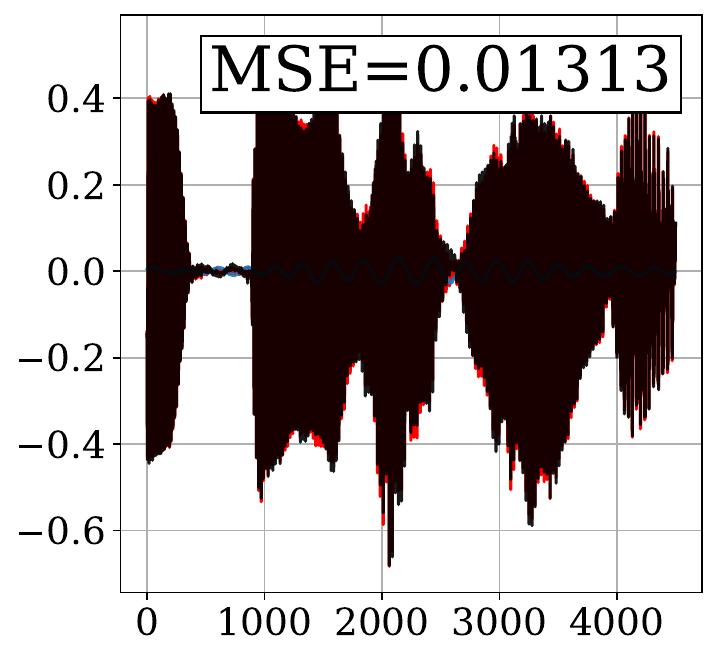}
        \caption{WIRE}
    \end{subfigure}\hspace{-0.5em}
    \begin{subfigure}{0.20\textwidth}
        \includegraphics[width=\linewidth]{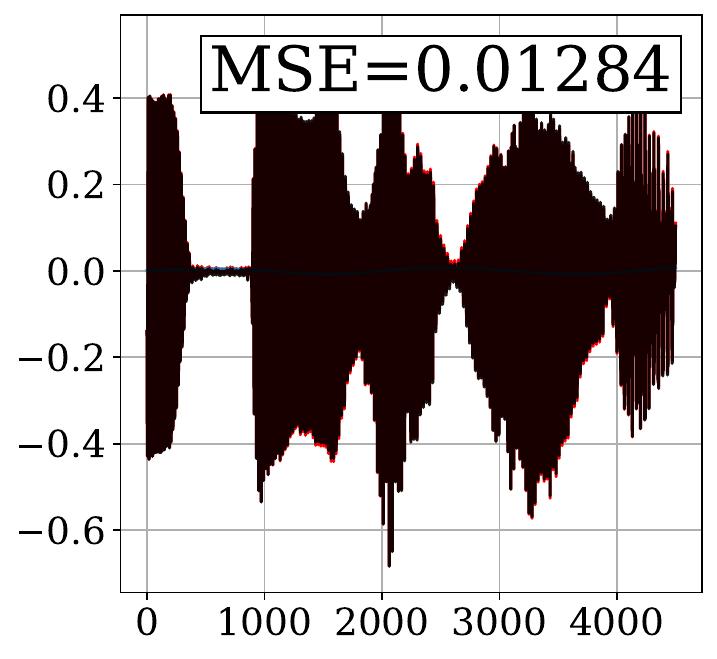}
        \caption{Gauss}
    \end{subfigure}

    \vspace{0.5em}

    \begin{subfigure}{0.20\textwidth}
        \includegraphics[width=\linewidth]{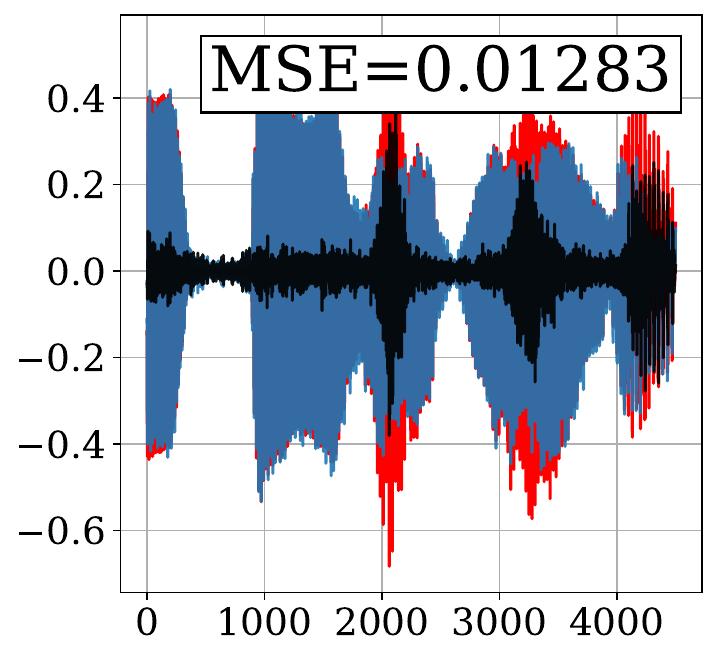}
        \caption{MIRE}
    \end{subfigure}\hspace{-0.5em}
    \begin{subfigure}{0.20\textwidth}
        \includegraphics[width=\linewidth]{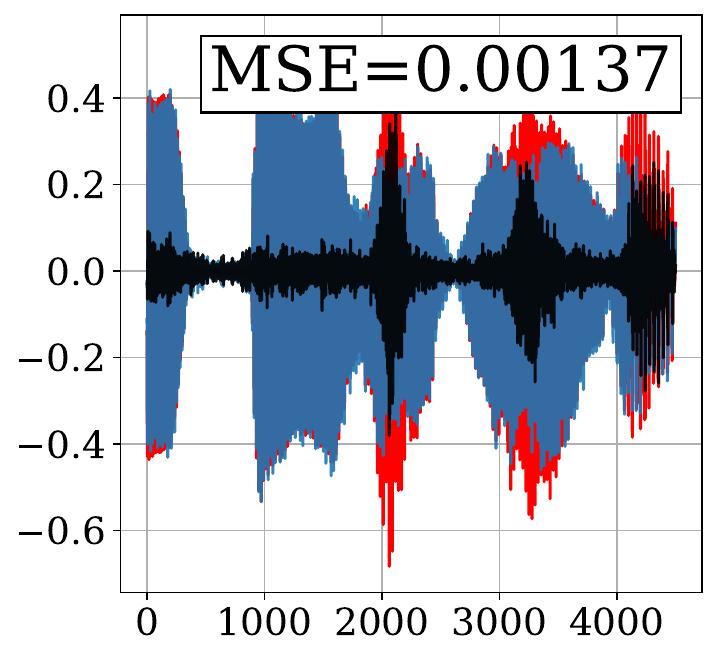}
        \caption{SPDER}
    \end{subfigure}\hspace{-0.5em}
    \begin{subfigure}{0.20\textwidth}
        \includegraphics[width=\linewidth]{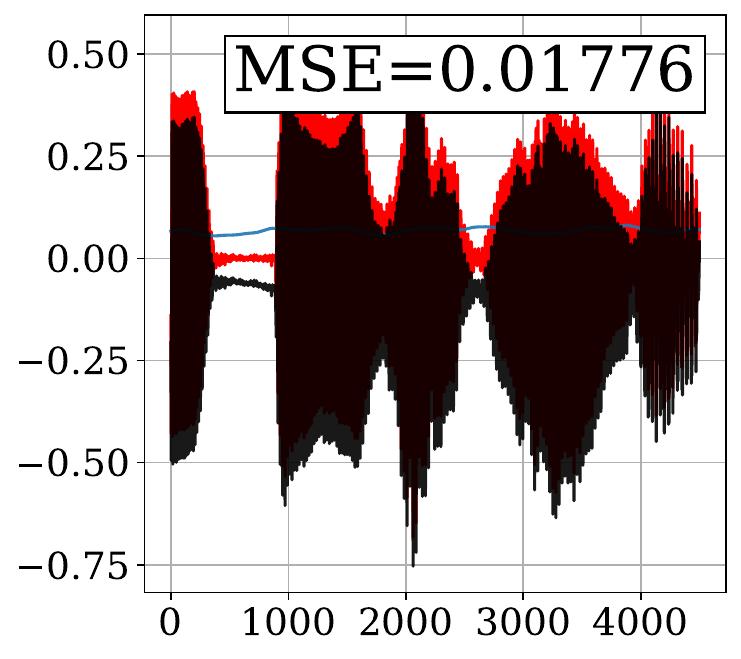}
        \caption{PE}
    \end{subfigure}\hspace{-0.5em}
    \begin{subfigure}{0.20\textwidth}
        \includegraphics[width=\linewidth]{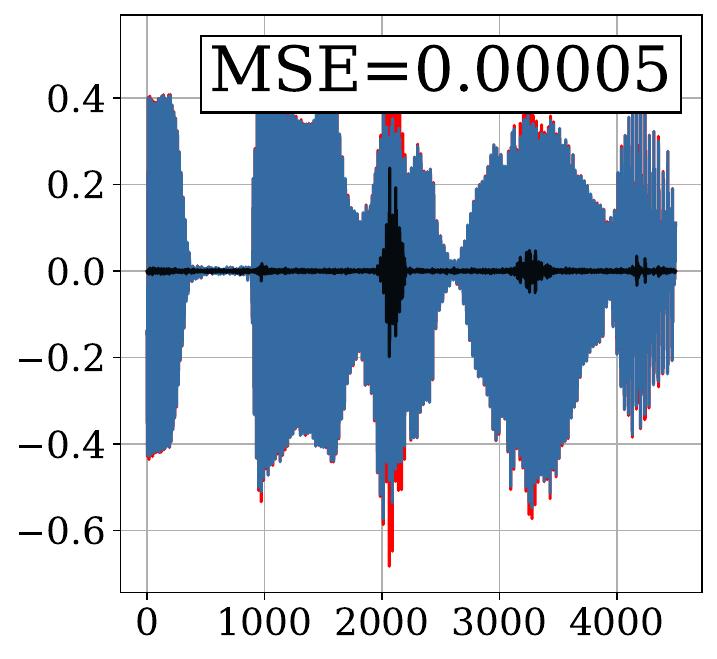}
        \caption{FM-SIREN}
    \end{subfigure}\hspace{-0.5em}
    \begin{subfigure}{0.20\textwidth}
        \includegraphics[width=\linewidth]{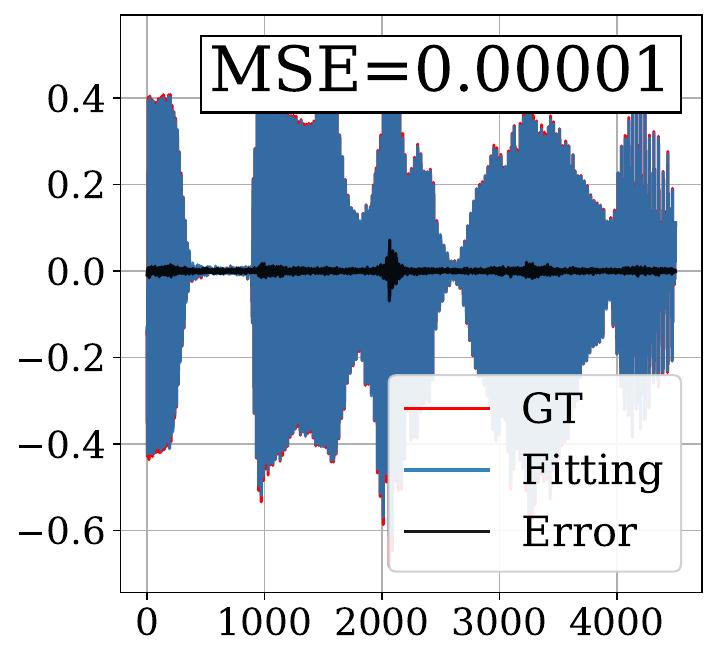}
        \caption{FM-FINER}
    \end{subfigure}

    \caption{Qualitative results of one-second audio reconstructions for the BBC Radio 1 clip in the Spoken English Wikipedia dataset \cite{spokenwikipedia}, using two-layer networks with different approaches. \colorbox{red}{\textcolor{white}{Red}}, \colorbox{customblue}{\textcolor{white}{medium blue}}, and \colorbox{black}{\textcolor{white}{black}} lines correspond to ground truth, reconstructed signal, and error signal, respectively. FM-SIREN and FM-FINER demonstrate superior performance in terms of MSE, as shown on the top-right of each subfigure.}
    \label{fig:audio-fit-2}
\end{figure}

\begin{figure}[t]
    \centering
    \begin{subfigure}{0.20\textwidth}
        \includegraphics[width=\linewidth]{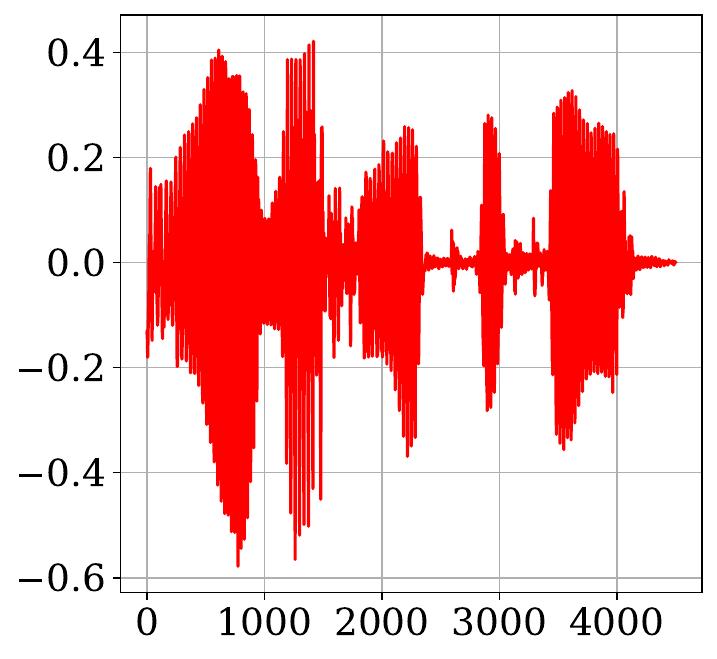}
        \caption{Ground Truth}
    \end{subfigure}
    \hspace{-0.5em}
    \begin{subfigure}{0.20\textwidth}
        \includegraphics[width=\linewidth]{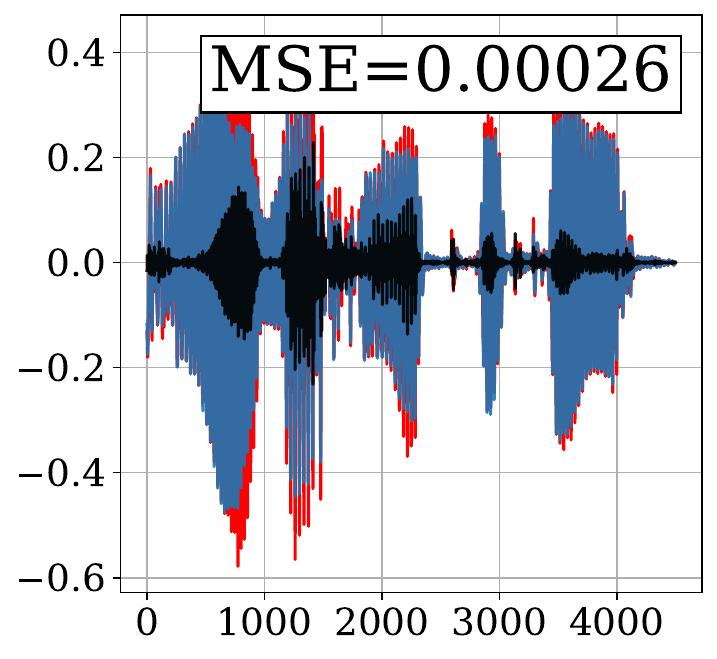}
        \caption{SIREN}
    \end{subfigure}\hspace{-0.5em}
    \begin{subfigure}{0.20\textwidth}
        \includegraphics[width=\linewidth]{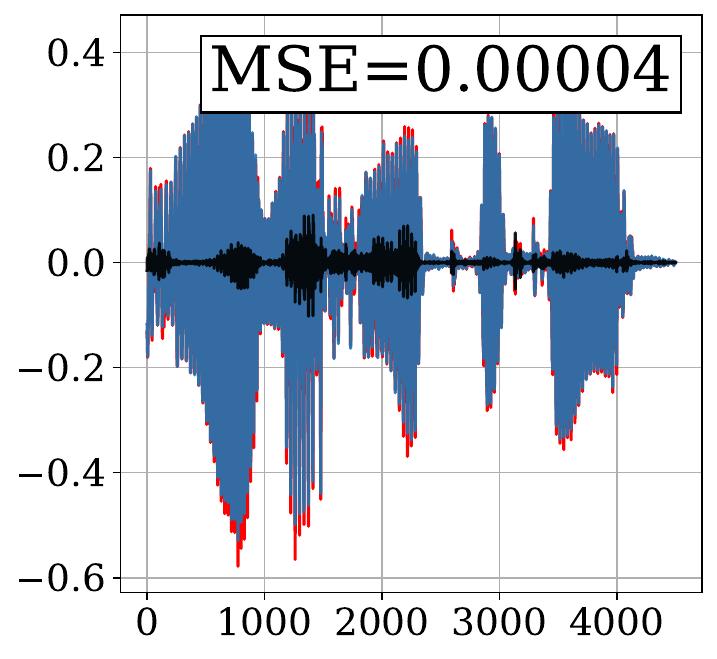}
        \caption{FINER}
    \end{subfigure}\hspace{-0.5em}
    \begin{subfigure}{0.20\textwidth}
        \includegraphics[width=\linewidth]{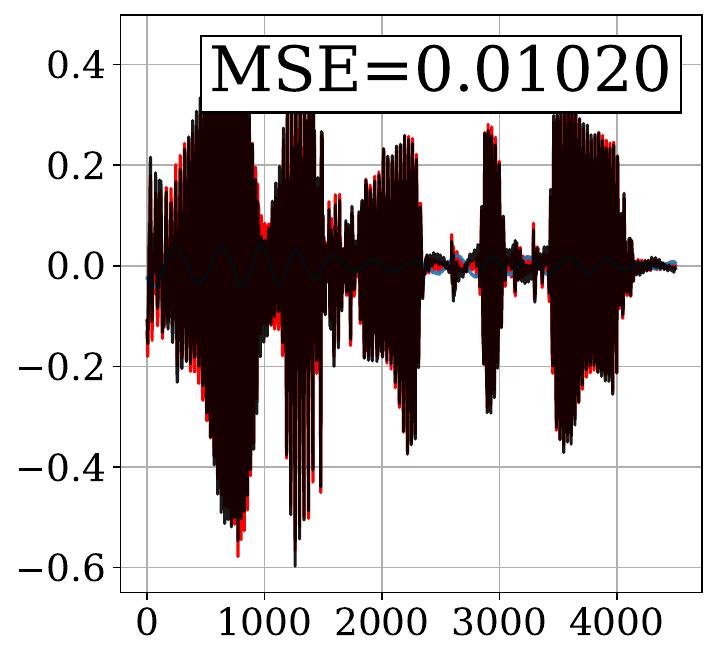}
        \caption{WIRE}
    \end{subfigure}\hspace{-0.5em}
    \begin{subfigure}{0.20\textwidth}
        \includegraphics[width=\linewidth]{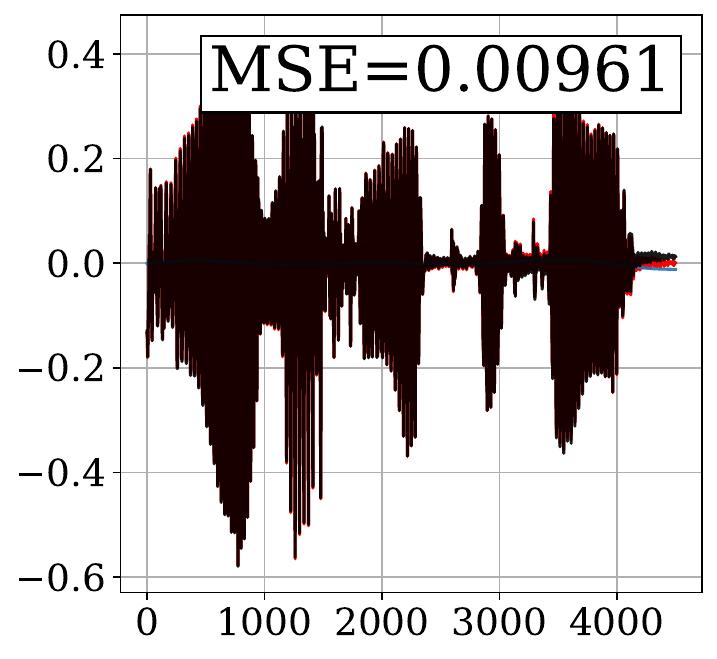}
        \caption{Gauss}
    \end{subfigure}

    \vspace{0.5em}

    \begin{subfigure}{0.20\textwidth}
        \includegraphics[width=\linewidth]{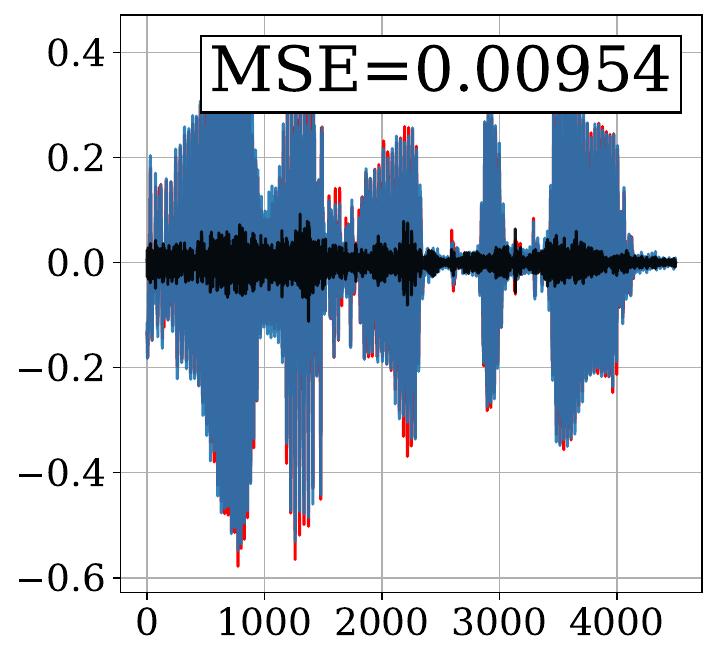}
        \caption{MIRE}
    \end{subfigure}\hspace{-0.5em}
    \begin{subfigure}{0.20\textwidth}
        \includegraphics[width=\linewidth]{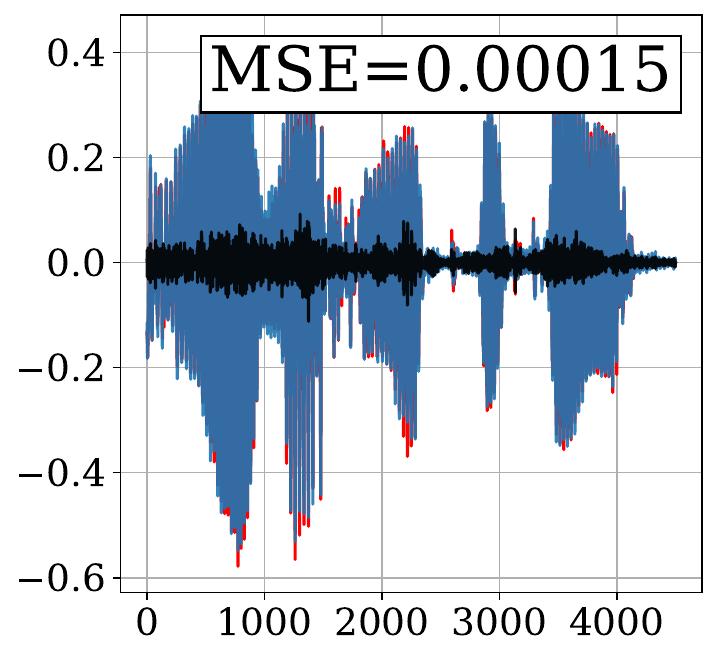}
        \caption{SPDER}
    \end{subfigure}\hspace{-0.5em}
    \begin{subfigure}{0.20\textwidth}
        \includegraphics[width=\linewidth]{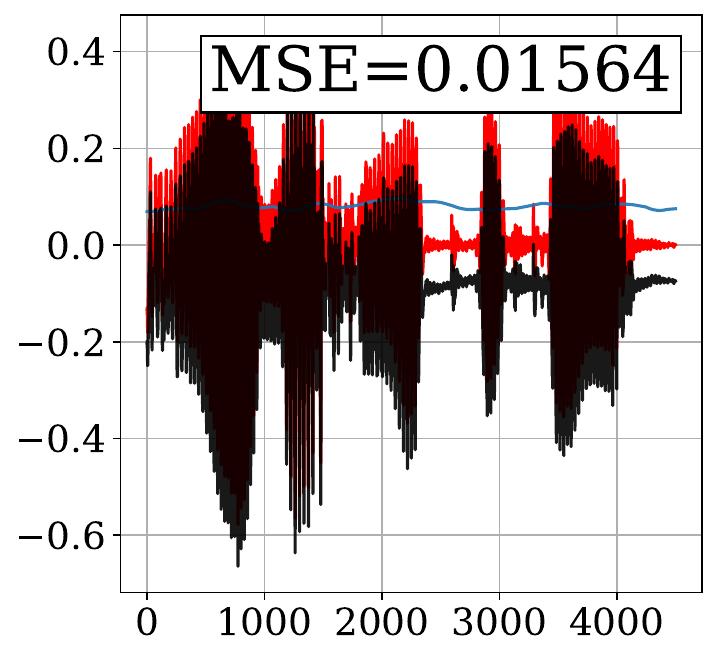}
        \caption{PE}
    \end{subfigure}\hspace{-0.5em}
    \begin{subfigure}{0.20\textwidth}
        \includegraphics[width=\linewidth]{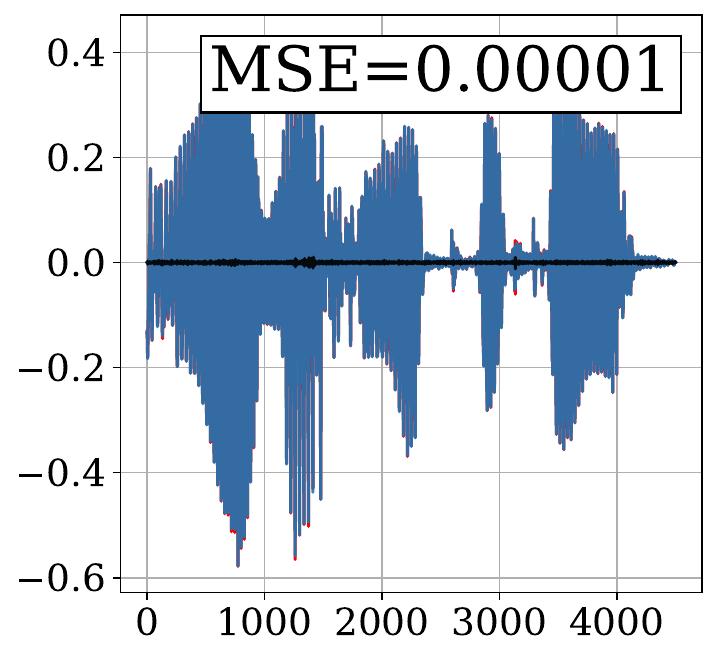}
        \caption{FM-SIREN}
    \end{subfigure}\hspace{-0.5em}
    \begin{subfigure}{0.20\textwidth}
        \includegraphics[width=\linewidth]{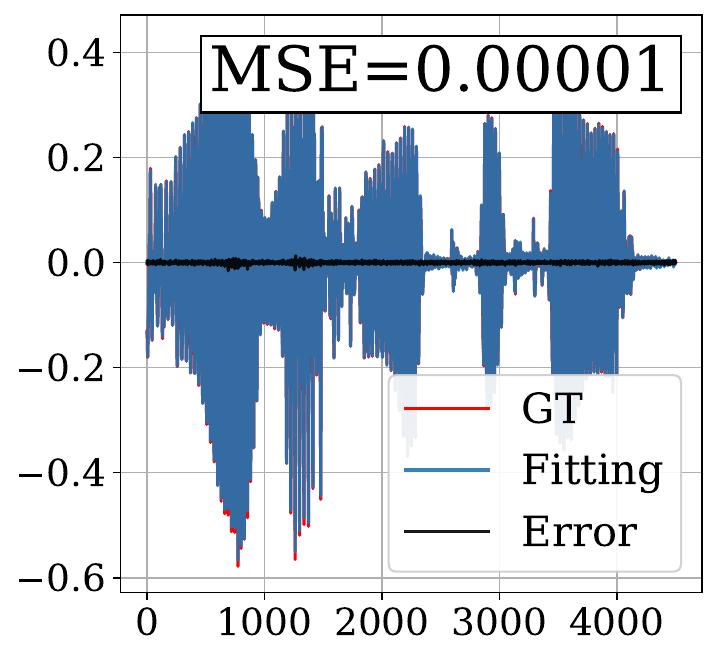}
        \caption{FM-FINER}
    \end{subfigure}

    \caption{Qualitative results of one-second audio reconstructions for the Munich clip in the Spoken English Wikipedia dataset \cite{spokenwikipedia}, using two-layer networks with different approaches. \colorbox{red}{\textcolor{white}{Red}}, \colorbox{customblue}{\textcolor{white}{medium blue}}, and \colorbox{black}{\textcolor{white}{black}} lines correspond to ground truth, reconstructed signal, and error signal, respectively. FM-SIREN and FM-FINER demonstrate superior performance in terms of MSE, as shown on the top-right of each subfigure.}
    \label{fig:audio-fit-3}
\end{figure}

\begin{figure}[t]
    \centering
    \begin{subfigure}{0.20\textwidth}
        \includegraphics[width=\linewidth]{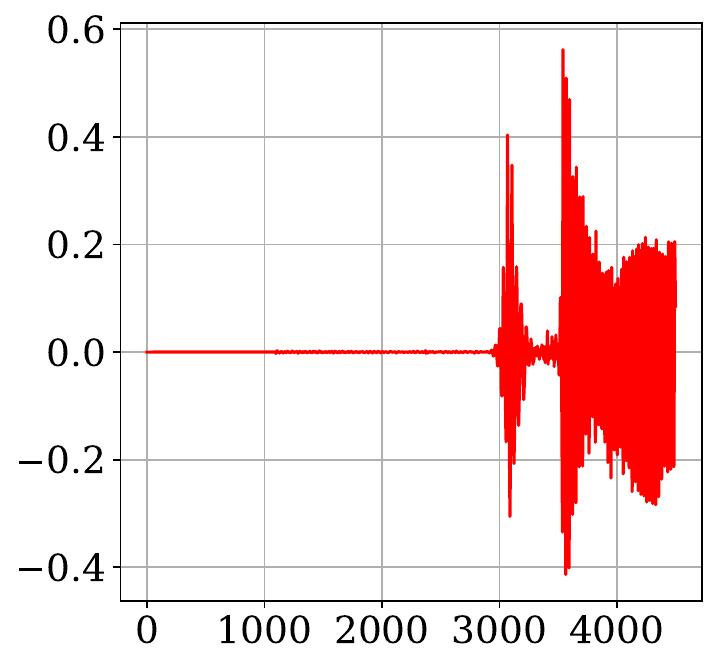}
        \caption{Ground Truth}
    \end{subfigure}
    \hspace{-0.5em}
    \begin{subfigure}{0.20\textwidth}
        \includegraphics[width=\linewidth]{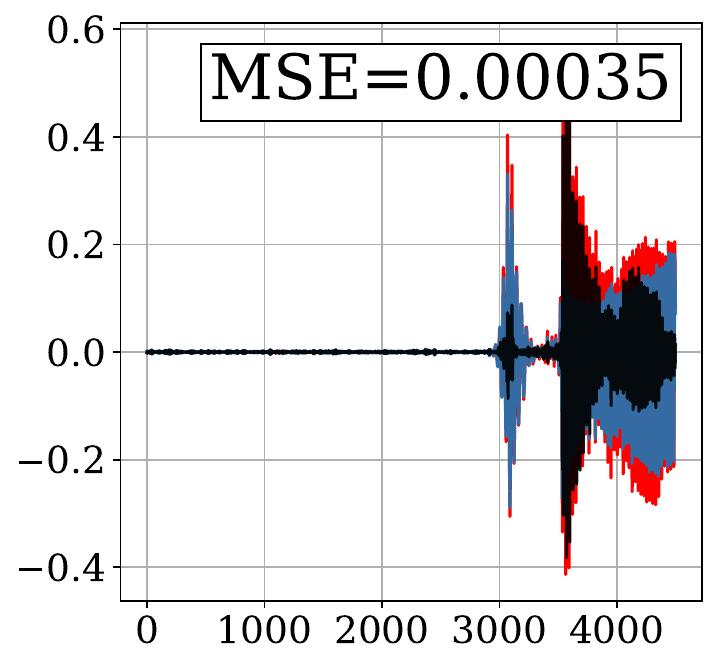}
        \caption{SIREN}
    \end{subfigure}\hspace{-0.5em}
    \begin{subfigure}{0.20\textwidth}
        \includegraphics[width=\linewidth]{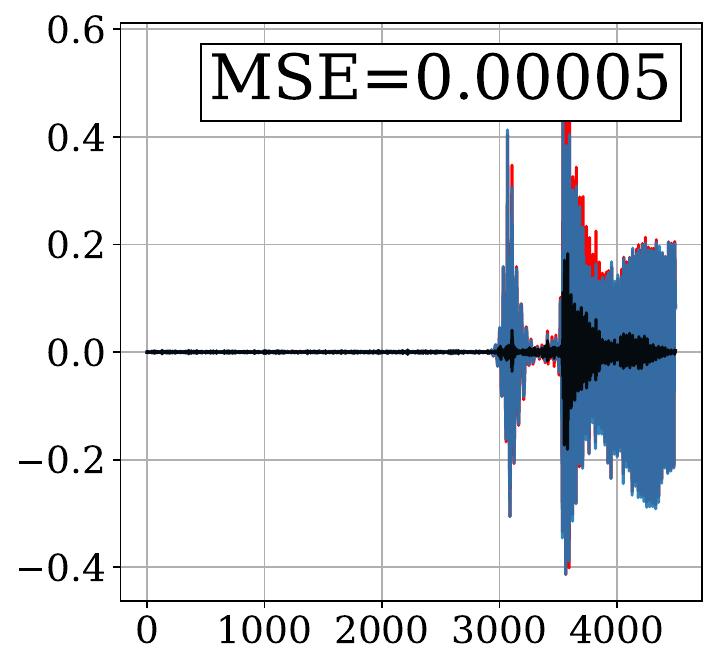}
        \caption{FINER}
    \end{subfigure}\hspace{-0.5em}
    \begin{subfigure}{0.20\textwidth}
        \includegraphics[width=\linewidth]{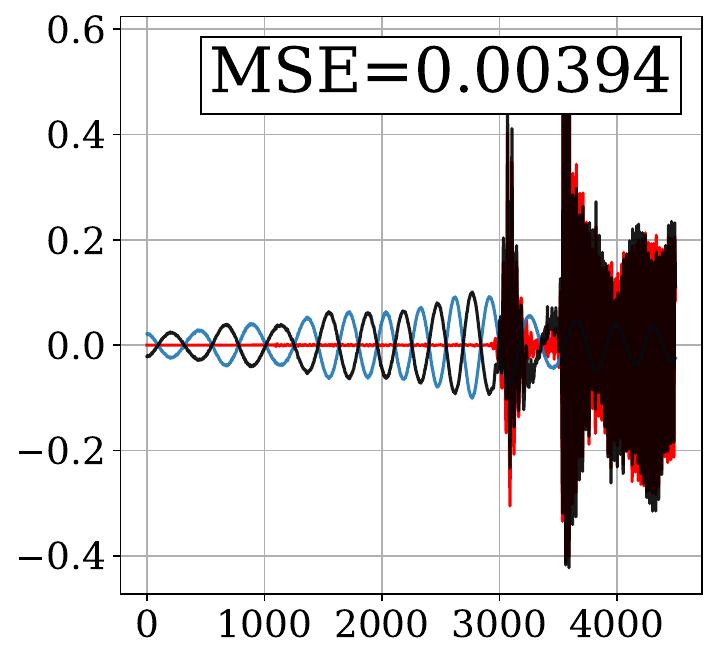}
        \caption{WIRE}
    \end{subfigure}\hspace{-0.5em}
    \begin{subfigure}{0.20\textwidth}
        \includegraphics[width=\linewidth]{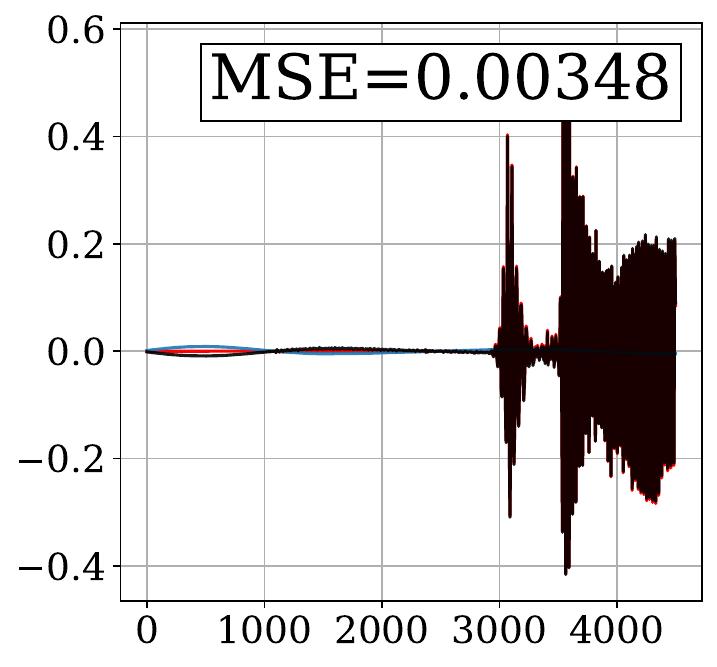}
        \caption{Gauss}
    \end{subfigure}

    \vspace{0.5em}

    \begin{subfigure}{0.20\textwidth}
        \includegraphics[width=\linewidth]{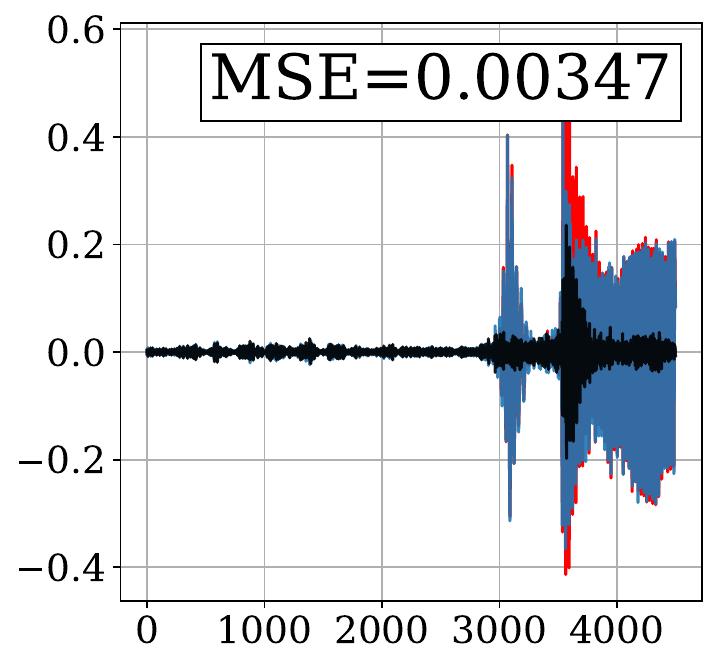}
        \caption{MIRE}
    \end{subfigure}\hspace{-0.5em}
    \begin{subfigure}{0.20\textwidth}
        \includegraphics[width=\linewidth]{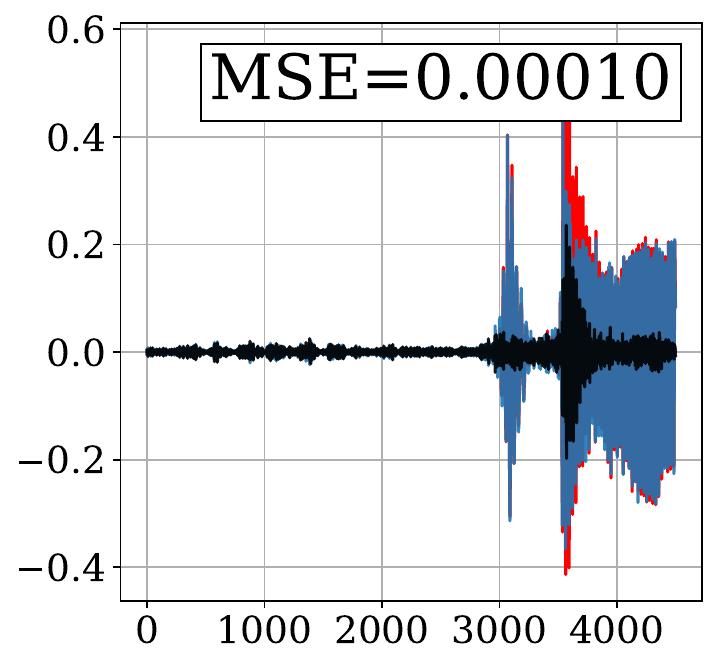}
        \caption{SPDER}
    \end{subfigure}\hspace{-0.5em}
    \begin{subfigure}{0.20\textwidth}
        \includegraphics[width=\linewidth]{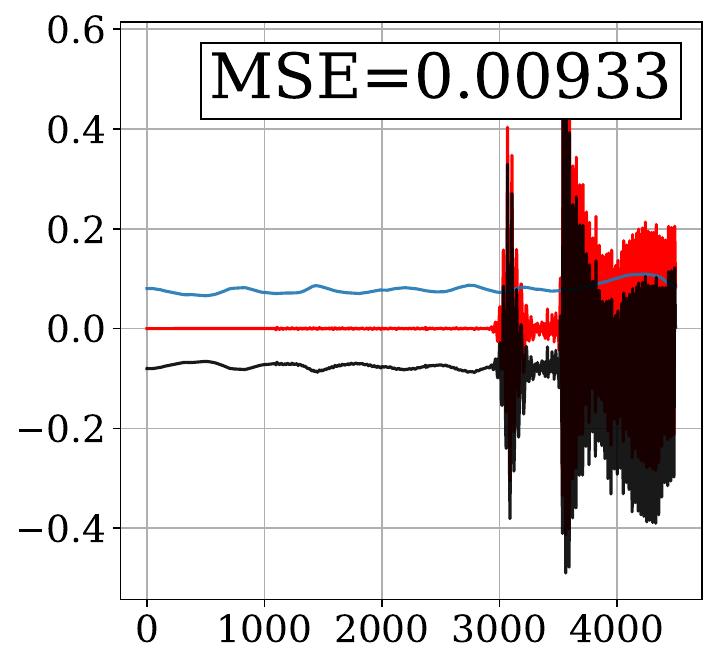}
        \caption{PE}
    \end{subfigure}\hspace{-0.5em}
    \begin{subfigure}{0.20\textwidth}
        \includegraphics[width=\linewidth]{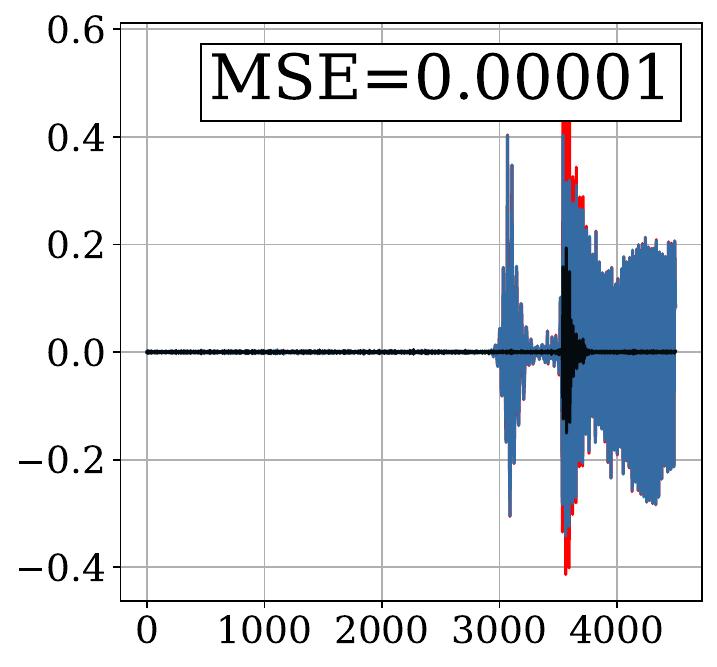}
        \caption{FM-SIREN}
    \end{subfigure}\hspace{-0.5em}
    \begin{subfigure}{0.20\textwidth}
        \includegraphics[width=\linewidth]{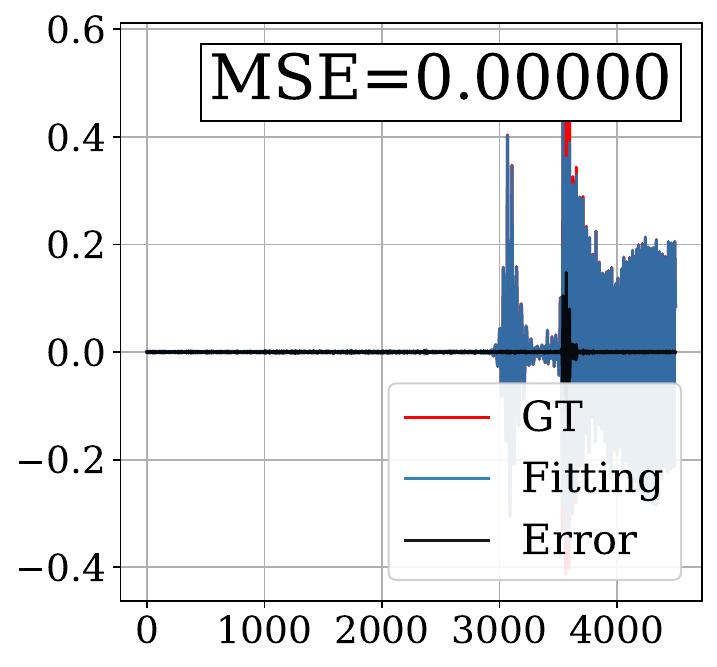}
        \caption{FM-FINER}
    \end{subfigure}

    \caption{Qualitative results of one-second audio reconstructions for the Anfield
    clip in the Spoken English Wikipedia dataset \cite{spokenwikipedia}, using two-layer networks with different approaches. \colorbox{red}{\textcolor{white}{Red}}, \colorbox{customblue}{\textcolor{white}{medium blue}}, and \colorbox{black}{\textcolor{white}{black}} lines correspond to ground truth, reconstructed signal, and error signal, respectively. FM-SIREN and FM-FINER demonstrate superior performance in terms of MSE, as shown on the top-right of each subfigure.}
    \label{fig:audio-fi-4}
\end{figure}

\begin{figure}[h]
    \centering
    \begin{subfigure}{0.24\textwidth}
        \centering
        \begin{minipage}{\linewidth}
            \includegraphics[width=\linewidth]{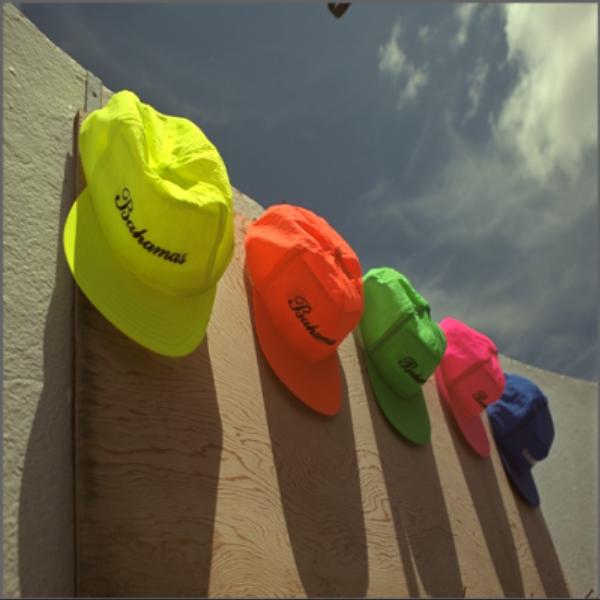}\\
            \includegraphics[width=\linewidth]{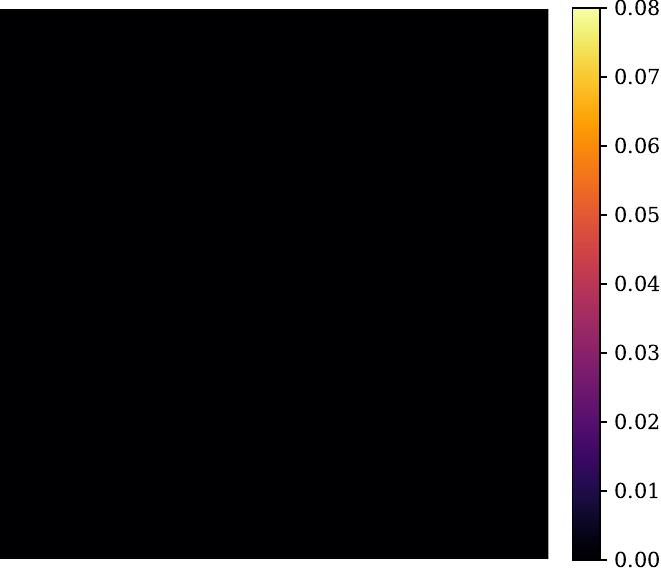}
        \end{minipage}
        \caption{Ground Truth}
    \end{subfigure}
    \begin{subfigure}{0.24\textwidth}
        \centering
        \begin{minipage}{\linewidth}
            \includegraphics[width=\linewidth]{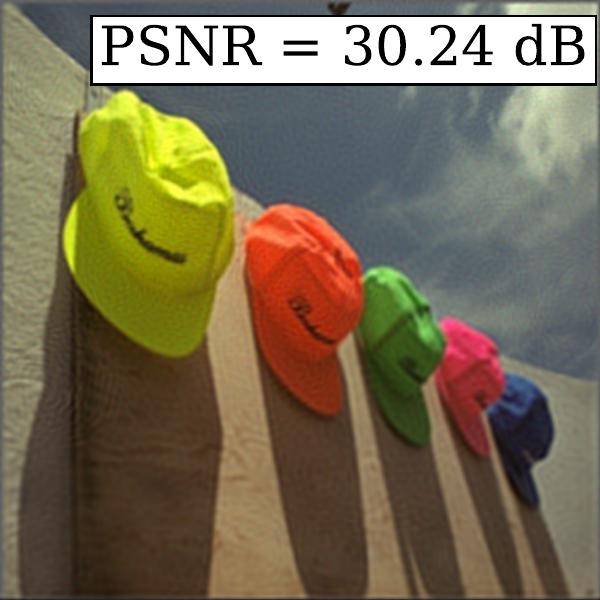}\\
            \includegraphics[width=\linewidth]{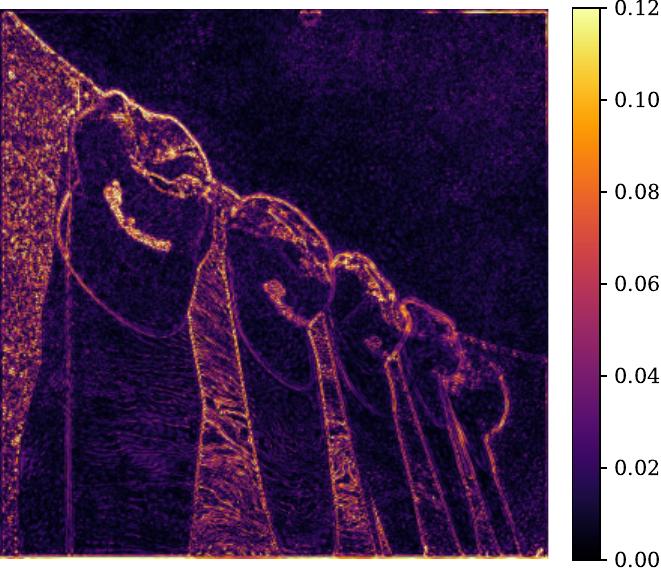}
        \end{minipage}
        \caption{SIREN}
    \end{subfigure}
    \begin{subfigure}{0.24\textwidth}
        \centering
        \begin{minipage}{\linewidth}
            \includegraphics[width=\linewidth]{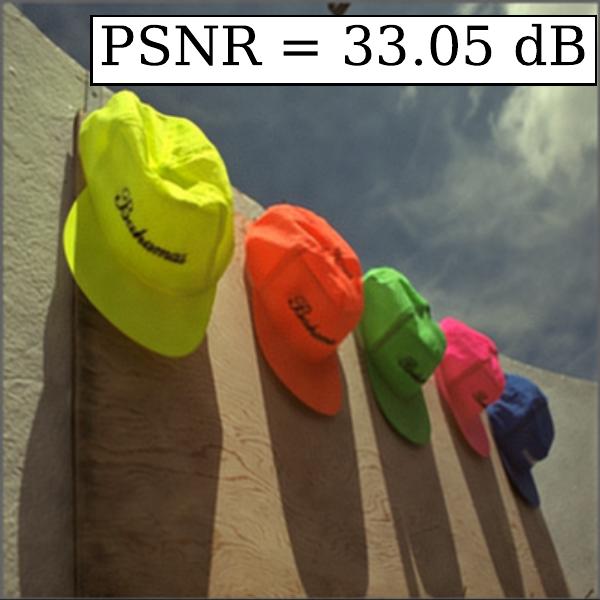}\\
            \includegraphics[width=\linewidth]{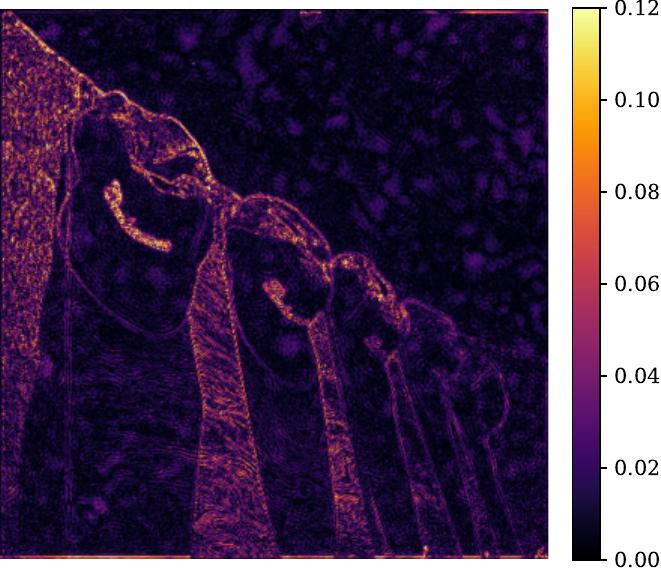}
        \end{minipage}
        \caption{FINER}
    \end{subfigure}
    \begin{subfigure}{0.24\textwidth}
        \centering
        \begin{minipage}{\linewidth}
            \includegraphics[width=\linewidth]{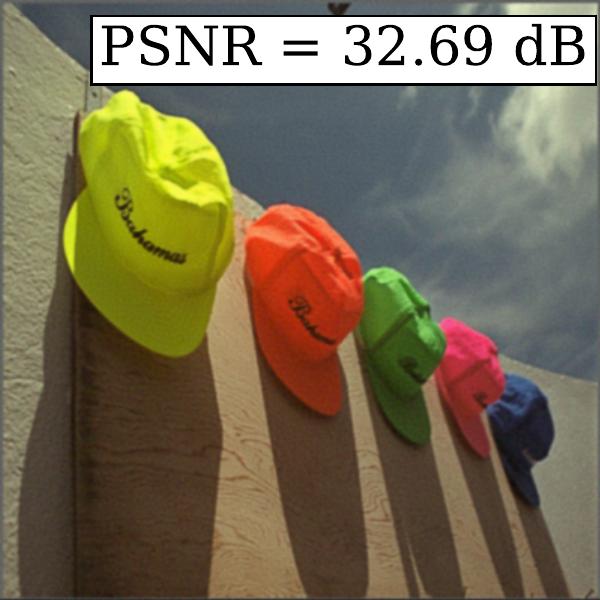}\\
            \includegraphics[width=\linewidth]{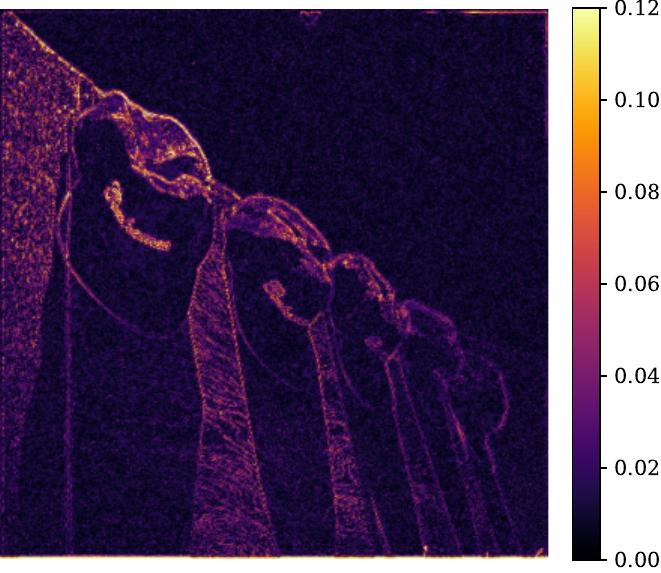
            }
        \end{minipage}
        \caption{PE}
    \end{subfigure}
    \begin{subfigure}{0.24\textwidth}
        \centering
        \begin{minipage}{\linewidth}
            \includegraphics[width=\linewidth]{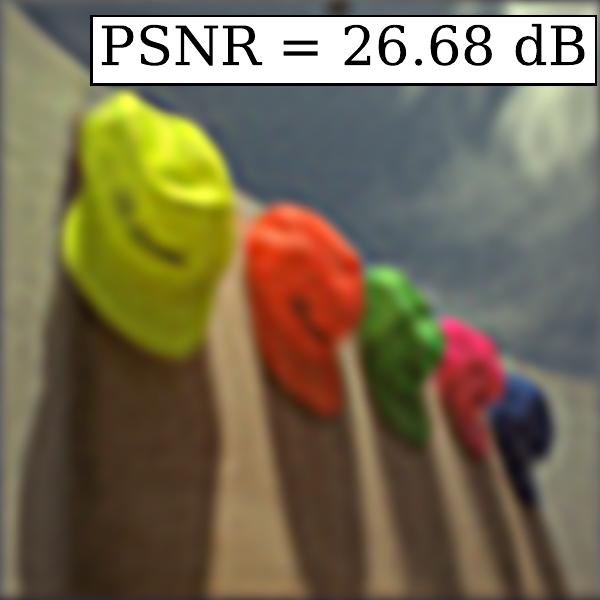}\\
            \includegraphics[width=\linewidth]{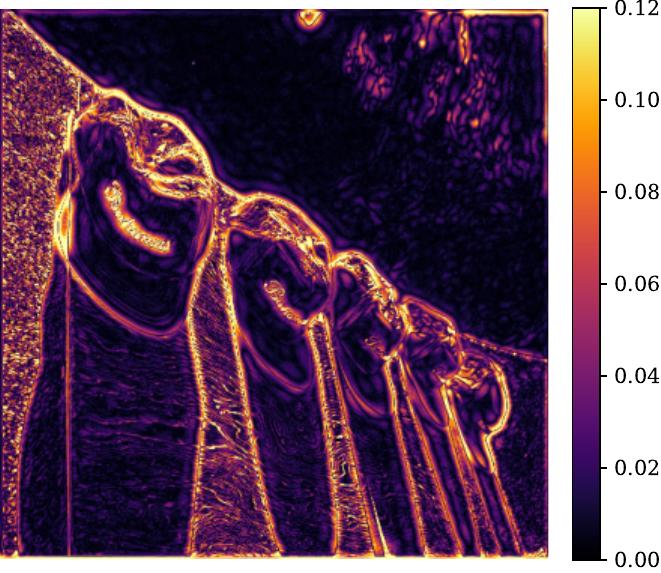}
        \end{minipage}
        \caption{Gauss}
    \end{subfigure}
    \begin{subfigure}{0.24\textwidth}
        \centering
        \begin{minipage}{\linewidth}
            \includegraphics[width=\linewidth]{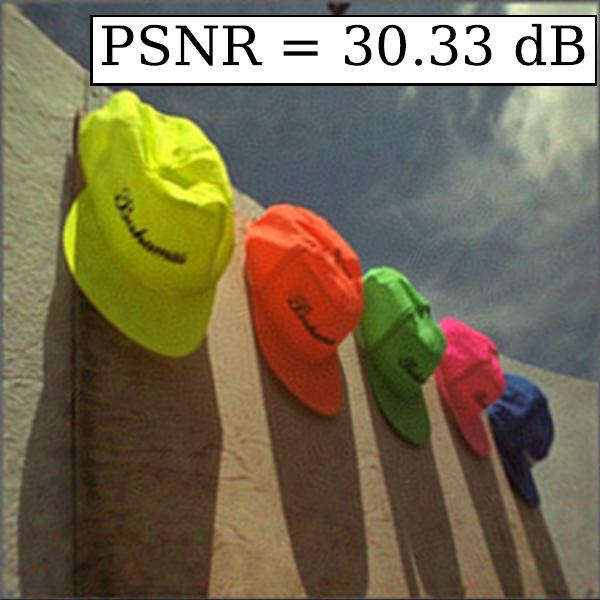}\\
            \includegraphics[width=\linewidth]{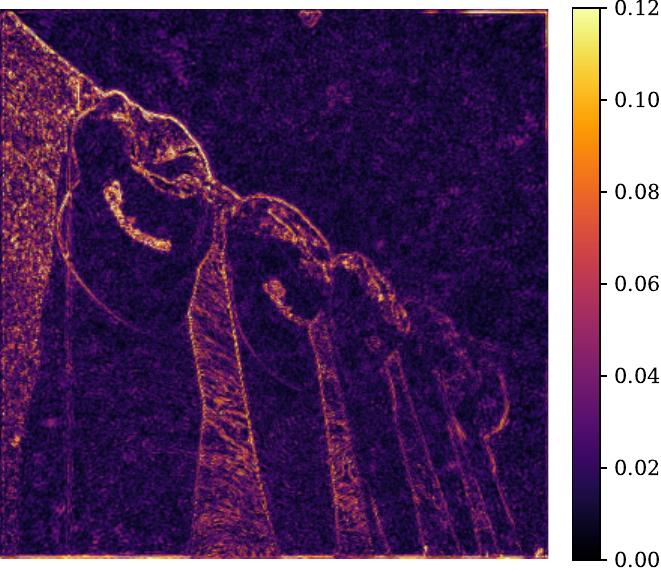}
        \end{minipage}
        \caption{WIRE}
    \end{subfigure}
    \begin{subfigure}{0.24\textwidth}
        \centering
        \begin{minipage}{\linewidth}
            \includegraphics[width=\linewidth]{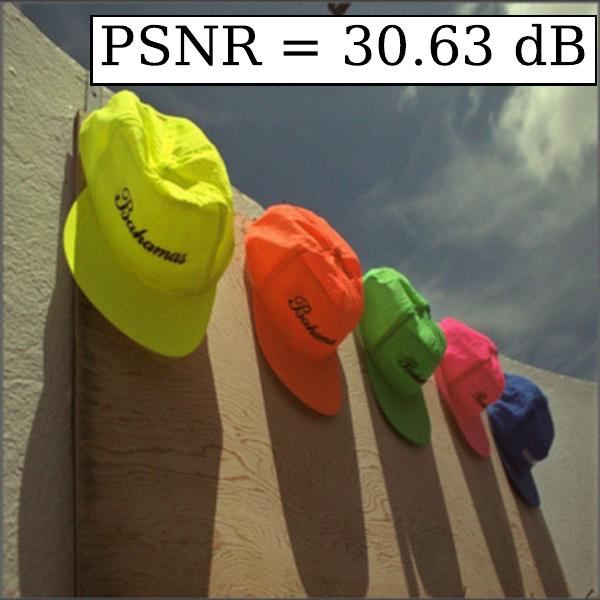}\\
            \includegraphics[width=\linewidth]{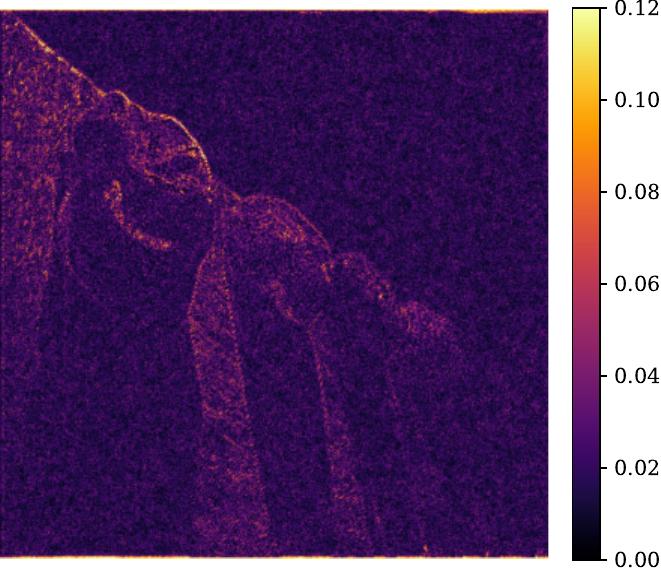}
        \end{minipage}
        \caption{TUNER}
    \end{subfigure}
    \begin{subfigure}{0.24\textwidth}
        \centering
        \begin{minipage}{\linewidth}
            \includegraphics[width=\linewidth]{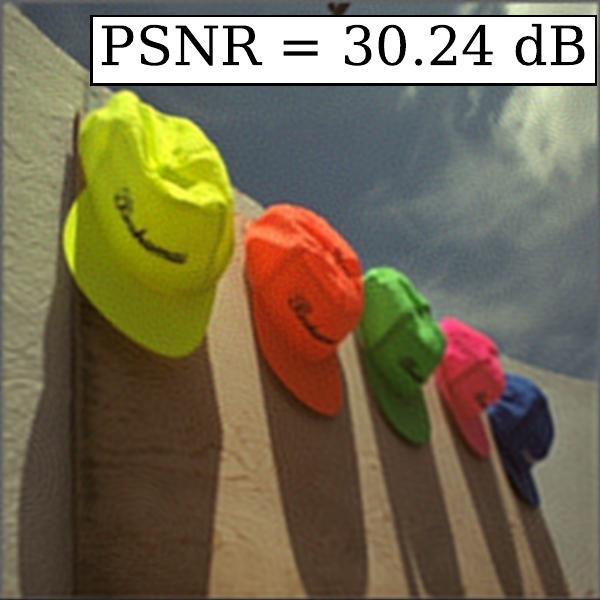}\\
            \includegraphics[width=\linewidth]{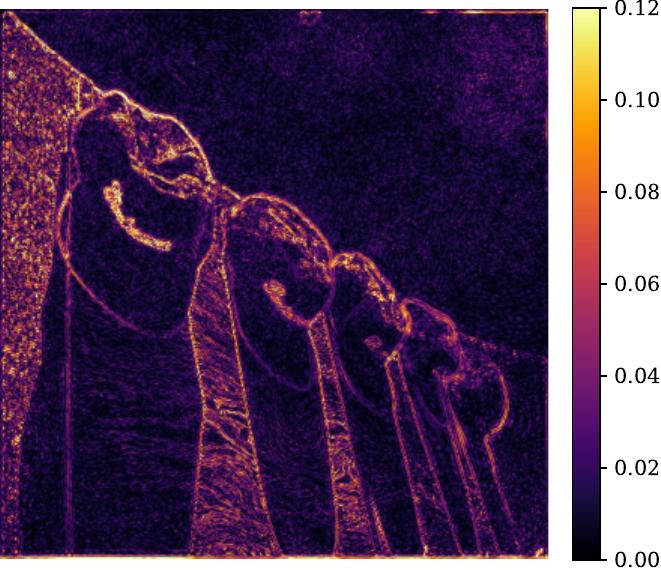}
        \end{minipage}
        \caption{FreSh}
    \end{subfigure}
    \begin{subfigure}{0.24\textwidth}
        \centering
        \begin{minipage}{\linewidth}
            \includegraphics[width=\linewidth]{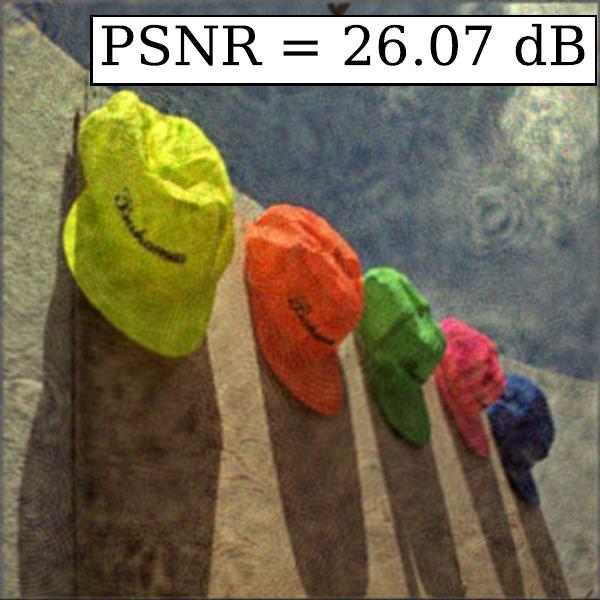}\\
            \includegraphics[width=\linewidth]{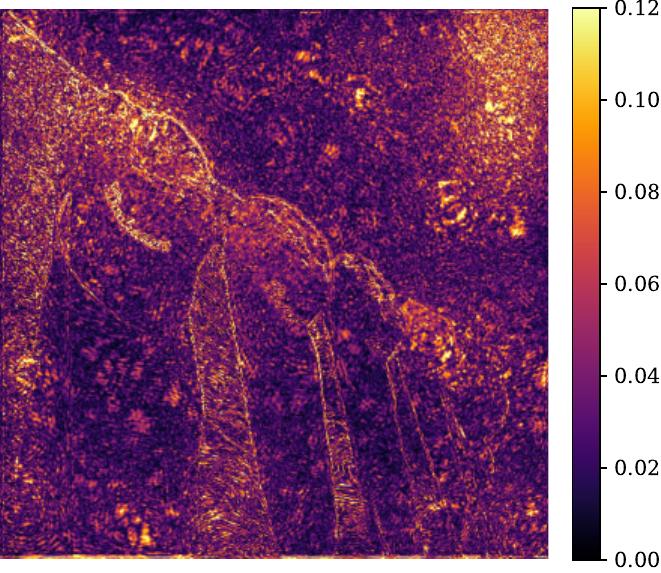}
        \end{minipage}
        \caption{SPDER}
    \end{subfigure}
    \begin{subfigure}{0.24\textwidth}
        \centering
        \begin{minipage}{\linewidth}
            \includegraphics[width=\linewidth]{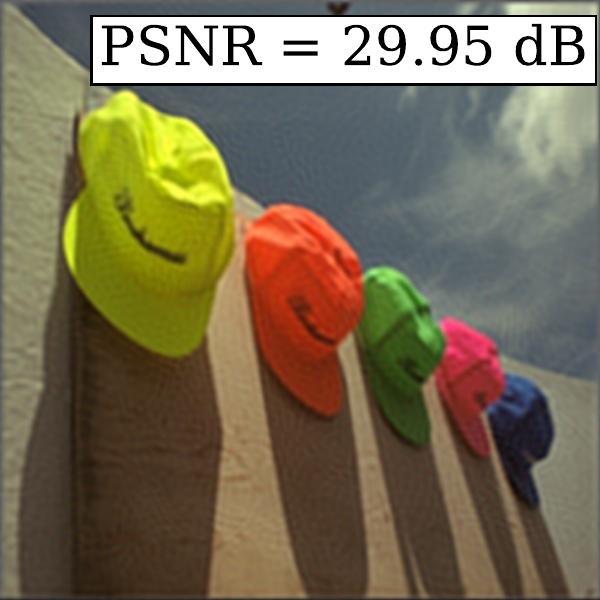}\\
            \includegraphics[width=\linewidth]{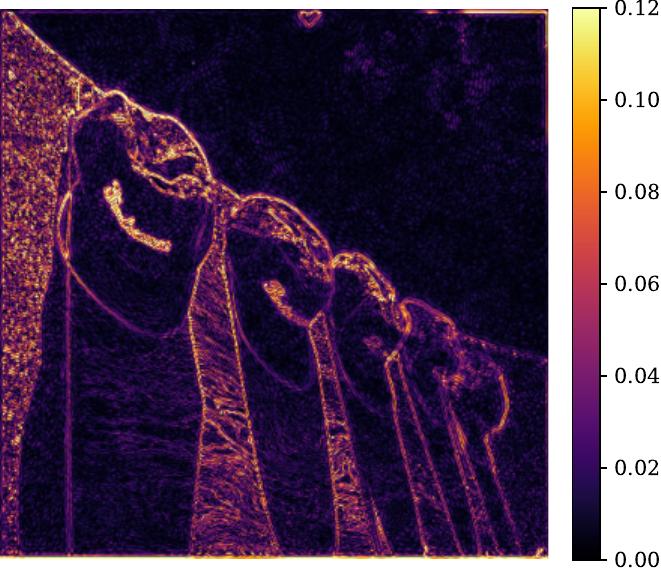}
        \end{minipage}
        \caption{MIRE}
    \end{subfigure}
    \begin{subfigure}{0.24\textwidth}
        \centering
        \begin{minipage}{\linewidth}
            \includegraphics[width=\linewidth]{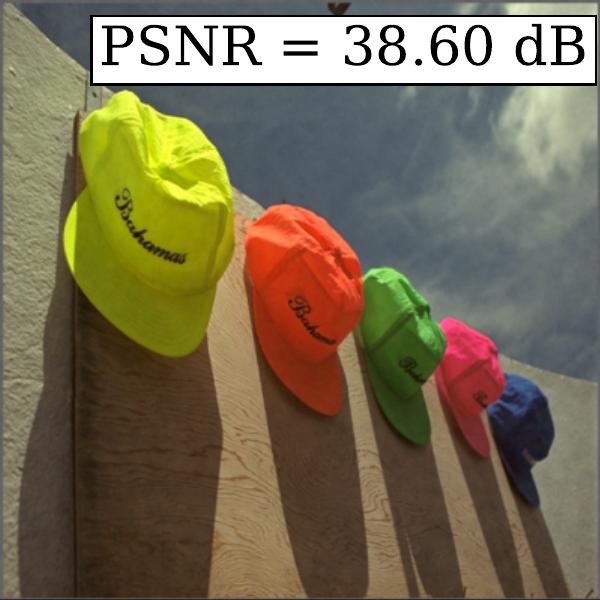}\\
            \includegraphics[width=\linewidth]{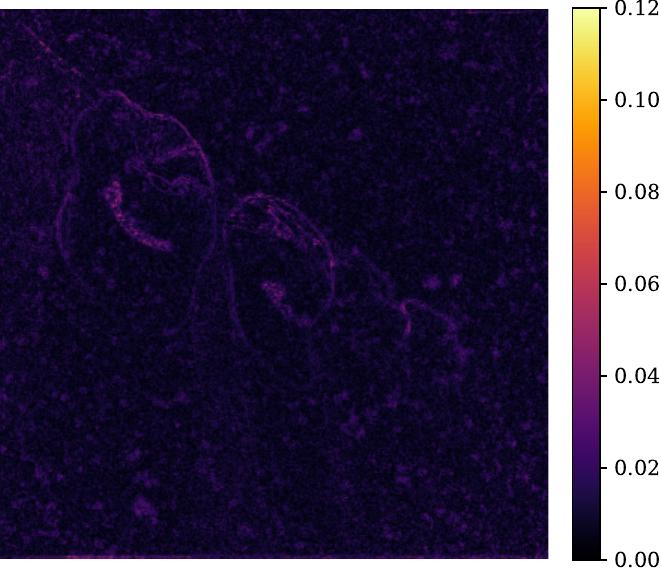}
        \end{minipage}
        \caption{FM-SIREN}
    \end{subfigure}
    \begin{subfigure}{0.24\textwidth}
        \centering
        \begin{minipage}{\linewidth}
            \includegraphics[width=\linewidth]{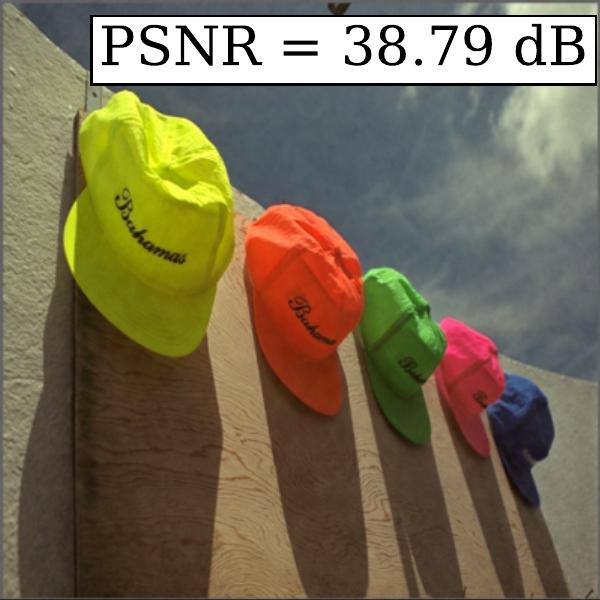}\\
            \includegraphics[width=\linewidth]{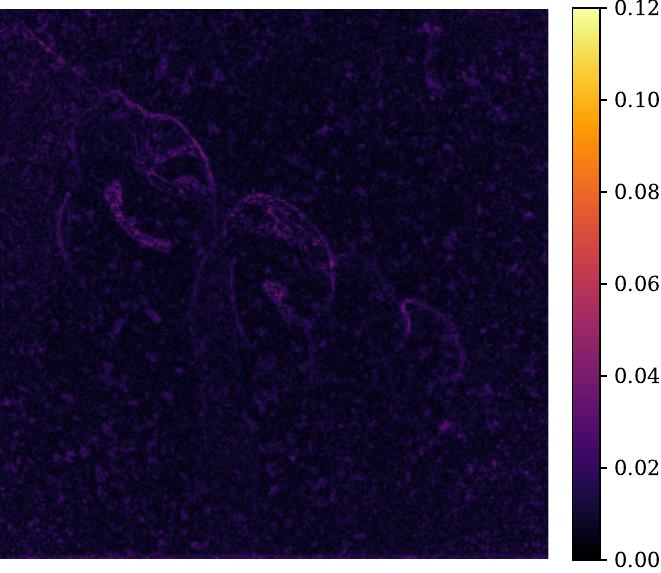}
        \end{minipage}
        \caption{FM-FINER}
    \end{subfigure}
    \caption{Qualitative results for kodim03 from Kodak Lossless True Color Image Suite \cite{kodak}. Top image in each subfigure is the reconstruction result while the bottom is the error map between reconstruction and ground truth. FM-SIREN and FM-FINER show significantly better reconstruction results as demonstrated by the PSNR values on the top-right of each subfigure. The error maps show that our models can learn both high frequency and low frequency structures with higher quality using small networks.}
    \label{fig:2Dfit_1}
\end{figure}

\begin{figure}[h]
    \centering
    \begin{subfigure}{0.24\textwidth}
        \centering
        \begin{minipage}{\linewidth}
            \includegraphics[width=\linewidth]{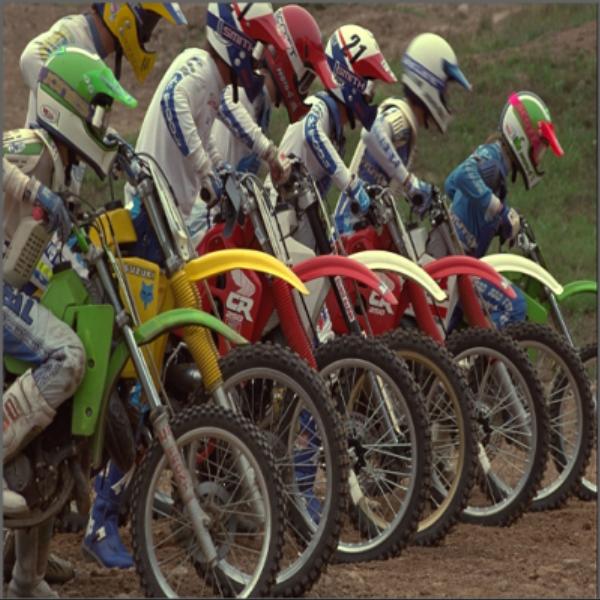}\\
            \includegraphics[width=\linewidth]{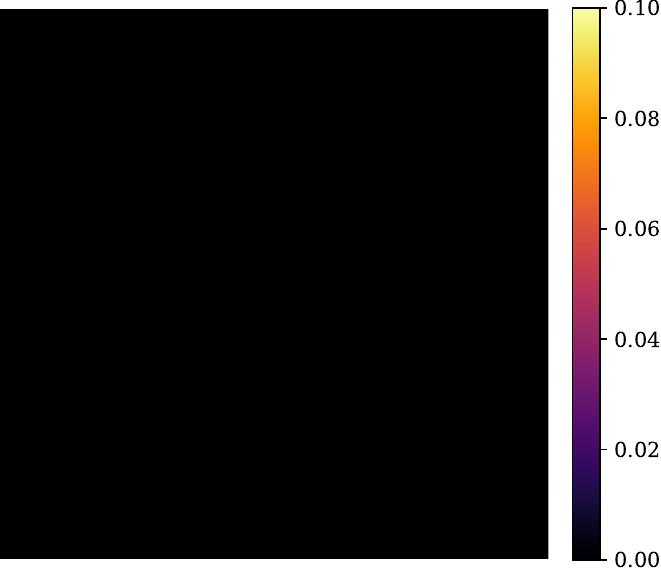}
        \end{minipage}
        \caption{Ground Truth}
    \end{subfigure}
    \begin{subfigure}{0.24\textwidth}
        \centering
        \begin{minipage}{\linewidth}
            \includegraphics[width=\linewidth]{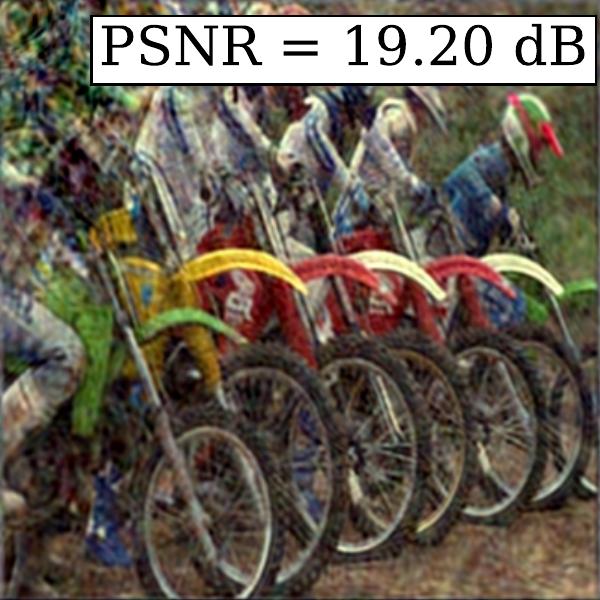}\\
            \includegraphics[width=\linewidth]{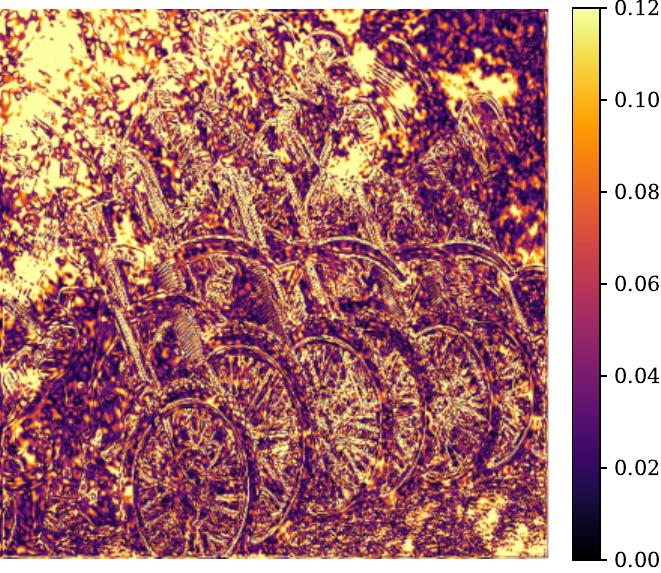}
        \end{minipage}
        \caption{SIREN}
    \end{subfigure}
    \begin{subfigure}{0.24\textwidth}
        \centering
        \begin{minipage}{\linewidth}
            \includegraphics[width=\linewidth]{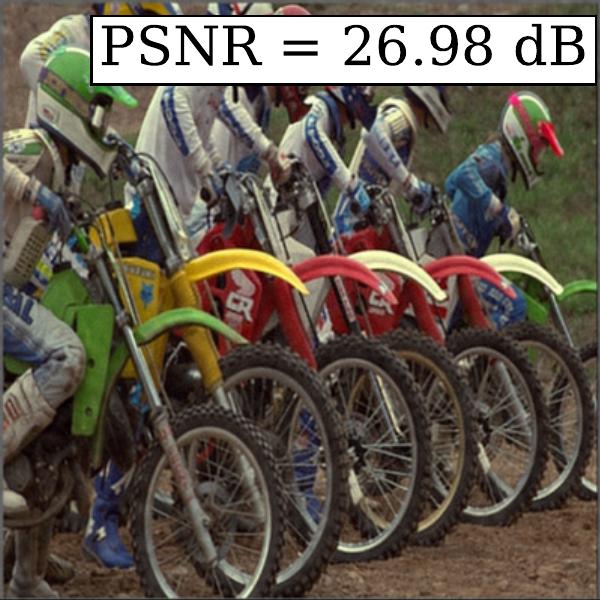}\\
            \includegraphics[width=\linewidth]{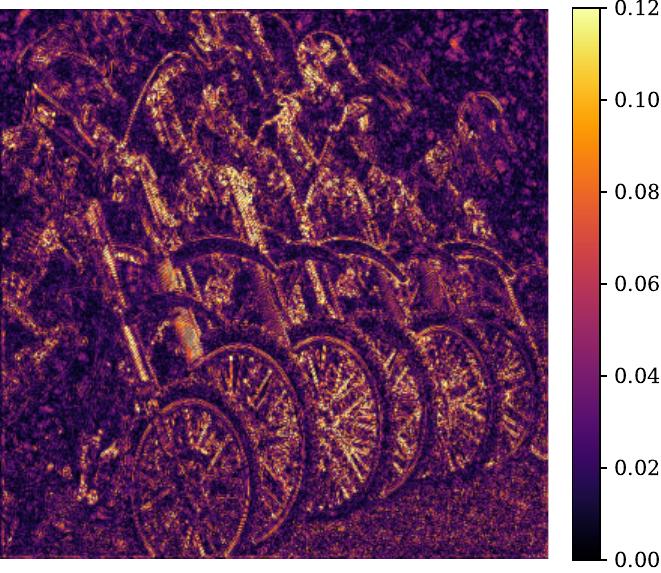}
        \end{minipage}
        \caption{FINER}
    \end{subfigure}
    \begin{subfigure}{0.24\textwidth}
        \centering
        \begin{minipage}{\linewidth}
            \includegraphics[width=\linewidth]{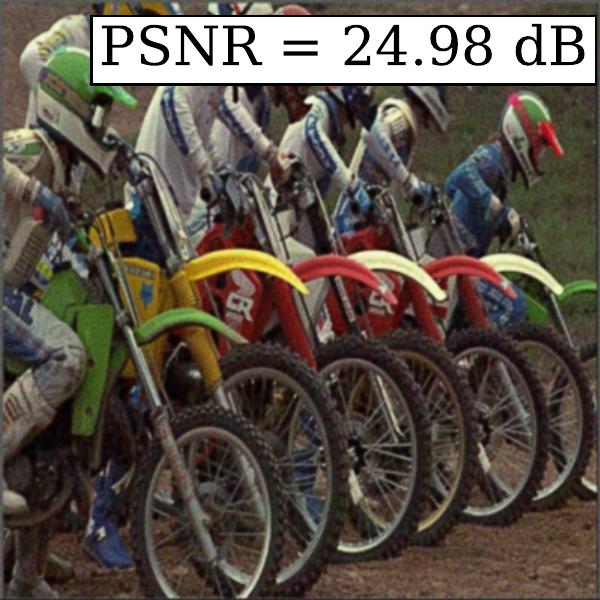}\\
            \includegraphics[width=\linewidth]{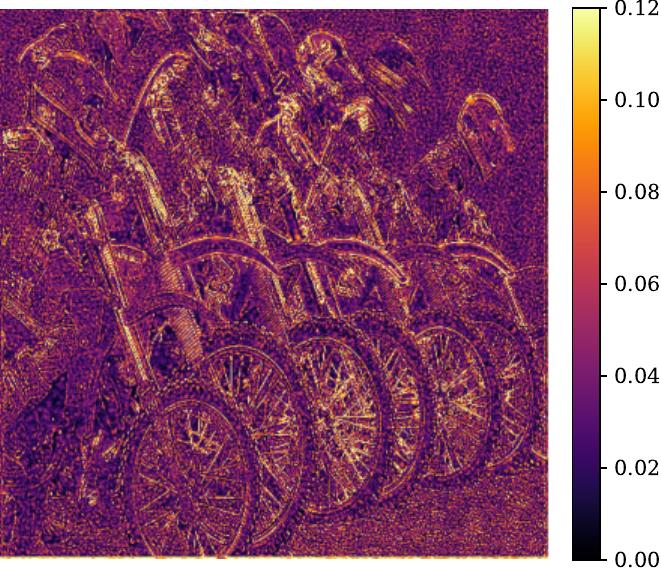
            }
        \end{minipage}
        \caption{PE}
    \end{subfigure}
    \begin{subfigure}{0.24\textwidth}
        \centering
        \begin{minipage}{\linewidth}
            \includegraphics[width=\linewidth]{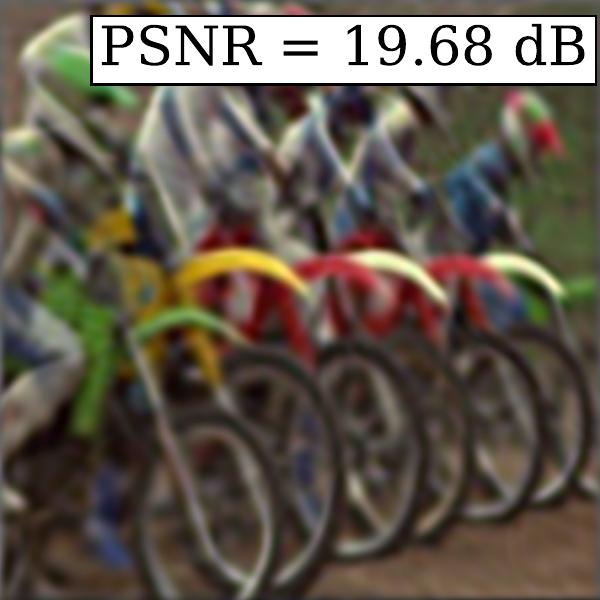}\\
            \includegraphics[width=\linewidth]{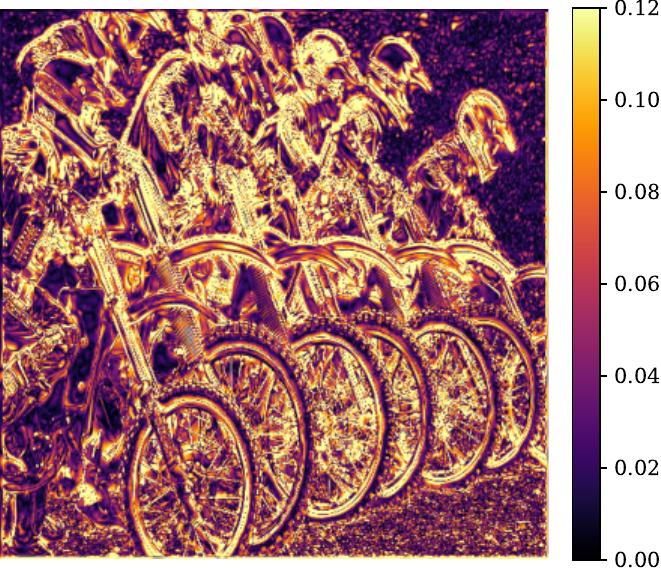}
        \end{minipage}
        \caption{Gauss}
    \end{subfigure}
    \begin{subfigure}{0.24\textwidth}
        \centering
        \begin{minipage}{\linewidth}
            \includegraphics[width=\linewidth]{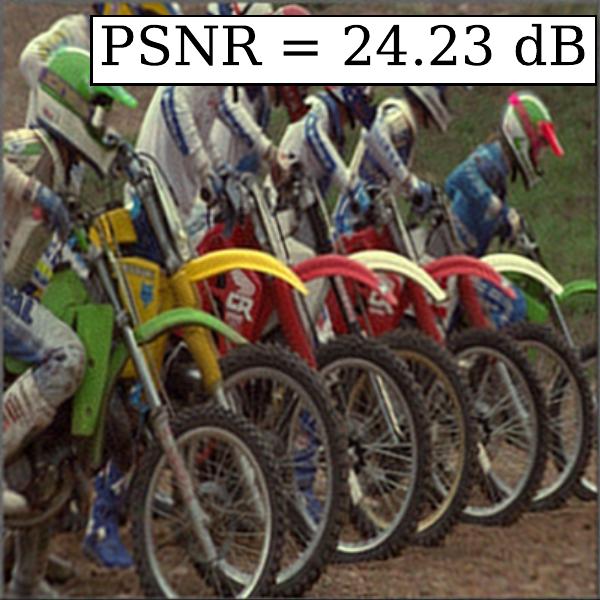}\\
            \includegraphics[width=\linewidth]{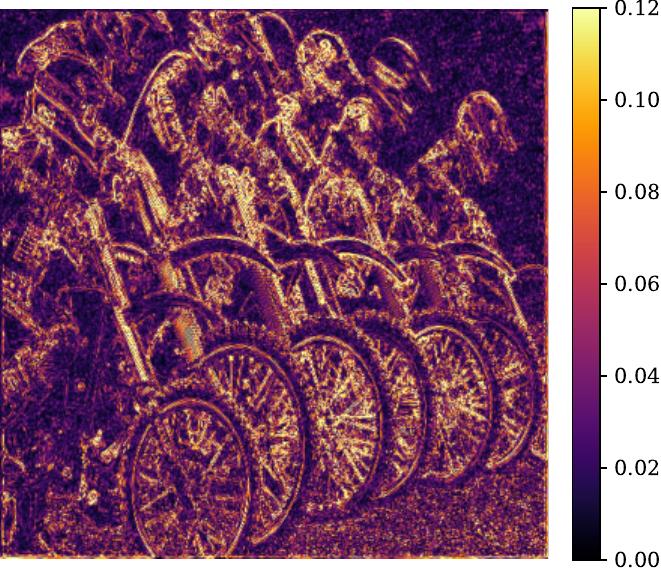}
        \end{minipage}
        \caption{WIRE}
    \end{subfigure}
    \begin{subfigure}{0.24\textwidth}
        \centering
        \begin{minipage}{\linewidth}
            \includegraphics[width=\linewidth]{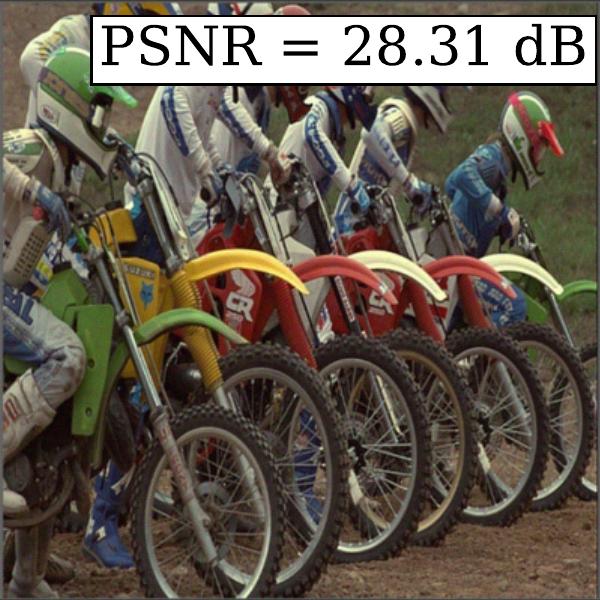}\\
            \includegraphics[width=\linewidth]{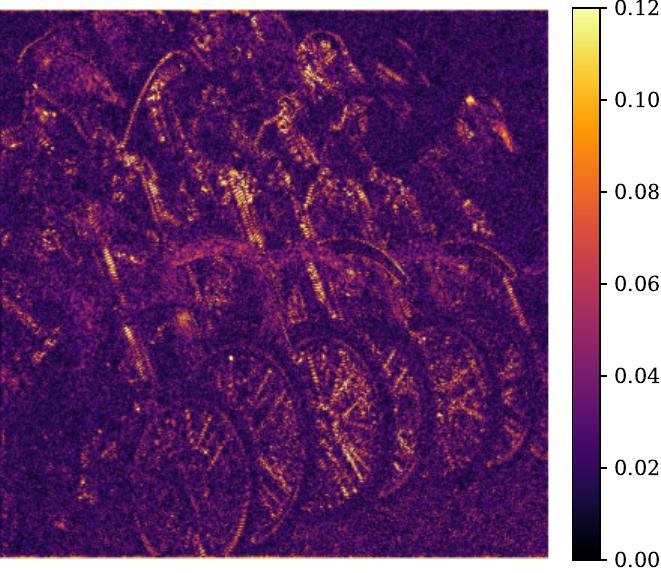}
        \end{minipage}
        \caption{TUNER}
    \end{subfigure}
    \begin{subfigure}{0.24\textwidth}
        \centering
        \begin{minipage}{\linewidth}
            \includegraphics[width=\linewidth]{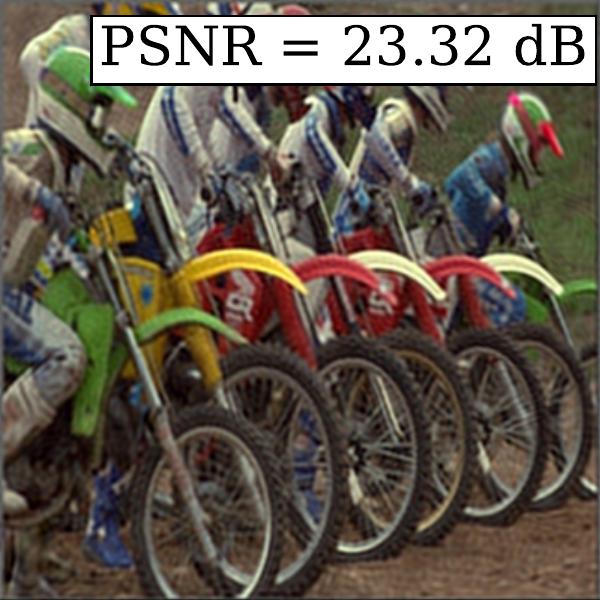}\\
            \includegraphics[width=\linewidth]{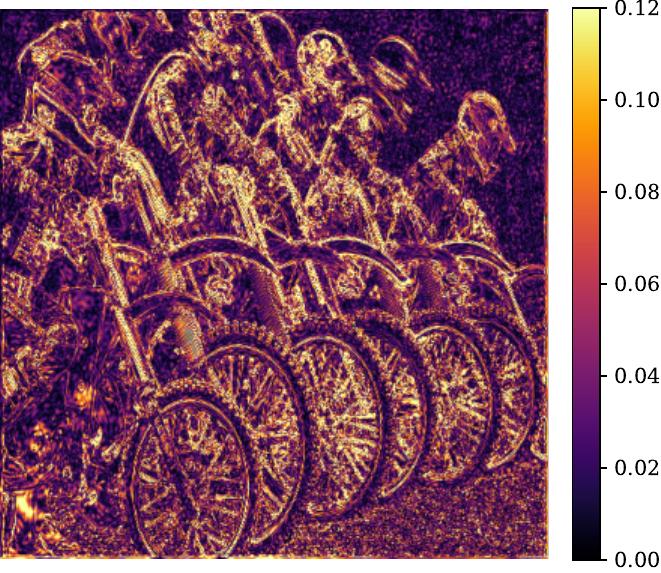}
        \end{minipage}
        \caption{FreSh}
    \end{subfigure}
    \begin{subfigure}{0.24\textwidth}
        \centering
        \begin{minipage}{\linewidth}
            \includegraphics[width=\linewidth]{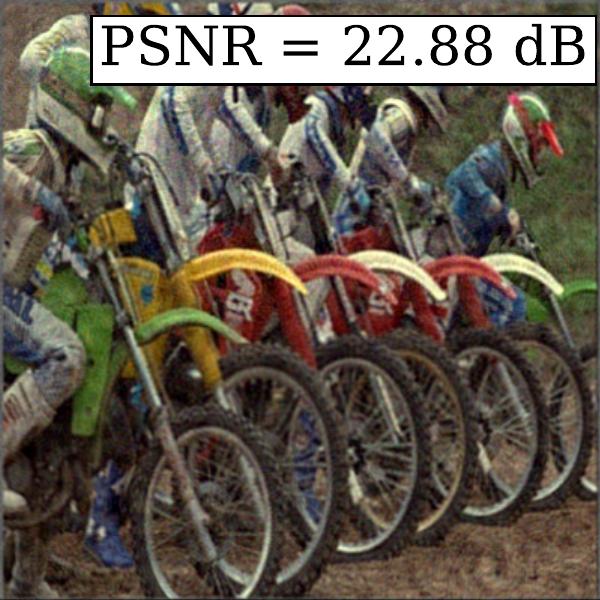}\\
            \includegraphics[width=\linewidth]{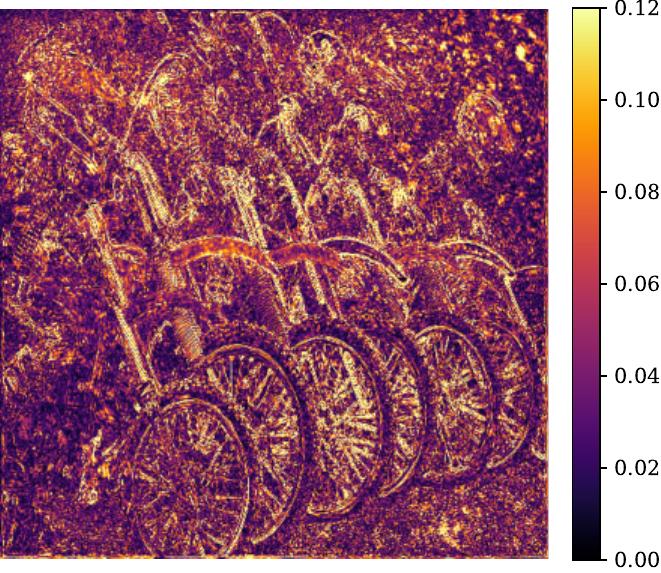}
        \end{minipage}
        \caption{SPDER}
    \end{subfigure}
    \begin{subfigure}{0.24\textwidth}
        \centering
        \begin{minipage}{\linewidth}
            \includegraphics[width=\linewidth]{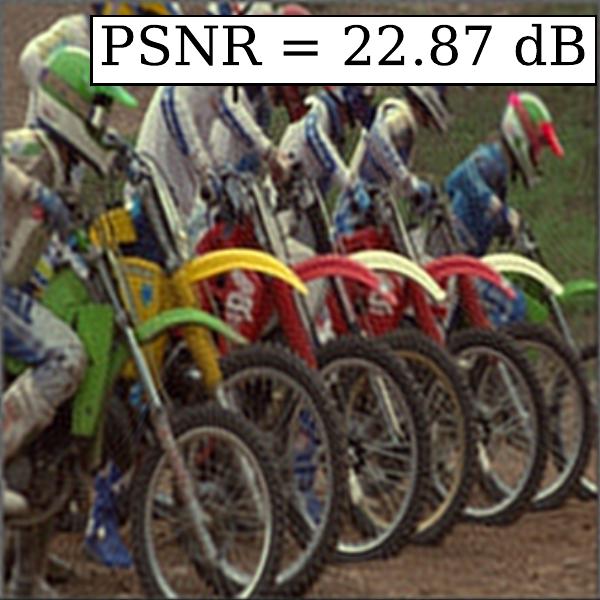}\\
            \includegraphics[width=\linewidth]{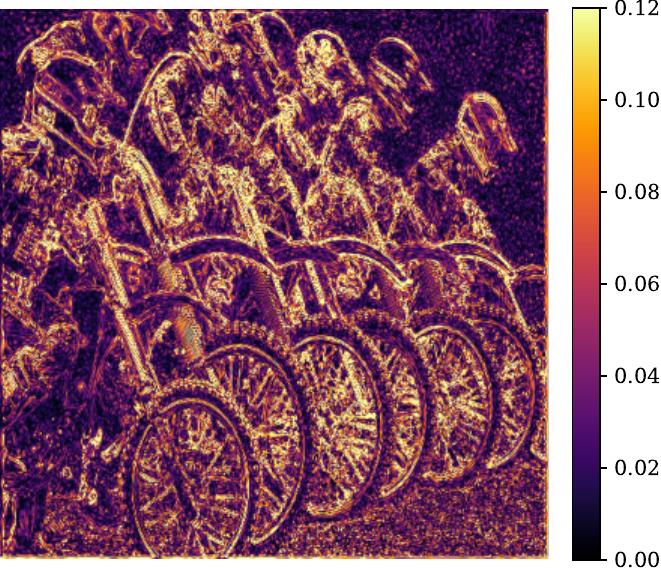}
        \end{minipage}
        \caption{MIRE}
    \end{subfigure}
    \begin{subfigure}{0.24\textwidth}
        \centering
        \begin{minipage}{\linewidth}
            \includegraphics[width=\linewidth]{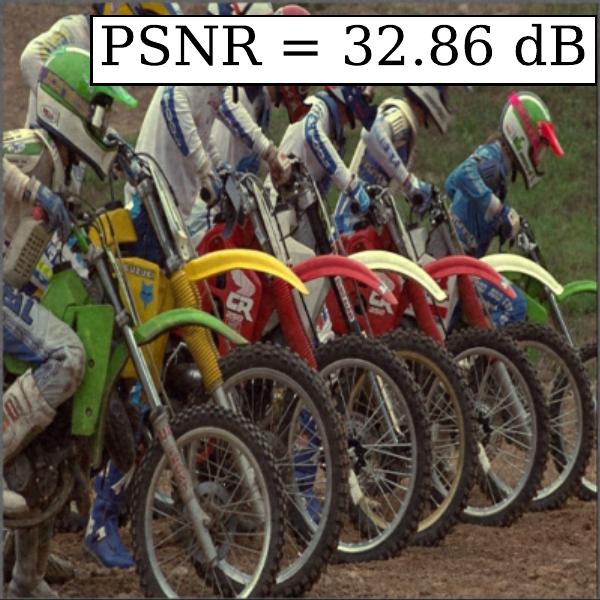}\\
            \includegraphics[width=\linewidth]{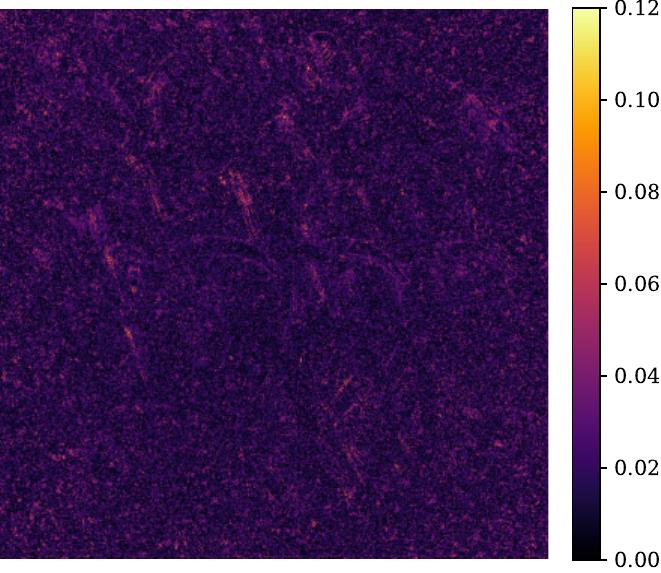}
        \end{minipage}
        \caption{FM-SIREN}
    \end{subfigure}
    \begin{subfigure}{0.24\textwidth}
        \centering
        \begin{minipage}{\linewidth}
            \includegraphics[width=\linewidth]{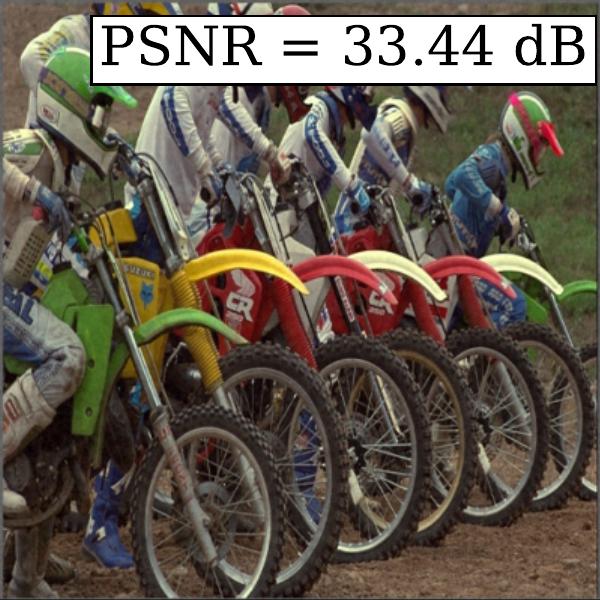}\\
            \includegraphics[width=\linewidth]{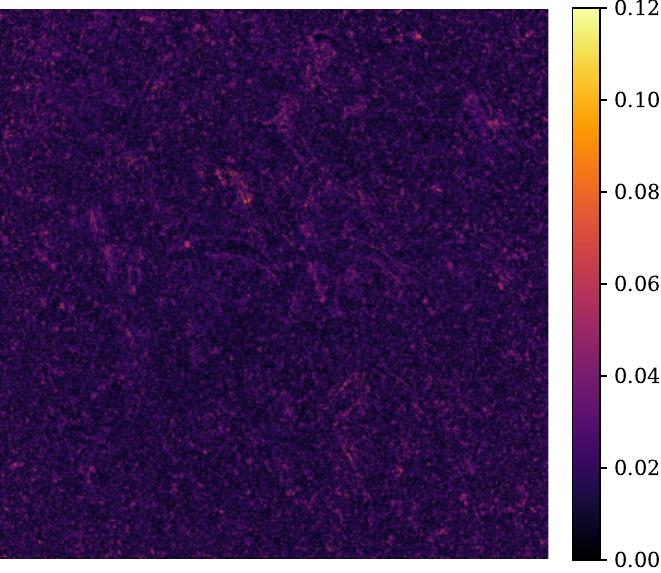}
        \end{minipage}
        \caption{FM-FINER}
    \end{subfigure}
    \caption{Qualitative results for kodim05 from Kodak Lossless True Color Image Suite \cite{kodak}. Top image in each subfigure is the reconstruction result while the bottom is the error map between reconstruction and ground truth. FM-SIREN and FM-FINER show significantly better reconstruction results as demonstrated by the PSNR values on the top-right of each subfigure. The error maps show that our models can learn both high frequency and low frequency structures with higher quality using small networks.}
    \label{fig:2Dfit_2}
\end{figure}

\begin{figure}[h]
    \centering
    \begin{subfigure}{0.24\textwidth}
        \centering
        \begin{minipage}{\linewidth}
            \includegraphics[width=\linewidth]{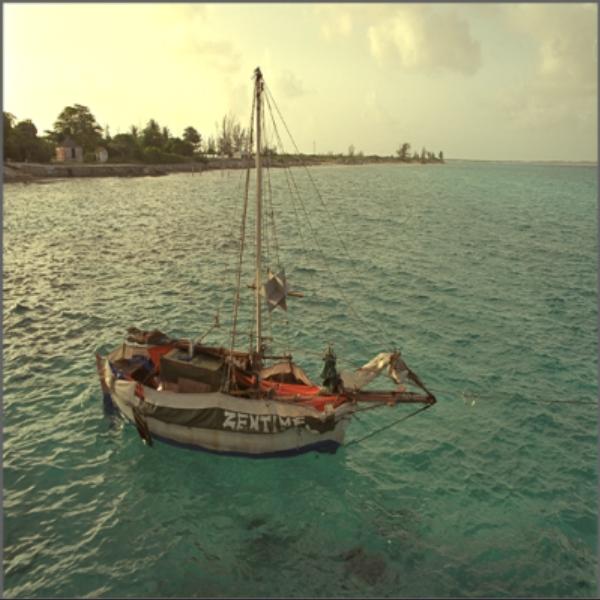}\\
            \includegraphics[width=\linewidth]{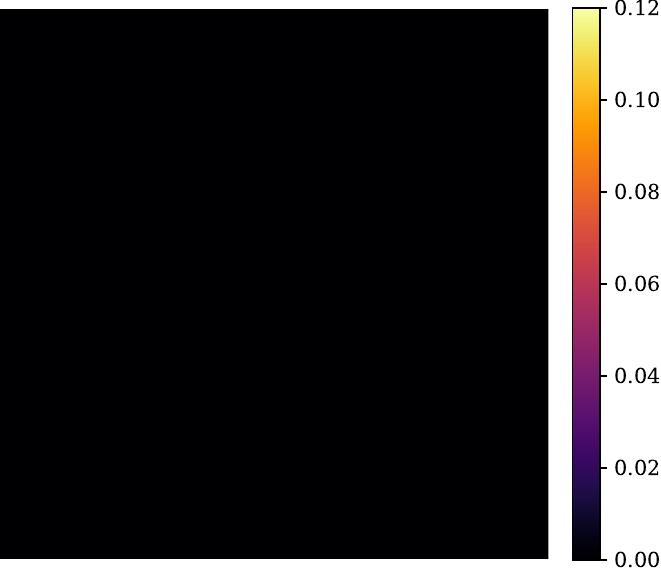}
        \end{minipage}
        \caption{Ground Truth}
    \end{subfigure}
    \begin{subfigure}{0.24\textwidth}
        \centering
        \begin{minipage}{\linewidth}
            \includegraphics[width=\linewidth]{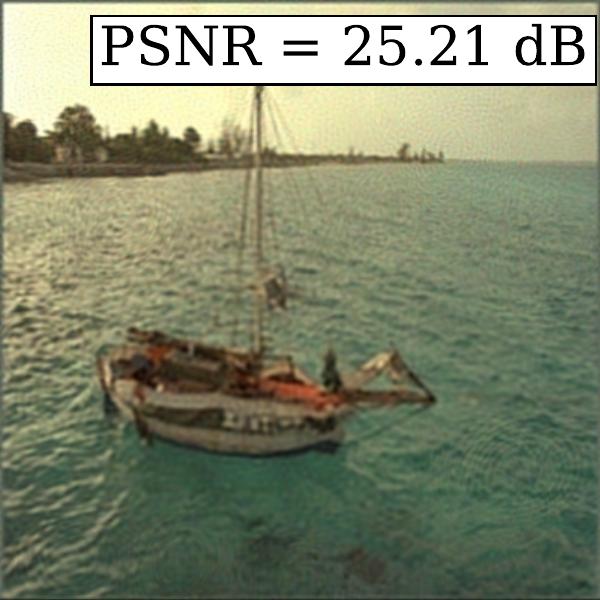}\\
            \includegraphics[width=\linewidth]{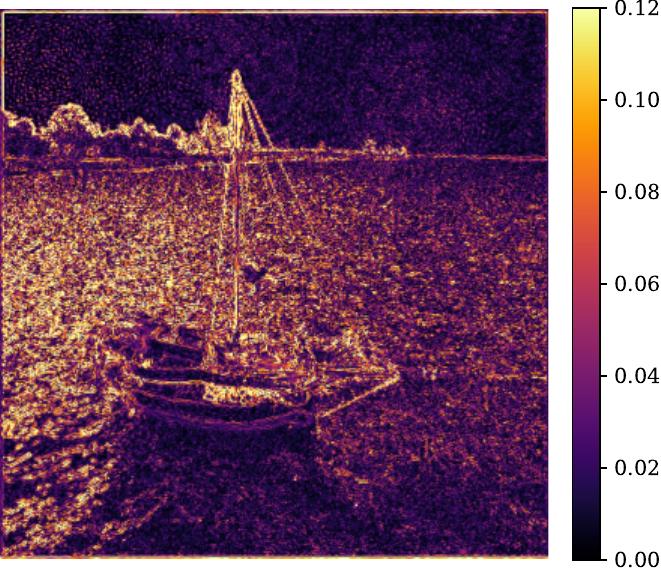}
        \end{minipage}
        \caption{SIREN}
    \end{subfigure}
    \begin{subfigure}{0.24\textwidth}
        \centering
        \begin{minipage}{\linewidth}
            \includegraphics[width=\linewidth]{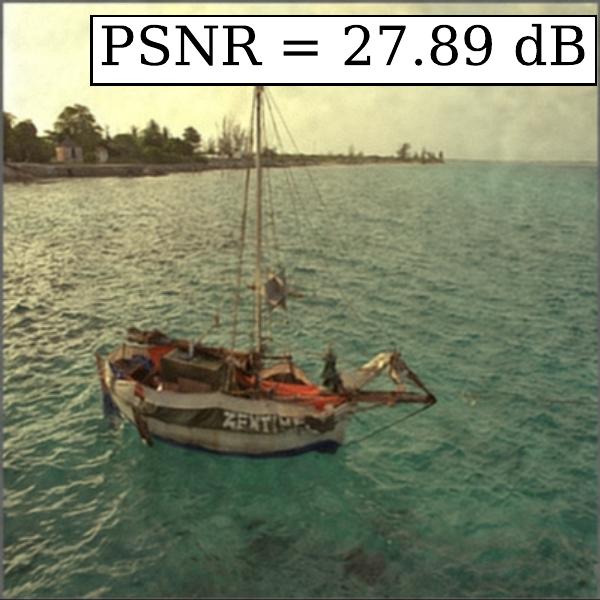}\\
            \includegraphics[width=\linewidth]{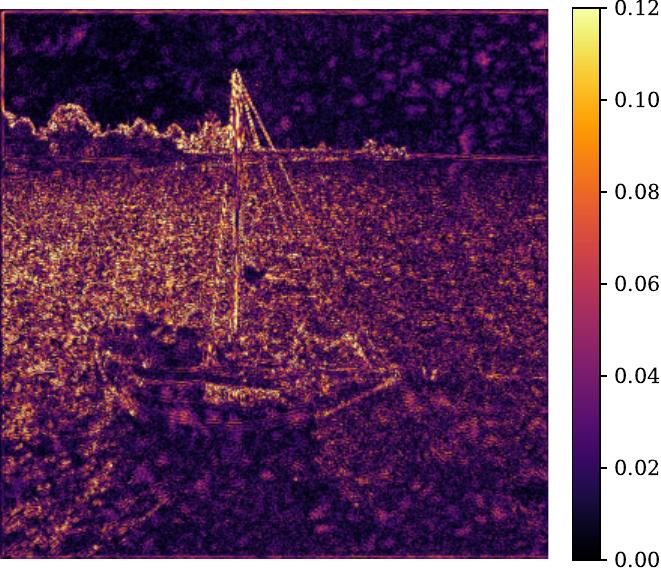}
        \end{minipage}
        \caption{FINER}
    \end{subfigure}
    \begin{subfigure}{0.24\textwidth}
        \centering
        \begin{minipage}{\linewidth}
            \includegraphics[width=\linewidth]{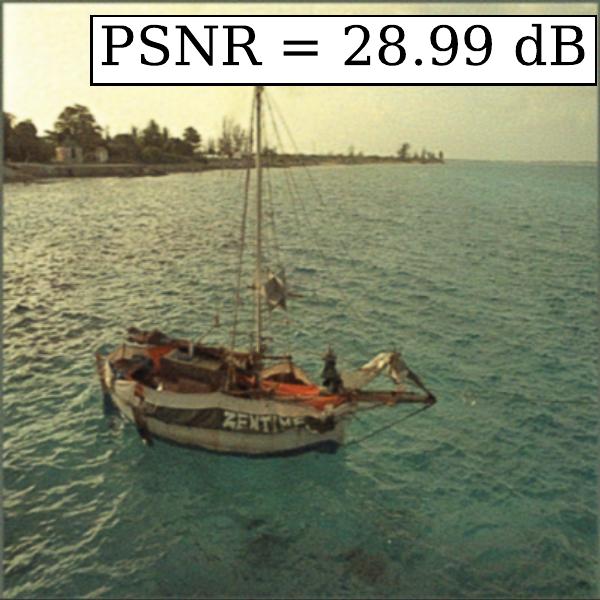}\\
            \includegraphics[width=\linewidth]{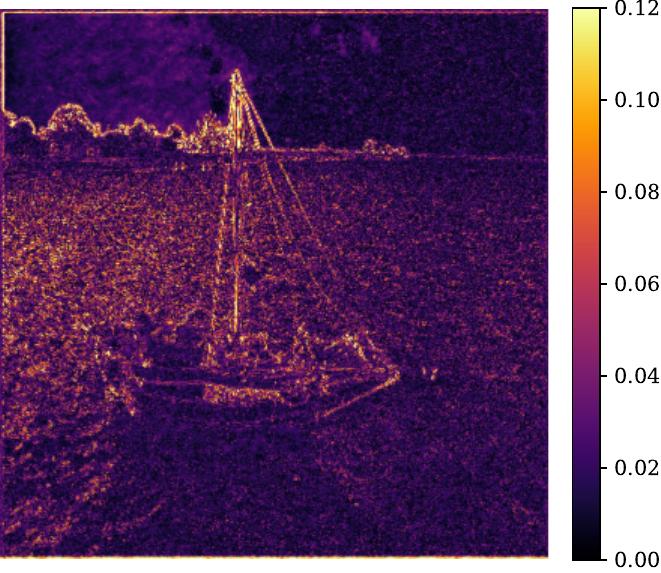
            }
        \end{minipage}
        \caption{PE}
    \end{subfigure}
    \begin{subfigure}{0.24\textwidth}
        \centering
        \begin{minipage}{\linewidth}
            \includegraphics[width=\linewidth]{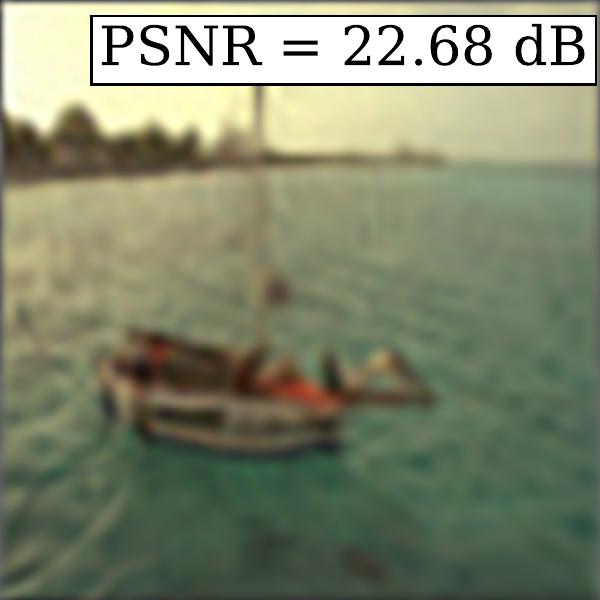}\\
            \includegraphics[width=\linewidth]{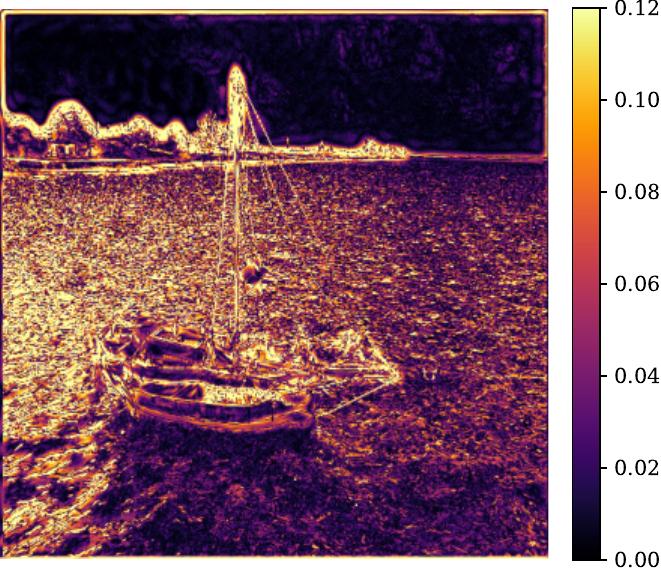}
        \end{minipage}
        \caption{Gauss}
    \end{subfigure}
    \begin{subfigure}{0.24\textwidth}
        \centering
        \begin{minipage}{\linewidth}
            \includegraphics[width=\linewidth]{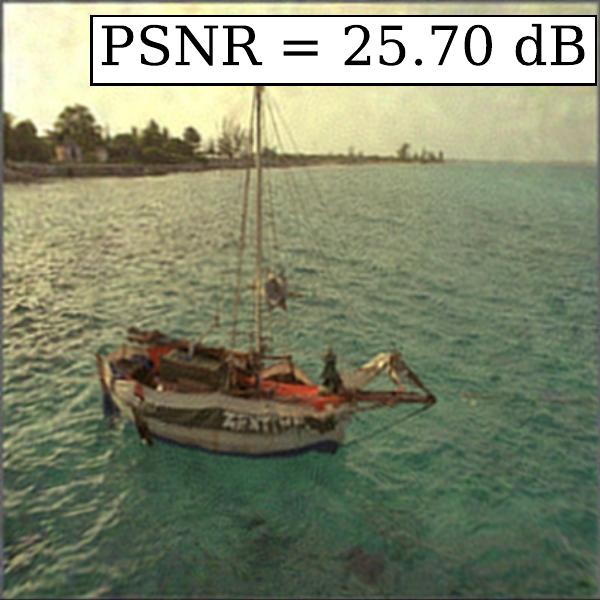}\\
            \includegraphics[width=\linewidth]{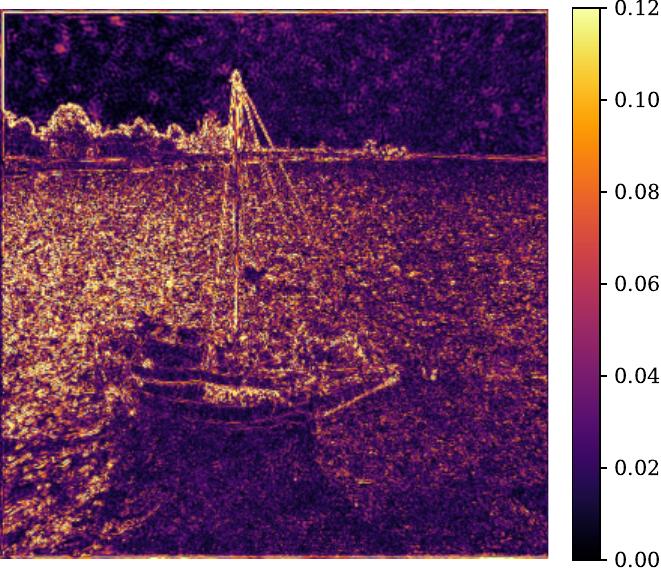}
        \end{minipage}
        \caption{WIRE}
    \end{subfigure}
    \begin{subfigure}{0.24\textwidth}
        \centering
        \begin{minipage}{\linewidth}
            \includegraphics[width=\linewidth]{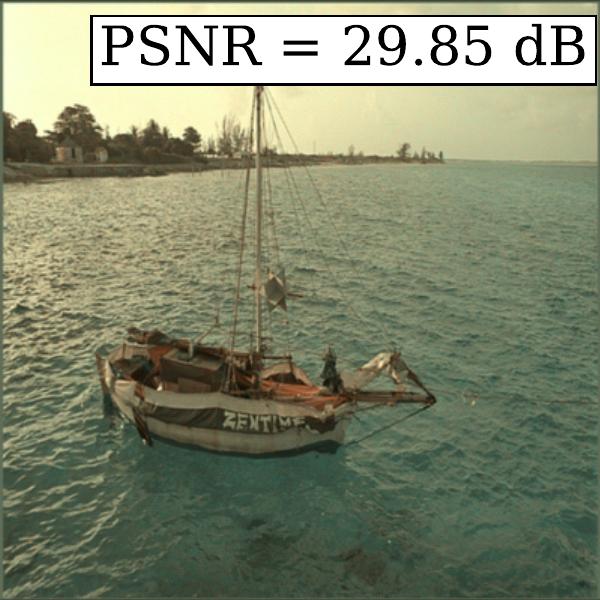}\\
            \includegraphics[width=\linewidth]{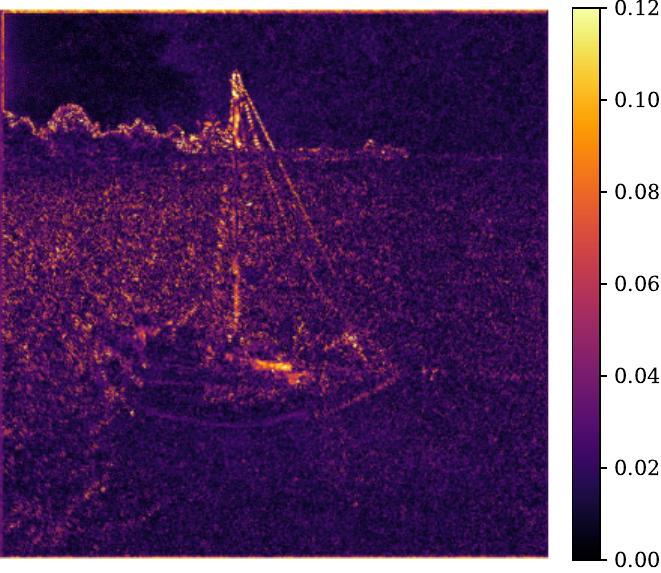}
        \end{minipage}
        \caption{TUNER}
    \end{subfigure}
    \begin{subfigure}{0.24\textwidth}
        \centering
        \begin{minipage}{\linewidth}
            \includegraphics[width=\linewidth]{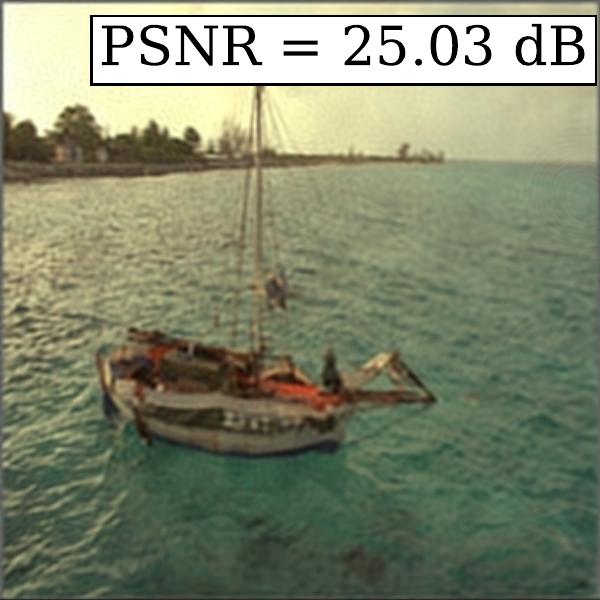}\\
            \includegraphics[width=\linewidth]{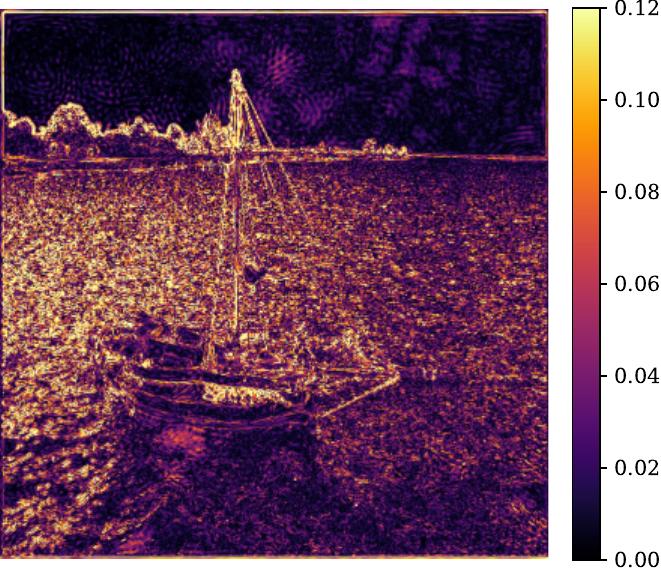}
        \end{minipage}
        \caption{FreSh}
    \end{subfigure}
    \begin{subfigure}{0.24\textwidth}
        \centering
        \begin{minipage}{\linewidth}
            \includegraphics[width=\linewidth]{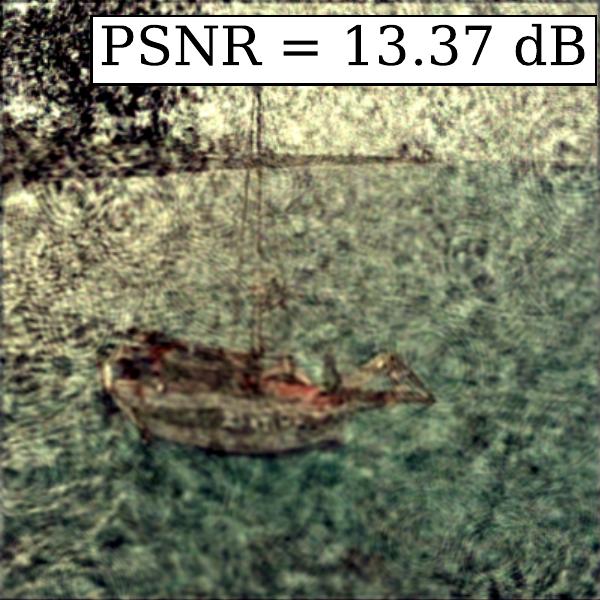}\\
            \includegraphics[width=\linewidth]{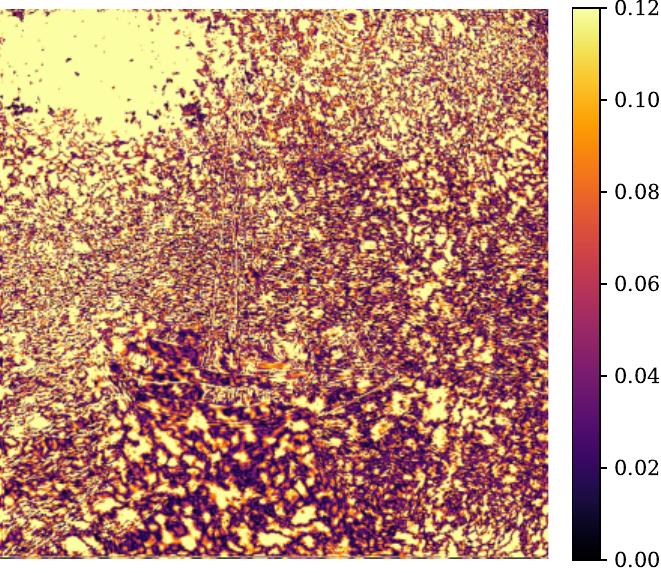}
        \end{minipage}
        \caption{SPDER}
    \end{subfigure}
    \begin{subfigure}{0.24\textwidth}
        \centering
        \begin{minipage}{\linewidth}
            \includegraphics[width=\linewidth]{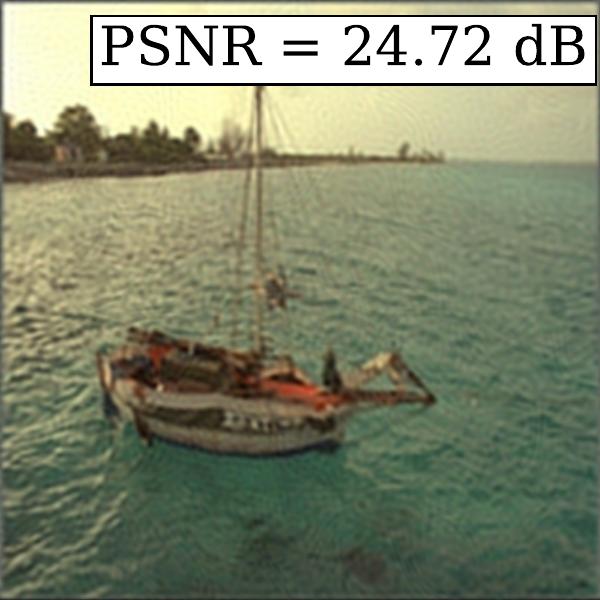}\\
            \includegraphics[width=\linewidth]{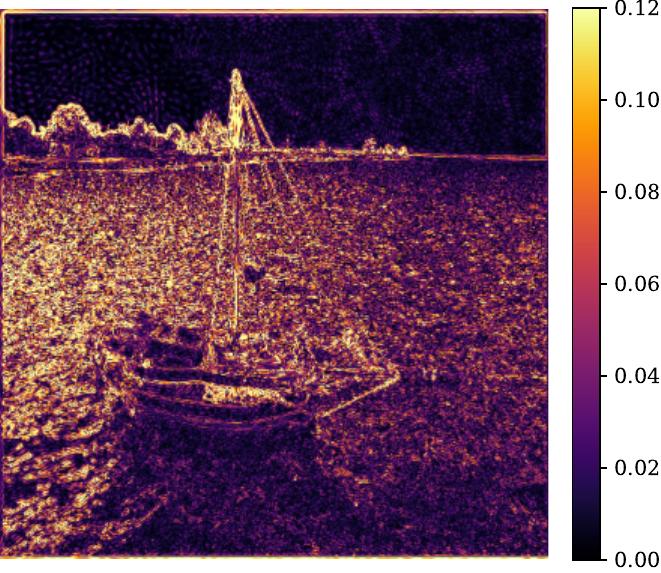}
        \end{minipage}
        \caption{MIRE}
    \end{subfigure}
    \begin{subfigure}{0.24\textwidth}
        \centering
        \begin{minipage}{\linewidth}
            \includegraphics[width=\linewidth]{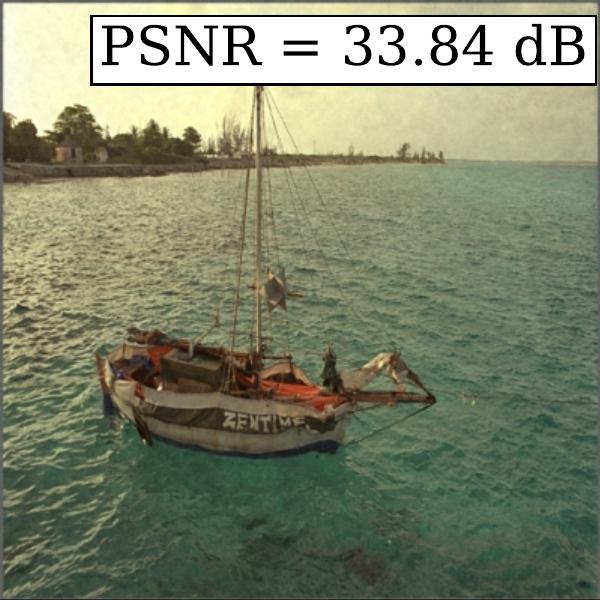}\\
            \includegraphics[width=\linewidth]{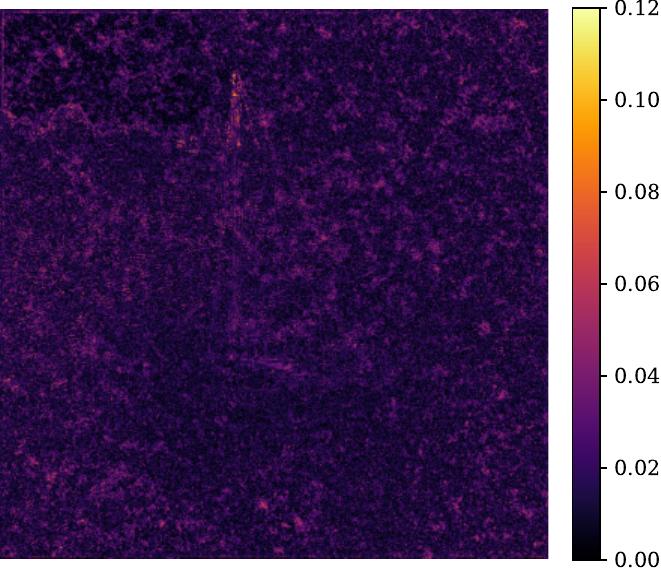}
        \end{minipage}
        \caption{FM-SIREN}
    \end{subfigure}
    \begin{subfigure}{0.24\textwidth}
        \centering
        \begin{minipage}{\linewidth}
            \includegraphics[width=\linewidth]{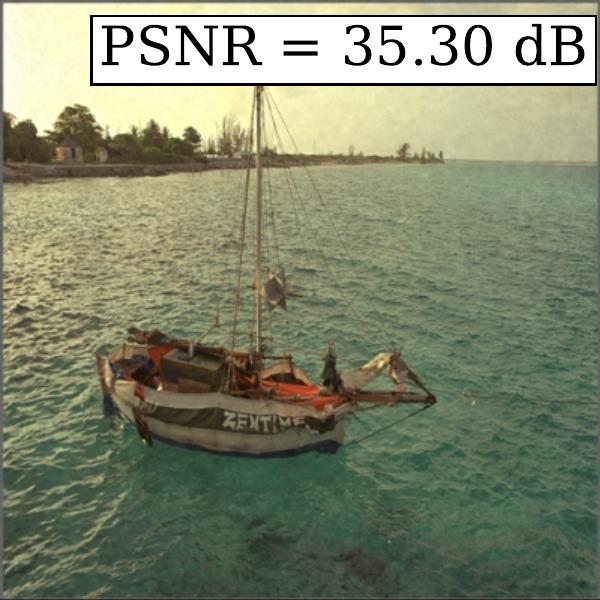}\\
            \includegraphics[width=\linewidth]{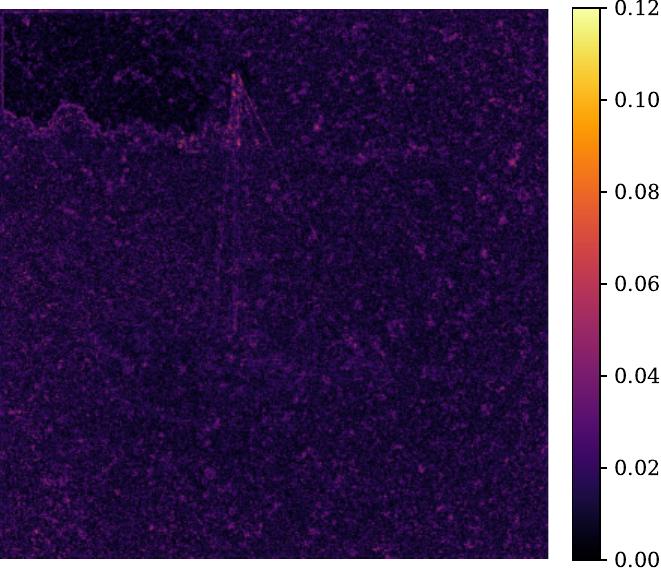}
        \end{minipage}
        \caption{FM-FINER}
    \end{subfigure}
    \caption{Qualitative results for kodim06 from Kodak Lossless True Color Image Suite \cite{kodak}. Top image in each subfigure is the reconstruction result while the bottom is the error map between reconstruction and ground truth. FM-SIREN and FM-FINER show significantly better reconstruction results as demonstrated by the PSNR values on the top-right of each subfigure. The error maps show that our models can learn both high frequency and low frequency structures with higher quality using small networks.}
    \label{fig:2Dfit_3}
\end{figure}

\begin{figure}[h]
    \centering
    \begin{subfigure}{0.24\textwidth}
        \centering
        \begin{minipage}{\linewidth}
            \includegraphics[width=\linewidth]{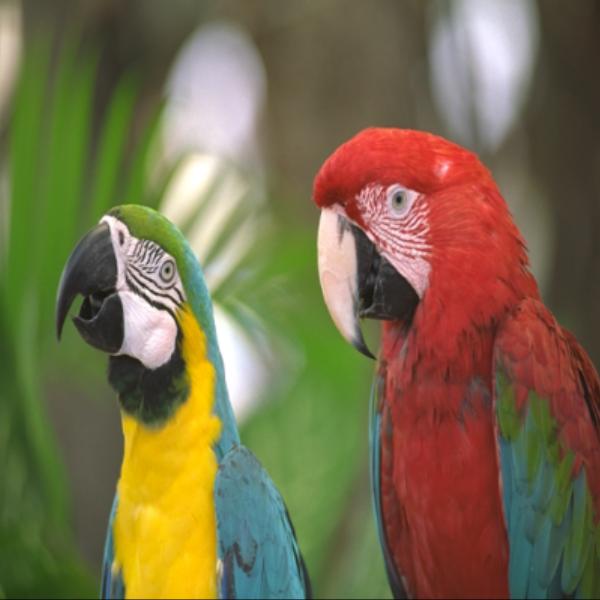}\\
            \includegraphics[width=\linewidth]{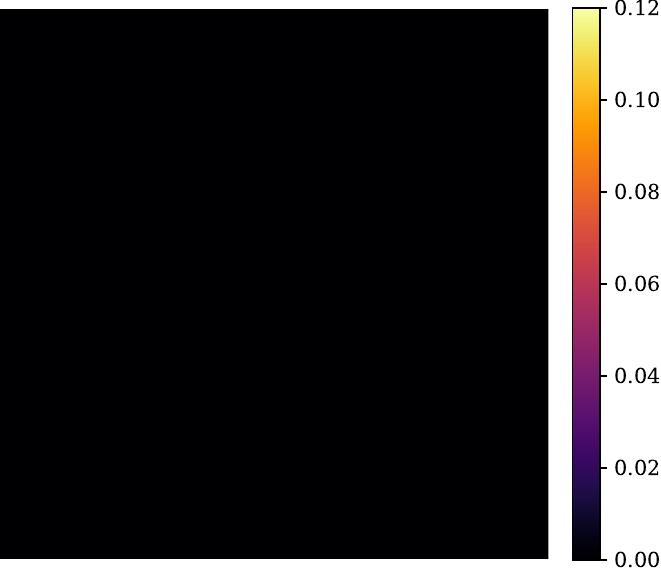}
        \end{minipage}
        \caption{Ground Truth}
    \end{subfigure}
    \begin{subfigure}{0.24\textwidth}
        \centering
        \begin{minipage}{\linewidth}
            \includegraphics[width=\linewidth]{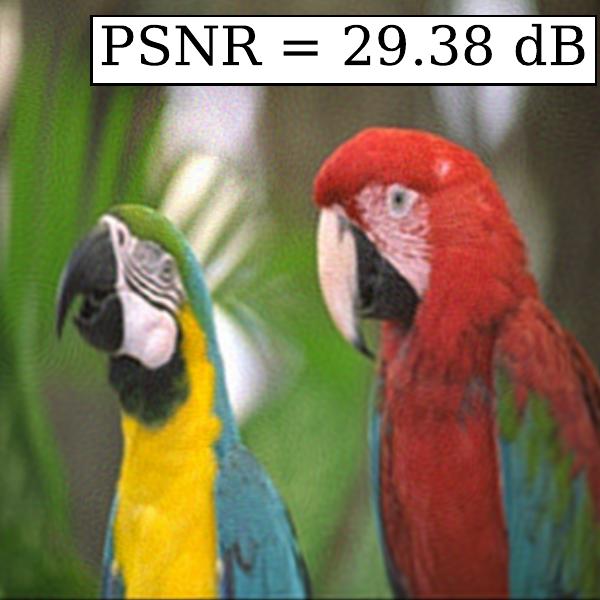}\\
            \includegraphics[width=\linewidth]{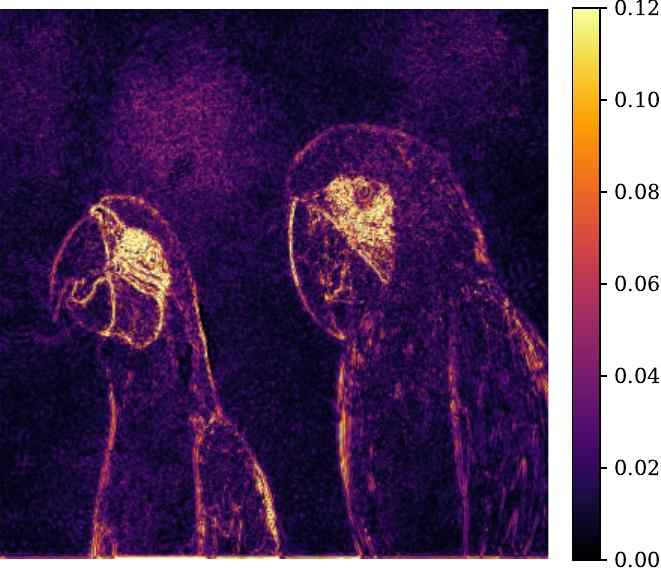}
        \end{minipage}
        \caption{SIREN}
    \end{subfigure}
    \begin{subfigure}{0.24\textwidth}
        \centering
        \begin{minipage}{\linewidth}
            \includegraphics[width=\linewidth]{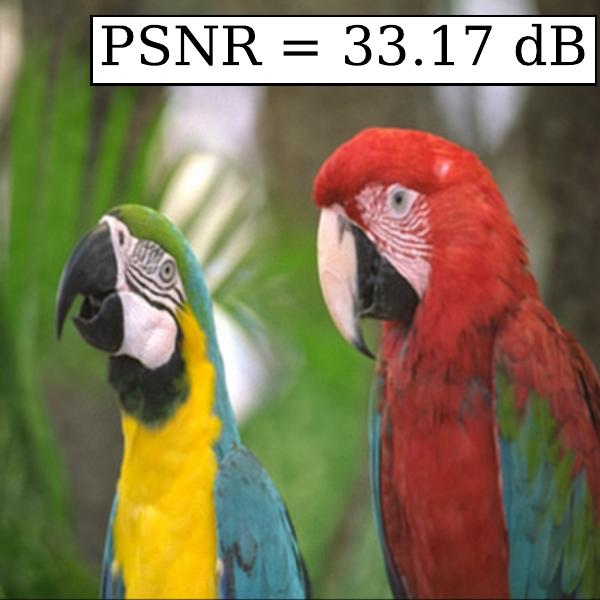}\\
            \includegraphics[width=\linewidth]{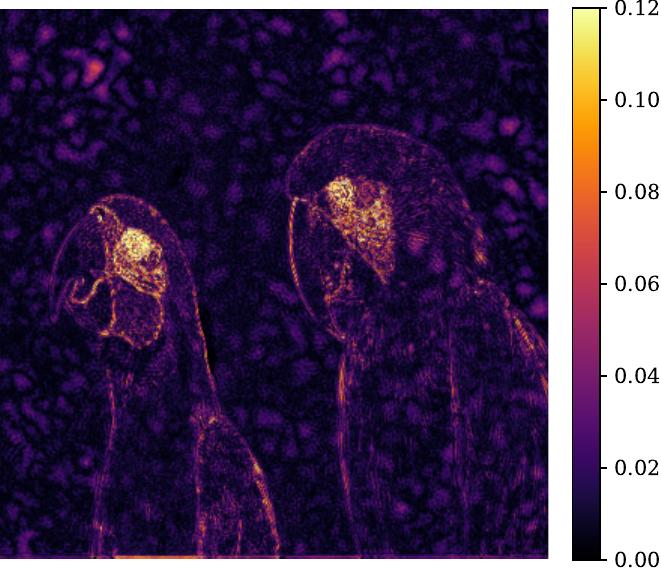}
        \end{minipage}
        \caption{FINER}
    \end{subfigure}
    \begin{subfigure}{0.24\textwidth}
        \centering
        \begin{minipage}{\linewidth}
            \includegraphics[width=\linewidth]{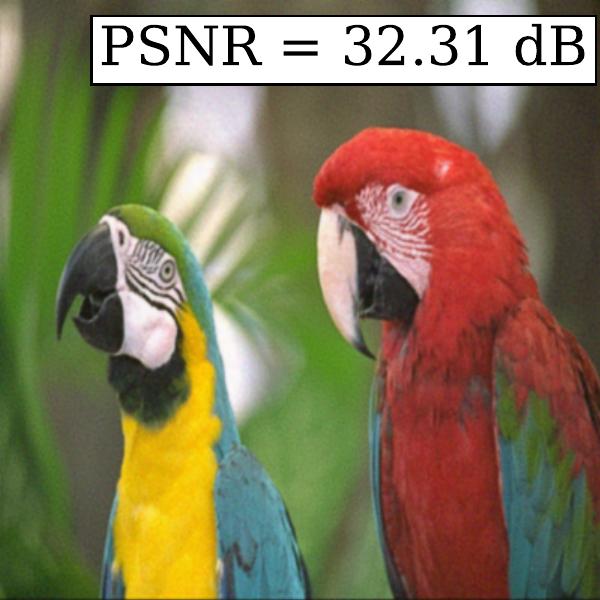}\\
            \includegraphics[width=\linewidth]{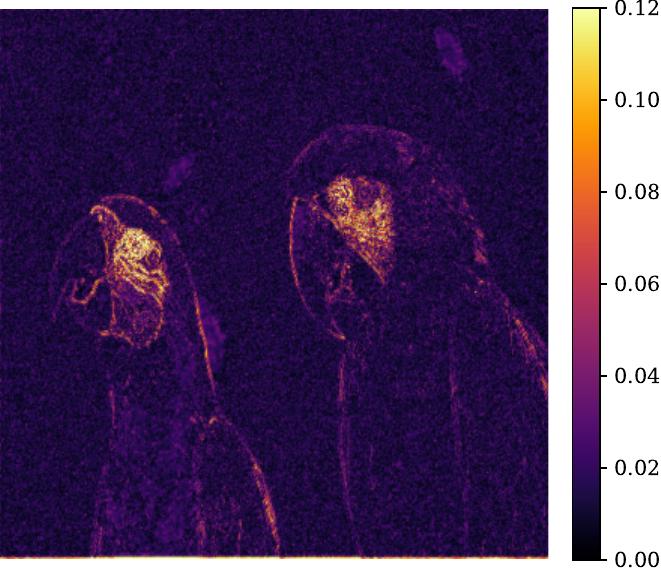
            }
        \end{minipage}
        \caption{PE}
    \end{subfigure}
    \begin{subfigure}{0.24\textwidth}
        \centering
        \begin{minipage}{\linewidth}
            \includegraphics[width=\linewidth]{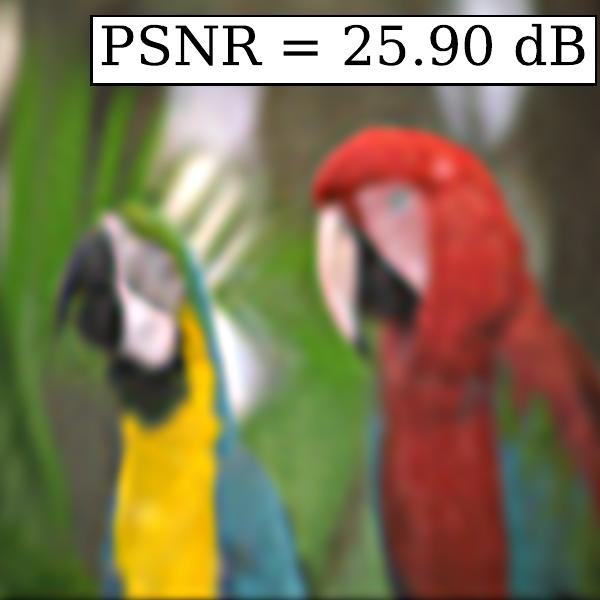}\\
            \includegraphics[width=\linewidth]{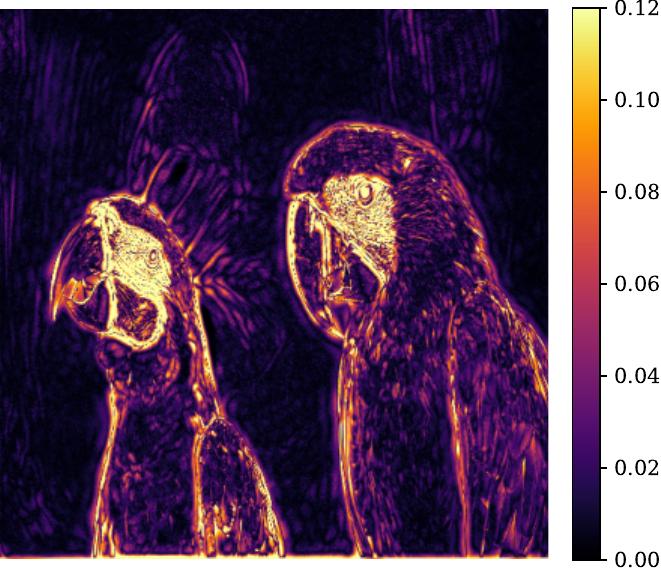}
        \end{minipage}
        \caption{Gauss}
    \end{subfigure}
    \begin{subfigure}{0.24\textwidth}
        \centering
        \begin{minipage}{\linewidth}
            \includegraphics[width=\linewidth]{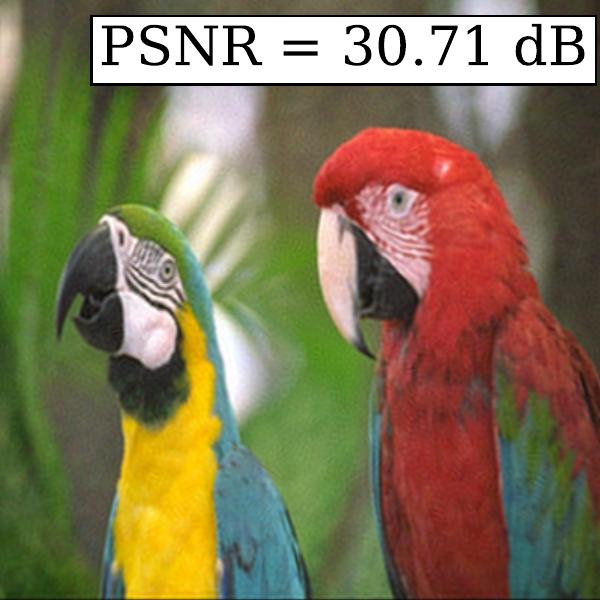}\\
            \includegraphics[width=\linewidth]{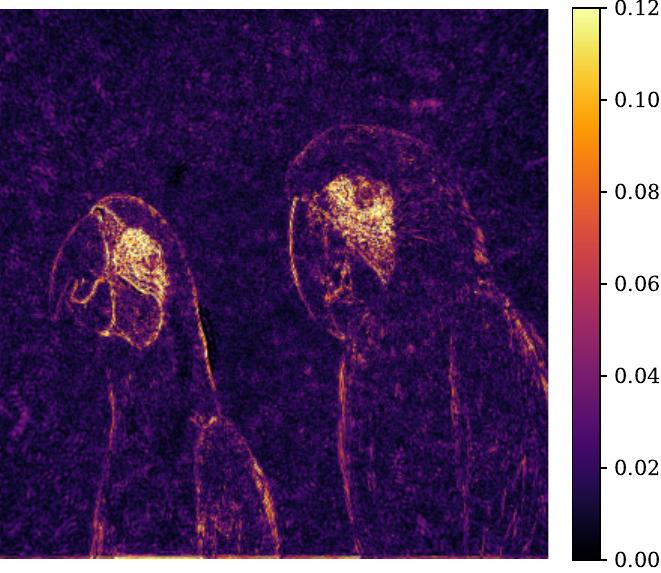}
        \end{minipage}
        \caption{WIRE}
    \end{subfigure}
    \begin{subfigure}{0.24\textwidth}
        \centering
        \begin{minipage}{\linewidth}
            \includegraphics[width=\linewidth]{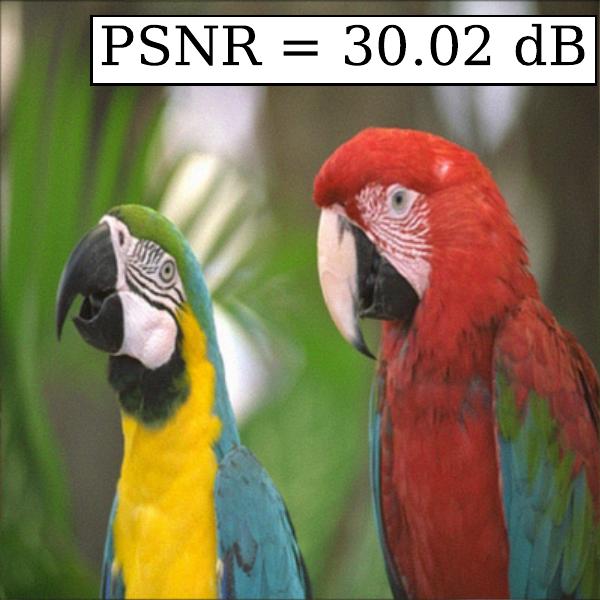}\\
            \includegraphics[width=\linewidth]{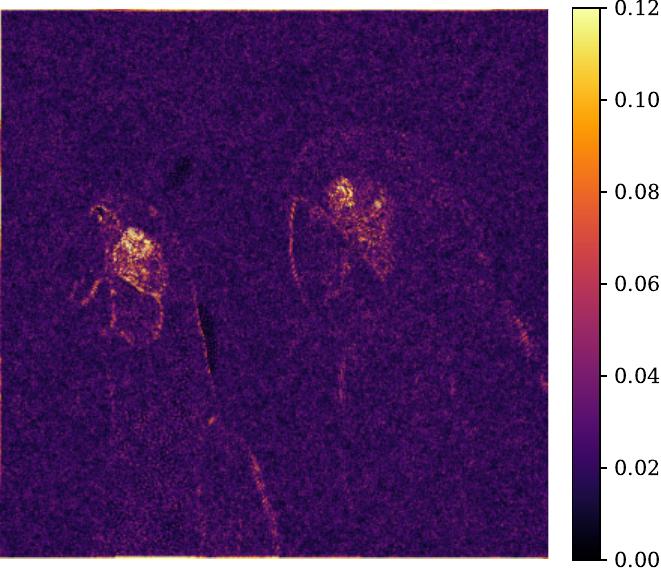}
        \end{minipage}
        \caption{TUNER}
    \end{subfigure}
    \begin{subfigure}{0.24\textwidth}
        \centering
        \begin{minipage}{\linewidth}
            \includegraphics[width=\linewidth]{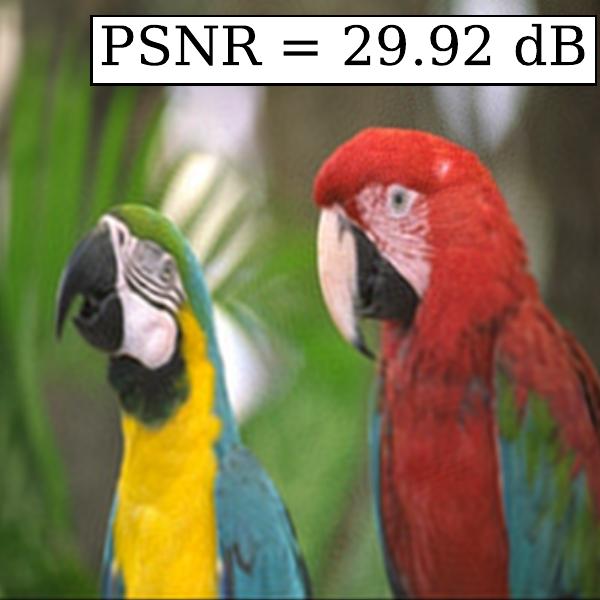}\\
            \includegraphics[width=\linewidth]{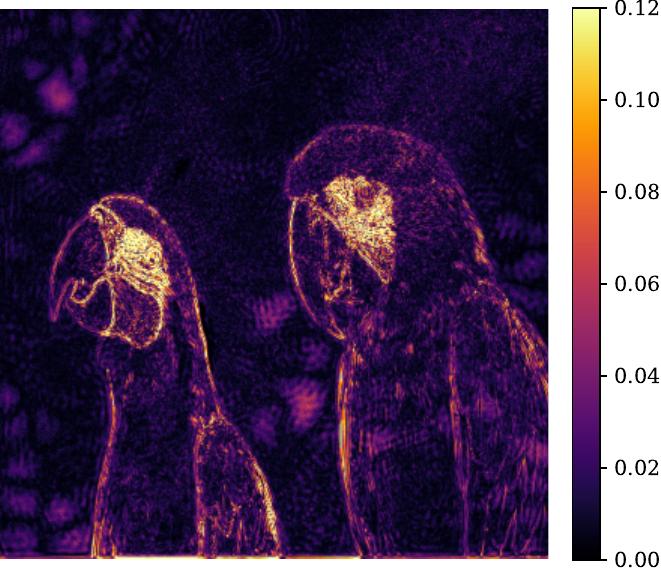}
        \end{minipage}
        \caption{FreSh}
    \end{subfigure}
    \begin{subfigure}{0.24\textwidth}
        \centering
        \begin{minipage}{\linewidth}
            \includegraphics[width=\linewidth]{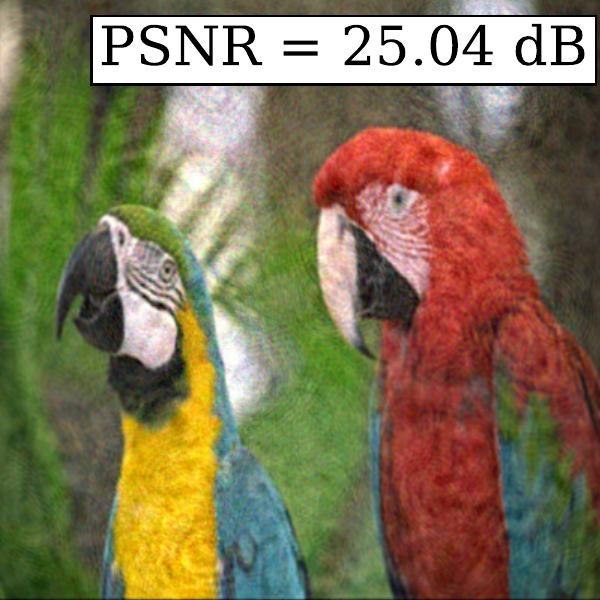}\\
            \includegraphics[width=\linewidth]{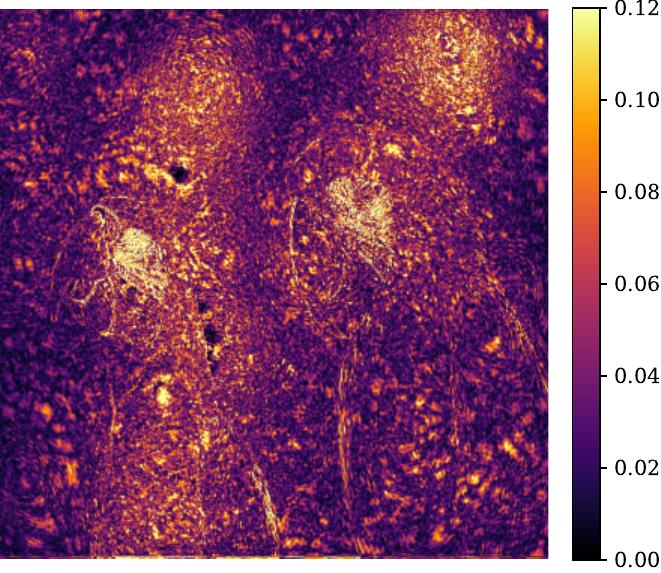}
        \end{minipage}
        \caption{SPDER}
    \end{subfigure}
    \begin{subfigure}{0.24\textwidth}
        \centering
        \begin{minipage}{\linewidth}
            \includegraphics[width=\linewidth]{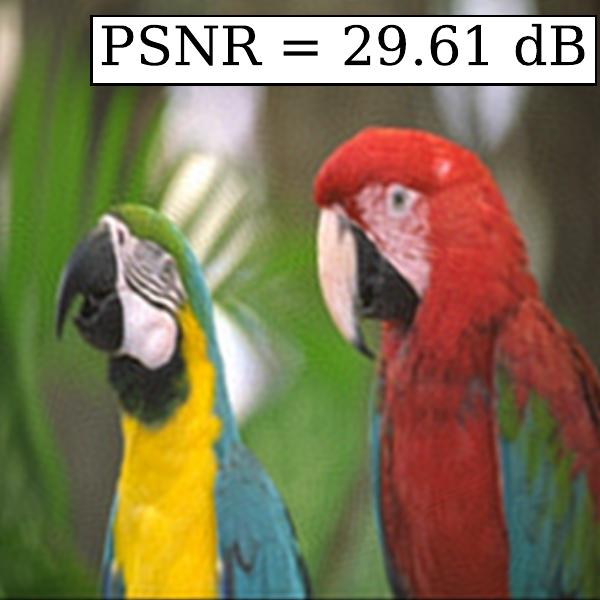}\\
            \includegraphics[width=\linewidth]{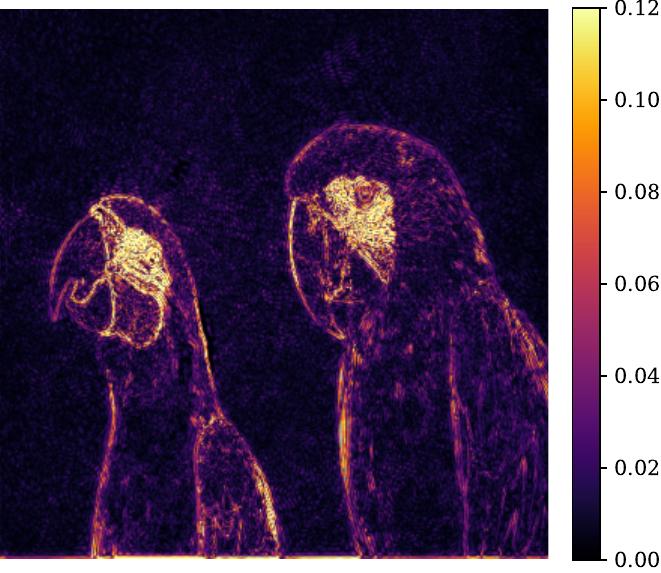}
        \end{minipage}
        \caption{MIRE}
    \end{subfigure}
    \begin{subfigure}{0.24\textwidth}
        \centering
        \begin{minipage}{\linewidth}
            \includegraphics[width=\linewidth]{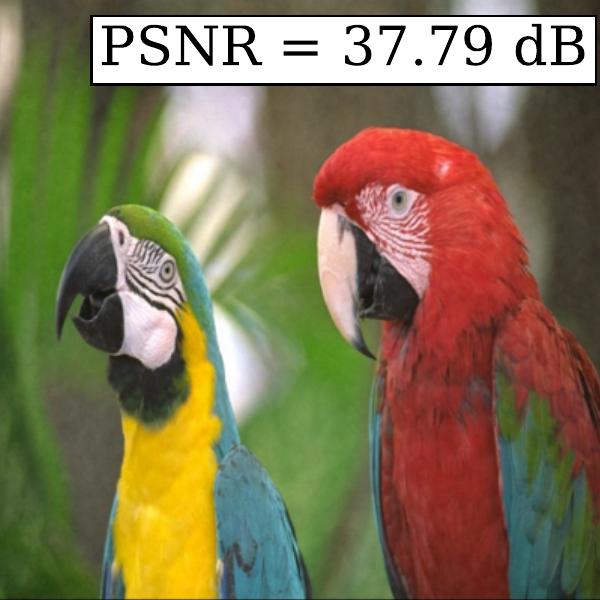}\\
            \includegraphics[width=\linewidth]{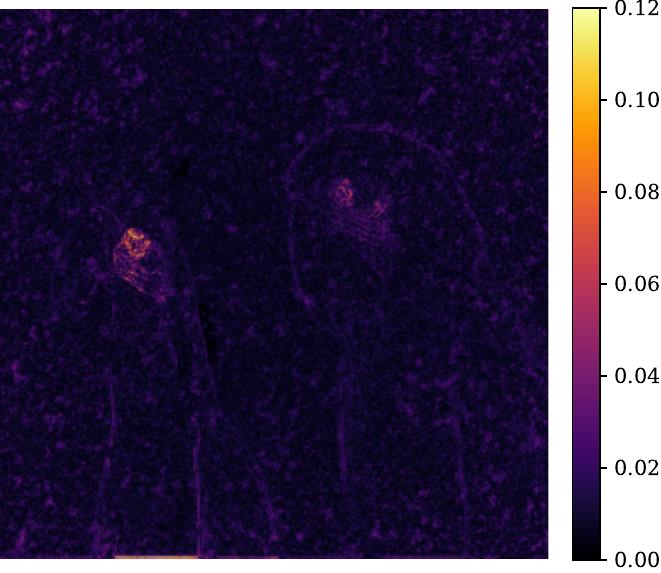}
        \end{minipage}
        \caption{FM-SIREN}
    \end{subfigure}
    \begin{subfigure}{0.24\textwidth}
        \centering
        \begin{minipage}{\linewidth}
            \includegraphics[width=\linewidth]{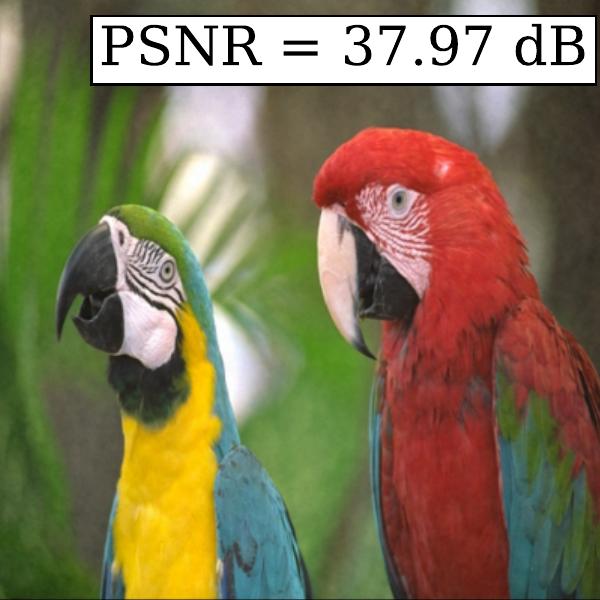}\\
            \includegraphics[width=\linewidth]{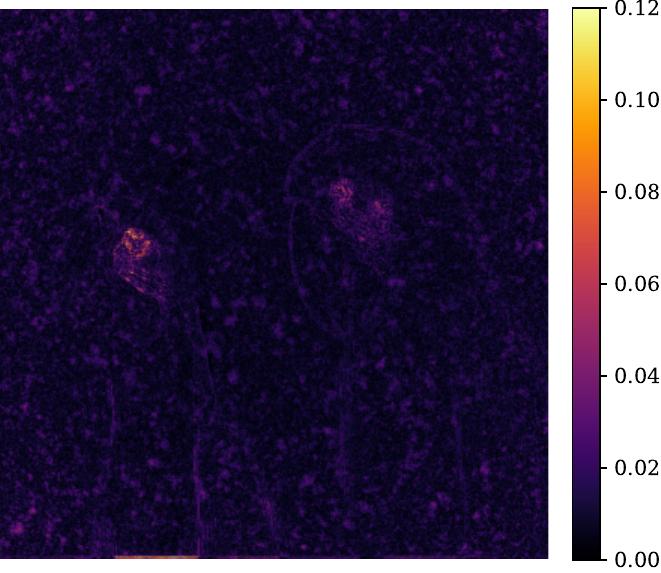}
        \end{minipage}
        \caption{FM-FINER}
    \end{subfigure}
    \caption{Qualitative results for kodim23 from Kodak Lossless True Color Image Suite \cite{kodak}. Top image in each subfigure is the reconstruction result while the bottom is the error map between reconstruction and ground truth. FM-SIREN and FM-FINER show significantly better reconstruction results as demonstrated by the PSNR values on the top-right of each subfigure. The error maps show that our models can learn both high frequency and low frequency structures with higher quality using small networks.}
    \label{fig:2Dfit_4}
\end{figure}

\begin{figure}[h]
    \centering
    \begin{subfigure}{0.24\textwidth}
        \includegraphics[width=\linewidth]{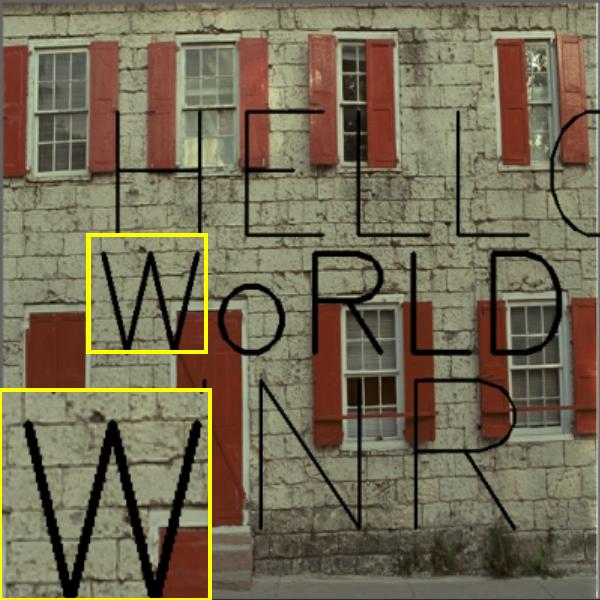}
        \caption{Masked Image}
    \end{subfigure}
    \begin{subfigure}{0.24\textwidth}
        \includegraphics[width=\linewidth]{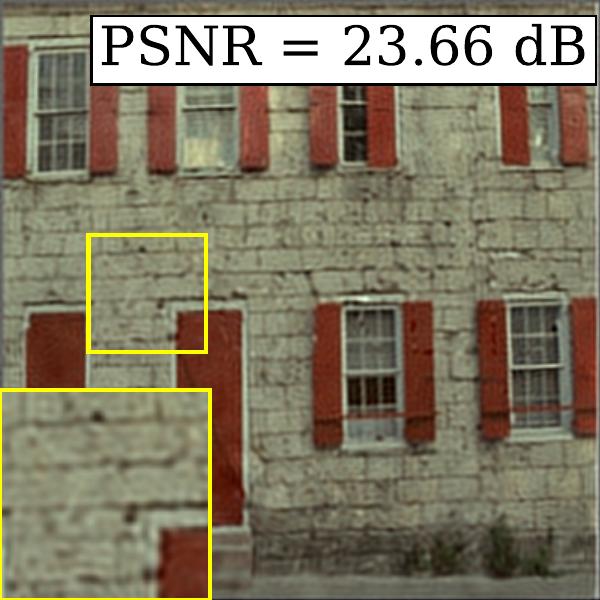}
        \caption{SIREN}
    \end{subfigure}
    \begin{subfigure}{0.24\textwidth}
        \includegraphics[width=\linewidth]{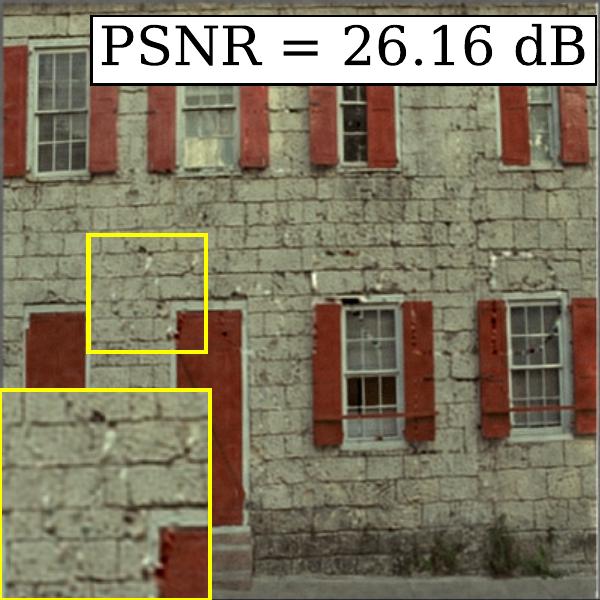}
        \caption{FINER}
    \end{subfigure}
    \begin{subfigure}{0.24\textwidth}
        \includegraphics[width=\linewidth]{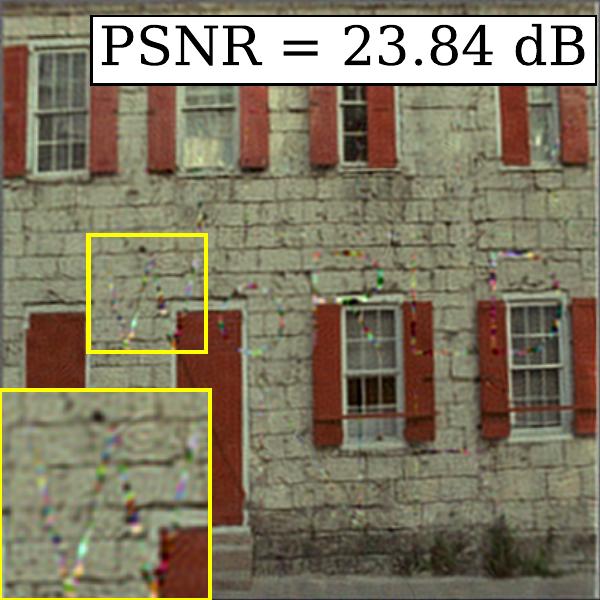}
        \caption{WIRE}
    \end{subfigure}
    \begin{subfigure}{0.24\textwidth}
        \includegraphics[width=\linewidth]{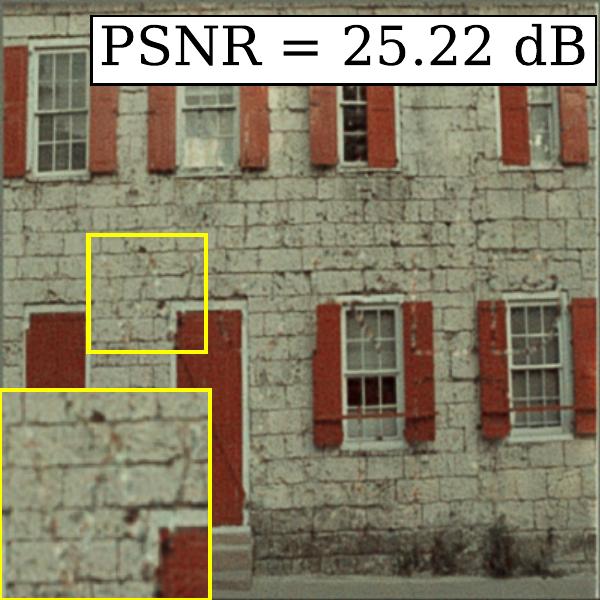}
        \caption{PE}
    \end{subfigure}
    \begin{subfigure}{0.24\textwidth}
        \includegraphics[width=\linewidth]{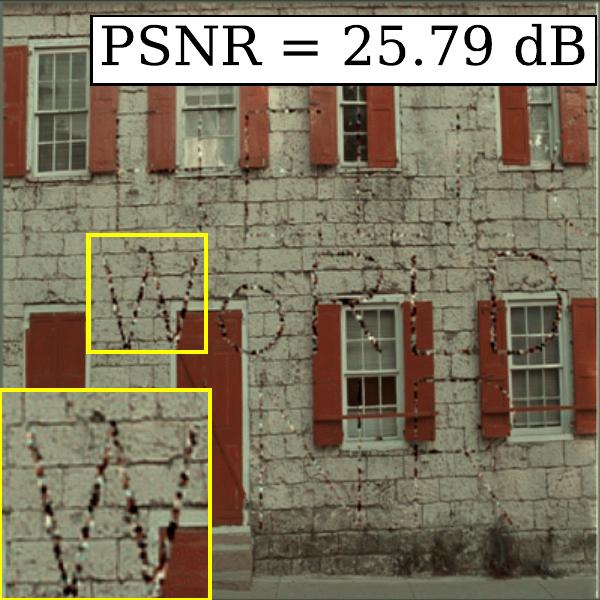}
        \caption{TUNER}
    \end{subfigure}
    \begin{subfigure}{0.24\textwidth}
        \includegraphics[width=\linewidth]{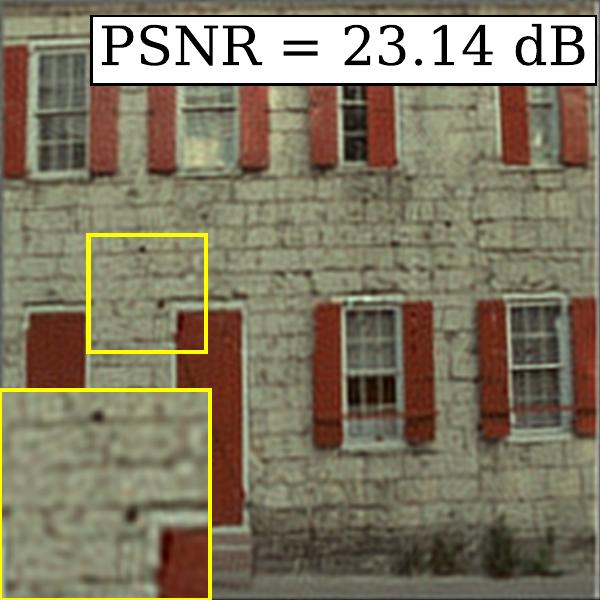}
        \caption{MIRE}
    \end{subfigure}
    \begin{subfigure}{0.24\textwidth}
        \includegraphics[width=\linewidth]{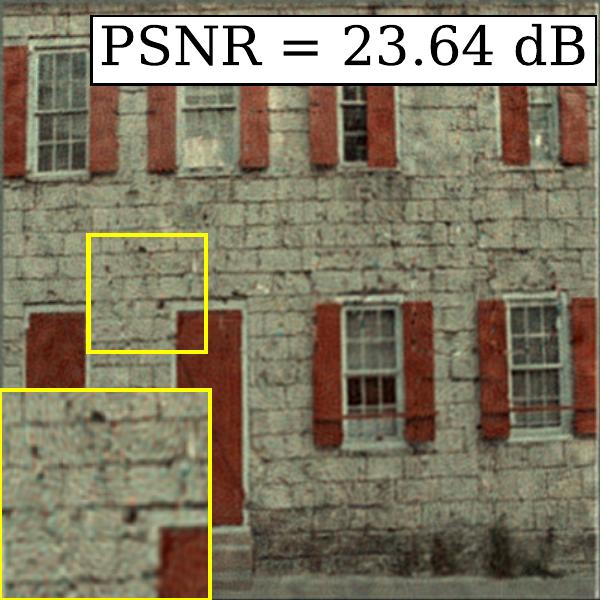}
        \caption{SPDER}
    \end{subfigure}
    \begin{subfigure}{0.24\textwidth}
        \includegraphics[width=\linewidth]{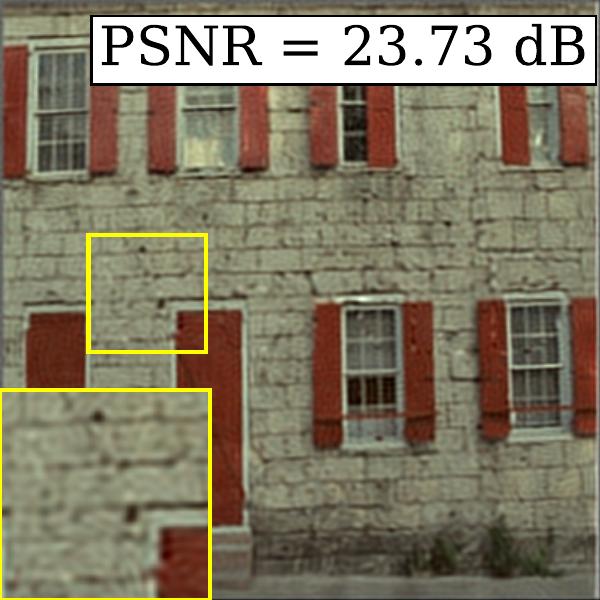}
        \caption{FreSh}
    \end{subfigure}
    \begin{subfigure}{0.24\textwidth}
        \includegraphics[width=\linewidth]{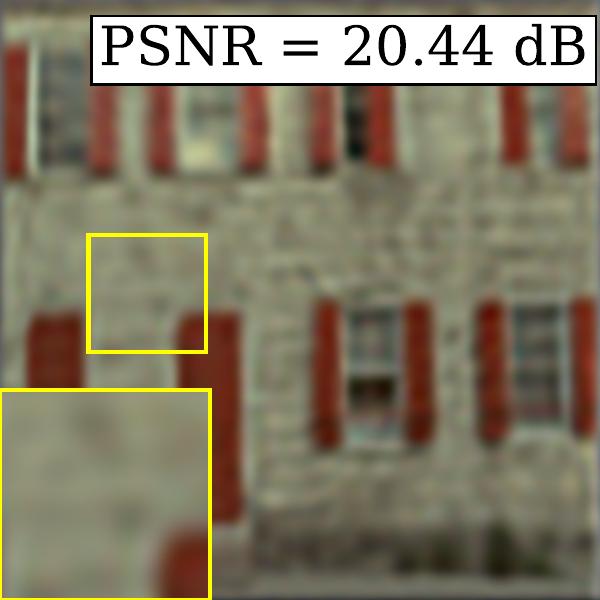}
        \caption{Gauss}
    \end{subfigure}
    \begin{subfigure}{0.24\textwidth}
        \includegraphics[width=\linewidth]{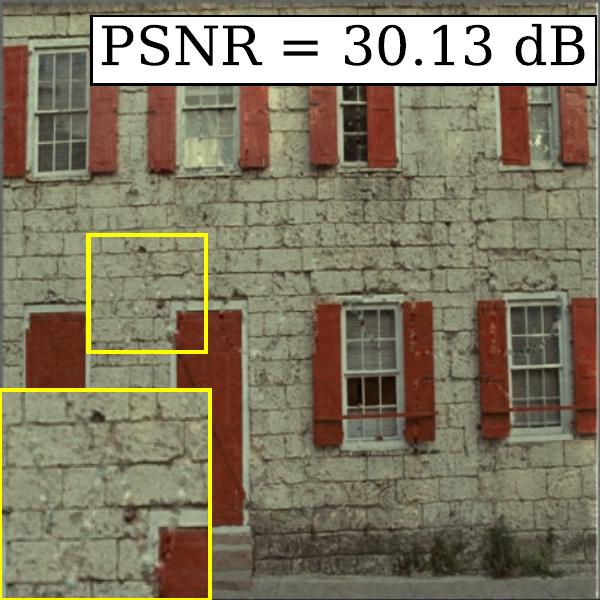}
        \caption{FM-SIREN}
    \end{subfigure}
    \begin{subfigure}{0.24\textwidth}
        \includegraphics[width=\linewidth]{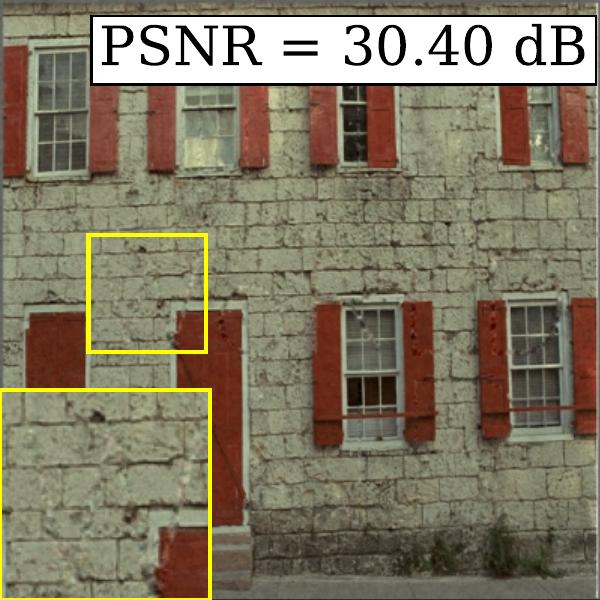}
        \caption{FM-FINER}
    \end{subfigure}
    \caption{Qualitative inpainting results of kodim01 for all baselines. Observe that the PSNR of the proposed FM-FINER and FM-SIREN schemes are superior to all other baselines, demonstrating excellent interpolation for missing pixel while preserving high reconstruction fidelity.}
    \label{fig:inpainting_qual_1}
\end{figure}

\begin{figure}[h]
    \centering
    \begin{subfigure}{0.24\textwidth}
        \includegraphics[width=\linewidth]{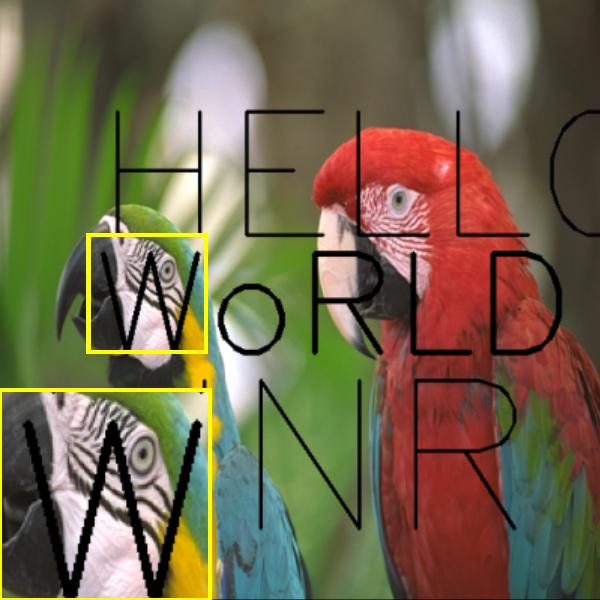}
        \caption{Masked Image}
    \end{subfigure}
    \begin{subfigure}{0.24\textwidth}
        \includegraphics[width=\linewidth]{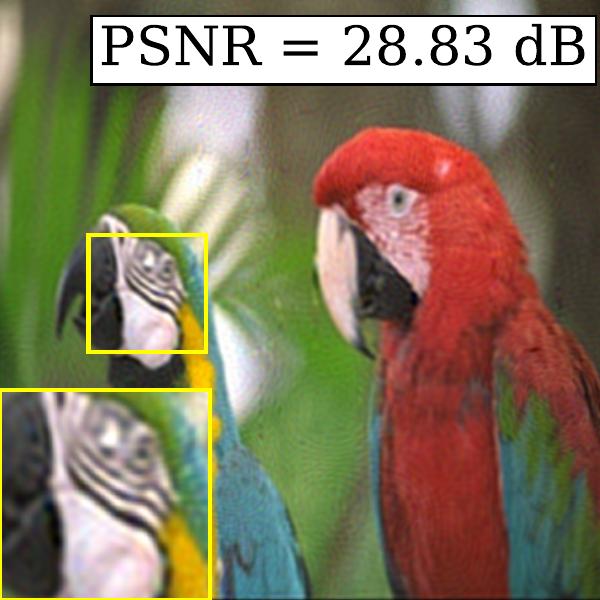}
        \caption{SIREN}
    \end{subfigure}
    \begin{subfigure}{0.24\textwidth}
        \includegraphics[width=\linewidth]{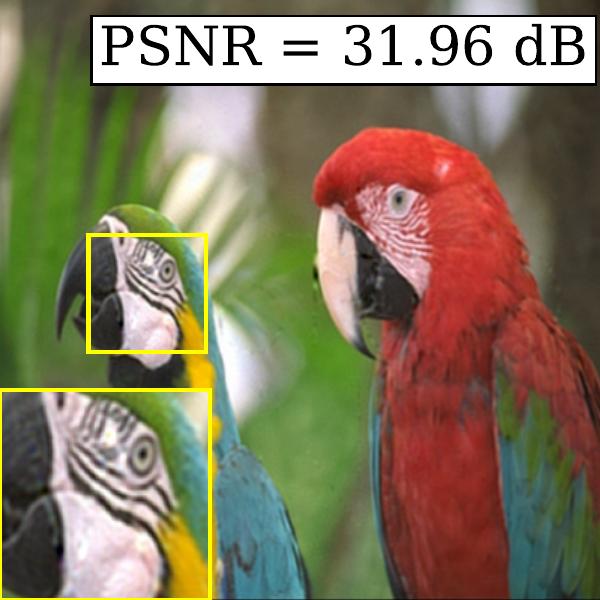}
        \caption{FINER}
    \end{subfigure}
    \begin{subfigure}{0.24\textwidth}
        \includegraphics[width=\linewidth]{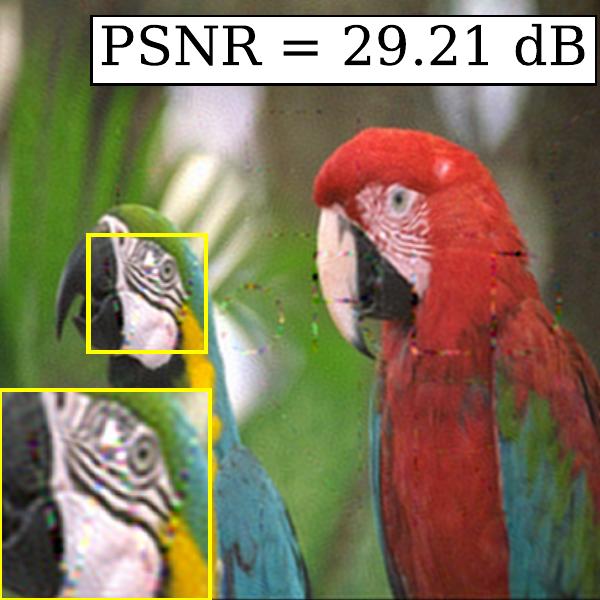}
        \caption{WIRE}
    \end{subfigure}
    \begin{subfigure}{0.24\textwidth}
        \includegraphics[width=\linewidth]{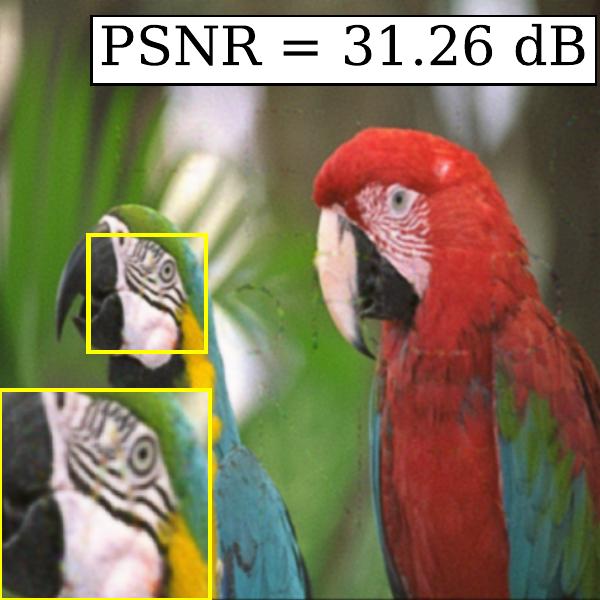}
        \caption{PE}
    \end{subfigure}
    \begin{subfigure}{0.24\textwidth}
        \includegraphics[width=\linewidth]{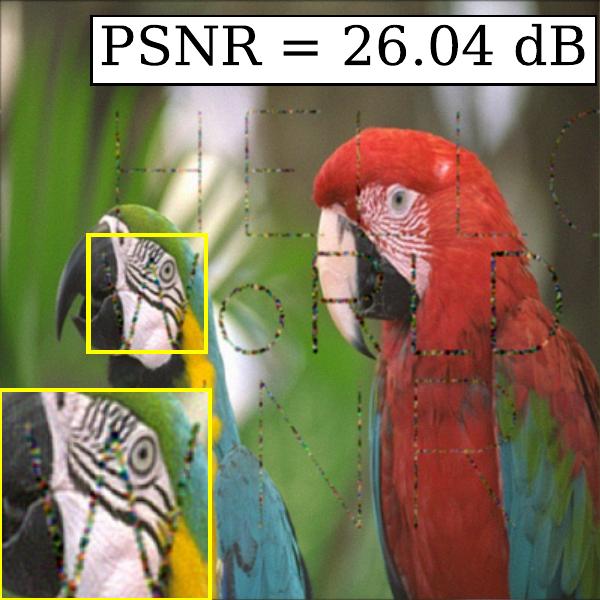}
        \caption{TUNER}
    \end{subfigure}
    \begin{subfigure}{0.24\textwidth}
        \includegraphics[width=\linewidth]{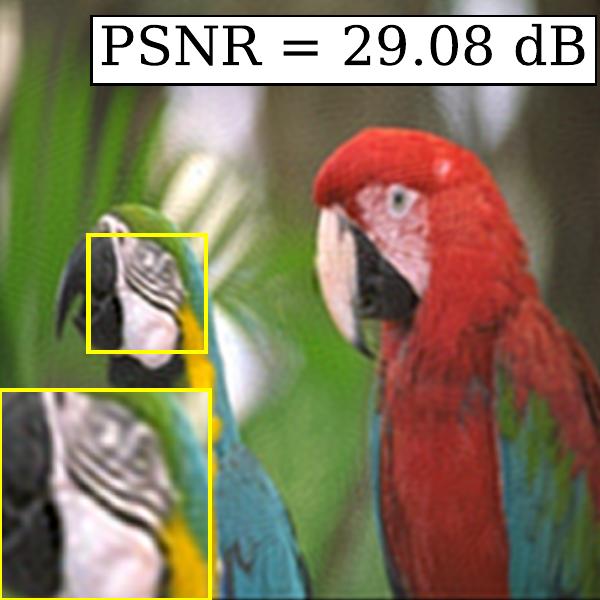}
        \caption{MIRE}
    \end{subfigure}
    \begin{subfigure}{0.24\textwidth}
        \includegraphics[width=\linewidth]{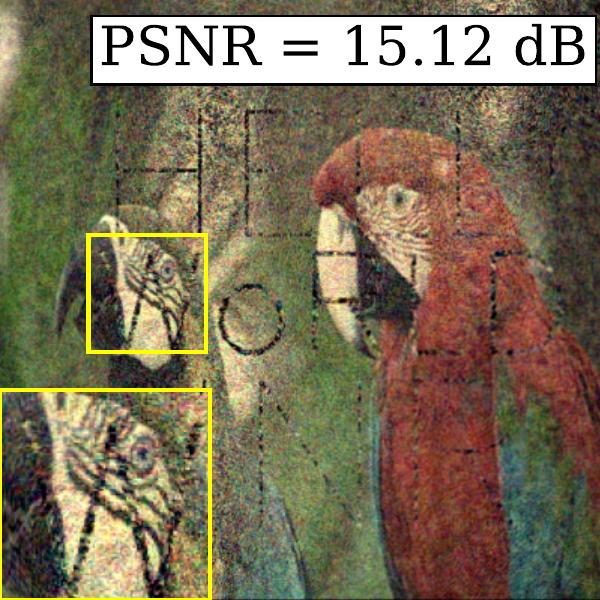}
        \caption{SPDER}
    \end{subfigure}
    \begin{subfigure}{0.24\textwidth}
        \includegraphics[width=\linewidth]{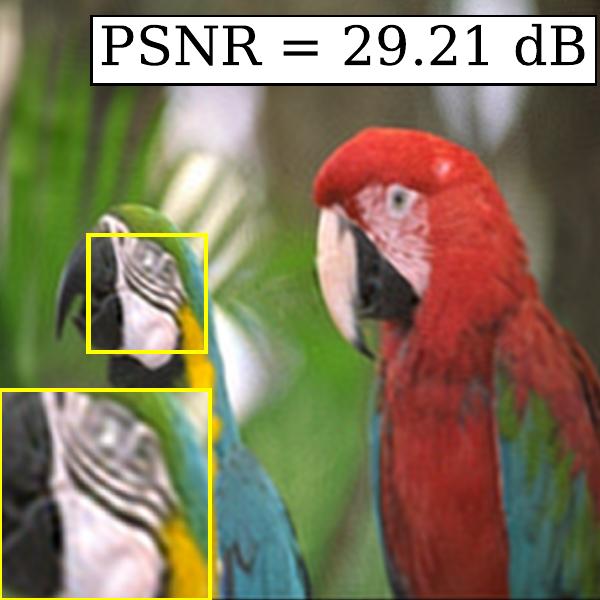}
        \caption{FreSh}
    \end{subfigure}
    \begin{subfigure}{0.24\textwidth}
        \includegraphics[width=\linewidth]{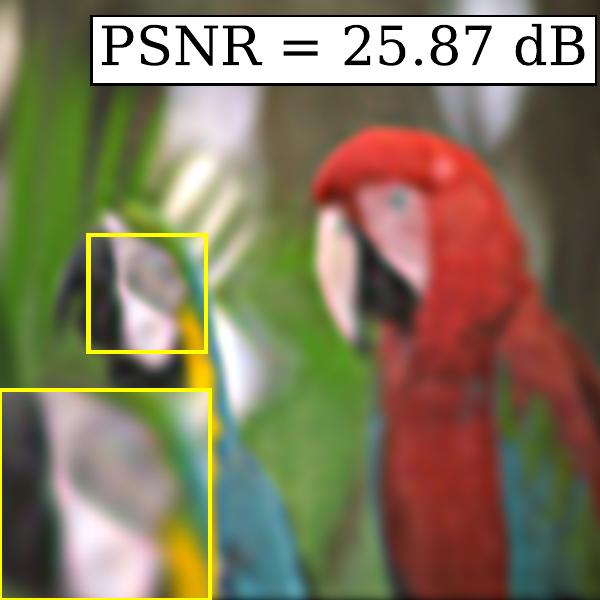}
        \caption{Gauss}
    \end{subfigure}
    \begin{subfigure}{0.24\textwidth}
        \includegraphics[width=\linewidth]{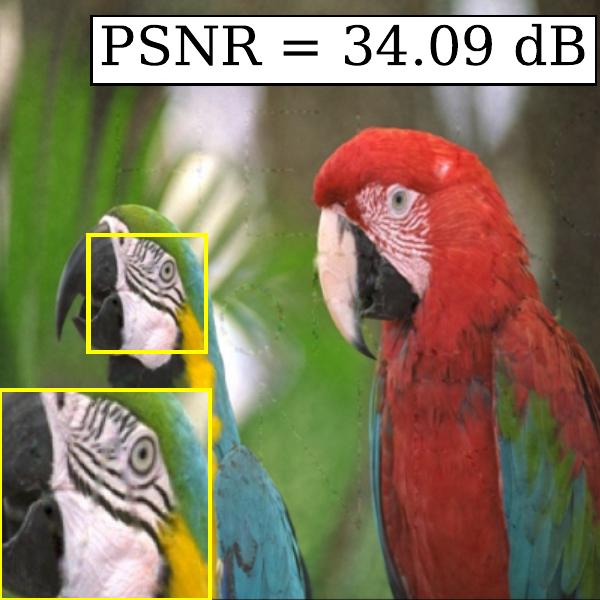}
        \caption{FM-SIREN}
    \end{subfigure}
    \begin{subfigure}{0.24\textwidth}
        \includegraphics[width=\linewidth]{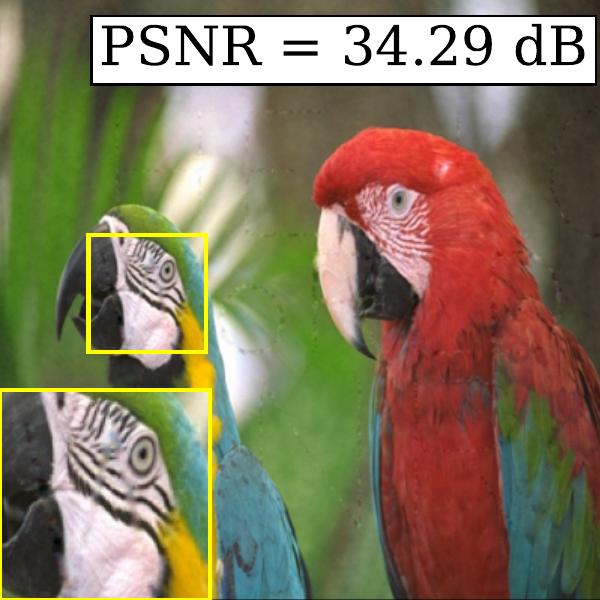}
        \caption{FM-FINER}
    \end{subfigure}
    \caption{Qualitative inpainting results of kodim23 for all baselines. Observe that the PSNR of the proposed FM-FINER and FM-SIREN schemes are superior to all other baselines, demonstrating excellent interpolation for missing pixel while preserving high reconstruction fidelity.}
    \label{fig:inpainting_qual_2}
\end{figure}

\begin{figure}[h]
    \centering
    \begin{subfigure}{0.24\textwidth}
        \includegraphics[width=\linewidth]{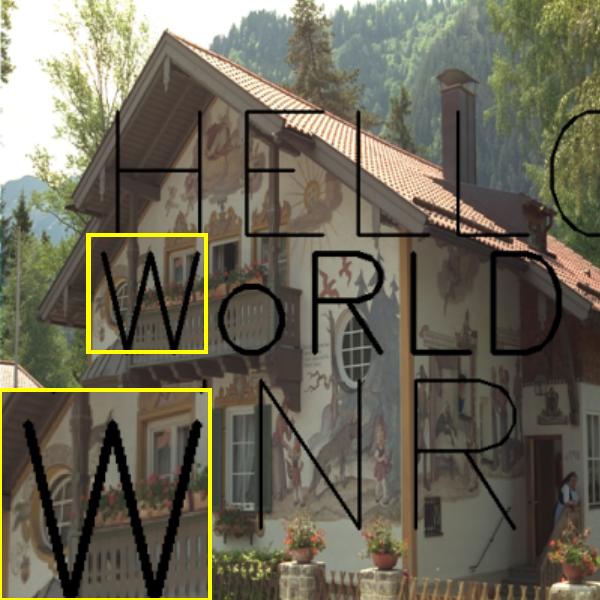}
        \caption{Masked Image}
    \end{subfigure}
    \begin{subfigure}{0.24\textwidth}
        \includegraphics[width=\linewidth]{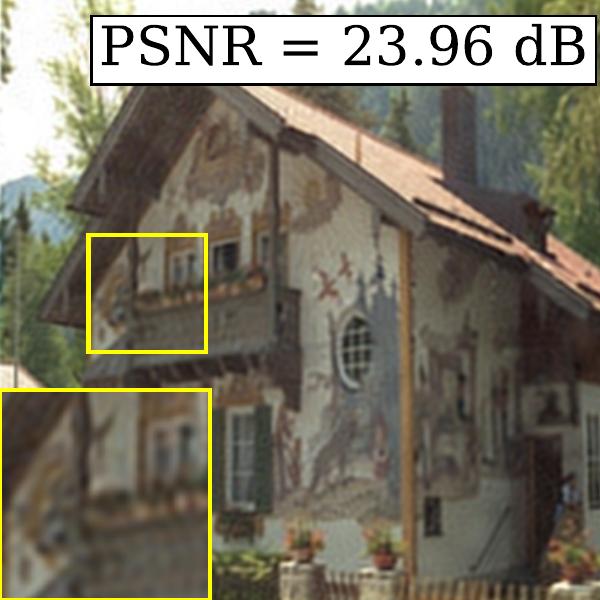}
        \caption{SIREN}
    \end{subfigure}
    \begin{subfigure}{0.24\textwidth}
        \includegraphics[width=\linewidth]{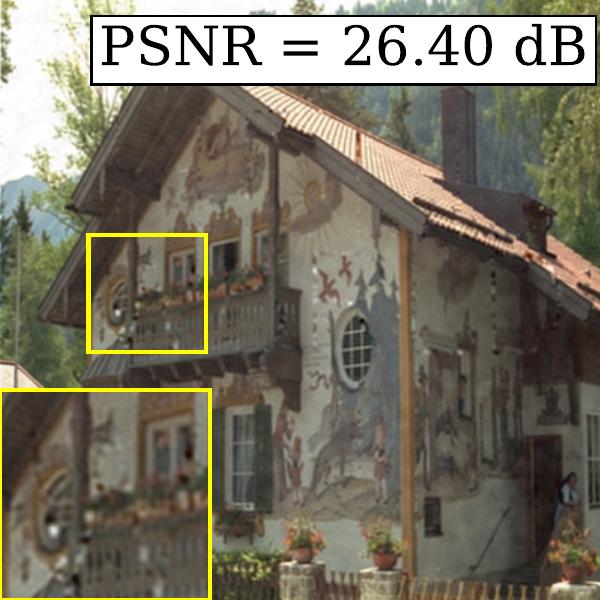}
        \caption{FINER}
    \end{subfigure}
    \begin{subfigure}{0.24\textwidth}
        \includegraphics[width=\linewidth]{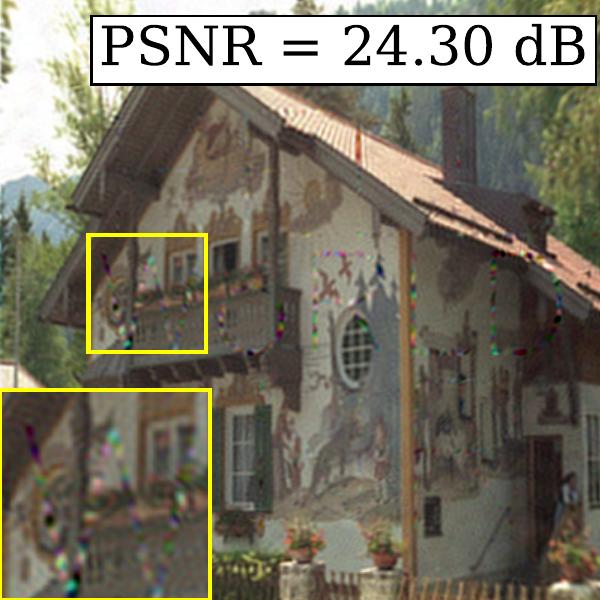}
        \caption{WIRE}
    \end{subfigure}
    \begin{subfigure}{0.24\textwidth}
        \includegraphics[width=\linewidth]{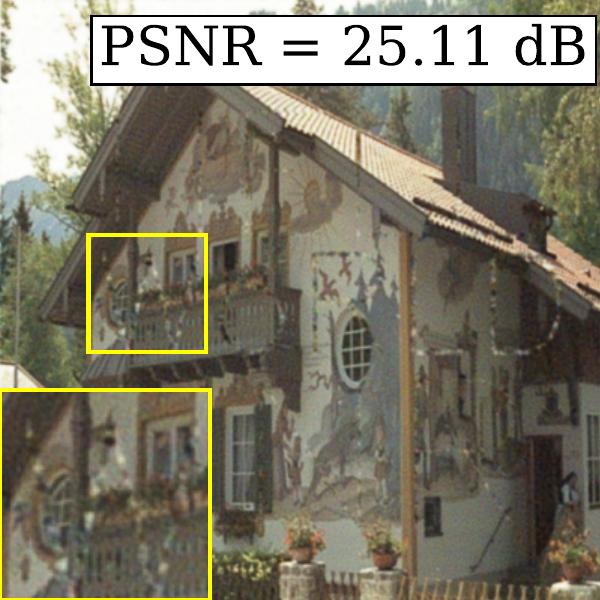}
        \caption{PE}
    \end{subfigure}
    \begin{subfigure}{0.24\textwidth}
        \includegraphics[width=\linewidth]{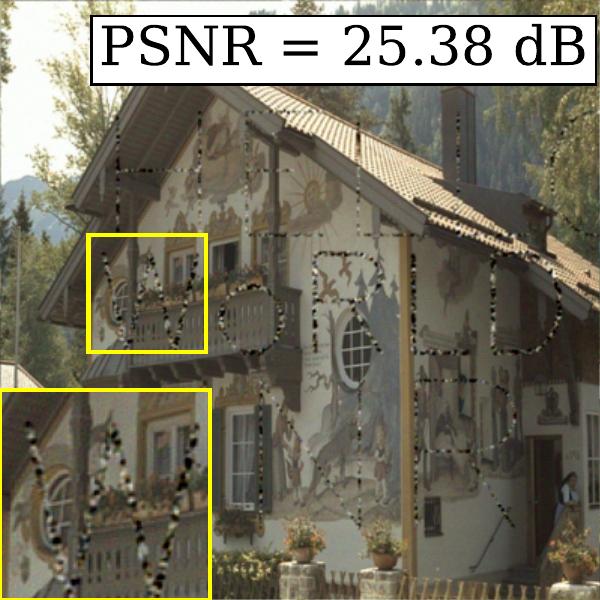}
        \caption{TUNER}
    \end{subfigure}
    \begin{subfigure}{0.24\textwidth}
        \includegraphics[width=\linewidth]{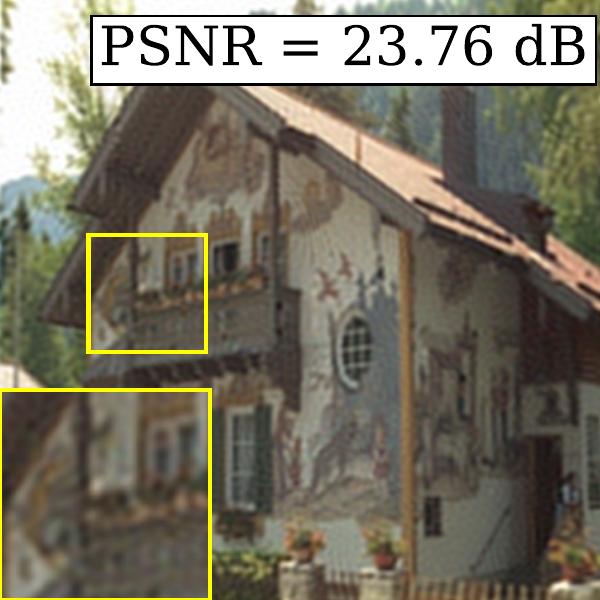}
        \caption{MIRE}
    \end{subfigure}
    \begin{subfigure}{0.24\textwidth}
        \includegraphics[width=\linewidth]{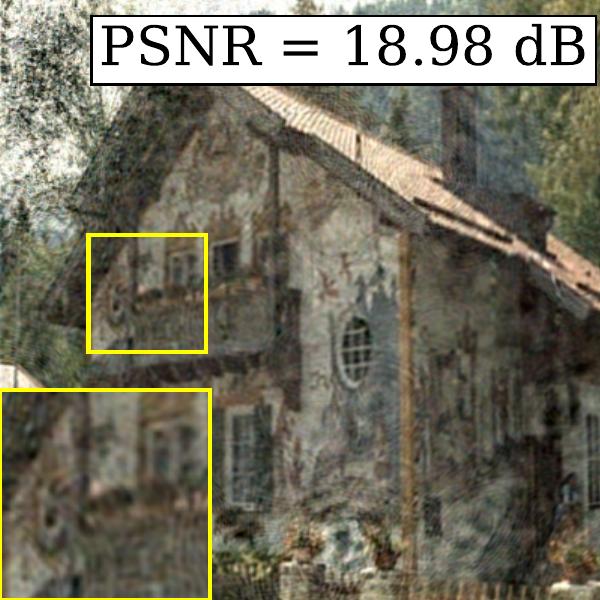}
        \caption{SPDER}
    \end{subfigure}
    \begin{subfigure}{0.24\textwidth}
        \includegraphics[width=\linewidth]{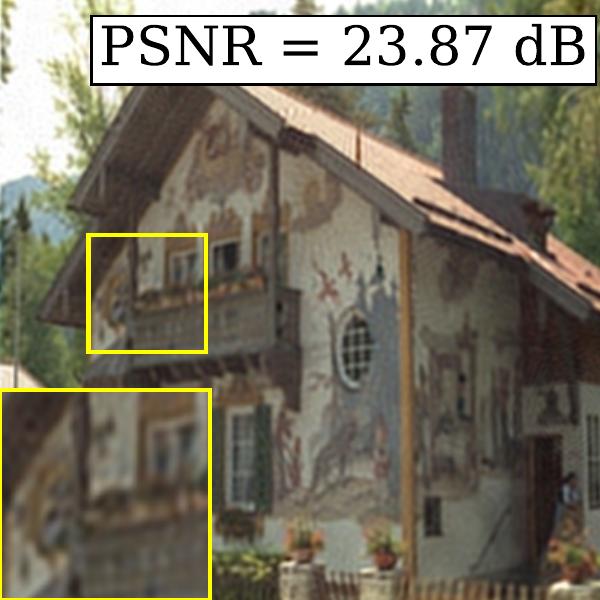}
        \caption{FreSh}
    \end{subfigure}
    \begin{subfigure}{0.24\textwidth}
        \includegraphics[width=\linewidth]{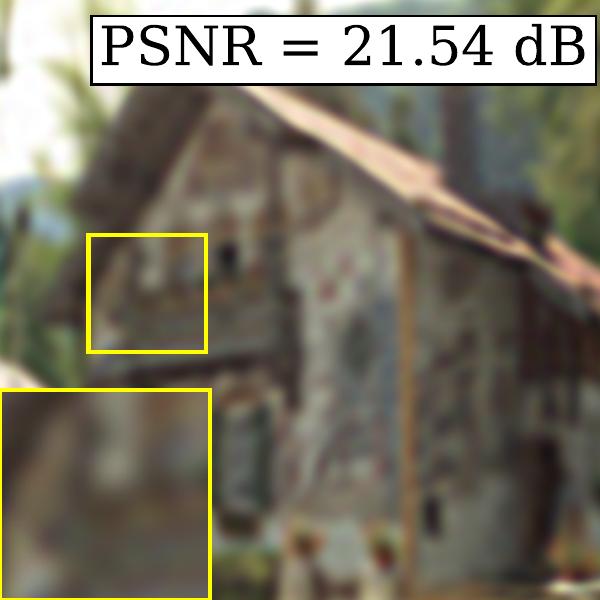}
        \caption{Gauss}
    \end{subfigure}
    \begin{subfigure}{0.24\textwidth}
        \includegraphics[width=\linewidth]{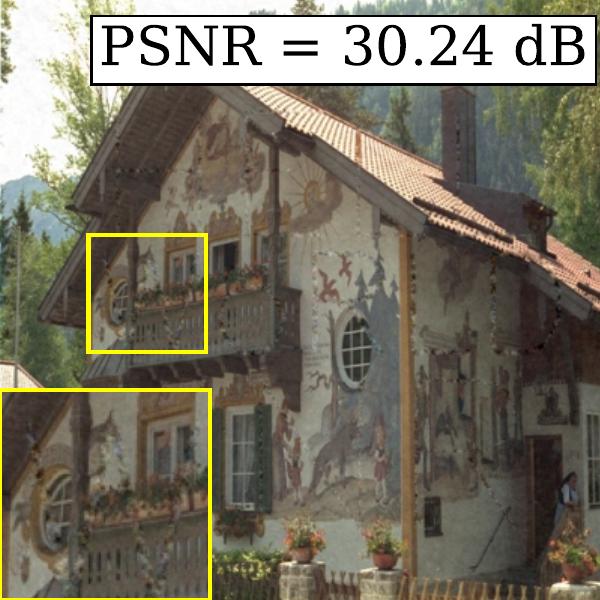}
        \caption{FM-SIREN}
    \end{subfigure}
    \begin{subfigure}{0.24\textwidth}
        \includegraphics[width=\linewidth]{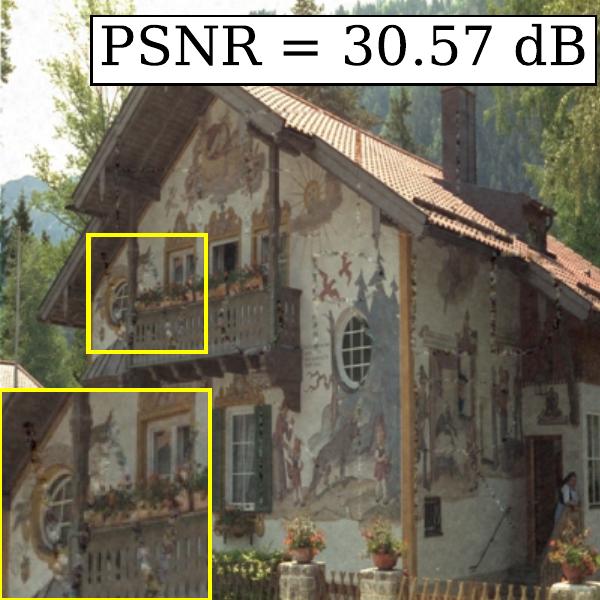}
        \caption{FM-FINER}
    \end{subfigure}
    \caption{Qualitative inpainting results of kodim23 for all baselines. Observe that the PSNR of the proposed FM-FINER and FM-SIREN schemes are superior to all other baselines, demonstrating excellent interpolation for missing pixel while preserving high reconstruction fidelity.}
    \label{fig:inpainting_qual_3}
\end{figure}

\begin{figure}[h]
    \centering
    \begin{subfigure}{0.19\textwidth}
        \includegraphics[width=\linewidth]{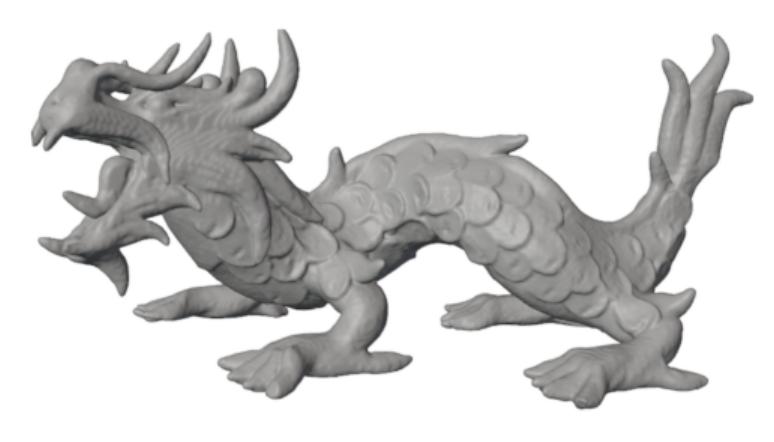}
        \caption{Ground Truth}
    \end{subfigure}
    \begin{subfigure}{0.19\textwidth}
        \includegraphics[width=\linewidth]{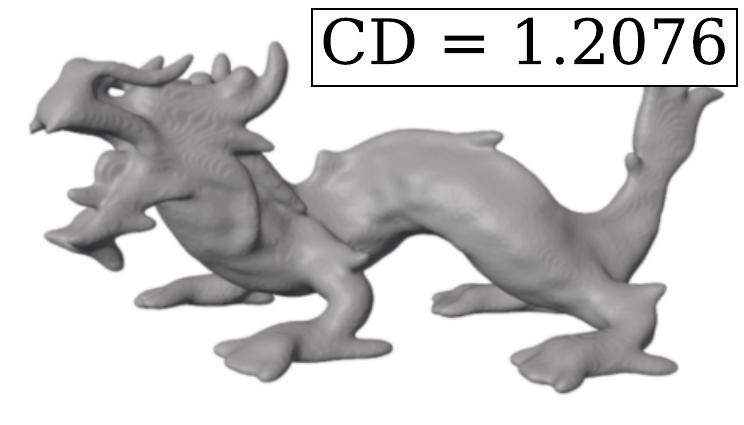}
        \caption{SIREN}
    \end{subfigure}
    \begin{subfigure}{0.19\textwidth}
        \includegraphics[width=\linewidth]{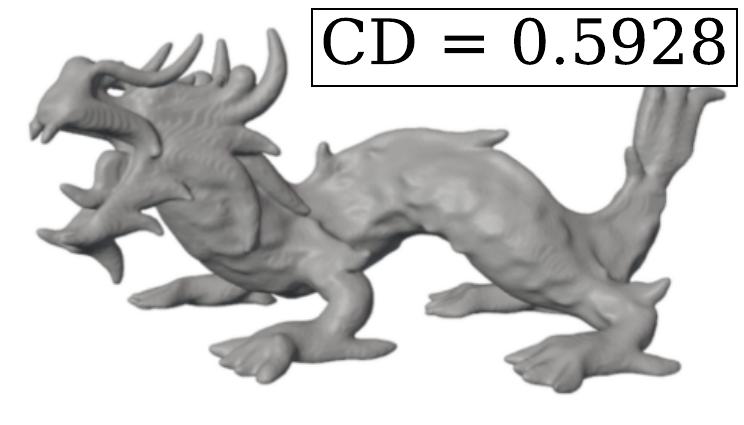}
        \caption{FINER}
    \end{subfigure}
    \begin{subfigure}{0.19\textwidth}
        \includegraphics[width=\linewidth]{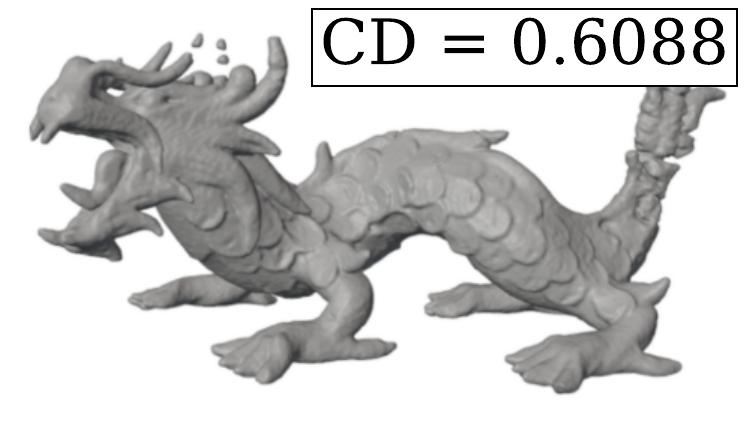}
        \caption{PE}
    \end{subfigure}
    \begin{subfigure}{0.19\textwidth}
        \includegraphics[width=\linewidth]{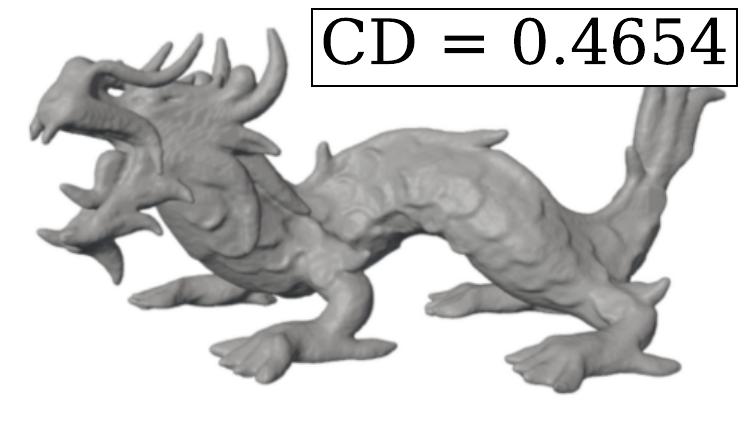}
        \caption{Gauss}
    \end{subfigure}
    \begin{subfigure}{0.19\textwidth}
        \includegraphics[width=\linewidth]{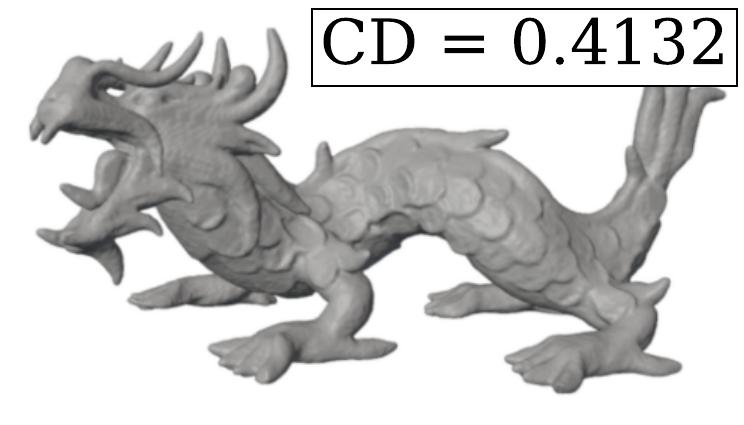}
        \caption{WIRE}
    \end{subfigure}
    \begin{subfigure}{0.19\textwidth}
        \includegraphics[width=\linewidth]{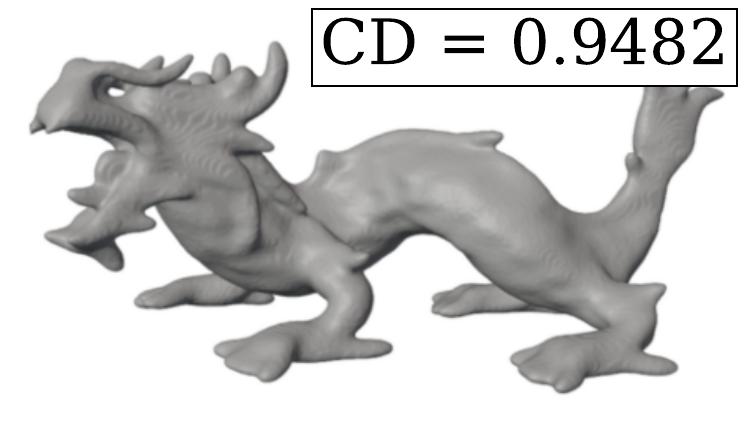}
        \caption{MIRE}
    \end{subfigure}
    \begin{subfigure}{0.19\textwidth}
        \includegraphics[width=\linewidth]{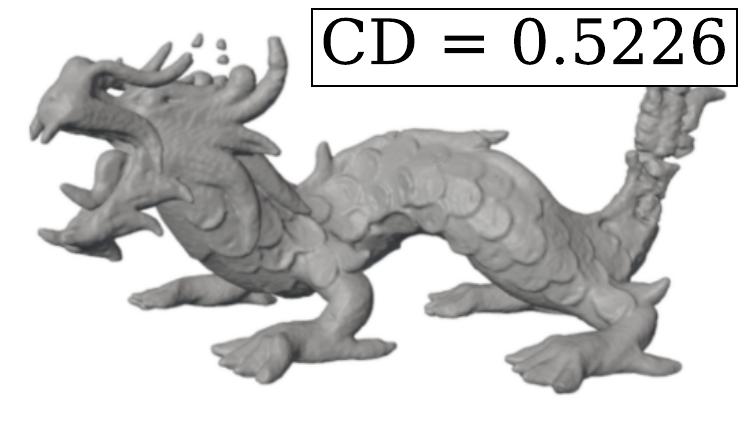}
        \caption{SPDER}
    \end{subfigure}
    \begin{subfigure}{0.19\textwidth}
        \includegraphics[width=\linewidth]{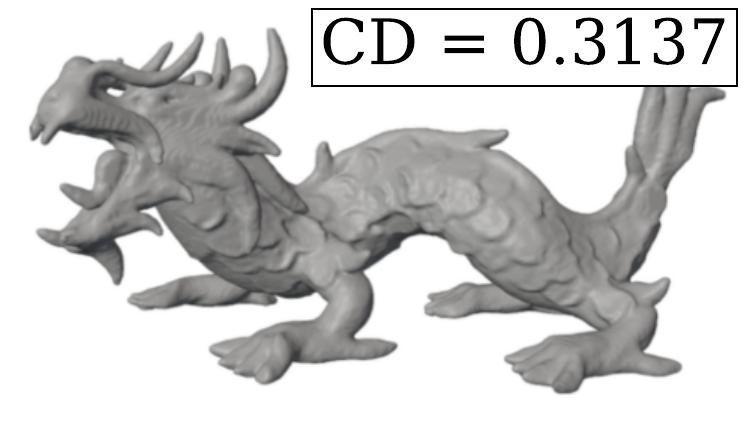}
        \caption{FM-SIREN}
    \end{subfigure}
    \begin{subfigure}{0.19\textwidth}
        \includegraphics[width=\linewidth]{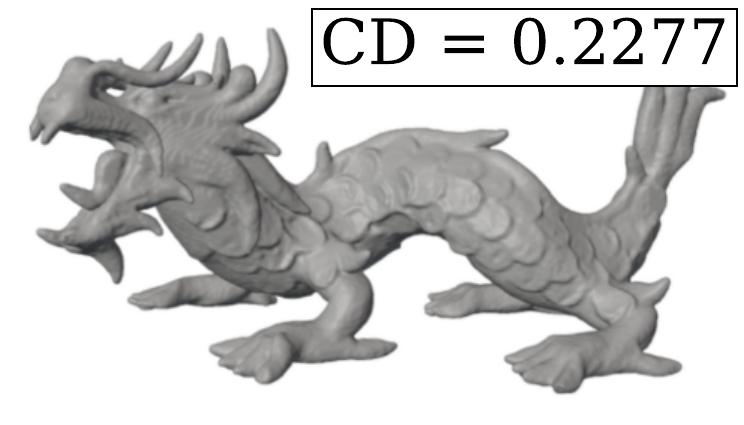}
        \caption{FM-FINER}
    \end{subfigure}
    \caption{Qualitative 3D reconstruction results of the Asian Dragon scene from the Stanford 3D Scanning Repository dataset \cite{stanford_3d_scan}. CD results are shown on the top-right of each subfigure where FM-SIREN and FM-FINER demonstrate superiority to other baselines.}
    \label{fig:3Dfit_1_appendix}
\end{figure}

\begin{figure}[h]
    \centering
    \begin{subfigure}{0.19\textwidth}
        \includegraphics[width=\linewidth]{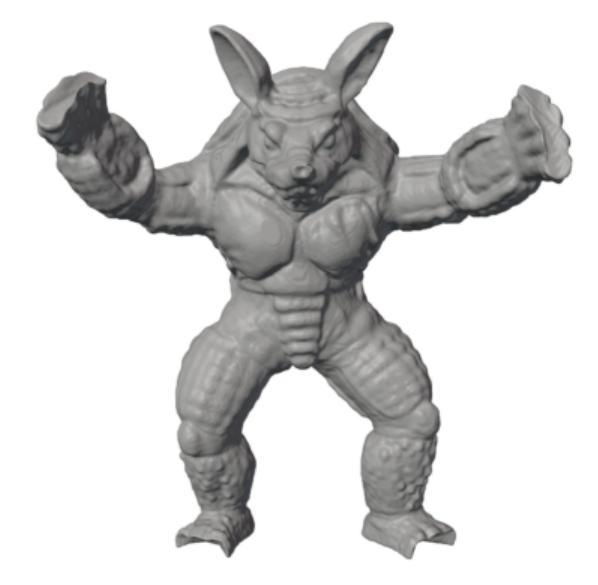}
        \caption{Ground Truth}
    \end{subfigure}
    \begin{subfigure}{0.19\textwidth}
        \includegraphics[width=\linewidth]{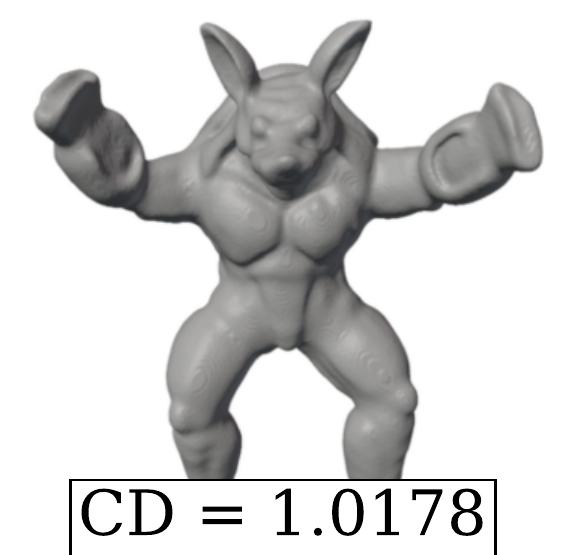}
        \caption{SIREN}
    \end{subfigure}
    \begin{subfigure}{0.19\textwidth}
        \includegraphics[width=\linewidth]{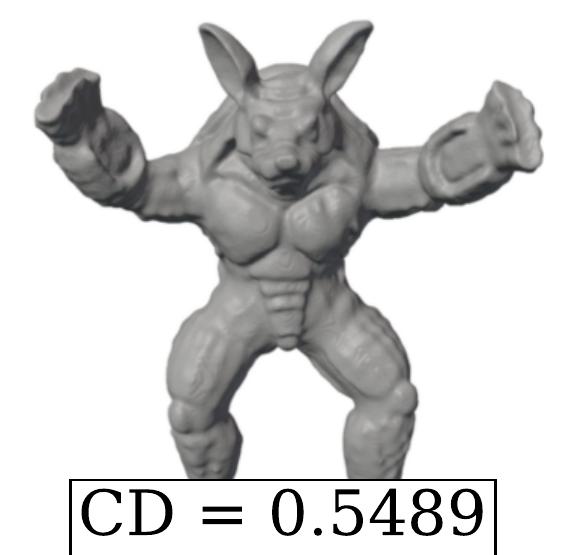}
        \caption{FINER}
    \end{subfigure}
    \begin{subfigure}{0.19\textwidth}
        \includegraphics[width=\linewidth]{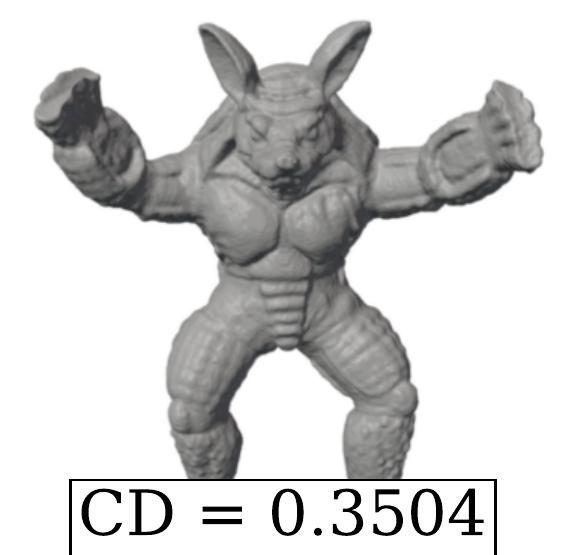}
        \caption{PE}
    \end{subfigure}
    \begin{subfigure}{0.19\textwidth}
        \includegraphics[width=\linewidth]{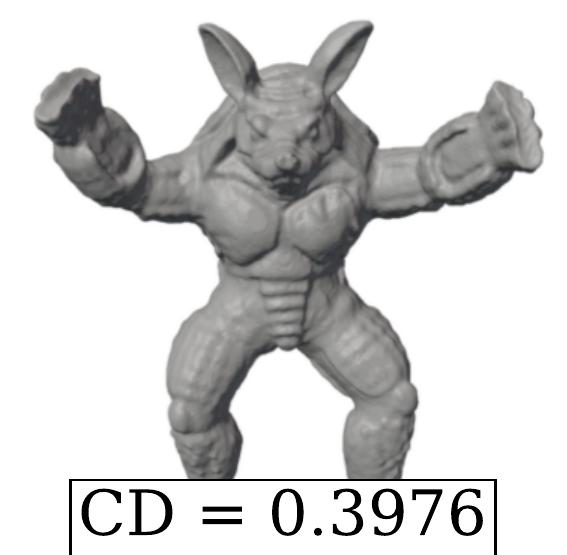}
        \caption{Gauss}
    \end{subfigure}
    \begin{subfigure}{0.19\textwidth}
        \includegraphics[width=\linewidth]{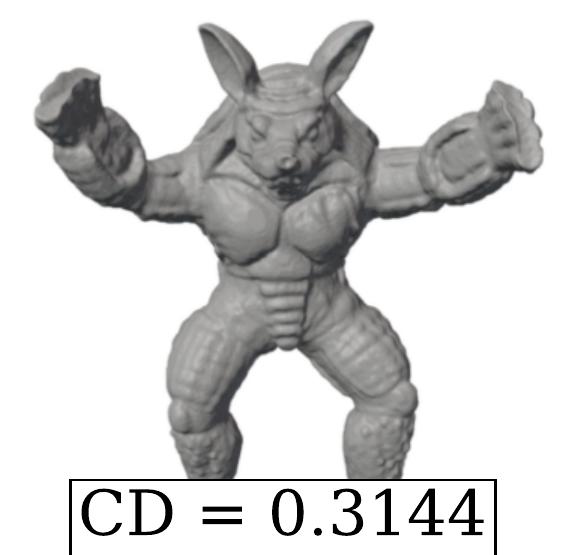}
        \caption{WIRE}
    \end{subfigure}
    \begin{subfigure}{0.19\textwidth}
        \includegraphics[width=\linewidth]{supplementary_material/figures_jpg/3D_shapes/armadillo/SIREN_cd.jpg}
        \caption{MIRE}
    \end{subfigure}
    \begin{subfigure}{0.19\textwidth}
        \includegraphics[width=\linewidth]{supplementary_material/figures_jpg/3D_shapes/armadillo/Gauss_cd.jpg}
        \caption{SPDER}
    \end{subfigure}
    \begin{subfigure}{0.19\textwidth}
        \includegraphics[width=\linewidth]{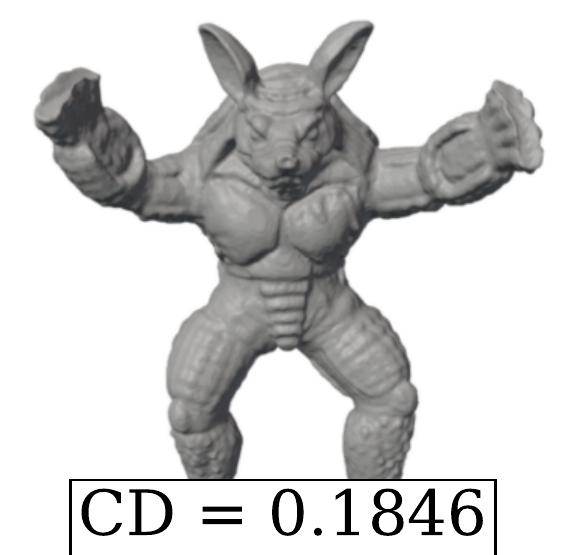}
        \caption{FM-SIREN}
    \end{subfigure}
    \begin{subfigure}{0.19\textwidth}
        \includegraphics[width=\linewidth]{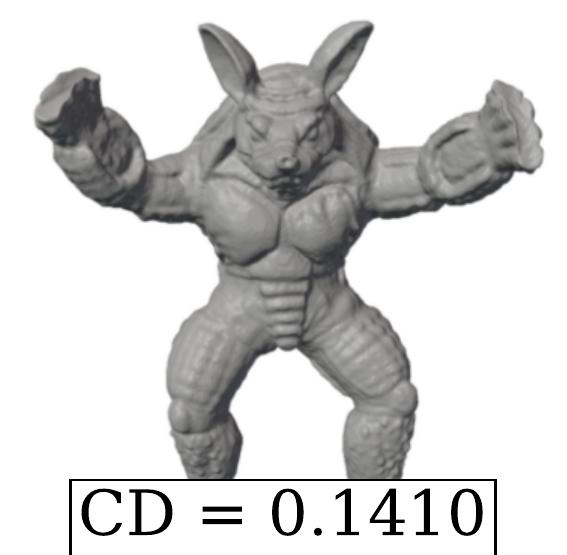}
        \caption{FM-FINER}
    \end{subfigure}
    \caption{Qualitative 3D reconstruction results of the Armadillo scene 
    from the Stanford 3D Scanning Repository dataset \cite{stanford_3d_scan}. CD results are shown on the top-right of each subfigure where FM-SIREN and FM-FINER demonstrate superiority to other baselines.}
    \label{fig:3Dfit_2_appendix}
\end{figure}

\begin{figure}[h]
    \centering
    \begin{subfigure}{0.19\textwidth}
        \includegraphics[width=\linewidth]{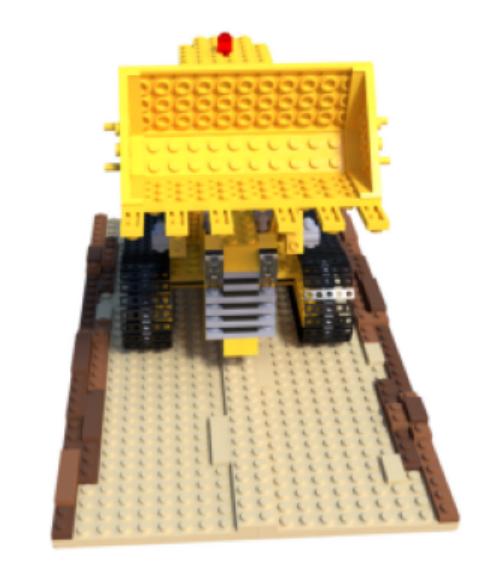}
        \caption{Ground Truth}
    \end{subfigure}
    \begin{subfigure}{0.19\textwidth}
        \includegraphics[width=\linewidth]{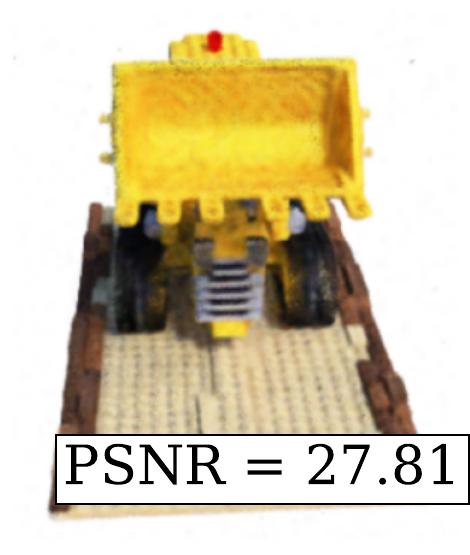}
        \caption{SIREN}
    \end{subfigure}
    \begin{subfigure}{0.19\textwidth}
        \includegraphics[width=\linewidth]{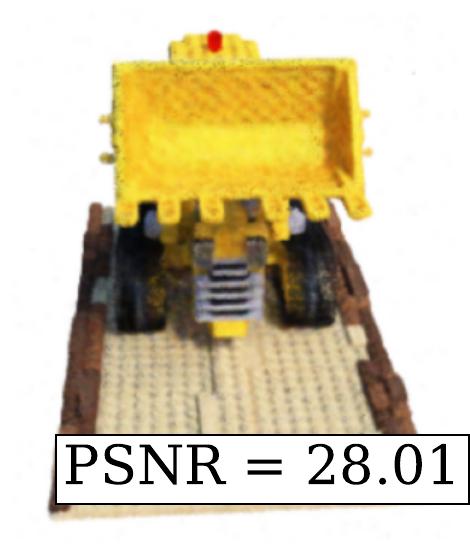}
        \caption{FINER}
    \end{subfigure}
    \begin{subfigure}{0.19\textwidth}
        \includegraphics[width=\linewidth]{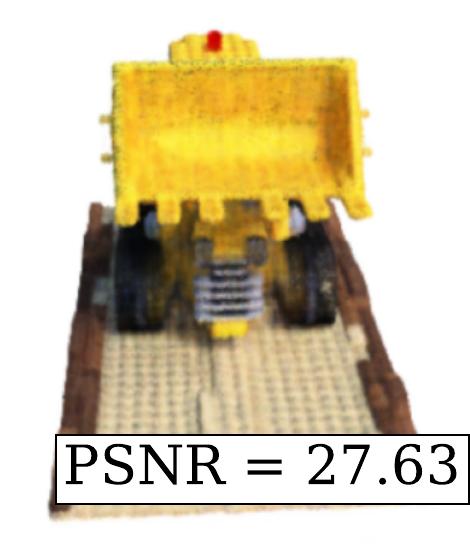}
        \caption{PE}
    \end{subfigure}
    \begin{subfigure}{0.19\textwidth}
        \includegraphics[width=\linewidth]{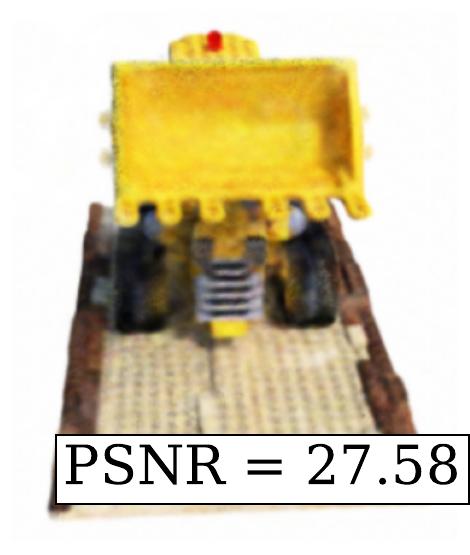}
        \caption{Gauss}
    \end{subfigure}
    \\
    \begin{subfigure}{0.19\textwidth}
        \includegraphics[width=\linewidth]{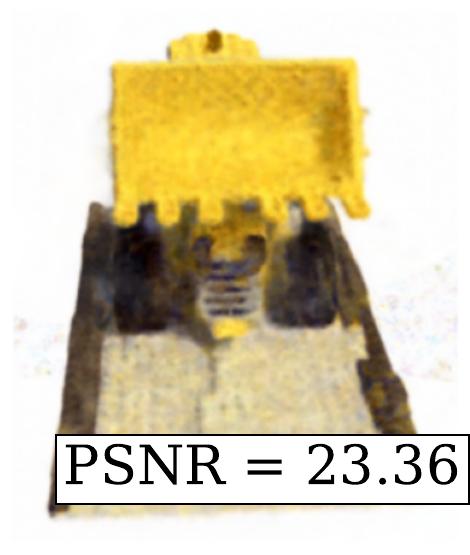}
        \caption{WIRE}
    \end{subfigure}
    \begin{subfigure}{0.19\textwidth}
        \includegraphics[width=\linewidth]{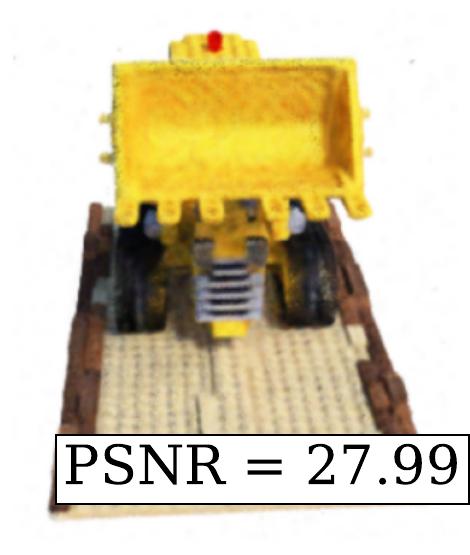}
        \caption{FreSh}
    \end{subfigure}
    \begin{subfigure}{0.19\textwidth}
        \includegraphics[width=\linewidth]{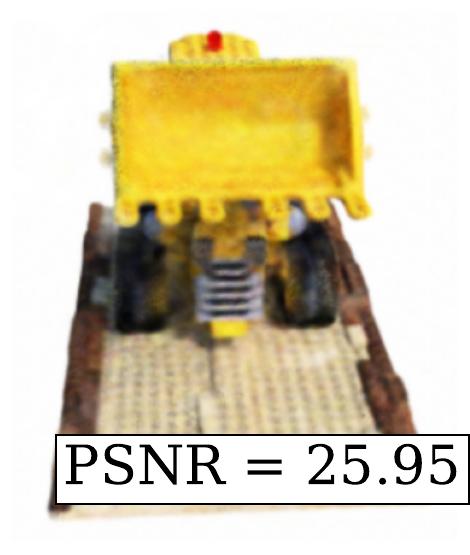}
        \caption{MIRE}
    \end{subfigure}
    \begin{subfigure}{0.19\textwidth}
        \includegraphics[width=\linewidth]{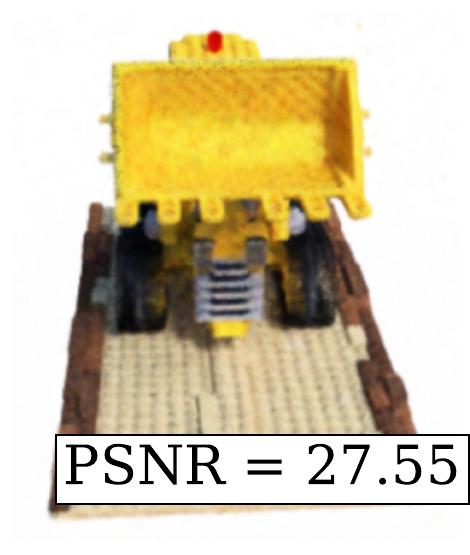}
        \caption{FM-SIREN}
    \end{subfigure}
    \begin{subfigure}{0.19\textwidth}
        \includegraphics[width=\linewidth]{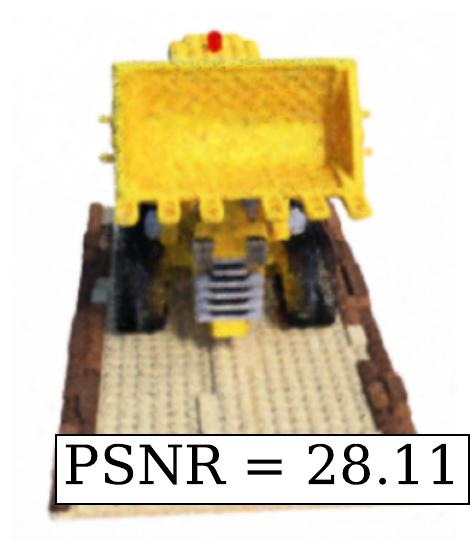}
        \caption{FM-FINER}
    \end{subfigure}
    \vspace{-5pt}
    \caption{Qualitative reconstruction results of the Chair scene from the Blender dataset \cite{mildenhall2021nerf}, using five-layer networks (three layers for the first block, one layer for density, and one layer for color). FM-FINER achieves improvements over all baselines, with exact PSNR values reported at the bottom of each subfigure.}
    \vspace{-10pt}
    \label{fig:NeRF_1}
\end{figure}

\begin{figure}[h]
    \centering
    \begin{subfigure}{0.19\textwidth}
        \includegraphics[width=\linewidth]{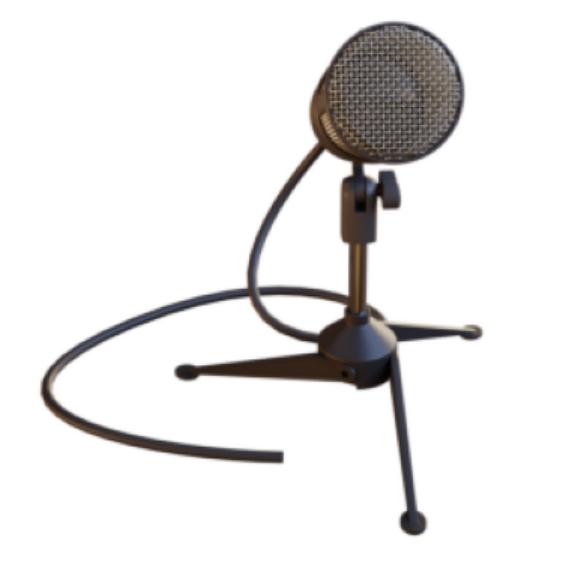}
        \caption{Ground Truth}
    \end{subfigure}
    \begin{subfigure}{0.19\textwidth}
        \includegraphics[width=\linewidth]{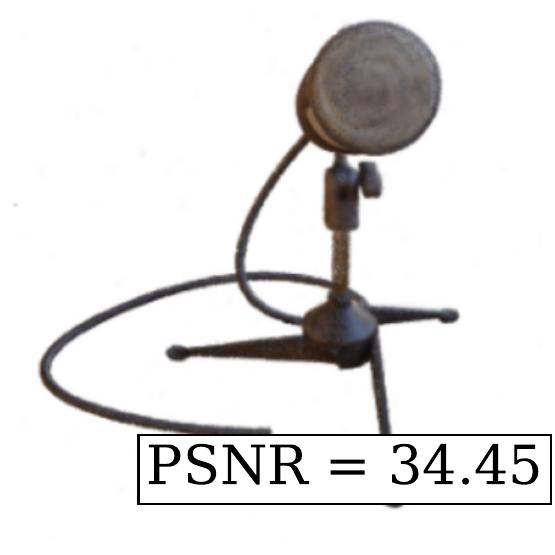}
        \caption{SIREN}
    \end{subfigure}
    \begin{subfigure}{0.19\textwidth}
        \includegraphics[width=\linewidth]{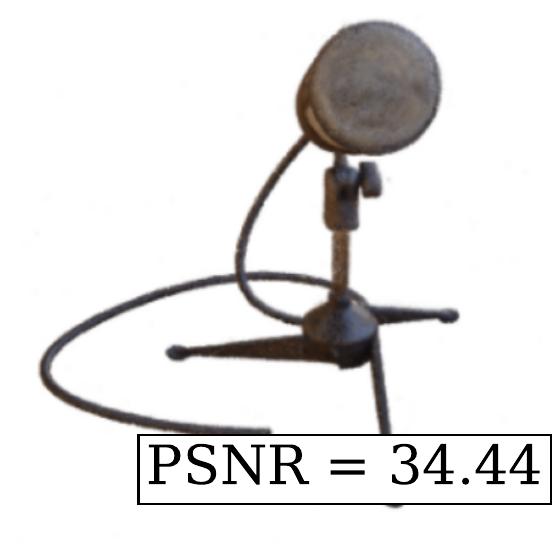}
        \caption{FINER}
    \end{subfigure}
    \begin{subfigure}{0.19\textwidth}
        \includegraphics[width=\linewidth]{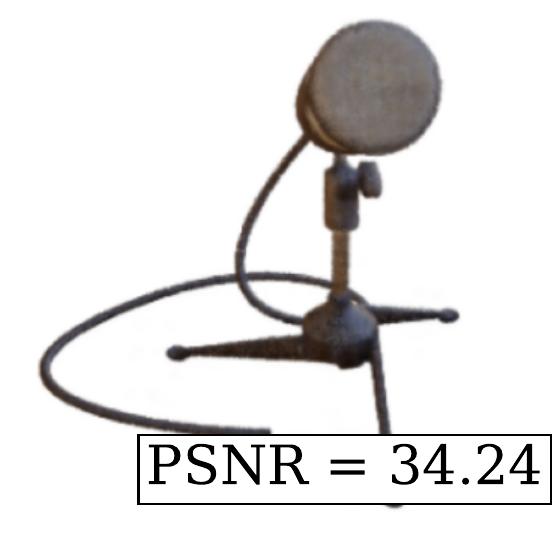}
        \caption{PE}
    \end{subfigure}
    \begin{subfigure}{0.19\textwidth}
        \includegraphics[width=\linewidth]{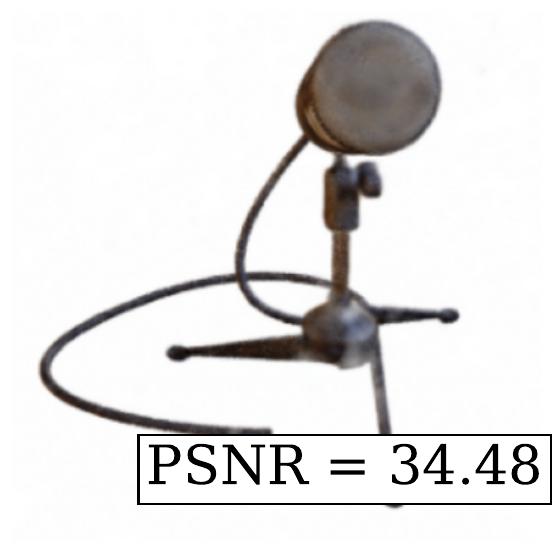}
        \caption{Gauss}
    \end{subfigure}
    \\
    \begin{subfigure}{0.19\textwidth}
        \includegraphics[width=\linewidth]{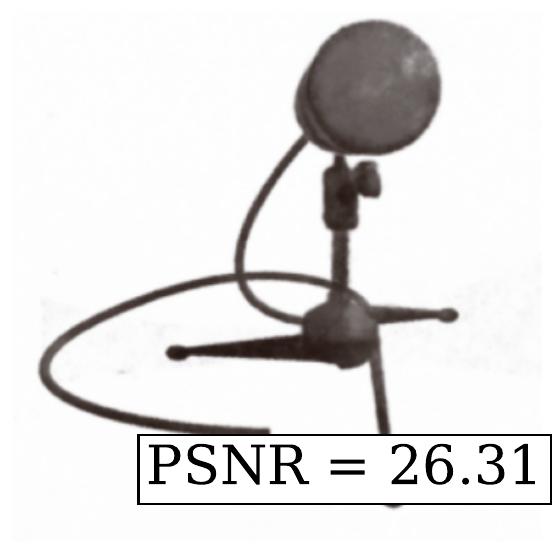}
        \caption{WIRE}
    \end{subfigure}
    \begin{subfigure}{0.19\textwidth}
        \includegraphics[width=\linewidth]{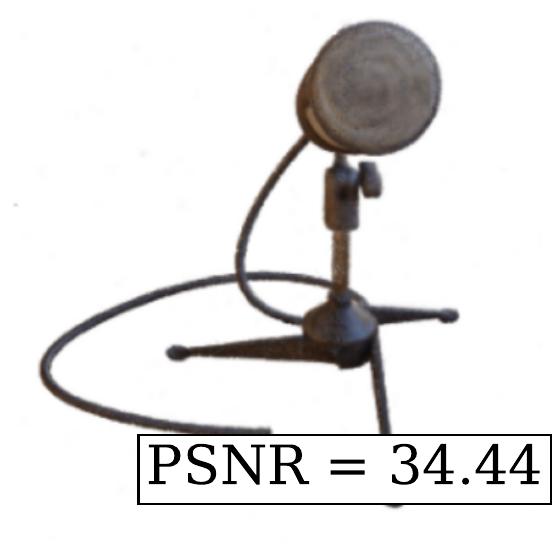}
        \caption{FreSh}
    \end{subfigure}
    \begin{subfigure}{0.19\textwidth}
        \includegraphics[width=\linewidth]{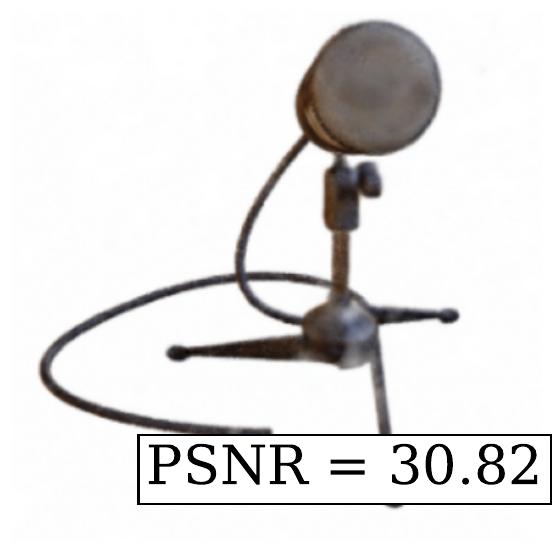}
        \caption{MIRE}
    \end{subfigure}
    \begin{subfigure}{0.19\textwidth}
        \includegraphics[width=\linewidth]{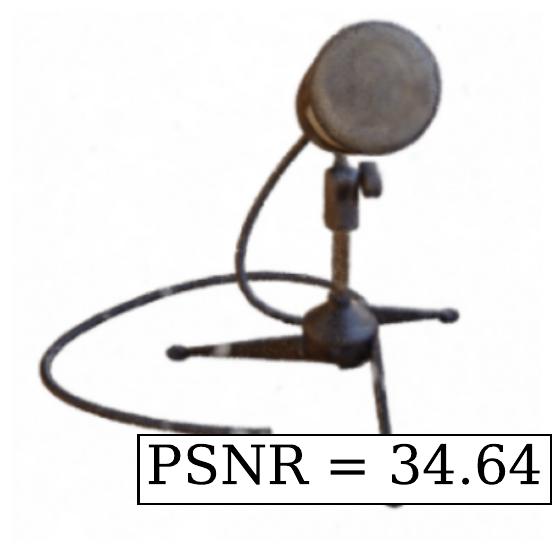}
        \caption{FM-SIREN}
    \end{subfigure}
    \begin{subfigure}{0.19\textwidth}
        \includegraphics[width=\linewidth]{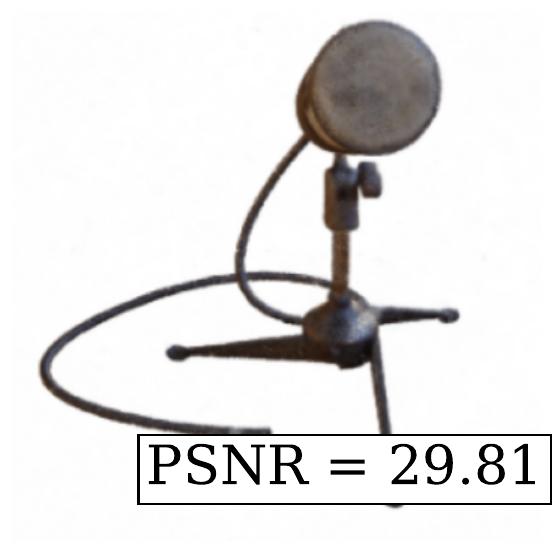}
        \caption{FM-FINER}
    \end{subfigure}
    \vspace{-5pt}
    \caption{Qualitative reconstruction results of the Chair scene from the Blender dataset \cite{mildenhall2021nerf}, using five-layer networks (three layers for the first block, one layer for density, and one layer for color). FM-SIREN achieves improvements over all baselines, with exact PSNR values reported at the bottom of each subfigure.}
    \vspace{-10pt}
    \label{fig:NeRF_2}
\end{figure}

In this section, we provide additional experimental results for all baselines. These results further demonstrate the effectiveness of our proposed frequency orthogonality approach in enhancing the fitting capabilities of implicit neural representation. Our quantitative audio fitting results in Table \ref{tab:appendix_audio} show a superior performance in terms of MSE, which is further supported by the visual comparisons in Figures \ref{fig:audio-fit-1} to \ref{fig:audio-fi-4}. For image fitting, we present supplementary quantitative results for both PSNR and SSIM in Table \ref{tab:3DIoU}, and Figures \ref{fig:2Dfit_1} to \ref{fig:2Dfit_4} visually confirm the higher quality of reconstruction achieved by our models. Additional image inpaiting results are shown in Figures \ref{fig:inpainting_qual_1} to \ref{fig:inpainting_qual_3}. Finally, qualitative results for 3D shape fitting and NeRF are shown in Figures \ref{fig:3Dfit_1_appendix} to \ref{fig:NeRF_2}.

\begin{table}[h!]
    \centering
    \caption{Sample MSE results for audio reconstruction from the Spoken English Wikipedia dataset \cite{spokenwikipedia}. \colorbox{Goldenrod}{Best}, \colorbox{Dandelion}{Second Best} results are highlighted.}
    \label{tab:appendix_audio}
    \begin{tabular}{@{}lcccccc@{}}
    \toprule
    \textbf{Model} & 
    \makecell{\textbf{500 Euro}\\\textbf{Note}\\($\times 10^{-3}$)} & 
    \makecell{\textbf{1995 Pacific}\\\textbf{Grand Prix}\\($\times 10^{-3}$)} & 
    \makecell{\textbf{Alleyway}\\($\times 10^{-3}$)} & 
    \makecell{\textbf{Anfield}\\($\times 10^{-3}$)} & 
    \makecell{\textbf{Munich}\\($\times 10^{-3}$)} & 
    \makecell{\textbf{Average}\\($\times 10^{-3}$)}\\
    \midrule
    FINER & $0.034$ & $0.237$ & $0.204$ & $0.049$ & $0.038$ & $0.1124$\\
    Gauss & $0.803$ & $1.588$ & $3.921$ & $3.493$ & $9.549$ & $3.8708$\\
    MIRE  & $0.791$ & $1.572$ & $3.910$ & $3.467$ & $9.544$ & $3.8568$\\
    PE    & $5.417$ & $6.819$ & $8.521$ & $7.752$ & $15.596$ & $8.8210$\\
    SIREN & $0.069$ & $1.319$ & $3.143$ & $0.276$ & $0.267$ & $1.0148$\\
    SPDER & $0.078$ & $0.972$ & $2.021$ & $0.181$ & $0.173$ & $0.6850$\\
    WIRE  & $1.035$ & $1.891$ & $4.236$ & $3.779$ & $9.883$ & $4.1648$\\
    \midrule
    \textbf{FM-SIREN} & \colorbox{Dandelion}{$0.001$} & \colorbox{Dandelion}{$0.031$} & \colorbox{Dandelion}{$0.020$} & \colorbox{Dandelion}{$0.010$} & \colorbox{Dandelion}{$0.010$} & \colorbox{Dandelion}{$0.0144$}\\
    \textbf{FM-FINER} & \colorbox{Goldenrod}{$0.0005$} & \colorbox{Goldenrod}{$0.007$} & \colorbox{Goldenrod}{$0.003$} & \colorbox{Goldenrod}{$0.004$} & \colorbox{Goldenrod}{$0.008$} & \colorbox{Goldenrod}{$0.0045$}\\
    \bottomrule
    \end{tabular}
    \vspace{20pt}
\end{table}

\begin{table}[h]
    \centering
    \caption{\label{tab:3DIoU} Sample PSNR (dB) results for image reconstruction on images from the Kodak dataset. 
    \colorbox{Goldenrod}{Best}, \colorbox{Dandelion}{Second Best} results are highlighted.}
    \begin{tabular}{@{}l cccccc c@{}}
    \toprule
    \textbf{Model} & \textbf{kodim02} & \textbf{kodim03} & \textbf{kodim07} & \textbf{kodim13} & \textbf{kodim14} & \textbf{kodim22} & \textbf{Average}\\
    \midrule
    FINER         & $32.75$ & $33.05$ & $32.45$ & $24.78$ & $28.74$ & $29.91$ & $30.28$ \\
    FreSh         & $19.18$ & $30.24$ & $28.70$ & $22.10$ & $25.34$ & $27.22$ & $25.46$ \\
    Gauss         & $27.00$ & $26.68$ & $23.06$ & $19.38$ & $21.91$ & $24.31$ & $23.72$ \\
    MIRE          & $29.83$ & $29.95$ & $28.13$ & $21.58$ & $24.96$ & $27.03$ & $26.91$ \\
    PE            & $32.13$ & $32.69$ & $29.93$ & $25.07$ & $28.99$ & $29.50$ & $29.72$ \\
    SIREN         & $20.66$ & $30.24$ & $28.37$ & $22.06$ & $25.19$ & $26.87$ & $25.57$ \\
    SPDER         & $28.80$ & $26.07$ & $27.29$ & $15.45$ & $11.28$ & $24.27$ & $22.19$ \\
    TUNER         & $32.03$ & $30.63$ & $31.07$ & $27.27$ & $29.16$ & $29.02$ & $29.86$ \\
    WIRE          & $30.15$ & $30.33$ & $29.80$ & $22.87$ & $26.61$ & $28.03$ & $27.97$ \\
    \midrule
    \textbf{FM-SIREN} & \colorbox{Dandelion}{$38.26$} & \colorbox{Dandelion}{$38.60$} & \colorbox{Dandelion}{$37.60$} & \colorbox{Dandelion}{$31.73$} & \colorbox{Dandelion}{$34.00$} & \colorbox{Dandelion}{$33.77$} & \colorbox{Dandelion}{$35.66$} \\
    \textbf{FM-FINER} & \colorbox{Goldenrod}{$38.44$} & \colorbox{Goldenrod}{$38.79$} & \colorbox{Goldenrod}{$37.70$} & \colorbox{Goldenrod}{$32.19$} & \colorbox{Goldenrod}{$34.63$} & \colorbox{Goldenrod}{$36.06$} & \colorbox{Goldenrod}{$36.30$} \\
    \bottomrule
    \end{tabular}
\end{table}




\clearpage

\end{document}